\documentclass[letterpaper, 10 pt, conference]{article}  % Comment this line out if you need a4paper
\usepackage [autostyle, english = american]{csquotes}
\MakeOuterQuote{"}
\usepackage{amsmath,array,graphicx}
\usepackage{graphics} % for pdf, bitmapped graphics files
\usepackage[]{subfig}
\usepackage{color}
\usepackage{multirow}
\usepackage{nicefrac}
\usepackage{enumerate}
\usepackage{graphics} % for pdf, bitmapped graphics files
\graphicspath{{images/},{images/hardware/},{images/ja_cases/}}
\newcommand\figref{Figure~\ref}
\usepackage[]{subfig}
\usepackage{makecell}
\usepackage[hyphens]{url}
\usepackage{hyperref}
\hypersetup{
    colorlinks=true,
    linkcolor=blue,
    filecolor=magenta,      
    urlcolor=cyan,
    citecolor=blue,
    breaklinks=true
}
\usepackage[margin=1in]{geometry}

\newcommand{\etal}{\textit{et al}.}

\begin{document}
\title{\LARGE \bf
Joint Attention in Driver-Pedestrian Interaction: from Theory to Practice}
\author{Amir Rasouli and John K. Tsotsos\\
\small Department of Electrical Engineering and Computer Science\\
\small York University, Toronto, ON, Canada\\
\tt\small \{aras,tsotsos\}@eecs.yorku.ca
}

\maketitle

\setcounter{page}{1}
\pagenumbering{arabic}

\begin{abstract}
Today, one of the major challenges that autonomous vehicles are facing is the ability to drive in urban environments. Such a task requires communication between  autonomous vehicles and other road users in order to resolve various traffic ambiguities. The interaction between road users is a form of negotiation in which the parties involved have to share their attention regarding a common objective or a goal (e.g. crossing an intersection), and coordinate their actions in order to accomplish it. 

In this literature review we aim to address the interaction problem between pedestrians and drivers (or vehicles) from joint attention point of view. More specifically, we will discuss the theoretical background behind joint attention, its application to traffic interaction and practical approaches to implementing joint attention for autonomous vehicles.     
 
\end{abstract}

\section{Introduction}
\begin{figure}[!t]
\centering
\includegraphics[width=0.8\textwidth]{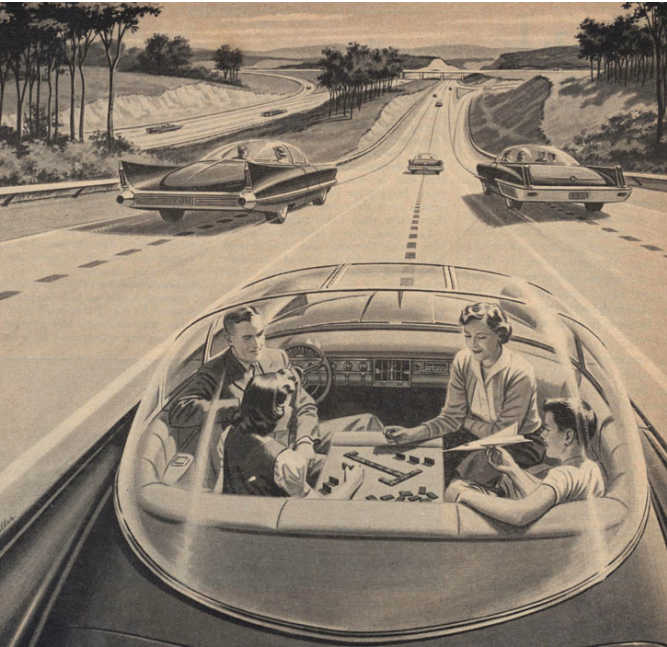}
\caption[A futuristic vision for autonomous driving.]{A view of a futuristic autonomous vehicle in which a family of four are playing a board game while enjoying a ride to their destination, 1956. Source: \cite{kroger2016automated}.}
\label{fig:utopia}
\end{figure}

Ever since the introduction of early commercial automobiles, engineers and scientists have been striving to achieve autonomy, that is removing the need for the human involvement from controlling the vehicles. The fascination with the autonomous driving technology is not new and goes back to the 1950s. In that era articles were appearing in the press featuring the autonomous vehicles in Utopian cities of the future (\figref{fig:utopia}) where drivers, instead of spending time controlling the vehicles, could interact with their family members or undertake other activities while enjoying the ride to their destinations \cite{kroger2016automated}.

Apart from the increased level of comfort for drivers, autonomous vehicles can positively impact society both at the micro and macro levels. One important aspect of autonomous driving is the elimination of driver involvement, which reduces the human errors (e.g. fatigue, misperception or inattention), and consequently, lowers the number of accidents (up to 93.5\%) \cite{winkle2016safety}. The reduction in human error can improve both the safety of the driver or the passengers of the vehicle and other traffic participants such as pedestrians.

At the macro level, fleets of autonomous vehicles can improve the efficiency of driving, better the flow of traffic and reduce car ownership (by up to 43\%) through car sharing, all of which can minimize the energy consumption, and as a result, lower the environmental impacts such as air pollution and road degradation \cite{litman2014autonomous}.

Throughout the past century, the automotive industry has witnessed many significant breakthroughs in the field of autonomous driving, ranging from simple lane following \cite{dickmanns1987curvature} to complex maneuvers and interaction with traffic in complex urban environments \cite{darms2008multisensor}. Today, autonomous driving has become one of the major topics of interest in technology. This field not only has attracted the attention of the major automotive manufacturers, such as BMW, Toyota, and Tesla, but also enticed a number of technology giants such as Google, Apple and Intel.

Despite the significant amount of interest in the field, there is still much to be done to achieve fully autonomous driving behavior in a sense of designing a vehicle capable of handling all dynamic driving tasks without any human involvement. One of the major challenges, besides developing efficient and robust algorithms for tasks such as visual perception and control, is communication with other road users in chaotic traffic scenes. Communication is a vital component in traffic interactions and is often relied upon by humans to resolve various ambiguities such as yielding to others or asking for the right of way. In order for communication to be effective, the parties require to understand each others' intentions as well as the context in which the communication is taking place. 

The aim of this paper is to address the aforementioned problems from autonomous driving perspective. Particularly, the focus is on understanding pedestrian behavior at crosswalks. For this purpose, we organize the rest of this paper into three main chapters. In Chapter one, we present a brief introduction to autonomous driving and discuss some of the major milestones, and unresolved challenges in the field. Chapter two focuses on the theories behind the problems, starting with joint attention and its role in human interaction, and continues by discussing some of the studies on nonverbal communication and behavior understanding, with a particular focus on pedestrian crossing behavior. In addition, in this chapter we elaborate on human general reasoning techniques to highlight how we, as humans, make decisions in traffic interactions. Chapter three, which comprises more than half of this report, addresses the practical challenges in pedestrian behavior understanding. To this end, this chapter reviews the state-of-the-art algorithms and systems for solving different aspects of the problem from two different perspectives, hardware and software. The hardware section describes various physical sensors used for these purposes, and the software section deals with processing the raw data from the sensors to perform tasks such as object detection, pose estimation and activity recognition, and decision-making tasks such as action planning and reasoning.  

\part{Autonomous Driving and Challenges Ahead}
\section{Autonomous Driving: From the Past to the Future}
Before reviewing the development of autonomous driving technologies, it is necessary to define what we mean by autonomy in the context of driving. Traditionally, there are four levels of autonomy including no autonomy (the driver is in the control of all driving aspects), advisory autonomy (such as warning systems in the vehicle which partially aid the driver), partial control (such as auto braking or lane adjustment) and full control (all aspects of the dynamic driving tasks are handled autonomously)\cite{bishop2000intelligent}.

\begin{figure}[!t]
\centering
\includegraphics[width=1\textwidth]{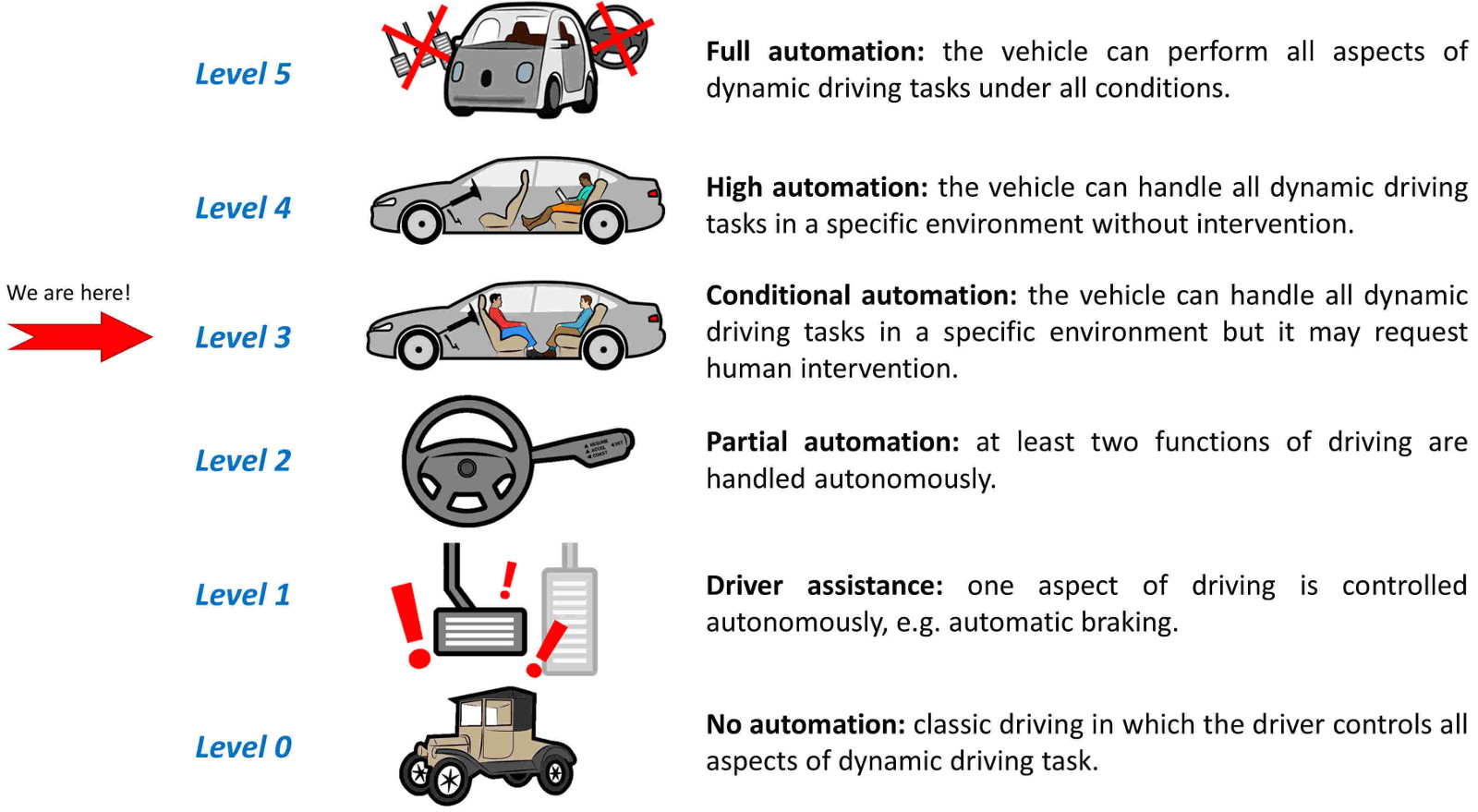}
\caption[Levels of autonomous driving.]{Six levels of driving autonmation. Today we have achieved level 3 autonomy. Source: \cite{Radovanovic2016}.}
\label{fig:level_autonomy}
\end{figure}

Today, the automotive industry further breaks down the levels of autonomy into six categories: (see \figref{fig:level_autonomy})\cite{sae2017}:\\

\noindent\textbf{Level 0:} \textit{No Automation}, where the human driver controls all aspects of the dynamic driving tasks. This level may include enhanced warning system but no automatic control is taking place.\\
\textbf{Level 1:} \textit{Driver Assistance}, where only one function of driving such as steering or acceleration/deceleration, using information about the driving environment, is handled autonomously. The driver is expected to control all other aspects of driving.\\
\textbf{Level 2:} \textit{Partial Automation}. In this mode, at least two functionalities of the dynamic driving tasks, in both steering and acceleration/deceleration, are controlled autonomously. \\
\textbf{Level 3:} \textit{Conditional Automation}, where the autonomous system can handle all aspects of the dynamic driving tasks in a specific environment, however, it may require human intervention in the cases of failure.\\
\textbf{Level 4:} \textit{High Automation}. This mode is similar to level 3 with the exception that no human intervention is required at any time during the environment specific driving task.\\
\textbf{Level 5:} \textit{Full Automation}. As the name implies, in this mode all aspects of the dynamic driving tasks under any environmental conditions are fully handled by an automated system.

The current level of autonomy available in the market, such as the one in Tesla, is level 2. Some manufacturers such as Audi are also promising autonomy level 3 capability on their newest models such as A8 \cite{Nguyen2017}.

\begin{figure}[!t]
\centering
\includegraphics[width=1\textwidth]{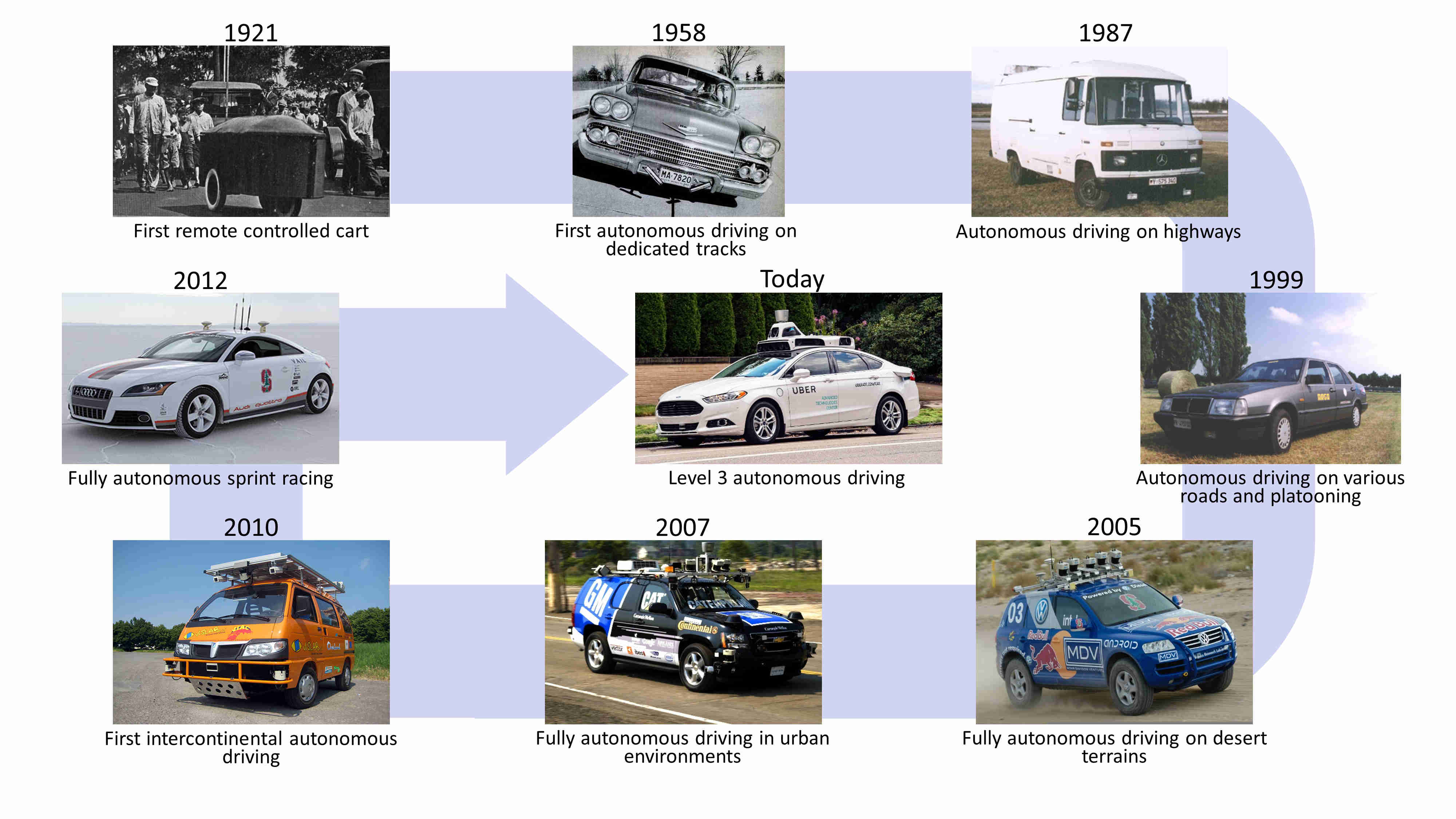}
\caption[A visual history of advancements in autonomous driving.]{A century of developments in driving automation. This timeline highlights some major milestones in autonomous driving technologies from the first attempts in the 1920s to today's modern autonomous vehicles. Source (in chronological order): \cite{cart1921,kroger2016automated,Oagana2016,broggi1999argo,stanley2005,boss2007,vislab2010,shelley2012,Davieg2016}}
\label{fig:autonomy_milestones}
\end{figure}

The following subsections will review the developments in the field of autonomous driving during the past century. A summary of some of the major milestones are illustrated in \figref{fig:autonomy_milestones}.

\subsection{The Beginning}
\begin{figure}[!t]
\centering
\subfloat[]{
\includegraphics[width=7cm,height=6cm]{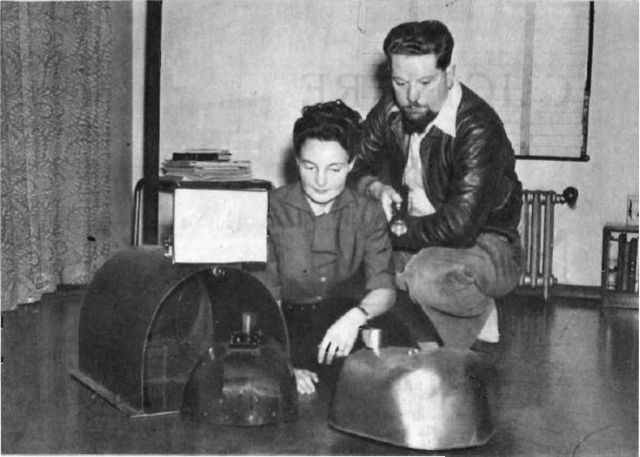}
\label{fig:walter_tortoise} }
\subfloat[]{
\includegraphics[width=4cm,height=6cm]{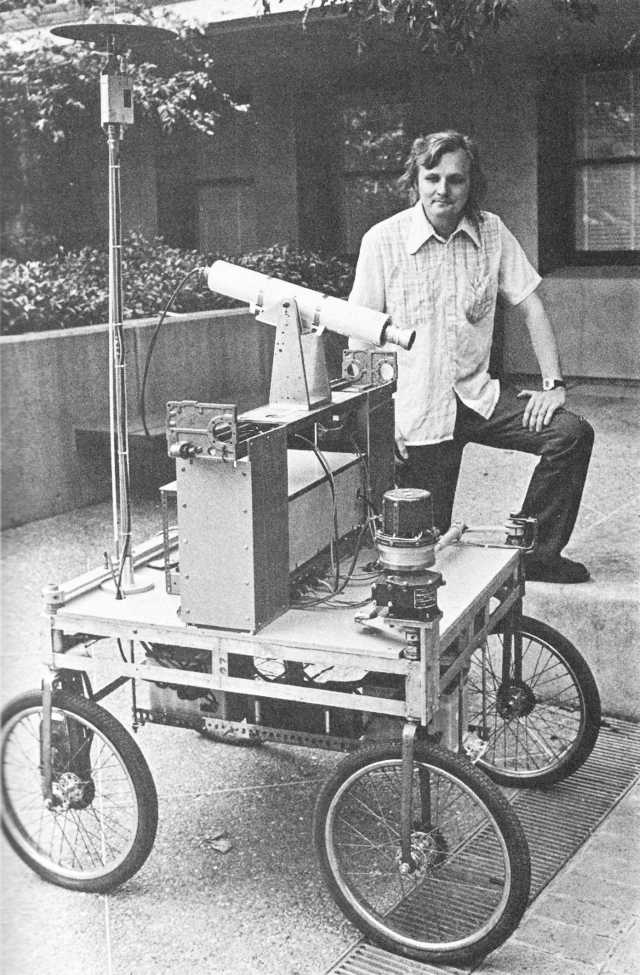}
\label{fig:moravec} }
\caption[Walter's Tortoises and Standford Cart.]{a) W. G. Walter and his Tortoises \cite{walter1948image}, and b) Hans Moravec and Stanford Cart \cite{moravecimage}.}
\label{fig:early_robots}
\end{figure}

Much of today's autonomous driving technology is owing to the pioneering works of roboticists such as Sir William Grey Walter, a British neurophysiologist who invented the robots Elsie and Elmer (also known as Tortoises)(\figref{fig:walter_tortoise}), in the year 1948 \cite{walter1948}. These simple robotic agents are equipped with light and pressure sensors and are capable of phototaxis by which they can navigate their way through the environment to their charging station. The robots are also sensitive to touch which allows them to detect simple obstacles on their path.

A more modern robotic platform capable of autonomous behaviors is Stanford Cart (\figref{fig:moravec}) \cite{moravec1980obstacle,moravec1983stanford}. This mobile platform is equipped with an active stereo camera and could perceive the environment, build an occupancy map and navigate its way around obstacles. In terms of performance, the robot successfully navigated a 20 m course in a room filled with chairs in just under 5 hours.

Autonomous vehicles also rely on similar techniques as in robotics to perform various perception and control tasks. However, since vehicles are used on roads, they generally require different and often stricter performance evaluations, in terms of robustness, safety, and real time reactions. In the remainder of this report we will particularly focus on robotic applications that are used in the context of autonomous driving.

\begin{figure}[!t]
\centering
\subfloat[]{
\includegraphics[width=6cm,height=4cm]{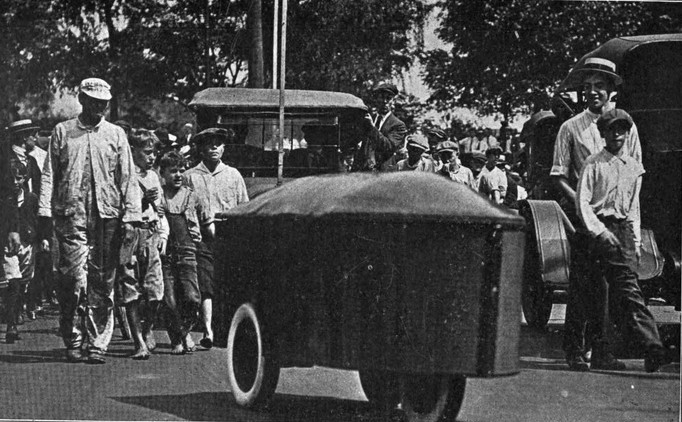}
\label{fig:cart} }
\subfloat[]{
\includegraphics[width=10cm,height=4cm]{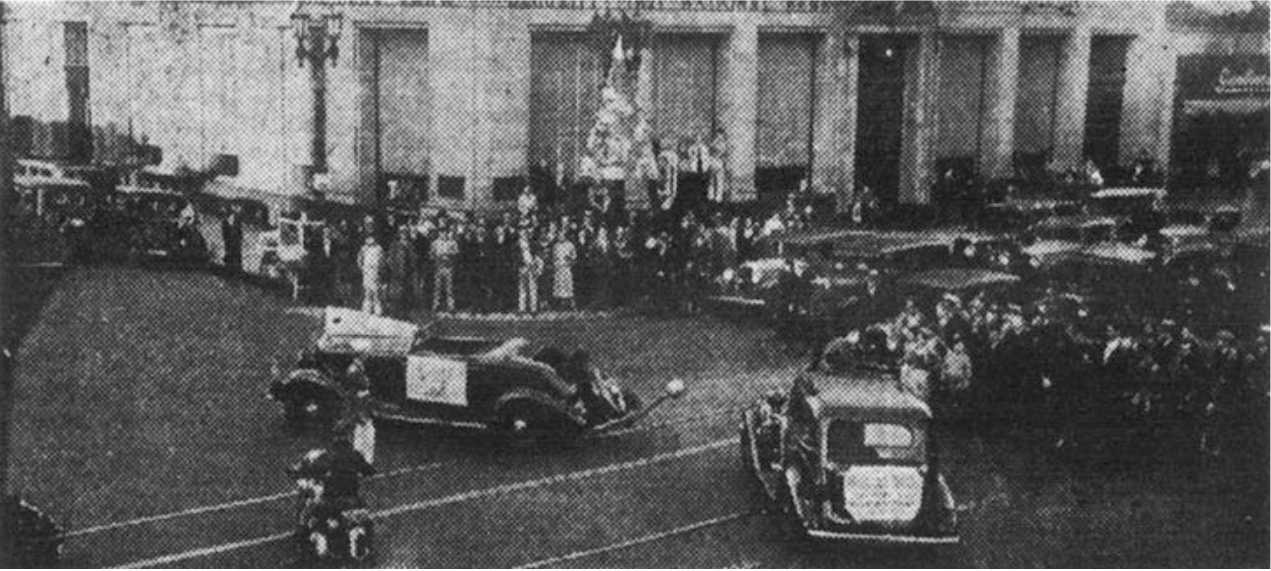}
\label{fig:safety_parade} }
\caption[Early remote-controlled vehicles.]{a) The first remote-controlled vehicle, 1921 \cite{cart1921}, and b) a more modern version of a commercial vehicle at Safety Parade 1936 \cite{kroger2016automated}.}
\label{fig:earily_autonomous_cars}
\end{figure}

Early attempts at developing autonomous driving technology go as far back as the first commercial vehicles. In this era, autonomous driving was realized in the form of remote-controlled vehicles removing the need for the driver to be physically present in the car.

In 1921, the first driverless car (\figref{fig:cart}) was developed by the McCook air force test base in Ohio \cite{kroger2016automated}. This 2.5 meter-long cart was controlled via radio signals transmitted from the distance of up to 30 m. In the 1930s, this technology was implemented on actual vehicles some of which were  exhibited in various parades (\figref{fig:safety_parade}) to promote the future of driveless cars and to show how they can increase driving safety \cite{kroger2016automated}.

\subsection{Hitting the Road}
The first instance of driving without human involvement was introduced in 1958 by General Motors (GM). The autonomous vehicle called "automatically guided automobile" was capable of autonomous driving on a test track with electric wires laid on the surface which were used to automatically guide the vehicle steering mechanism \cite{kroger2016automated}.

In the late 1980s, one of the pioneers of modern autonomous driving, E. D. Dickmanns \cite{dickmanns1987curvature,mysliwetz1987distributed}, alongside his team of researchers at Daimler, developed the first visual algorithm for road detection in real time. They employed two active cameras to scan the road, detect its boundaries, and then measure its curvature. To reduce computation time, a Kalman filter was used to estimate the changes in the curvature as the car was traversing the road.

\begin{figure}[t]
\centering
\includegraphics[width=1\textwidth]{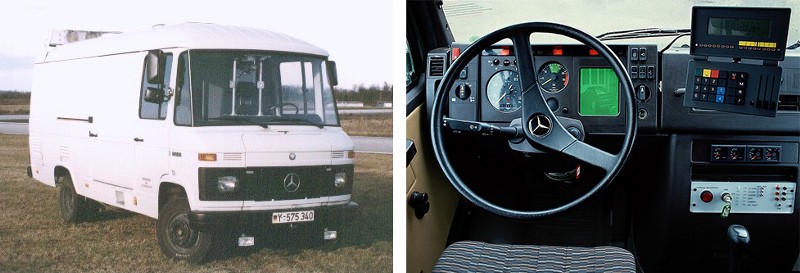}
\caption[VaMoRs, the autonomous car.]{VaMoRs and a view of its interior \cite{Mueller2017}.}
\label{fig:vamors}
\end{figure}

In the early 1990s, the team at Daimler enhanced the algorithm by adding obstacle detection capability. This algorithm identifies the parts of the road as obstacles if their height is more than a certain elevation threshold above the 2D road surface \cite{dickmanns1990integrated}. In the same year, the visual perception algorithm was tested on an actual Mercedes van, VaMoRs (\figref{fig:vamors}). Using the algorithm in conjunction with an automatic steering mechanism, VaMoRs was able to drive up to the speed of 100 km/h on highways, and up to 50 km/h on regional roads. The vehicle could also perform basic lane changing maneuvers and safely stop before obstacles when driving up to 40 km/h.

Throughout the same decade, we witnessed the emergence of learning algorithms such as neural nets which were designed to handle various driving tasks. ALVINN is an example of such systems that was developed as part of the NAVLAB autonomous car project by Carnegie Melon University (CMU). The system uses a neural net algorithm to learn and detect different types of roads (e.g. dirt or asphalt) and obstacles \cite{pomerleau1991combining,pomerleau1992progress,pomerleau1996neural, baluja1996evolution}. The algorithm, besides passive camera sensors, relies on laser range finders and laser reflectance sensors (for surface material assessment) to achieve a more robust detection.

To guide the vehicle, a similar learning technique is used by the NAVLAB team to learn driving controls from recordings collected from an expert driver \cite{thorpe1991toward}. An extension of this project uses an online supervised learning method to deal with illumination changes, and a neural net to identify more complex road structures such as intersections \cite{jochem1995vision}. The NAVLAB project is implemented on a U.S. Army HMMWV and is capable of obstacle avoidance and autonomous driving up to 28 km/h on rugged terrains and 88 km/h on regular roads.

Despite the fact that learning algorithms achieved promising performance in various visual perception tasks, in the late 90s, the traditional vision algorithms still remained popular. Methods such as color thresholding \cite{broggi1999argo} and various edge detection filters such as Sobel filters \cite{yim2003three} or model-based algorithms for road boundary estimation and prediction \cite{kluge1994extracting} were widely used.

In the mid-90s, autonomous assistive technologies have become standard features in a number of commercial vehicles. For instance, an extension of the lane detection and following algorithms developed by Dickmanns' team \cite{dickmanns1994seeing,franke1994daimler} was used in the new lines of Mercedes-Benz vehicles \cite{bohrer1995integrated}. This new extension, in addition to road detection, can detect cars by identifying symmetric patterns of their rear views. Using the knowledge of the road, the automatic system adjusts the position of the vehicle within the lanes and performs emergency braking if the vehicle gets too close to an obstacle. An interesting feature of this system is the ability to track objects, allowing the vehicle to autonomously follow a car in the front, i.e. the ability to platoon.

The new millennium was the time in which autonomous vehicles started to enjoy the technological advancements in both the design of sensors and increase in computation power. At this time, we observe an increase in the use of high power sensors such as GPS, LIDAR, high-resolution stereo cameras \cite{hong2000intelligent} and IMU \cite{behringer2004darpa}. The information from various sources of sensors was commonly used by autonomous vehicles, thanks to the availability of high computation power, which allowed them to achieve a better performance in tasks such as assessment of the environment, localization, and navigation of the vehicle and mapping. The emergence of such features brought the automotive industry one step closer to achieving full autonomy.

\subsection{Achieving Autonomy}
In 2004 the Defense Advanced Research Projects Agency (DARPA) organized one of the first autonomous driving challenges in which the vehicles were tasked to traverse a distance of 240 km between Las Vegas and Los Angeles \cite{behringer2004darpa}. In this competition, none of the 15 finalists were able to complete the course, and the longest distance traversed was only 11.78 km by the team Red from CMU.

\begin{figure}[!t]
\centering
\subfloat[]{
\includegraphics[width=8cm,height=6cm]{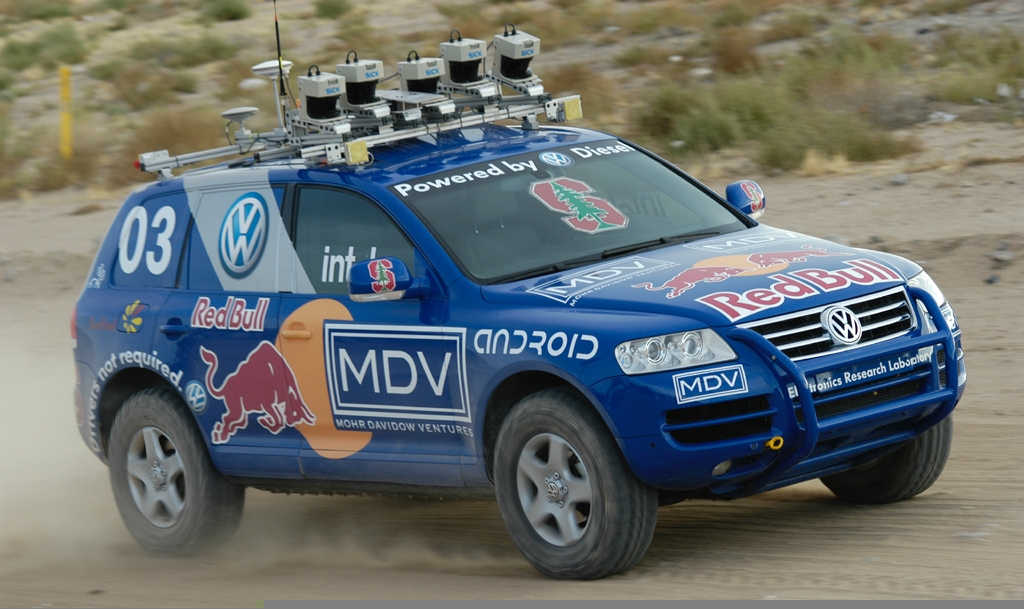}
\label{fig:stanley} }
\subfloat[]{
\includegraphics[width=8cm,height=6cm]{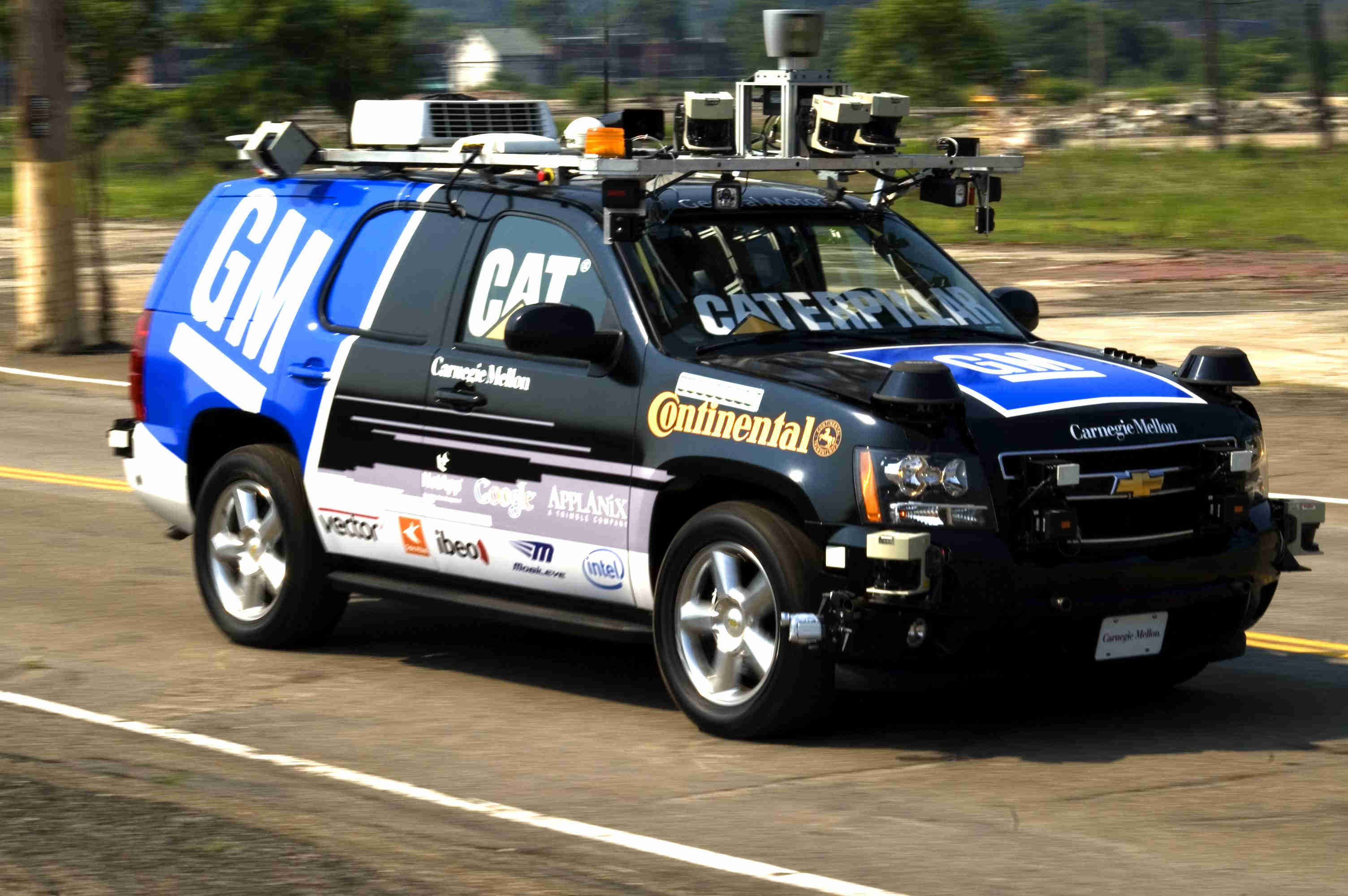}
\label{fig:boss} }
\caption[Stanley and BOSS at DARPA challenge.]{a) Stanley from Stanford in DARPA 2005 \cite{stanley2005}, and b) BOSS from CMU in DARPA 2007 \cite{boss2007}.}
\label{fig:darpa}
\end{figure}

The following year a similar challenge was held over the course of 212 km on a desert terrain between California and Nevada \cite{dang2006path}. In this year, however, 5 cars finished the entire course (one of them over the 10 hours limit), out of 23 teams that participated in the final event. Stanley (\figref{fig:stanley}), the winning car from Stanford, finished the race under 6 hours and 53 minutes while maintaining an average speed of 30 km/h throughout the race \cite{thrun2006stanley}.

Stanley benefited from various sources of sensory input including a mono color camera for road detection and assessment, GPS for global positioning and localization, and RADAR and laser sensors for long and short-range detection of the road respectively. The  Stanley project produced a number of state-of-the-art algorithms for autonomous driving such as the probabilistic traversable terrain assessment method \cite{thrun2006probabilistic}, a supervised learning algorithm for driving on different surfaces \cite{hoffmann2007autonomous} and a dynamic path planning technique to deal with challenging rugged roads \cite{dolgov2008practical}.

In the year 2007, DARPA hosted another challenge, and this time it took place in an urban environment. The goal of this competition was to test vehicles' ability to drive a course of 96 kilometers under 6 hours on urban streets while obeying traffic laws. The cars had to be able to negotiate with other traffic participants (vehicles), avoid obstacles, merge into traffic and park in a dedicated spot. In addition to robot cars, some professional drivers were also hired to drive on the course.

Among the 11 finalists, BOSS (\figref{fig:boss}) from CMU \cite{urmson2008autonomous} won the race. Similar to Stanley, BOSS benefited from a wide range of sensors and was able to demonstrate safe driving in traffic at the speed of up to 48 km/h.

Ever since the DARPA challenges, continuous improvements have been made in various tasks that contribute to achieving full autonomy, such as high-resolution and accurate mapping \cite{dolgov2009autonomous, kummerle2009autonomous}, and complex control algorithms capable of estimating traffic behavior and responding to it \cite{wei2009robust,wei2013towards,brechtel2014probabilistic}.

Autonomous vehicles have also been put to the test on larger scales. In the year 2010, VisLab held an intercontinental challenge by setting the goal of driving the  distance of over 13000 km from Parma in Italy to Shanghai in China \cite{bertozzi2010vislab}. Four autonomous vans each with 5 engineers on board participated in the challenge over the course of three months. One unique feature of this challenge was that the autonomous vehicles, for most of the course, performed platooning in which one vehicle led the way and assessed the road while the others followed it.

Furthermore, autonomous cars have found their way into racing. Shelley from Stanford \cite{shelleydef} is one of the first autonomous vehicles that autonomously drove the 20 km world-famous Pikes Peak International Hill Climb in only 27 minutes while reaching a maximum speed of 193 km/h.

\subsection{Today's Autonomous Vehicles}

\begin{figure}[!th]
\centering
\subfloat[]{
\includegraphics[width=8cm,height=5cm]{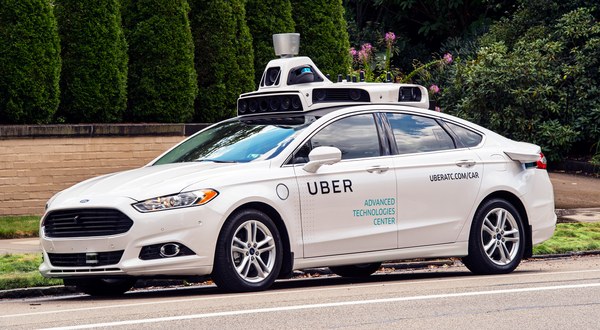}
\label{fig:uber}}
\subfloat[]{
\includegraphics[width=8cm,height=5cm]{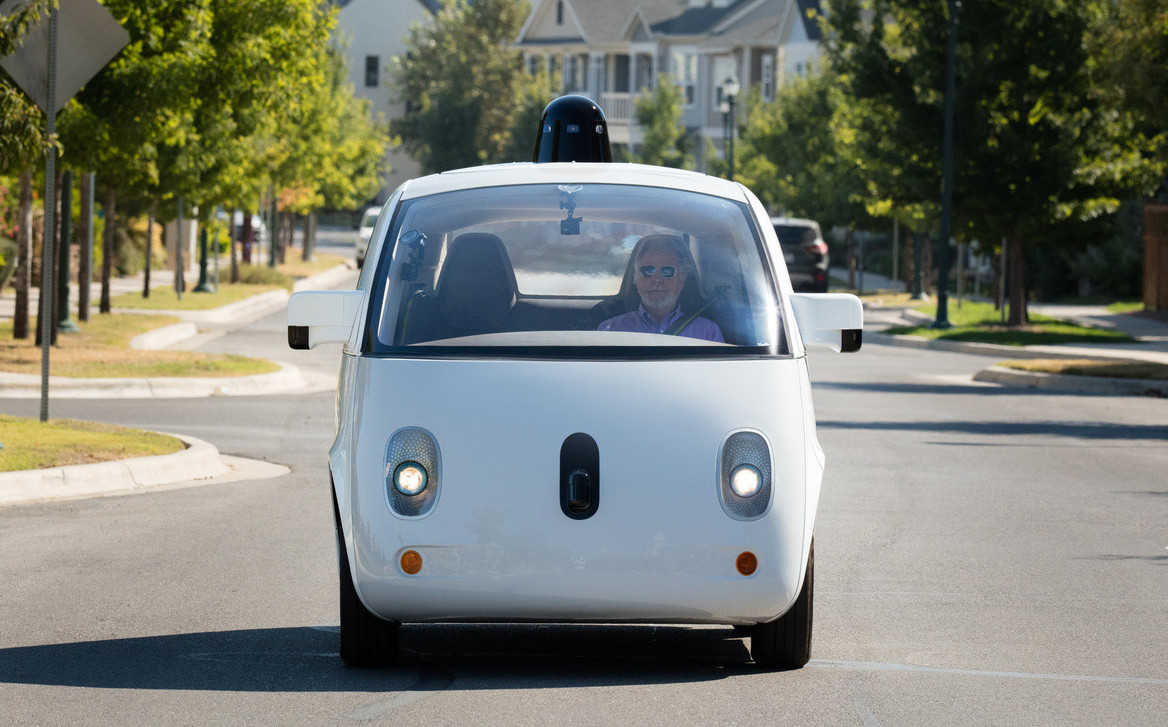}
\label{fig:waymo} } \\
\subfloat[]{
\includegraphics[width=8cm,height=5cm]{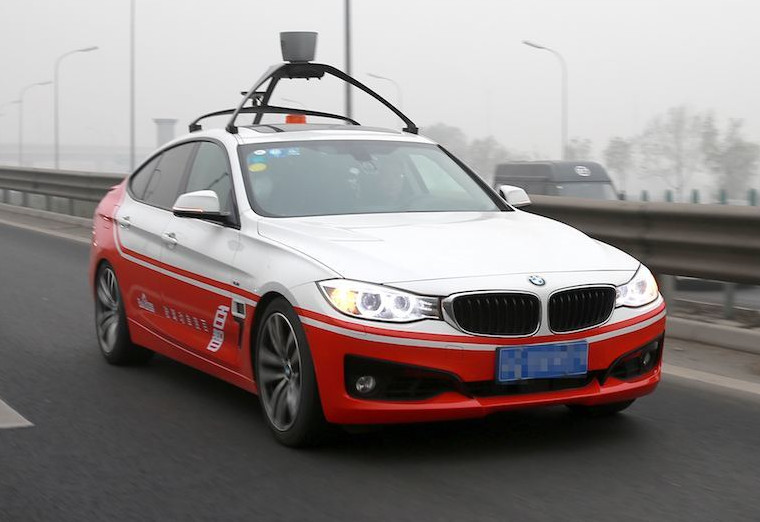}
\label{fig:baidu} }
\subfloat[]{
\includegraphics[width=8cm,height=5cm]{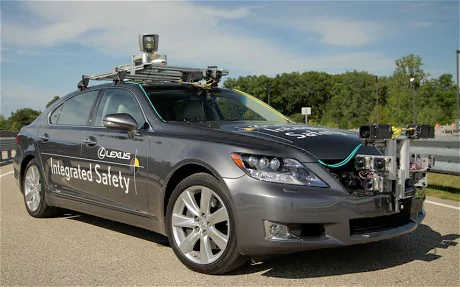}
\label{fig:toyota }}
\caption[Modern autonomous cars.]{Modern autonomous cars: a) Uber \cite{Davieg2016}, b) Waymo (from Google) \cite{waymo}, c) Baidu \cite{baidu}, and d) Toyota \cite{toyota}}.
\label{fig:autocars}
\end{figure}

Today, more than 40 companies are actively working on autonomous vehicles \cite{companiesauto} including Tesla \cite{tesla}, BMW \cite{nica2016}, Mercedes \cite{ziegler2014making}, and Google \cite{waymo}. Although most of the projects run by these companies are at the research stage, some are currently being tested on actual roads (see \figref{fig:autocars}). Few companies such as Tesla already sell their newest models with level 2 autonomy capability and claim that these vehicles have all the hardware needed for fully autonomous driving.

Autonomous driving research is not limited to passenger vehicles. Recently, Uber has successfully tested its autonomous truck system, Otto, to deliver 50,000 cans of beer by driving the distance of over 190 km \cite{truck}. The Reno lab, at the University of Nevada, also announced that they are working on an autonomous bus technology and are planning to put it to test on the road by 2019 \cite{bus}. Autonomous driving technology is even coming to ships. In a recent news release, Rolls-Royce has disclosed its plans on starting a joint industry project in Finland, called Advanced Autonomous Waterborne Applications (AAWA), to develop a fully autonomous ship technology by no later than 2020 \cite{ship}.

\subsection{What's Next?}
How far we are from achieving fully autonomous driving technology, is a subject of controversy. Companies such as Tesla \cite{teslafull} and BMW \cite{nica2016} are optimistic and claim that they will have their first fully autonomous vehicles entering the market by 2020 and 2022 respectively. Other companies such as Toyota are more skeptical and believe that we are nowhere close to achieving level 5 autonomy \cite{toyotaauto}.

So the question is, what are the challenges that we need to overcome in order to achieve autonomy? Besides challenges associated with developing suitable infrastructures \cite{Friedrich2016} and regulating autonomous cars \cite{Gasser2016}, technologies currently used in autonomous vehicles are not robust enough to handle all traffic scenarios such as different weather or lighting conditions (e.g. snowy weather), road types (e.g. driving on roads without clear marking or bridges) or environments (e.g. cities with large buildings) \cite{muoio2016}. Relying on active sensors for navigation significantly constraints these vehicles, especially in crowded areas. For instance, LIDAR, which is commonly used as a range finder, has a high chance of interference if similar sensors are present in the environment \cite{ron2016}.

Some of the consequences associated with technological limitations are evident in recent accidents reports involving autonomous vehicles. Cases that have been reported include minor rear-end collisions \cite{lambert2016,teslapolice2017}, car flipping over \cite{uberaccident2017}, and even fatal accidents \cite{teslafatalt1,teslafatalt2}.

Moreover, autonomous cars are facing another major challenge, namely interaction with other road users in traffic scenes \cite{wolf2016interaction}. The interaction involves understanding the intention of other traffic participants, communicating with them and predicting what they are going to do next.

But why is the interaction between road users so important? The answer to this question is threefold:

\begin{enumerate}
\item \textbf{It ensures the flow of traffic}. We as humans, in addition to official traffic laws, often rely on some forms of informal laws (or social norms) to interact with other road users. Such norms influence the way we perceive others and how we interpret their actions \cite{farber2016communication}. In addition, as part of interaction, we often communicate with each other to disambiguate certain situations. For example, if a car wants to turn at a non-signalized intersection on a heavily trafficked street, it might wait for another driver's signal indicating the right of way. Failure to understand the intention of others, in an autonomous driving context, may result in accidents some of which were reported in the last year involving some of Google's autonomous vehicles \cite{richtel2016,anthony2016}.

\item \textbf{It improves safety}. Interaction can guarantee the safety of road users, in particular, pedestrians as the most vulnerable traffic participants. For instance, at the point of crossing, pedestrians often establish eye contact with drivers or wait for an explicit signal from them. This assures the pedestrians that they have been seen, therefore if they commence crossing, the drivers will slow down or stop before them \cite{gough2016}. How the crossing takes place or the way pedestrians will behave, however, may vary significantly depending on various factors such as demographics and social factors (e.g. presence of other pedestrians) \cite{sun2002modeling} as well as environmental (e.g. weather conditions) and dynamic (e.g. the speed of the vehicles, traffic congestion) factors \cite{parkes1995potential}.

\item \textbf{It prevents from being subject to malicious behaviors}. Given that autonomous cars may potentially commute without any passengers on board, they are subject to being disrupted or bullied \cite{farber2016communication}. For example, people may step in front of the car to force it to stop or change its route. Such instances of bullying have been reported involving some autonomous robots currently being used in malls. Some of these robots were defaced, kicked or pushed over by drunk pedestrians \cite{bully}.
\end{enumerate}

\part{Joint Attention, Interaction and Behavior Understanding}
\section{Joint Attention in Human Interaction}
\label{jt section}

\begin{figure}[!t]
\centering
\includegraphics[width=1\textwidth]{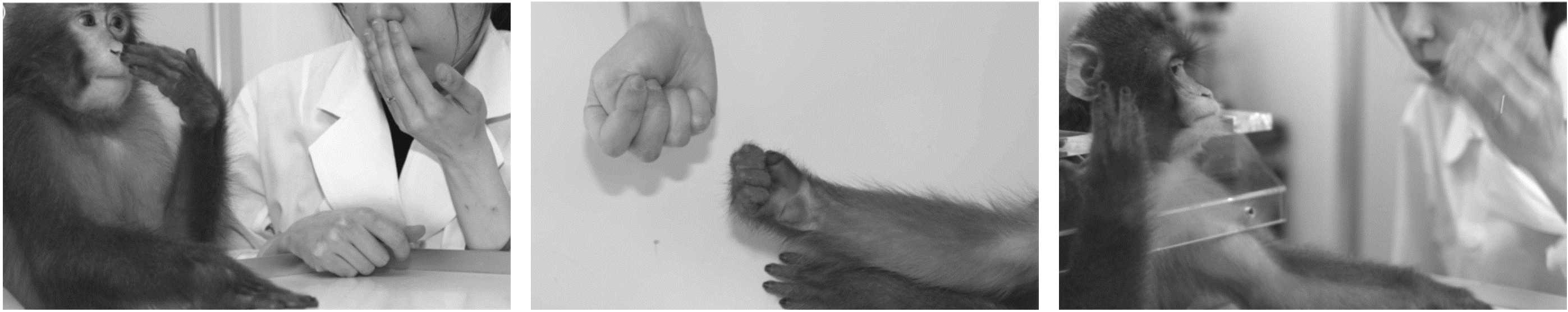}
\caption[The monkey is imitating the human experimenter's gestures]{The monkey is imitating the human experimenter's gestures. Source: \cite{kumashiro2003natural}}
\label{fig:jt_monkey}
\end{figure}

The precursor to any form of social interaction between humans (or primates \cite{kumashiro2003natural}, see \figref{fig:jt_monkey}) is the ability to coordinate attention \cite{kidwell2007joint}, which means the interacting parties at very least should be able to pay attention to one another, discern the relevant objects and events of each other's attentional focus, and implement their own lines of action by taking into account where and toward what others may be attending.

In developmental psychology, the ability to share attention and to coordinate behavior is defined under the \textit{joint attention} framework, or as some scholars term it, \textit{shared attention} \cite{kidwell2007joint} or \textit{visual co-orientation} \cite{butterworth1980towards}. Traditionally, joint attention has been studied as a visual perception mechanism in which two or more interacting parties establish eye contact and/or follow  one another's gaze towards an object or an event of interest \cite{scaife1975capacity,tomasello1983joint}. More recently, joint attention has also been investigated in different sensory modalities such as touch \cite{botero2016tactless} or even remotely via the web \cite{sheldrake2016joint}. Since the objective of this report is  visual perception in autonomous driving, in the following chapters we only focus on the problem of joint visual attention and simply address it as joint attention.

What does joint attention really mean? Joint attention is often defined as a triadic relationship between two interacting parties and a shared object or event \cite{moore1997role,mundy1997joint,dube2004toward,kidwell2007joint}. Simply speaking, joint attention means the simultaneous engagement of two or more individuals with the same external thing \cite{holth2005operant}.

In the traditional definitions, an important part of joint attention is the ability to reorient and follow the gaze of another subject to an object or an event \cite{moore1997role,dube2004toward}. However, in more recent interpretations of joint attention, the gaze following requirement is relaxed and replaced by terms such as "mental focus" \cite{holth2005operant} or "shared intentionality" \cite{tomasello2007shared}. This means joint attention constitutes the ability of a person to engage with another for the sake of a common goal or task, which may not involve explicit gaze following action.

\subsection{Joint Attention in Early Childhood Development}

In 1975, Scaife and Bruner \cite{scaife1975capacity} were the first to discover the joint visual attention mechanism and its role in early developmental processes in infants. They observed that infants below the age of 4 months were able to respond to the gaze changes of the adults in interactive sessions about one-third of the times. In comparison, the older infants, above the age of 11 months, almost always responded to the changes and could follow the gaze of the adults while interacting with them. In addition, at this age, infants can follow the eye movements of the adults as well as their head movements.

Butterworth and Cochran \cite{butterworth1980towards} further investigate the joint attentional behavior and reveal that infants between the age of 6 to 18 months adjust their line of gaze with those of the adult's focus of attention, however, they act only if the adult is referring to loci within the infant's visual space. As a result, if the adult looked behind the infant, the infant only scans the space in front of them. The authors add that at early stages infants do not follow the gaze to the intended object, instead they turn their head to the corresponding side but focus their own gaze on the first object that comes in their field of view. The authors conclude that it is only in the second year when infants are able to focus on the same object that is intended by the adult.

In a subsequent study by Moore \textit{et al.} \cite{moore1997role} it is shown that while sharing attention, the actual movement is critical in gaze following. Through experimental evaluations, the authors illustrate that if only the final focus of the adult is presented to the infant they would not necessarily focus on the same object.

\subsection{Joint Attention in Social Cognition Development}
Joint attention has been linked to the development of social cognitive abilities such as learning of artifacts and environments \cite{charman2000testing,carpenter1998social}. More specifically, joint attention is a fundamental component in language development through which infants learn to describe their surroundings \cite{butterworth1980towards,tomasello1983joint,mundy1998individual}. In a study by Tomasello and Todd \cite{tomasello1983joint}, it is argued that the lexical development of children depends on the way the joint attention activity is administered between the adult and the infant. It is shown that when mothers initiated interaction by directing their child's attention, rather than following it, their child learned fewer object labels and more personal-social words, i.e. they were more expressive (and vice versa).

In short, Tomasello and Carpenter \cite{tomasello2007shared} summarize the social cognitive skills that are acquired through joint attention into four groups:

\begin{enumerate}
\item Gaze following and joint attention
\item Social manipulation and cooperative communication
\item Group activity and collaboration
\item Social learning and instructed learning
\end{enumerate}

The importance of joint attention is not limited to early childhood and is believed to be vital in social competence at all ages. Adolescents and adults who cannot follow, initiate, or share attention in social interactions may be impaired in their capacity for relationship \cite{kidwell2007joint}. There are a large number of studies on the effects of joint attention incapabilities and social disorders, in particular in people with autism \cite{mundy1997joint,macdonald2006behavioral,dube2004toward,deroche2016joint}. For instance, autistic children are found to have minor deficits in responding to joint attention while struggle to initiate joint attention. The effect of aging on joint attention has also been investigated \cite{deroche2016joint}. It is shown that adults tend to get slower in gaze-cuing as they age.

\subsection{Moving from Imitation to Coordination}

The traditional view of joint attention, as discussed earlier, focuses on the role of joint attention as a means whereby infants interact with adults and imitate their behavior to learn about their surroundings.

As adults, however, we often engage in more complex interactions, which can take many forms such as competition, conflict, coercion, accommodation and cooperation \cite{goffman1978presentation}. Cooperation, as in the context of traffic interaction, refers to a social process in which two or more individuals or groups intentionally combine their activities towards a common goal \cite{badis1979}, e.g. crossing an intersection. 
%Here, the goal can be either a path-goal (e.g. enjoying a dance) or final-goal (e.g. win a dance contest).

\begin{figure}[!t]
\centering
\includegraphics[width=0.5\textwidth]{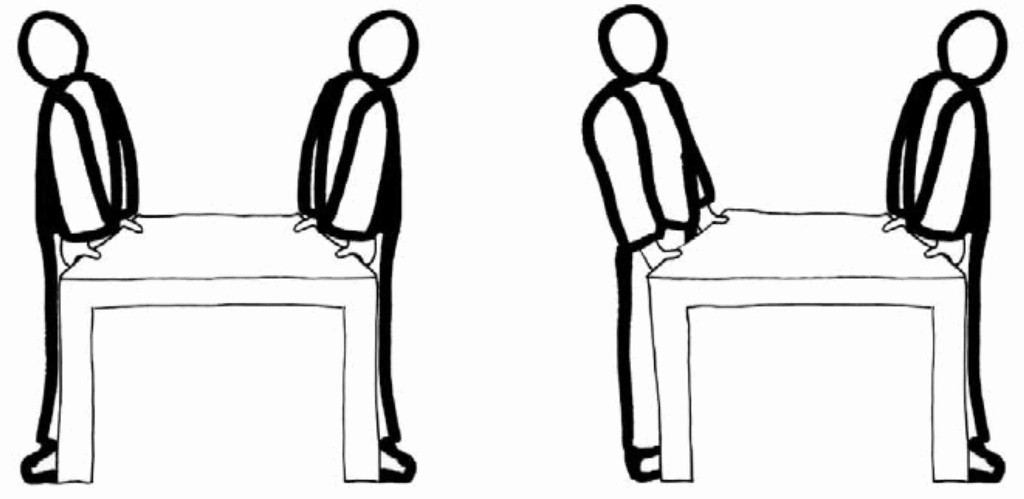}
\caption[Coordination between humans for carrying a table.]{Coordination been humans for carrying a table. Rather than imitating each other's actions, (left image), people must sometimes perform complementary actions to reach a common goal (right image). Source: \cite{sebanz2006joint}}
\label{fig:table}
\end{figure}

What makes a complex coordination possible? Of course, the immediate answer is that a form of joint attention has to take place so the parties involved focus on a common objective. However, joint attention in its classical definition does not fully satisfy the requirements for cooperation. First, although certain cooperative tasks can be resolved by imitation (e.g. make the same movements to balance a table while carrying it), in some scenarios complementary actions are required to accomplish the task (e.g. the person at the front watches for obstacles while the one behind carries the table) \cite{sebanz2006joint}, as shown in \figref{fig:table}. Second, even though involved parties focus on a common object or event, this does not mean that they also share the same intention. In this regard, in the context of cooperation, some scholars use the term \textit{intentional joint attention} \cite{fiebich2013joint} indicating that the agents are not only mutually attending to the same entity, but they also intend to do so.

This type of cooperation that involves a form of attention sharing is often referred to as joint action \cite{fiebich2013joint,sebanz2006joint}. In some literature, joint attention is considered as the simplest form of joint action \cite{fiebich2013joint}. However, for simplicity, throughout the rest of this report, we address the whole phenomenon as joint attention.

\subsection{How Humans Coordinate?}

As described earlier, joint attention provides a mechanism for sharing the same perceptual input and directing the attention to the same event or object \cite{sebanz2006joint}. Here, perhaps, the most crucial components to trigger this attentional shift are eyes, because they naturally attract the observer's attention even if they are irrelevant to the task. Of course, other means of communication such as hand gesture or body posture changes can be used for seeking attention \cite{nuku2008joint}. This is particularly true in the case of traffic interaction where nonverbal communication between road users is the main means of establishing joint attention.

\begin{figure}[!t]
\centering
\includegraphics[width=1\textwidth]{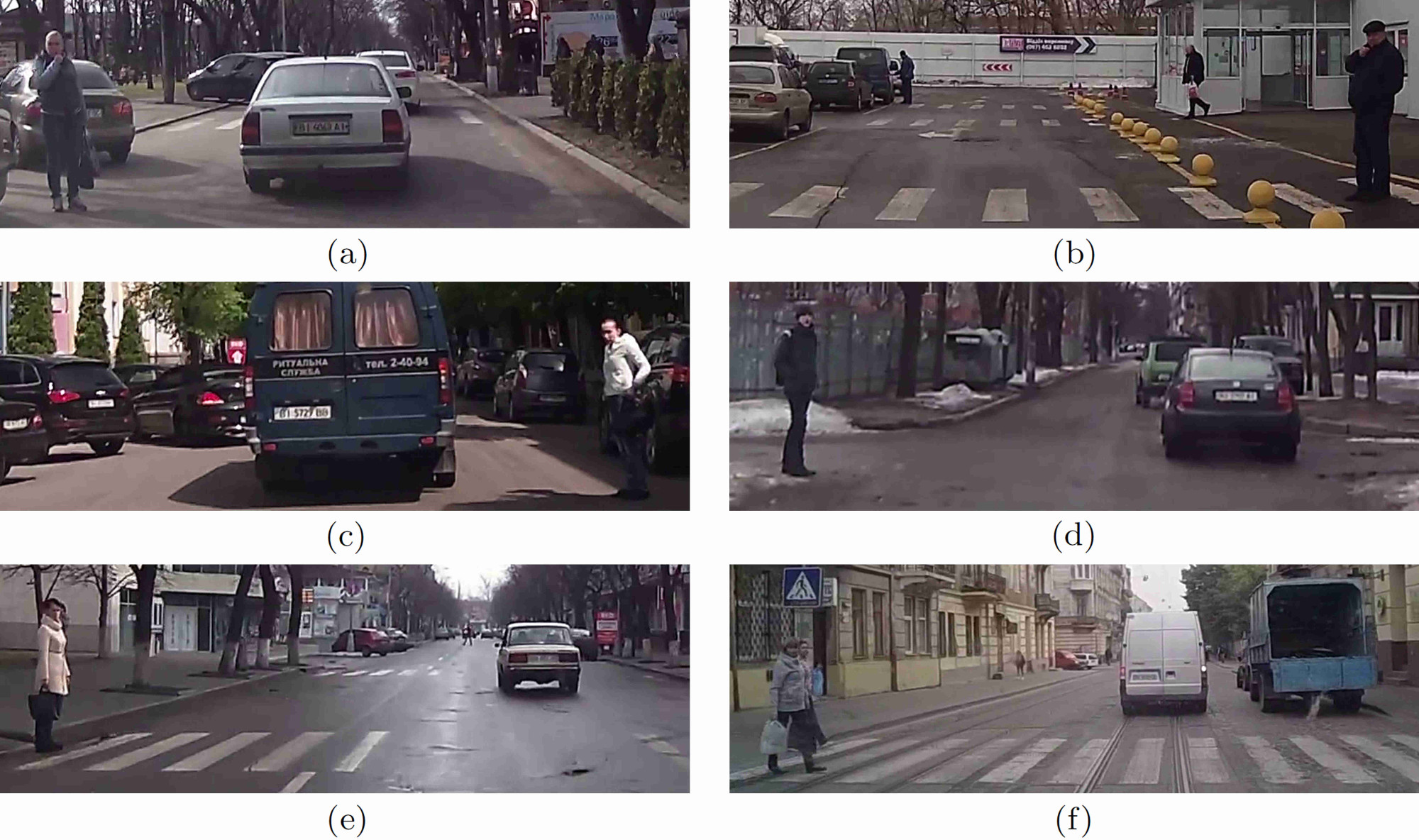}
\caption{From joint attention to crossing. Are these pedestrians going to cross?\protect\footnotemark}
\label{fig:att_crossing}
\end{figure}

Next, the interacting parties have to understand each other's intention in order to cooperate \cite{fiebich2013joint}. In some scenarios establishing joint attention might convey a message indicating the intention of the parties. For instance, at crosswalks, pedestrians often establish eye contact with the drivers indicating their intention of crossing \cite{sucha2017pedestrian}. However, joint attention on its own is not sufficient for understanding the intention of others (see \figref{fig:att_crossing} for an example) or what they are going to do next. A more direct mechanism is action observation \cite{sebanz2006joint}.

%\subsection{Joint Attention in Traffic Interaction}

%Before concluding this Section, it is beneficial to define the problem of driver-pedestrian interaction in traffic scenes in the context of joint attention. Here, the driver and pedestrian,

%\begin{enumerate}
%\item share the same intentionality, that is passing through the intersection.
%\item communicate (share attention) to transmit their intention.
%\item focus on the same task until it is accomplished.
%\end{enumerate}

\section{Observation or Prediction?}
\subsection{A Biological Perspective}

When it comes to understanding observed actions of others, humans do not rely entirely on vision \cite{fogassi2005parietal,umilta2001know}. In a study by Umilta \textit{et al.} \cite{umilta2001know}, the authors found that there is a set of neurons (referred to as mirror neurons) in the ventral premotor cortex (part of the motor cortex involved in the execution of voluntary movements) of macaque monkeys that fire both during the execution of the hand action and observing the same action in others. The authors show that a subset of these neurons become active during action presentation, even though the part of the action that is crucial in triggering the response is hidden and can therefore only be inferred. This implies that the neurons in the observer action system are the basis of action recognition.

Such anticipatory behaviors are also observed in humans. Humans in general have limited visual processing capability (even more so due to the loss of information during saccadic eye movements), especially when it comes to observing moving objects. Therefore, they actively anticipate the future pose of the object to interpret a perceived activity \cite{verfaillie2002representing}. In an experiment by Flanagan and Johansson \cite{flanagan2003action}, the authors showed a video of a person stacking blocks to a number of human subjects and measured their eye movements. They noticed that the gaze of the observers constantly preceded the action of the person stacking blocks and predicted a forthcoming grip in the same way they would perform the same task themselves. The authors then concluded that when observing an action, the human behavior is \textit{predictive} rather than \textit{reactive}.

\footnotetext{a) no, b) no, c) no, d) yes, e) yes, and f) no.}

It is necessary to note that such predictive behaviors in action observation has biological advantages for humans (and also for machines). In addition to dealing with visual processing limitations, anticipatory behaviors can help to deal with visual interruptions due to occlusion in the scene.

\subsection{A Philosophical Perspective}

From a philosophical perspective, it can also be shown that behavior (or action) prediction is the only way to engage in social interaction. For this purpose we refer to the arguments presented by Dennet \cite{dennett1981brainstorms} and the comments on the topic by Baron-Cohen \cite{baron1995mindblindness}.

The ability to find "an explanation of the complex system's behavior and predicting what it will do next", which may include beliefs, thoughts, and intentions \cite{dennett1981brainstorms}, or as Baron-Cohen terms it "mindreading" is crucial in both making sense of one's behavior and communication. Dennet argues that mindreading, or as he calls it adopting "intentional stance", is the only way to engage in social interaction. He further elaborates that the two alternatives to intentional stance, namely physical stance and design stance, are not sufficient for interpretation of one's intentions or actions.

According to Dennet, physical stance refers to our understanding of systems whose physical stance we know about, for instance, we know cutting skin results in bleeding. In terms of understanding complex behavior, however, in order to rely on physical properties, we need to know millions of different physiological (brain) states that give rise to different behaviors. As a result, mindreading is an infinitely simpler and more powerful solution.

Design stance, on the other hand, tries to understand the system in terms of the functions of its observable parts. For example, one does not need to know anything about the microprocessor's internal design to understand the function of pressing the Delete key on the keyboard. Similarly, design stance can explain some aspects of human behavior, such as blinking reflex in response to blowing on eye surface, but it does not suffice to make sense of complex behaviors. This is primarily due to the fact that people have very few external operational parts for which one could work out a functional or design description.

In addition to behavioral understanding, Baron-Cohen \cite{baron1995mindblindness} argues that mindreading is a key element in communication. Apart from decoding communication cues or words, we try to understand the underlying communicative intention. In this sense, we try to find the "relevance" of the communication by asking questions such as what that person "means" or "intends me to understand". For instance, if someone gestures towards a doorway with an outstretched arm and an open palm, we immediately assume that they mean (i.e., intend us to understand) that we should go through the door.

\subsection{The Role of Context in Behavior Prediction}
\label{the_role_of_context}
Now that humans are constantly relying on predicting forthcoming actions of one another in social contexts, the question is, what does enable us to predict behavior? we answer this question in two parts: making sense of one's action and interpreting communication cues. Although these two components are inherently similar, not necessarily in all scenarios actions follow a form of communication, i.e. one might simply observe another person without interaction.

According to Humphrey \cite{humphrey1984consciousness}, when observing someone's action, we first need to perceive the current state of being by relying on our sensory inputs. Next, we need to understand the meaning of the action by relating it to the knowledge of the task (e.g. crossing the street). This knowledge is either biologically encoded in our brain, for instance, people have a very accurate knowledge of human body and how it moves \cite{verfaillie2002representing}, or, in more complex scenarios, it requires knowing the stimulus conditions (context) under which an individual performs an action. The context may include various physical or behavioral attributes present in the scene. Humphrey also emphasizes that in order to understand others, we need to predict the consequences of our actions and realize how they can influence their behavior \cite{humphrey1984consciousness}.

The role of context is also highlighted in communication and how it can influence the way we convey communication  cues. Sperber and Wilson \cite{sperber1987precis}, in their theory of relevance, argue that communication is achieved either by encoding and decoding messages via a code system which pairs internal messages with external signals, or by using the evidence from the context to infer the communicator's intention. Although this theory is originally developed for verbal communication, it has implications which can certainly be relevant to nonverbal communication as well.

The scholars behind the theory of relevance claim that \textit{code model} does not explain the transmission of semantic representations and thoughts that are actually communicated. They believe that there is a need for an alternative model of communication, what they call \textit{inferential model}. In an inferential process, there is a set of premises as input and a set of conclusions as output which follow logically from the premises. When engaging in inferential communication, a communicator intentionally modifies the environment of his audience by providing a stimulus that takes two forms: the informative intention that informs the audience of something, and the communicative intention that informs the audience of the communicator's informative intention. On the other hand, the communicatee makes an inference using his background knowledge that he is sharing with the communicator, i.e. their knowledge of context in which the communication is taking place \cite{sperber1987precis}.

To characterize the shared knowledge involved in communication, the authors use the term \textit{cognitive environment}, which refers to a set of facts that are manifested to an individual. Intuitively speaking, the total cognitive environment of an individual consists of all the facts that he is aware of as well as all the facts that he is capable of becoming aware of at that time and place \cite{sperber1987precis}.

Sperber and Wilson use the term "relevance" to connect context to communication. They argue that any assumption or phenomena (as part of cognitive environment) are relevant in communication if and only if it has some effect in that context. They add that the word "relevance" signifies that the contextual effect has to be large in the given context and at the same time requires small effort to be processed \cite{sperber1987precis}. The amount of processing required to understand the context, however, is a subject of controversy.

In traffic interactions context can be quite broad involving various elements such as dynamic factors, e.g. speed of the cars, the distance of the pedestrians; social factors, e.g. demographics, social norms; and environment configuration, e.g. street structure, traffic signals. Traffic context and its impact on pedestrian and driver behavior will be discussed in more detail in Section \ref{traffic_context}. 

\section{Nonverbal Communication: How the Human Body Speaks to Us}

Nonverbal communication cues such as focusing on gaze direction, pointing gestures and postural movements play an important role in establishing joint attention and interacting with others \cite{nuku2008joint}.

\begin{figure}[!t]
\centering
\includegraphics[width=0.8\textwidth]{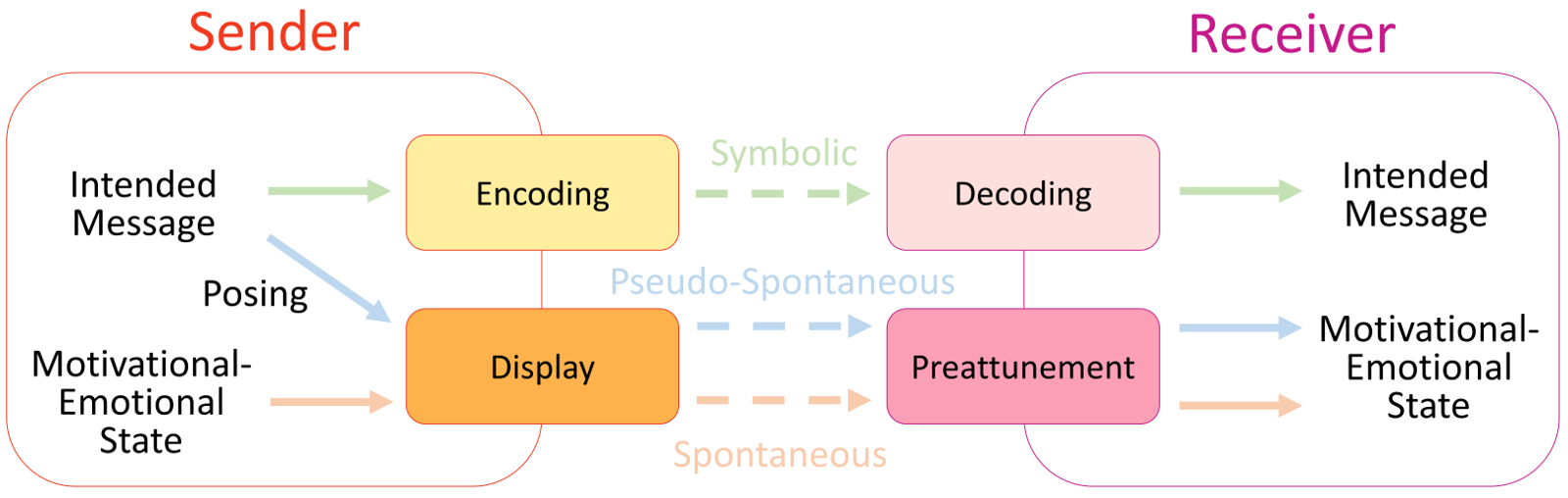}
\caption[Nonverbal communication forms.]{A simplified view of nonverbal communication forms. Source: \cite{buck2002verbal}}
\label{fig:comm_forms}
\end{figure}

In general, nonverbal communication refers to communication styles that do not include the literal verbal content of communication \cite{briton1995beliefs}, i.e. it is affected by means other than words \cite{hecht1999nonverbal}. Buck and Vanlear \cite{buck2002verbal} argue that nonverbal communication comes in three types (see \figref{fig:comm_forms}):

\begin{enumerate}

\item Spontaneous: This form is based on biologically shared signal system and nonvoluntary movements. Spontaneous communication may include facial expressions, micro gestural movements and postures.

\item Symbolic communication: This type of communication is deliberate and has arbitrary relationship with its referent and knowledge of what should be shared by sender and receiver. For instance, symbolic communication may include system of sign language, body movements or facial expressions associated with language.

\item Pseudo-spontaneous: This form involves the intentional and propositional manipulation by the sender of the expressions that are virtually identical to spontaneous displays from the point of view of the receiver. This may include acting or performing.

\end{enumerate}

In traffic context all three types of nonverbal communication are observable. It is intuitive to imagine the occurrence of the first two types of communication. For example, pedestrians may perform various spontaneous movements including yawning, scratching their head, stretching their muscles, etc. As for symbolic communication, humans use various forms of nonverbal signals to transmit their intentions such as waving hands, nodding, or any other form of bodily movements. Symbolic communication in traffic interactions will be discussed in more detail in Section \ref{nonverbal_context}. 

Distinguishing between pseudo-spontaneous and spontaneous nonverbal communication in traffic scenes is not easy, even for humans. It requires the knowledge of the context in which the behavior is observed, and, to some extent, the personality of the person who is communicating the signal. Although rare, the occurrence of pseudo-spontaneous movements is still a possibility in traffic context. People, for example, may perform various bodily movements to distract the driver (or the autonomous vehicle) as a joke or a prank.  

\subsection{Studies of Nonverbal Communication: An Overview}

The modern study of nonverbal communication is dated back to the late 19 century. Darwin, in his book \textit{Expression of the Emotions in Man and Animals} \cite{darwin1998expression}, was the first to focus on the possible modifying effects of body and facial expressions in communication. Darwin argues that nonverbal expressions and bodily movements have specific evolutionary functions, for instance, wrinkling the nose reduces the inhalation of bad odor.

In more recent studies, behavioral ethologists point that in humans, throughout their evolutionary history, these nonverbal bodily movements had gained communicative values \cite{krauss1996nonverbal}. In fact, it is estimated that 55\% of communication between humans is trough facial expressions \cite {mehrabian1976public}. According to Birdwhistell \cite{birdwhistell2010kinesics}, humans are capable of making and recognizing about 250,000 different facial expressions.

Scientists in behavioral psychology measured the importance of bodily movements in various interaction scenarios. For example, Dimatteo \textit{et al.} \cite{dimatteo1980predicting} show that the ability to understand and express emotions through nonverbal communication significantly improves the level of patient satisfaction in a physician visit.

Comprehension or expression of nonverbal communication is linked to various factors. For instance, in a work by Nowicki and Duke \cite{nowicki1994individual}, it is shown that the accuracy of emotional comprehension increases with age and academic achievement. Gender also plays a role in nonverbal communication. In general, women are found to engage in eye contact more often than men \cite{argyle1965eye} and are also better at sending and receiving nonverbal signals \cite{briton1995beliefs}. Another important factor is culture which determines how people engage in nonverbal communication. For example, in Western culture, eye contact is much less of a taboo compared to Middle Eastern culture \cite{argyle1965eye}.

\subsection{Studying Nonverbal Communication}

Measuring behavioral responses is usually administered by showing a sequence of images or videos containing human faces to subjects. Then the subjects are either asked about how comfortable they feel making eye contact with the human in the picture \cite{argyle1965eye} or their emotions are directly observed  \cite{nowicki1994individual}. In another method known as Profile of Nonverbal Sensitivity (PONS), in addition to assessment of emotions, participants are asked to express certain emotions. The expressions are then shown to independent observers who are asked to identify the emotions they represent, for example, whether they imply sadness, anger or happiness. The final score is the combination of both assessment and expression of emotions by the participants \cite{dimatteo1980predicting}. In some studies, fMRI is used to measure brain activities of participants during nonverbal communication \cite{senju2009eye}.

In the context of autonomous driving, however, communication is mainly studied through naturalistic observations \cite{sucha2017pedestrian,rothenbucher2016ghost}. The observation is sometimes combined with other methods to minimize subjectivity. For instance, pedestrians are instructed to perform a certain behavior, e.g. engage in eye contact, and then the behavior of the drivers (who are unaware of the scenario) are observed naturalistically \cite{gueguen2015pedestrian}. The observees sometimes are interviewed to find out how they felt regarding the communication that took place between them and the other road users \cite{risser1985behavior}.

\subsection{Eye Contact: Establishing Connection}

Eye contact, perhaps, is the most important part and the foundation of communication and social interaction. In fact, scientists argue that eye contact creates a phenomenon in the observer called "eye contact effect" which modulates the concurrent and/or immediately following cognitive processing and/or behavioral response \cite{senju2009eye}. Putting it differently, direct eye contact increases physiological arousal in humans, triggering the sense of trying to understand the other party's intention by asking questions such as "why are they looking at me?" \cite{baron1995mindblindness}.

Depending on the context, in the course of social interaction, eye contact may serve different functions, which according to Argyle and Dean \cite{argyle1965eye} can be one of the following:
\begin{enumerate}

\item Information-seeking: It is possible to obtain a great deal of feedback by careful inspection of other's face, especially in the region of the eyes. Various mental states such as noticing one, desire, trust, caring, etc. can be interpreted from the eyes \cite{baron1995mindblindness}.

\item Signaling that the channel is open: Through eye contact a person knows that the other is attending to him, therefore further interaction is possible.

\item Concealment and exhibitionism:
Some people like to be seen, and eye contact is an evidence of them being seen. In contrast some people don't like to be seen, and eye contact is an evidence of they are being depersonalized.

\item Establishment and recognition of social relationship:
Eye contact may establish a social relationship. For example, if person A wants to dominate person B, he stares at B with the appropriate expression. Person B may accept person A's dominance by a submissive expression or deny it by looking away.

\item The affiliative conflict theory: People may engage in eye contact for both approaching or avoiding contact with others.

\end{enumerate}

Since the communication between road users is a form of social interaction, eye contacts in traffic scenes might serve any of the functions mentioned above. However, in the context of traffic interaction, the first two functions are particularly important. In most cases, prior to crossing,  pedestrians assess their surroundings to check the state of approaching vehicles, traffic signals or road conditions. Likewise, drivers continuously observe the road for any potential hazards. It is also common that pedestrians engage in eye contact with drivers to transmit, for example, their intention of crossing. The role of eye contact in pedestrian crossing will be elaborated in Section \ref{nonverbal_context}.   

\subsection{Understanding Motives Through Bodily Movements}

\begin{figure}[!t]
\centering
\includegraphics[width=0.7\textwidth]{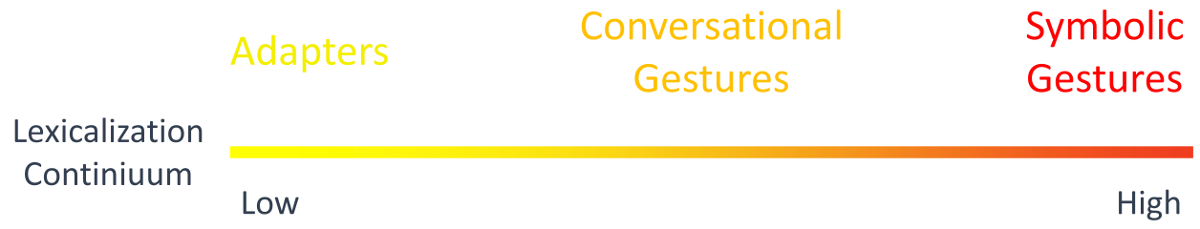}
\caption[The function of hand gesture depending on its lexical meaning.]{The function of hand gesture depending on its lexical meaning.}
\label{fig:comm_lex}
\end{figure}

Besides eye contact, humans often rely on other forms of bodily movements for further message passing. For instance, hand gestures are commonly used during both nonverbal and verbal communication. Although all hand gestures are hand movements, all hand movements are not necessarily hand gestures. This depends on the movement and how the movement is done. Krauss \textit{et al.} \cite{krauss1996nonverbal} group hand gestures into three categories (see \figref{fig:comm_lex}):
\begin{enumerate}

\item Adapters: Aka body-focused movements or self-manipulation. These are the types of gestures that do not convey any particular meaning and have pure manipulative purposes, e.g. scratching, rubbing or tapping.

\item Symbolic gestures: Purposeful motions to transfer a conversational meaning. Such motions are often presented in the absence of speech. Symbolic gestures are highly influenced by cultural background.

\item Conversational gestures: Are hand movements that often accompany speech.

\end{enumerate}

In addition to hand gestures, body posture and positioning may also convey a great deal of information regarding one's intention. Scheflen \cite{scheflen1964significance} lists three functionalities of postural configuration in different aspects of communication:

\begin{enumerate}

\item Distinguishes the contribution of individual behavior in the group activity.

\item Indicates how the contributions are related to one another.

\item Defines steps and order in interaction.

%, i.e. the "program"

\end{enumerate}

\subsection{Nonverbal Communication in Traffic Scenes}
\label{nonverbal_context}
The role of nonverbal communication in resolving traffic ambiguities is emphasized by a number of scholars \cite{wilde1980immediate,clay1995driver,sucha2017pedestrian,rasouliunderstanding}. In this context, any kind of signals between road users constitutes communication. In traffic scenes, communication is particularly precarious because first, there exists no official set of signals and most of them are ambiguous, and second, the type of communication may change depending on the atmosphere of the traffic situation, e.g. city or country \cite{risser1985behavior}.

The lack of communication or miscommunication can greatly contribute to traffic conflicts. It is shown that more than a quarter of traffic conflicts is due to the absence of effective communication between road users. In a recent study it was found that out of conflicts caused by miscommunication, 47\% of the cases occurred with no communication, 11\% was due to the lack of necessary communication and 42\% happened during communication \cite{risser1985behavior}.

Traffic participants use different methods to communicate with each other. For example, pedestrians use eye contact (gazing/staring), a subtle movement in the direction of the road, handwave, smile or head wag. Drivers, on the other hand, flash lights, wave hands or make eye contact \cite{sucha2017pedestrian}. Some researchers also point out that the speed changes of the vehicle can be an indicator of the driver's intention. For example, in a case study by Varhelyi \cite{varhelyi1998drivers} it is shown that drivers use high speed as a signal to communicate to pedestrians that they do not intend to yield. 

Among different forms of nonverbal communication, eye contact is particularly important. Pedestrians often establish eye contact with drivers to make sure they are seen \cite{klienke1977compliance}. Drivers also often rely on eye contact and gazing at the face of other road users to assess their intentions \cite{walker2007drivers}. In addition, a number of studies show that establishing eye contact between road users increases compliance with instructions and rules \cite{hamlet1984eye,klienke1977compliance}. For instance, drivers who make eye contact with pedestrians will more likely yield the right of way at crosswalks \cite{gueguen2015pedestrian}.

\begin{figure}[!t]
\centering
\includegraphics[width=0.8\textwidth]{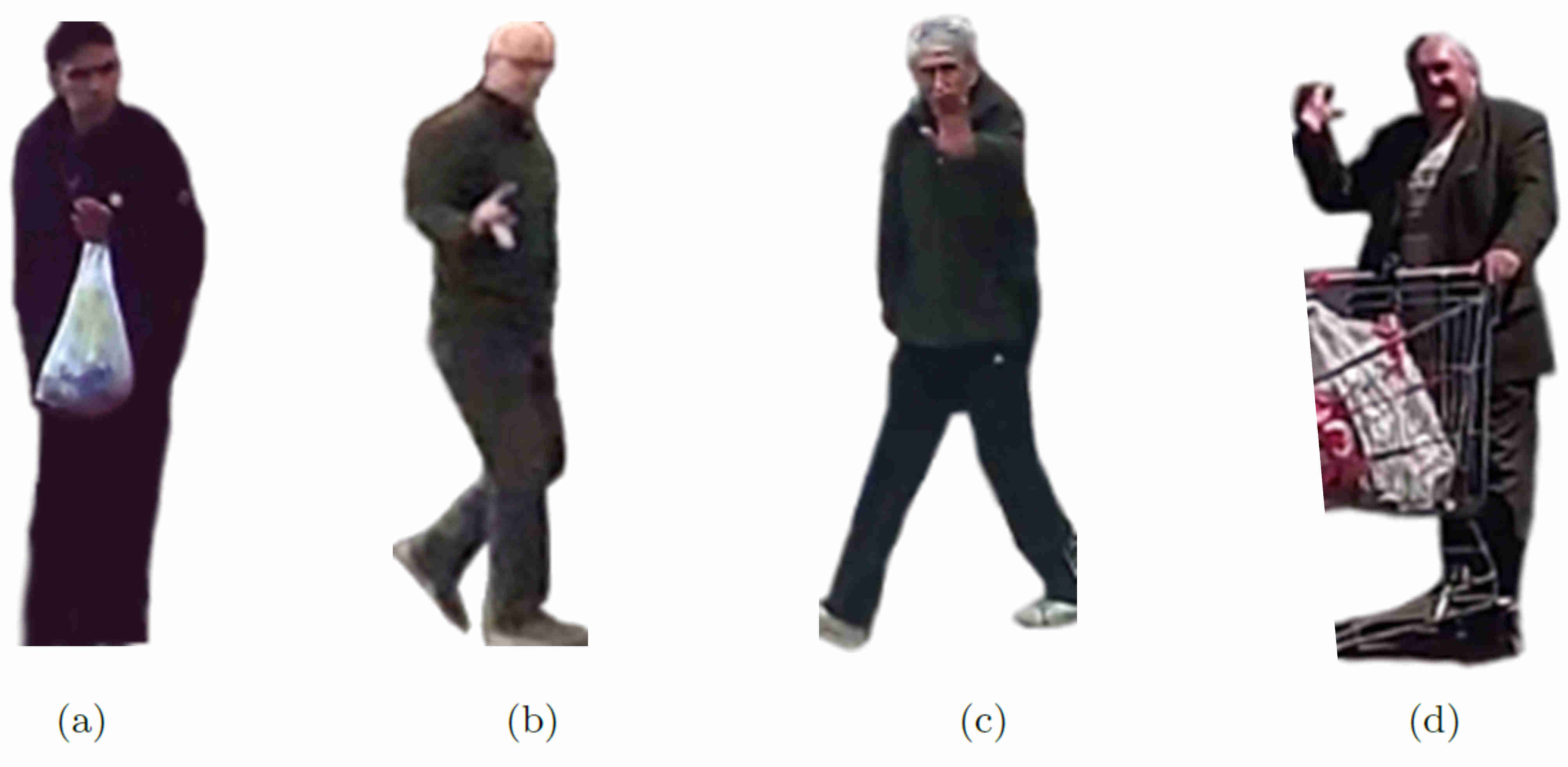}
\caption[Pedestrian hand gestures in traffic scenes]{Examples of pedestrian hand gestures in a traffic scene without context. What are these pedestrians trying to say?\protect\footnotemark}.
\label{fig:hand_context}
\end{figure}

In order to understand the meaning of nonverbal signals, care should be taken while interpreting them. For example, handwave by a pedestrian may be a sign of request for the right of way or showing gratitude. An illustrative example can be seen in \figref{fig:hand_context}.\footnotetext{a) Yielding, b) asking for right of way, c) showing gratitude, and d) greeting a person on the other side of the street.}

\section{Context and Understanding Pedestrian Crossing Behavior}
\label{sec:pedestrian_beh}
It is important to note that context highly depends on the task, which means a particular element that has relevance in one task, may be irrelevant in the other. Hence, in this section we only focus on traffic context and present a review of some of the past studies in the field of traffic behavioral analysis with a particular focus on factors that influence pedestrian crossing behavior.

\subsection{Studying Pedestrian Behavior}

The methods of studying human behavior (in traffic scenes) have transformed throughout the history as new technological advancements have emerged. Traditionally, questionnaires in written forms \cite{wilde1980immediate,price2000relationship} or direct interviews \cite{crundall1999driving} are widely used to collect information from traffic participants or authorities monitoring the traffic. These forms of studies, however, have been criticized for a number of reasons such as the bias people have in answering questions, the honesty of participants in responding or even how well the interviewees are able to recall a particular traffic situation.

Traffic reports have also been widely used in a number of studies. These reports are mainly generated by professionals such as police force after accidents \cite{sullivan2011differences}. The advantage of traffic reports is that they provide good details regarding the elements involved in a traffic accident, albeit not being able to substantiate the underlying reasons.

In addition, behavior can be analyzed via direct observation by someone who is either present in the vehicle \cite{risser1985behavior} or stands outside \cite{tom2011gender} while recording the behavior of the road users. The drawback of this method is the strong observer bias, which can be caused by both the observer's misperception of the traffic scene or his subjective judgments.

\begin{figure}[!t]
\centering
\subfloat[]{
\includegraphics[width=6cm,height=5cm]{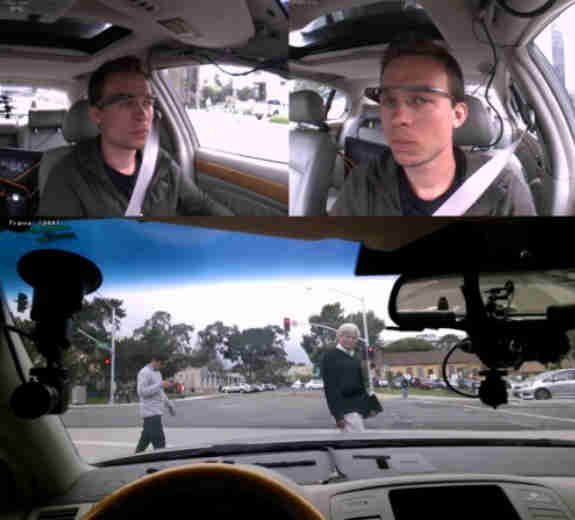}
\label{fig:eyetrack} }
\subfloat[]{
\includegraphics[width=6cm,height=5cm]{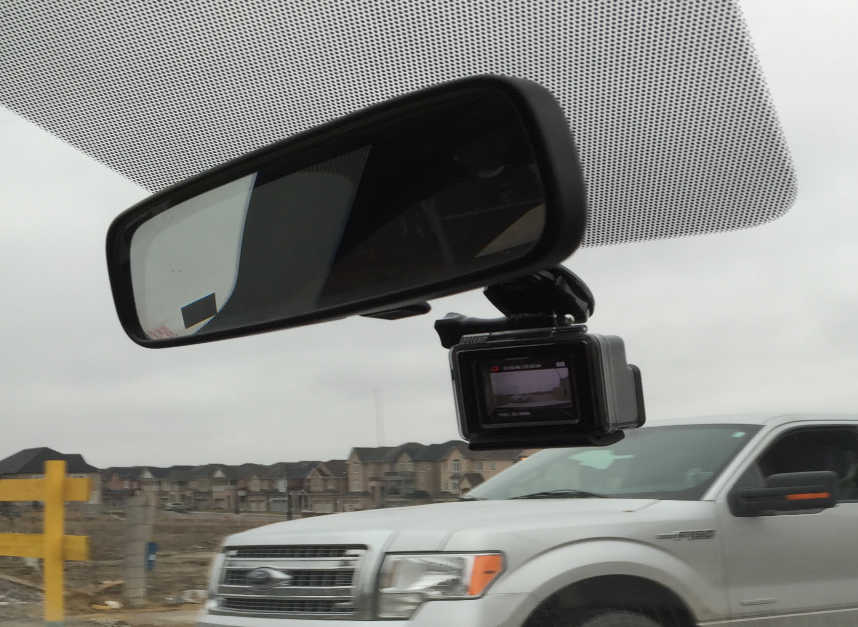}
\label{fig:natural} }
\caption[Technology in behavioral studies.]{Examples of using technology for behavioral studies: a) eyetracker for driver behavior analysis \cite{tawari2014attention}, and b) the placement of a concealed camera for naturalistic traffic recoding.}
\label{fig:beh_tech}
\end{figure}

New technological developments in the design of sensors and cameras gave rise to different modalities of recording traffic events. Eye tracking devices are one of such systems that can be placed on the participants' heads to record their eye movements (see \figref{fig:eyetrack}) during the course of driving \cite{clay1995driver}. Computer simulations \cite{reed2008intersection} and video recordings \cite{price2000relationship} are also widely used to study the behavior of drivers in laboratory environments. These methods, however, have been criticized for not providing the real driving conditions, therefore the observed behaviors may not necessarily reflect the ones exhibited by road users in a real traffic scenario.

Naturalistic studies, perhaps, are one of the most effective methods used in traffic behavior understanding. Although the first instances of such studies are dated back to almost half a century ago \cite{heimstra1969experimental}, they have gained tremendous popularity in the recent years. In this method of study, a camera (or a network of cameras) are placed in either the vehicles  \cite{neale2005overview,eenink2014udrive} (see \figref{fig:natural}) or outdoor on road sides \cite{sun2002modeling,wang2010study}. Since the objective is to record the natural behavior of the road users, the cameras are located in inconspicuous places not visible to the observees. In the context of recording driving habits, although the presence of the camera might be known to the driver, it does not alter the driver's behavior in a long run. In fact, studies show that the presence of camera may only influence the first 10-15 minutes of the driving, hence the beginning of each recording is usually discarded at the time of analysis \cite{risser1985behavior}.

Despite being very effective, naturalistic studies have some disadvantages. For example, researcher bias might affect the analysis. Moreover, in some cases it is hard to recognize certain behaviors, e.g. whether a pedestrian notices the presence of the car or looks at the traffic signal in the scene. To remedy this issue, it is a common practice to use multiple observers to analyze the data and use an average score for the final analysis \cite{heimstra1969experimental}. In some studies, a hybrid approach is employed by combining naturalistic recordings with on-site interviews \cite{sucha2017pedestrian}. Using this method, after recording a behavior, the researcher approaches the corresponding road user and asks whether, for example, they looked at the signal prior to crossing. Overall, the hybrid approach can help lowering the ambiguities observed in certain behaviors.

\subsection{Pedestrian Behavior and Context}
\label{traffic_context}
\begin{figure}[!t]
\centering
\includegraphics[width=1\textwidth]{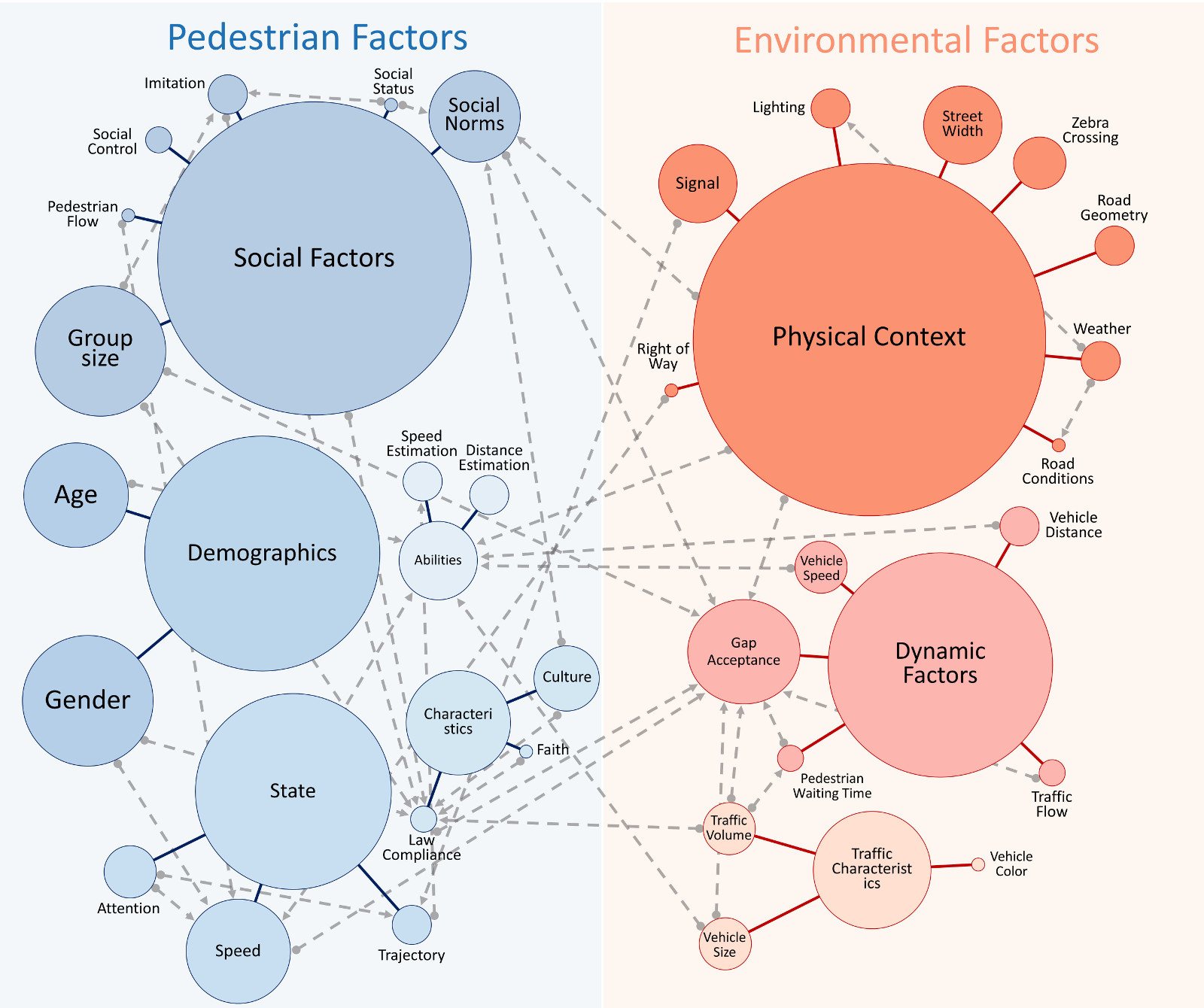}
\caption[Factors involved in pedestrian crossing behavior.]{Factors involved in pedestrian decision-making process at the time of crossing. The size of each circle indicates how many times the factors are found relevant in the literature. The branches with solid lines indicate the sub-factors of each category and the dashed lines show the interconnection between different factors and and arrows show the direction of influence.}
\label{fig:ped_factors}
\end{figure}

The factors that influence pedestrian behavior can be divided into two main groups, the ones that directly relate to the pedestrian (e.g. demographics) and the ones that are environmental (e.g. traffic conditions). A summary of these factors and how they relate to one another can be found in \figref{fig:ped_factors}.

\subsubsection{Pedestrian Factors}

\textbf{\textit{Social Factors}}. Among the social factors, perhaps, \textit{group size} is one of the most influential ones. Heimstra \textit{et al.} \cite{heimstra1969experimental} conducted a naturalistic study to examine the crossing behavior of children and found that they commonly (more than 80\%) tend to cross as a group rather than individually. \textit{Group size}  also changes both the behavior of the drivers with respect to the pedestrians and the way the pedestrians act at crosswalks. For instance, it is shown that drivers more likely yield to a large group of pedestrians (3 or more) than individuals \cite{wilde1980immediate,sun2002modeling}.

Moreover, crossing as a group, pedestrians tend to be more careless at crosswalks and often accept shorter gaps between the vehicles to cross \cite{wang2010study} or do not look for the upcoming traffic \cite{sucha2017pedestrian}. \textit{Group size} also impacts the way pedestrians comply with the traffic laws, i.e. \textit{group size} exerts some form of \textit{social control} over individual pedestrians \cite{rosenbloom2009crossing}. It is observed that individuals in a group are less likely to follow a person who is breaking the law, e.g. crossing on the red light \cite{wilde1980immediate}.

In addition, \textit{group size} influences \textit{pedestrian flow} which determines how fast pedestrians would cross the street. Ishaque and Noland \cite{ishaque2008behavioural} indicate that if there is no interaction between the pedestrians, there is a linear relationship between \textit{pedestrian flow} and \textit{pedestrian speed}. This means, in general, pedestrians walk slower in denser groups.

\textit{Social norms} or, as some experts call it "informal rules" \cite{farber2016communication}, play a significant role in how traffic participants behave and how they predict each other's intention \cite{wilde1980immediate}. The difference between \textit{social norms} and legal norms (or formal rules) can be illustrated using the following example: formal rules define the speed limit of a street, however, if the majority of drivers exceed this limit, the \textit{social norm} is then quite different \cite{wilde1980immediate}.

The influence of \textit{social norms} is so significant that merely relying on formal rules does not guarantee safe interaction between traffic participants. This fact is highlighted in a study by Johnston \cite{johnston1973road} in which he describes the case of a 34-year old married woman who was extremely cautious (and often hesitant) when facing yield and stop signs. In a period of four years, this driver was involved in 4 accidents, none of which she was legally at fault. In three out of four cases the driver was hit from behind, once by a police car. This example clearly depicts how disobeying \textit{social norms}, even if it is legal,  can interrupt the flow of traffic.

\textit{Social norms} even influence the way people interpret the law. For example, the concept of "psychological right of way" or "natural right of way" has been widely studied \cite{wilde1980immediate}. This concept describes the situation in which drivers want to cross a non-signalized intersection. The law requires the drivers to yield to the traffic from the right. However, in practice drivers may do quite the opposite depending on the \textit{social status} (or configuration) of the crossing street. It is found that factors such as \textit{street width}, \textit{lighting} conditions or the presence of shops may determine how the drivers would behave \cite{gheri1963blickverhalten}.

\textit{Imitation} is another social factor that defines the way pedestrians (as well as drivers \cite{vsucha2014road}) would behave. A study by Yagil \cite{yagil2000beliefs} shows that the presence of a law-adhering (or law-violating) pedestrian increases the likelihood of other pedestrians to obey (or disobey) the law. This study shows that the impact of law violation is often more significant.

The probability of \textit{imitation} occurrence may vary depending on the \textit{social status} of the person who is being imitated. In the study by Leftkowitz \textit{et al.} \cite{lefkowitz1955status} a confederate was assigned by the experimenter to cross or stand on the sidewalk. The authors observed that when the confederate was wearing a fancy outfit, there was a higher chance of other pedestrians imitate his actions (either breaking the law or complying).

\textbf{\textit{Demographics}}. Arguably, \textit{gender} is one of the most influential factors that define the way pedestrians behave \cite{heimstra1969experimental,wilde1980immediate,papadimitriou2009critical,twisk2012understanding}. In a study of children behavior at crosswalks, Heimstra \textit{et al.} \cite{heimstra1969experimental} show that girls in general are more cautious than boys and look for traffic more when crossing. Similar pattern of behavior is also observed in adults \cite{yagil2000beliefs} and in some sense it defines the way men and women obey the law. In general, men tend to break the law (e.g. red-light crossing) more frequently than women \cite{ishaque2008behavioural,tom2011gender}.

Furthermore, \textit{gender} differences affect the motives of pedestrians when complying with the law. Yagil \cite{yagil2000beliefs} argues that crossing behavior in men is mainly predicted by normative motives (the sense of obligation to the law) whereas in women it is better predicted by instrumental motives (the perceived danger or risk). He adds that women are  more influenced by social values, e.g. how other people think about them, while men are mainly concerned with physical conditions, e.g. the structure of the street.

Men and women also differ in the way they assess the environment before or during crossing. For instance, Tom and Granie \cite{tom2011gender} show that prior to and during a crossing event, men more frequently look at vehicles whereas women look at traffic lights and other pedestrians. In addition, compared with women, male pedestrians tend to cross with a higher \textit{speed} \cite{ishaque2008behavioural}.

\textit{Age} impacts pedestrian behavior in obvious ways. Generally, elderly pedestrians are physically less capable compared to adults, as a result, they walk slower \cite{ishaque2008behavioural}, have more variation in \textit{walking pattern} (e.g. do not have steady velocity) \cite{goldhammer2014analysis} and are more cautious in terms of accepting gap in traffic to cross \cite{sun2002modeling,waizman2015micro,rasouliagree}. Furthermore, the elderly and children are found to have a lesser ability to assess the speed of vehicles, hence are more vulnerable \cite{clay1995driver}. At the same time, this group of pedestrians has a higher \textit{law compliance} rate than adults \cite{ishaque2008behavioural}.

\textbf{\textit{State}}. The \textit{speed} of pedestrians is thought to influence their visual perception of dynamic objects. Oudejans \textit{et al.} \cite{oudejans1996cross} argue that while walking, pedestrians have better optical flow information and have a better sense of \textit{speed and distance estimation}. As a result, walking pedestrians are less conservative to cross compared to when they are standing.

\begin{figure}[!t]
\centering
\subfloat[Zebra crossing]{
\includegraphics[width=5cm,height=3cm]{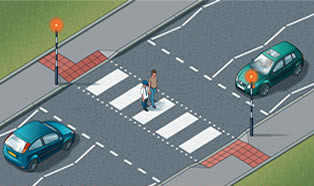}
\label{fig:zebra} }
\subfloat[Crossing with pedestrian refuge island]{
\includegraphics[width=5cm,height=3cm]{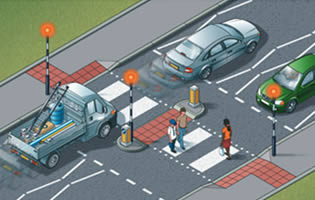}
\label{fig:island} }
\subfloat[Pelican crossing]{
\includegraphics[width=5cm,height=3cm]{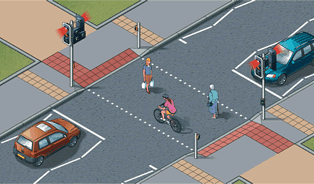}
\label{fig:pelican} }
\caption[Different types of crosswalks.]{Different types of crosswalks. Source: \cite{crossing2010}}
\label{fig:crossings}
\end{figure}

Pedestrian \textit{speed} may vary depending on the situation. For instance, pedestrians tend to walk faster during crossing compared to when walk on side walk \cite{tian2013pilot}. Crossing \textit{speed} also varies in different types of intersections. Crompton \cite{crompton1979pedestrian} reports pedestrian mean speed at different crosswalks as follows: 1.49 m/s at zebra crossings, 1.71 m/s as crossing with pedestrian refuge island and 1.74 m/s at pelican crossings (see \figref{fig:crossings}).

The effect of \textit{attention} on traffic safety has been extensively studied in the context of driving \cite{underwood2003visual,klauer2005driver,underwood2007visual,barnard2016study}. Inattention of drivers is believed to be one of the leading causes of traffic accidents (up to 22\% of the cases) \cite{sabey1975interacting}. In the context of pedestrian crossing, it is shown that inattention significantly increases the chance that the pedestrian is hit by a car  \cite{nasar2008mobile,schwebel2012distraction}.

Hymann \textit{et al.} \cite{hyman2010did} investigate the effect of \textit{attention} on pedestrian walking \textit{trajectory}. They show that the pedestrians who are distracted by the use of electronics, such as mobile phones, are 75\% more likely to display inattentional blindness (not noticing the elements in the scene). The authors also point out that while using electronic devices pedestrians often change their walking direction and, on average, tend to walk slower than undistracted pedestrians.

\textit{Trajectory} or pedestrians walking direction is another factor that plays a role in the way pedestrians make crossing decision. Schmidt and Farber \cite{schmidt2009pedestrians} argue that when pedestrians are walking in the same direction as the vehicles, they tend to make riskier decisions regarding whether to cross. According to the authors, walking direction can alter the ability of pedestrians to estimate speed. In fact, pedestrians have a more accurate speed estimation when the approaching cars are coming from the opposite direction.

\textbf{\textit{Characteristics}}. Without doubt one of the most influential factors altering pedestrian behavior is \textit{culture}. It defines the way people think and behave, and forms a common set of social norms they obey  \cite{lindgren2008requirements}. Variations in traffic \textit{culture} not only exist between different countries, but also within the same country, e.g. between town and countryside or even between different cities \cite{bjorklund2005driver}.

Attempts have been made to highlight the link between culture and the types of behavior that road users exhibit. Lindgren \textit{et al.} \cite{lindgren2008requirements} compare the behaviors of Swedish and Chinese drivers and show that they assign different levels of importance to various traffic problems such as speeding or jaywalking. Schmidt and Farber \cite{schmidt2009pedestrians} point out the differences in \textit{gap acceptance} of Indians (2-8s) versus Germans (2-7s). Clay \cite{clay1995driver} indicates the way people from different culture perceive and analyze a situation. She notes that when judging people during interaction, Americans focus more on pedestrian characteristics whereas Indians rely on contextual factors.

Some researchers go beyond \textit{culture} and study the effect of \textit{faith} or religious beliefs on pedestrian behavior. Rosenbloom \textit{et al.} \cite{rosenbloom2004heaven} gather that ultra-orthodox pedestrians in an ultra-orthodox setting are three times more likely to violate traffic laws than secular pedestrians.

Generally speaking, pedestrian level of \textit{law compliance} defines how likely they would break the law (e.g. crossing at red light). In addition to demographics, \textit{law compliance} can be influenced by physical factors which will be discussed later in this report.

\textbf{\textit{Abilities}}. Pedestrians' abilities, namely \textit{speed estimation} and \textit{distance estimation}, can influence the way they perceive the environment and consequently the way they react to it. In general, pedestrians are better at judging \textit{vehicle distance} than \textit{vehicle speed} \cite{sun2015estimation}. They can correctly estimate \textit{vehicle speed} when the vehicle is moving below the speed of 45 km/h, whereas \textit{vehicle distance} can be correctly estimated when the vehicle is moving up to a speed of 65 km/h.

\subsubsection{Environmental Factors}

\textbf{\textit{Physical context}}. The presence of street delineations, including traffic \textit{signals} or \textit{zebra crossings}, has a major effect on the way traffic participants behave \cite{wilde1980immediate} or on their degree of \textit{law compliance}. Some scholars distinguish between the way traffic \textit{signals} and \textit{zebra crossings} influence yielding behavior. For example, traffic signals (e.g. traffic light) prohibit vehicles to go further and force them to yield to crossing pedestrians. At non-signalized zebra crossings, however, drivers usually yield if there is a pedestrian present at the curb who either clearly communicates their intention of crossing (often by eye contact) or starts crossing (by stepping on the road) \cite{sucha2017pedestrian}.

\textit{Signals} also alter pedestrians level of cautiousness. In a study by Tom and Granie \cite{tom2011gender}, it is shown that pedestrians look at vehicles 69.5\% of the time at signalized and 86\% at unsignalized intersections. In addition, the authors point out that pedestrians' \textit{trajectory} differs at unsignalized crossing. They tend to cross diagonally when no signal is present. Tian \textit{et al.} \cite{tian2013pilot} also adds that when vehicles have \textit{the right of way}, pedestrians tend to cross faster.

\textit{Road geometry} (e.g. presence of pedestrian refuge in the middle of the road) and \textit{street width} impact the level of crossing risk (or affordance), and as a result, pedestrian behavior \cite{oudejans1996cross}. In particular, these elements alter pedestrian \textit{gap acceptance} for crossing. The narrower the street is, the smaller gap is required to cross \cite{schmidt2009pedestrians}.

\textit{Weather} or \textit{lighting} conditions affect pedestrian behavior in many ways \cite{harrell1991factors}. For instance, in bad \textit{weather} conditions (e.g. rainy weather) pedestrians' \textit{speed estimation} is poor, and they tend to be more conservative while crossing \cite{sun2015estimation}. Moreover, lower illumination level (e.g. nighttime) reduces pedestrians' major visual functions (e.g. resolution acuity, contrast sensitivity and depth perception), thus they tend to make riskier decisions. Another direct effect of \textit{weather} would be on \textit{road conditions}, such as slippery roads due to rain, that can impact movements of both drivers and pedestrians \cite{lin2016impact}.

\textbf{\textit{Dynamic factors}}. One of the key dynamic factors is \textit{gap acceptance} or how much, generally in terms of time, gap in traffic pedestrians consider safe to cross. \textit{Gap acceptance} depends on two dynamic factors, \textit{vehicle speed} and \textit{vehicle distance} from the pedestrian. The combination of these two factors defines Time To Collision (or Contact) (TTC), or how far the approaching vehicle is from the point of impact \cite{yangpedestrian}. The average pedestrian \textit{gap acceptance} is between 3-7s, i.e. usually pedestrians do not cross when TTC is below 3s and very likely cross when it is higher than 7s \cite{schmidt2009pedestrians}. As mentioned earlier, \textit{gap acceptance} may highly vary depending on social factors (e.g. \textit{demographics} \cite{wang2010study}, \textit{group size} \cite{wilde1980immediate}, \textit{culture} \cite{schmidt2009pedestrians}), level of \textit{law compliance} \cite{ishaque2008behavioural}, and the \textit{street width}.

The effects of \textit{vehicle speed} and \textit{vehicle distance} are also studied in isolation. In general, it is shown that increase in \textit{vehicle speed} deteriorates pedestrians' ability to estimate speed \cite{clay1995driver} and distance \cite{sun2015estimation}. In addition, Schmidth and Farber \cite{schmidt2009pedestrians} show that pedestrians tend to rely more on distance when crossing, i.e. within the same TTC, they tend to cross more often when the speed is higher.

Some scholars look at the relationship between pedestrian \textit{waiting time} prior to crossing and \textit{gap acceptance}. Sun \textit{et al.} \cite{sun2002modeling} argue that the longer pedestrians wait, the more frustrated they become, and as a result, their   \textit{gap acceptance} lowers. The impact of \textit{waiting time} on crossing behavior, however, is controversial. Wang \textit{et al.} \cite{wang2010study} dispute the role of \textit{waiting time} and mention that in isolation, \textit{waiting time} does not explain the changes in \textit{gap acceptance}. They add that to be considered effective, it should be studied in conjunction with other factors such as personal characteristics.

Although \textit{traffic flow} is a byproduct of \textit{vehicle speed and distance}, on its own it can be a predictor of pedestrian crossing behavior \cite{schmidt2009pedestrians}. By seeing the overall pattern of vehicles movements, pedestrians might form an expectation about what other vehicles approaching the crosswalk might do next.

\textbf{\textit{Traffic characteristics}}. \textit{Traffic volume} or density is shown to affect pedestrian \cite{vsucha2014road} and driver behavior \cite{schmidt2009pedestrians} significantly. Essentially, the higher the density of traffic, the lower the chance of the pedestrian to cross \cite{ishaque2008behavioural}. This is particularly true when it comes to \textit{law compliance}, i.e. pedestrians less likely cross against the signal (e.g. red light) if the traffic volume is high. The effect of \textit{traffic volume}, however, is stronger on male pedestrians than women \cite{yagil2000beliefs}.

The effects of vehicle characteristics such as \textit{vehicle size} and \textit{vehicle color} on pedestrian behavior have also been investigated. Although \textit{vehicle color} is shown not to have a significant effect, \textit{vehicle size} can influence crossing behavior in two ways. First, pedestrians tend to be more cautious when facing a larger vehicle \cite{papadimitriou2009critical}. Second, the size of the vehicle impacts pedestrian \textit{speed and distance estimation} abilities. In an experiment involving 48 men and women, Caird and Hancock \cite{caird1994perception} reveal that as the size of the vehicle increases, there is a higher chance that people will underestimate its arrival time.

\section{Reasoning and Pedestrian Crossing}
To better understand the pedestrian behavior, it is important to know the underlying  reasoning and decision-making processes during interactions. In particular, we need to identify how pedestrians process sensory input, reason about their surroundings and infer what to do next. In the following subsections we start by reviewing the classical views of decision-making and reasoning and talk about various types of reasoning involved in traffic interaction.

\subsection{The Economic Theory of Decision-Making}

The early works in the domain of logic and reasoning define the problem of decision-making in terms of selecting an action based on its utility value, i.e. what positive or negative returns are obtained from performing the action. In the context of economic reasoning, the utility of actions is calculated based on the monetary cost they incur versus the amount of return they promise \cite{edwards1954theory}.

Some scientists consider decision-making process and reasoning to inherently be alike because they both depend on the construction of mental models, and so they should both give rise to similar phenomena \cite{legrenzi1993focussing}. Conversely, a number of scholars believe the similarity is only metaphorical as there are different rules to establish the validity of premises \cite{simon1959theories}.

For the sake of this paper, we do not intend to go any further into differences between decision-making and reasoning. Rather, given that reasoning is thought to be a fundamental component of intelligence \cite{carroll1993human}, we focus the rest of our discussion only on reasoning and try to identify its variations.

\subsection{Classical Views of Reasoning}

There are numerous attempts in the literature to identify the types of reasoning, especially from human cognitive perspective \cite{sternberg1978toward,barlow1974inductive,carroll1993human}, and the way they are programmed in our subconscious (e.g. rule-based, model-based or probabilistic)\cite{legrenzi1993focussing,johnson2015logic,oaksford2001probabilistic}.

\begin{figure}[!t]
\centering
\includegraphics[width=0.7\textwidth]{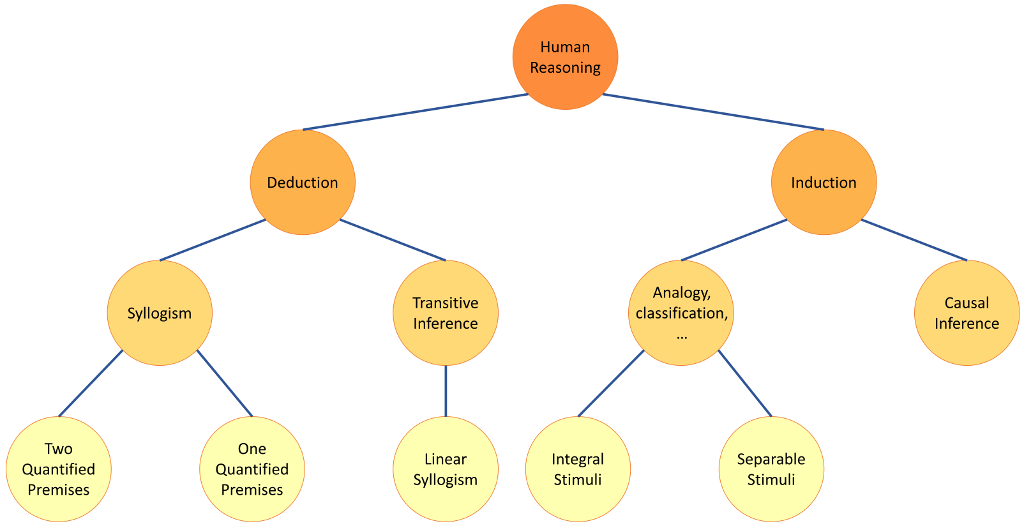}
\caption[Sternberg's view of human reasoning.]{The classical view of human reasoning according to Sternberg \cite{sternberg1978toward}.}
\label{fig:reasoning_dia}
\end{figure}

In the early works of psychology, there are two dominant types of reasoning identified: \textit{Deductive reasoning} in which there is a deductively certain solution and \textit{inductive reasoning} where no logically valid solution exists but rather there is an inductively probable one \cite{sternberg1978toward}. In his definition of reasoning, Sternberg \cite{sternberg1978toward} divides deduction and induction into subcategories including \textit{syllogistic} and \textit{transitive inference} (in deduction), and \textit{causal inference} and either of \textit{analogical, classificational, serial, topological or metaphorical} (in induction) (see \figref{fig:reasoning_dia}).

Syllogism, in particular, is well studied in the literature and refers to a form of deduction that involves two premises and a conclusion \cite{johnson2015logic}. The components can be either categorical (exactly three categorical propositions) or conditional (if ... then) \cite{sternberg1978toward}. Transitive inference is also a form of linear syllogism that contains two premises and a question. Each premise describes a relationship between two items out of which one is overlapping between two premises.

Causal inference, as the name implies, presents a series of causations based on which a conclusion is made about what causes what. As for the other components of inductive reasoning, they depend on the way the reasoning task is presented. For instance, consider the task of selecting analogically which option (a) Louis XIV, (b) Robespierre completes the sequence, Truman : Eisenhower :: Louis XIII : ?. Performing the same task in the classification format would look like this: Out of groups (a) and (b) in (a) Eisenhower, Truman (b) Louis XIII, Louis XIV, which group Robespierre belongs to? For more examples of different types of reasoning please refer to \cite{sternberg1978toward}.

In other categorizations of reasoning, in addition to deduction and induction, researchers consider a third group, \textit{abductive reasoning}. Generally speaking, abduction refers to a process invoked to explain a puzzling observation. For instance, when a doctor observes a symptom in a patient, he hypothesizes about its possible causes based on his knowledge of the causal relations between diseases and symptoms \cite{aliseda2006abductive}. According to Fischer \cite{fischer2001abductive}, there are three stages in abduction: 1) a phenomenon to be explained or understood is presented, 2) an available or newly constructed hypothesis is introduced, 3) by means of which the case is abducted.

Whether abduction should be considered as a separate class of reasoning or a variation of induction, or induction be a subsidiary of abduction, is a subject of controversy among scientists \cite{aliseda2006abductive}. Quoting from Peirce \cite{aliseda2006abductive}, who distinguishes between three different types of reasoning: "Deduction proves that something must be; induction shows that something actually is operative; abduction merely suggests that something may be". More precisely, if one wants to separate abduction from induction the following characteristics should be considered: abduction reasons from a single observation whereas induction from enumerative samples to general statements. Consequently, induction explains a set of observations while abduction explains only a single one. In addition, induction predicts further observations but abduction does not directly account for later observations. Finally, induction needs no background theory necessarily, however, abduction heavily relies on a background theory to construct and test its abductive explanations.

In the psychology literature, some researchers, in addition to deduction and induction, identify the third group of reasoning as quantitative reasoning \cite{carroll1993human}. This type of reasoning involves the identification of quantitative relationships between phenomena that can be described mathematically.

Reasoning abilities, in particular, deduction and induction, are widely used in various daily tasks such as crossing. For example, deductive abilities help to make sense of various pictorial or verbal premises and may help to solve various cognitive tasks such as general inferences or syllogisms that help one to reason about selecting a proper route (this task is commonly referred to as ship destination in the literature). Inductive reasoning may also be used in judging people or situations based on the past experiences, e.g. understanding social norms, nonverbal cues, etc. 

\subsection{Variations in Reasoning}
Listing all types of reasoning (subcategories of main reasoning groups) is a daunting task because the terminology, and reasoning tasks and definitions may vary significantly in different communities (e.g. AI and psychology). Some sources attempt to enumerate types of reasoning by distinguishing between the tasks they are applied to. For instance, in \cite{typesreason} over 20 categories of reasoning are enumerated such as backward reasoning (start from what we want and work back), butterfly logic (one induces a way of thinking in others), exemplar reasoning (using examples), criteria reasoning (comparing to established criteria), and more.

In this subsection, however, we list and discuss the types of reasoning based on two criteria: there exist a significant support for that type of reasoning and relevance to the task of traffic interaction.

\textbf{Probabilistic reasoning}

Probabilistic reasoning aka approximate reasoning is a branch of reasoning that challenges the traditional logic reasoning \cite{oaksford2001probabilistic,johnson2015logic,zadeh1975fuzzy}. It refers to a process of finding an approximate solution to the relational assignment of equations. What distinguishes probabilistic reasoning from the traditional reasoning is its fuzziness and non-uniqueness of consequents. A common example to illustrate probabilistic reasoning is as follows: most men are vain; Socrates is a man; therefore, it is very likely that Socrates is vain \cite{zadeh1975fuzzy}.

To explain the advantage of probabilistic reasoning over traditional logic reasoning, probabilistic logicians argue that people are irrational beings and systematically make large errors (as evident in various logical tasks such as Wason selection task \cite{cosmides1989logic}). However, at the same time, they seem very successful in everyday rationality in guiding thoughts and actions \cite{oaksford2001probabilistic}.

In addition, probabilistic reasoning argues against both mental-logic and mental-model approaches\footnote{\textbf{Mental-logic} argues that deduction depends on an unconscious system of formal rules of inference akin to those in proof-theoretic logic, i.e. we use a series of rules to deduce a conclusion.\\
\textbf{Mental-model} postulates that people understand discourse and construct mental models of the possibilities consistent with the discourse. Then a conclusion is reached by selecting a possibility over the others.} by stating that working memory is limited in terms of storing a length of a formal derivation or the number of mental models. In this sense, errors in so-called logical tasks can be explained by the fact that people use probabilistic reasoning in everyday tasks and generalize them to apply to logic.

In the traffic context, the decision process of road users is highly probabilistic, for instance, they reason in approximate terms to decide whether to cross (e.g. by estimating the behavior of drivers) or which route to take.

\textbf{Spatial reasoning}

Spatial reasoning refers to the ability to plan a route, to localize entities and visualize objects from descriptions of their arrangement. An example task can be as follows: B is on the right of A, C is on the left of B, then A is on the right of C \cite{byrne1989spatial}. Applications that benefit from spatial reasoning include (but not limited to) various navigational tasks, path finding, geographical localization, unknown object registration, etc.\cite{frank1992qualitative}.

It is intuitive to link spatial reasoning to various traffic tasks. We often use such reasoning abilities to plan our trip, to choose the direction of crossing or to estimate risk by measuring our position with respect to the other road users.

\textbf{Temporal reasoning}

As the name implies, temporal reasoning deals with time and change in state of premises. Formally speaking, temporal reasoning consists of formalizing the notion of time and providing means to represent and reason about the temporal aspect of knowledge \cite{vila1994survey}. Temporal reasoning accounts for change and action \cite{pani2001temporal}, i.e. it allows us to describe change and the characteristics of its occurrence (e.g. during interaction with others). Temporal reasoning typically involves three elements, \textit{explanation}, to produce a description of the world at some past time, \textit{prediction}, to determine the state of the world at a given future time and \textit{learning about physics}, to be able to describe the world at different times and to produce a set of rules governing change which account for the regularities in the world \cite{vila1994survey}.

Temporal reasoning can be used in various tasks such as medical diagnosis, planning, industrial processes, etc. \cite{vila1994survey}. In the context of traffic interaction, as discussed earlier, people deal with the prediction of each others' intentions, which, in addition to static context (e.g. physical characteristics), highly depends on the past observed actions of others and changes in their behavior (e.g. nonverbal communication). Therefore, a form of temporal reasoning is necessary to make a decision in such situations.

\textbf{Qualitative (physical) reasoning}

There is a vast amount of literature focusing on qualitative (or as some call it physical) reasoning \cite{klenk2005solving,baillargeon1995physical,cohn1997qualitative,davis2008,frank1992qualitative}. The main objective of qualitative reasoning is to make sense of the physical world \cite{klenk2005solving}, i.e. reason about various types of physical phenomena such as support or collision phenomena \cite{baillargeon1995physical}.

In physical reasoning, factors such as space, time or spacetime are divided into physically significant intervals (or histories or regions) bounded by significant events (or boundaries) \cite{davis2008}.
All values and relations between numerical quantities and functions are discretized and represented symbolically based on their relevance to the behavior that is being modeled \cite{cohn1997qualitative}. Although such qualitative discretization may result in the loss of precision, it significantly simplifies the reasoning process and allows deduction when precise information is not available \cite{frank1992qualitative}.

Qualitative reasoning can be applied to a wide range of applications such as Geographical Information Systems (GIS), robotic navigation, high-level vision, the semantics of spatial prepositions in natural languages, commonsense reasoning about physical situations, and specifying visual language syntax and semantics \cite{cohn1997qualitative}. It is also easy to see the role of qualitative reasoning role in traffic scenarios. For example, when observing a vehicle, the exact distance or speed of the vehicle is not known to the pedestrian. Rather, the pedestrian reasons about the state of the vehicle using terms such as the vehicle is very near, near, far or very far, or in the case of speed, the vehicle is moving very slow, slow, fast or very fast.

\textbf{Deep neural reasoning}

Recent developments in machine-learning techniques gave rise to deep neural reasoning \cite{jaeger2016artificial,peng2015towards,socher2013reasoning,graves2016hybrid}. In this method, the reasoning task is performed via a neural network which is trained using many solved examples \cite{jaeger2016artificial}. As for the working memory, which is a necessity in extensive reasoning tasks such as language comprehension, question answering or activity understanding, deep neural reasoning models either implicitly keep a representation of the previous events, e.g. long short term memory (LSTM) networks \cite{sutskever2014sequence}, or have a dedicated memory that interacts with the neural net (i.e. network only acts as a "controller") \cite{graves2014neural,graves2016hybrid}. These reasoning techniques show promising performance in tasks such as question answering, finding the shortest path between two points or moving blocks puzzle \cite{graves2016hybrid}.

\textbf{Social reasoning}

In psychology literature, the term social reasoning \cite{sichman1998social,de1965social} is widely used to refer to the abilities that involve reasoning about others. More specifically, social reasoning refers to a mechanism that uses information about others in order to infer some conclusions. It is thought to be essential in any intelligent agent in order to react properly for various interactive situations \cite{sichman1998social}.

To highlight the importance of social reasoning (also known as social intelligence or social cognition) in the human interactions, we will refer to Humphrey \cite{humphrey1984consciousness} who believes that social intelligence is the foundation of holding society together. In Humphrey's opinion, the essential part of social intelligence is abstract reasoning, which never was needed before in performing other intelligent tasks. He adds that in order to interact social primates should be calculating beings, which means they must be able to calculate the consequences of their own behavior, the likely behavior of others, and the balance of advantage and loss.

An important implication of Humphrey's discourse is the necessity of prediction and forward planning in social interaction (which also was discussed earlier in Section \ref{jt section}). Humphrey refers to social interaction as "social chess" and argues that like chess, in addition to the cognitive skills, which are required merely to perceive the current state of play, the interacting parties must be capable of a special sort of forward planning.

\textbf{Sensory reasoning}

Reasoning may also be involved in the way we use or control our sensory input. In humans, two types of visual reasoning are commonly used to assess surroundings \cite{carroll1993human}. These are dynamic spatial reasoning which helps to infer where a moving object is going and when it will arrive at its predicted destination, and ecological spatial reasoning, which refers to an individual's ability to orient himself in the real world. These reasoning abilities can help a pedestrian to estimate the movement of the traffic and calculate a safe gap for crossing.

\section{Attention and its Role in Human Visual System}
\label{attenion_role}

The prerequisite to understanding pedestrian behavior is to identify relevant contextual elements and visually attend to them. If we consider visual perception as the process of constructing an internal representation of the local environment \cite{wright1998visual}, visual attention can be seen as perceptual operations responsible for selecting important or task-relevant objects for further detailed visual processing such as identification \cite{tsotsos1995toward}. Generally speaking, perception helps analyzing lower level physical properties of visual stimuli (e.g. color, size, motion), while attention facilitates such processing by directing perception towards a particular object \cite{wright1998visual}. Attention is thought to be the foundation of visual perception without which we simply do not have the capacity to make sense of the environment.

Whether visual attention is stimulus-based (activated by distinctive characteristics of objects) or task-oriented has been a subject of controversy for many years. Despite such disagreement on the source of attention, the majority of recent studies strongly support the dominant role of the task in triggering attention \cite{bruce2015computational}. In fact, experimental evidence shows that the nature of the task, i.e. the objective of the observer, influences the way attention is taking place. This may include the priming of visual receptors at early stages of perception to the properties of the object or higher level processing of the features and eye movements to fixate on where the object may be located \cite{wright1998visual}.

\subsection{Attention and Driving}

In behavioral psychology, a large body of research has been dedicated to studying the role of attention in driving safety. Some of these studies were introduced earlier, for example, studies on how frequently pedestrians look prior to crossing or how they engage in eye contact \cite{lee2009interaction}. In addition, a number of scholars have studied the attention patterns of drivers (e.g. eye movements) with the purpose of designing assistive systems for vehicles.

On the importance of visual perception in driving, studies show that over 90\% of the information during driving is received through the visual system \cite{parkes1995potential}. In this context, a vital component of visual perception is attention, lack of which accounts for 78\% of all crashes and 65\% of all near-crashes. Of interest to autonomous driving systems, non-specific eye glances to random objects in the scene caused 7\% of all crashes and 13\% of all near-crashes \cite{klauer2006impact}. 

Moreover, some findings suggest that the way drivers allocate their attention greatly contributes  to the safety of driving \cite{underwood2003visual,underwood2007visual}. For instance, in \cite{underwood2003visual} the authors measured and compared the driving habits of novice and experts drivers and the way they pay attention to the environment by recording their eye movements while driving. They revealed that novice drivers frequently  had more fixation transitions from objects in the environment to the road ahead. Expert drivers, on the other hand, performed fewer fixations suggesting that they typically relied on their peripheral vision to monitor the road. Expert drivers had  also a better road monitoring strategy by performing horizontal scanning on mid range road (1 to 2 seconds away from the vehicle) whereas novice drivers were mainly focusing on the road far ahead. In addition, expert drivers are thought to be more aware of their surroundings by keeping track of various objects around the vehicle. 

More recent works focus on the effects of technology on driver's attention.  Llaneras \etal  \cite{llaneras2013human} show the effect of Limited Ability Autonomous Driving Systems (LAADS)
such as adaptive cruise control and lane keeping assist on drivers' attention. They found that when the drivers had the opportunity to relinquish vehicle speed maintenance and lane control functions to a reliable autonomous system, they were frequently looking away from the forward roadway (about 33\% of the time) while driving. Takeda \etal   \cite{takeda2016electrophysiological} also evaluate driver attention in level 3 self-driving vehicles compared to when they are in control of driving tasks. They show that while not driving, drivers tend to have longer blinks and fewer saccadic eye movements to monitor the environment. The authors add that such changes in behavioral attention can potentially be hazardous as level 3 autonomous vehicles may require the intervention of the driver at certain critical moments.

\part{Practical Approaches to Pedestrian Behavior Understanding}

\section{Joint Attention in Practical Systems}
The majority of practical joint attention systems are done in social and assistive robotics, in particular, for tasks involving close encounters with humans. Although most of these works do not have a direct application to traffic interaction, some of their subcomponents and ideas can potentially be used in autonomous driving. Hence, in the following subsections we briefly introduce some of the main practical systems that take advantage of joint attention mechanisms.
 
\subsection{Mindreading through Joint Attention}
The joint attention model introduced by Baron-Cohen \cite{baron1995mindblindness} is the backbone of many practical systems capable of establishing joint attention such as Cog and Kismet \cite{scassellati1996mechanisms,scassellati1999imitation}.

Baron-Cohen argues that a shared attention model, or as he terms it mindreading, requires at least four modules, which can be organized into three tiers depending on the complexity of the task they are handling. The overall structure of his system is illustrated in \figref{fig:jt_diagram} and is briefly described below:

\begin{figure}[!t]
\centering
\includegraphics[width=0.7\textwidth]{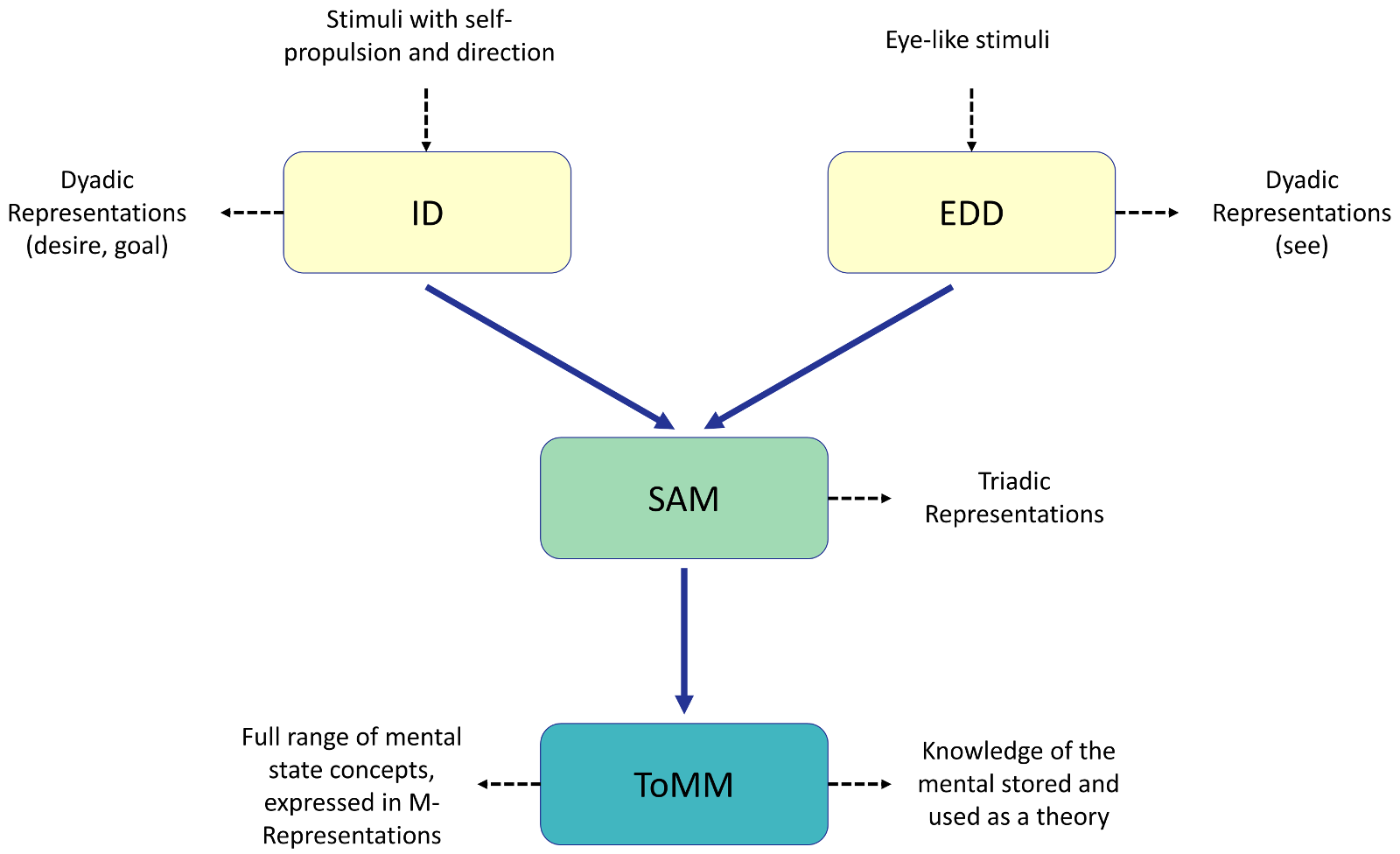}
\caption[Baron-Cohen's joint attention (mindreading) system.]{Baron-Cohen's joint attention (mindreading) system. The term M-Representations refers to propositional attitudes expressed as e.g. believes, thinks. Source: \cite{baron1995mindblindness}}
\label{fig:jt_diagram}
\end{figure}

\textbf{The Intentional Detector (ID)}. According to Baron-Cohen, ID is "a perceptual device that interprets
motion stimuli in terms of the primitive volitional mental states of goal and desire". This module is activated whenever any perceptual input (vision, touch, or audition) identifies something as an agent. This may include any agent-like entity (defined as anything with self-propelled motion) such as a person, a butterfly or even a billiard ball, which is initially considered as a query agent with goals and desires. It should be noted that after discovering that an object is not an agent (i.e. its motion is not self-caused), the initial reading is revised.

\textbf{The Eye-Direction Detector (EDD)}. Baron-Cohen lists three functionalities for EDD including detecting the presence of eyes or eye-like stimuli, computing whether eyes are directed towards it or towards something else, and inferring that if another organism's eyes are directed at something then that organism sees that thing. The main difference of EDD and ID is that ID interprets stimuli in terms of the volitional mental states of desire and goal, whereas EDD does so in terms of what an agent sees. Therefore, EDD is a specialized part of the human visual system.

\textbf{The Shared-Attention Mechanism (SAM)}. This is the module responsible for building triadic representation (between the agent, self and an object). This representation can be generated by using EDD, through which the agent can follow the gaze direction of the agent and identify the object of interest. However, SAM is limited in terms of its power to infer more complex relationship when the object is not within reach of both the agent and self, for instance, when communicating with someone who is not present in the scene.

\textbf{The Theory-of-Mind Mechanism (ToMM)}. This system is needed for inferring the full range of mental states from behavior, thus employing a "theory of mind". ToMM is much more than simply confirming the intention of others or observing their gaze changes. It deals with two major tasks: representing epistemic mental states (e.g. pretending, thinking and deceiving) and combining them with other mental states (the volitional and the perceptual) to help to understand how mental states and actions are related.  

\subsection{Practical Systems}
The majority of the socially interactive systems that employ a form of joint attention use this mechanism for the purpose of learning, such as gaze control, similar to infants. In addition, most of these applications are designed for close interactions which usually involve a physical object. This means these models have little (if any) implication for joint attention and intention estimation in the context of traffic interaction. As a result, we only briefly discuss some of the most known systems developed during the past couple of decades. A more comprehensive list of socially interactive systems that employ the joint attention mechanism can be found in \cite{breazeal2008social,fong2003survey}.

\begin{figure}[!t]
\centering
\subfloat[Cog]{
\includegraphics[width=5cm,height=5cm]{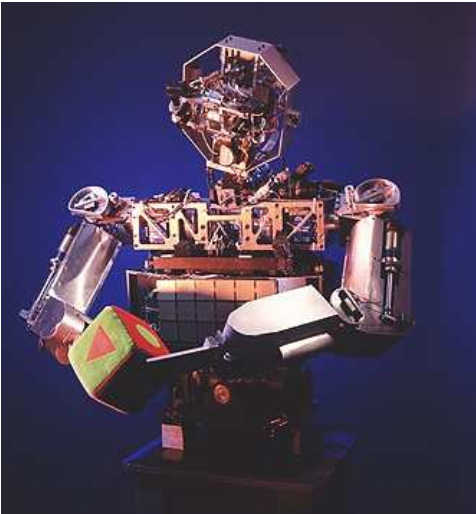}
\label{fig:cog}}
\hspace{0.5cm}
\subfloat[Kismet]{\includegraphics[width=5cm,height=5cm]{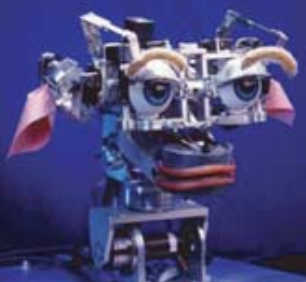}
\label{fig:kismet}}\\
\caption[Examples of robotic platforms with joint attention capability for learning.]{Examples of robotic platforms with joint attention capability for learning. Source: a) \cite{scassellati1999knowing} and b) \cite{breazeal2005robot}.}
\label{fig:learning bots}
\end{figure}

\subsubsection{Socially Interactive Systems}

Cog (\figref{fig:cog}) \cite{scassellati1996mechanisms,scassellati1999knowing,scassellati1999imitation} is one of the early social robots that has the mechanism of joint attention. The robot has 21 degrees of freedom (DoF) and a variety of sensory modalities such as vision and audition. Its physical structure consists of a movable torso, arms, neck, and eyes which give it a human-like motion. The robot is designed to perform gaze monitoring and to engage in shared attention as well as to perform certain actions (such as pointing) to request shared attention to an object.

A descendant of Cog is Kismet (\figref{fig:kismet}) \cite{scassellati1999knowing,breazeal1999robot,brooks1999cog}, a robot head that uses a similar active vision approach as Cog for eye movements and engages in various social activities. Kismet has 3 DoF in eyes, one in its neck and 11 DoF in its face including eyebrows (each with two DoF to lift and arch), ears (each with two DoF to lift and rotate), eyelids (each with one DoF to open or close), and a mouth (with one degree of freedom to open or close). Such flexibility allows Kismet to communicate, via facial expressions, a range of emotions such as anger, fatigue, fear, disgust, excitement, happiness, interest, sadness, and surprise in response to perceptual stimuli.

A particular feature of Kismet that enables the detection of joint attention is its two-part perceptual system containing an attentional system and gaze detection model. The attentional system is designed to detect salient objects (objects of interest in joint attention). The model contains a bottom-up component that finds interesting regions using face and motion detection algorithms as well as low-level color information (using opponent axis space to find saturated main colors). There is also a top-down component in the attentional system controlled by the emotions of the robot. For instance, if the robot is bored the focus of attention shifts more towards the face of the instructor than the object \cite{breazeal1999robot}.

The gaze finding algorithm in Kismet has a hierarchical structure. First, through detection of motion, regions that may contain the instructor's face are selected, then a face detection algorithm is run to locate the face of the instructor. Once identified, the robot fixates on the face and a second camera with zoom capability captures a close-up image of the face. The close range image allows the robot to find the instructor's eyes and as a result determine her gaze direction \cite{scassellati1999knowing}.

\begin{figure}[!t]
\centering
\subfloat[Robovie]{
\includegraphics[width=4cm,height=4cm]{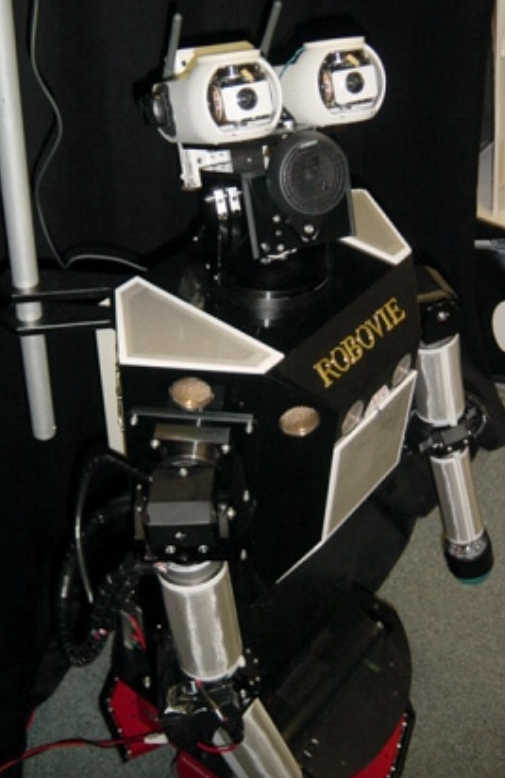}
\label{fig:robovie}}
\hspace{0.5cm}
\subfloat[Geminoid]{
\includegraphics[width=4cm,height=4cm]{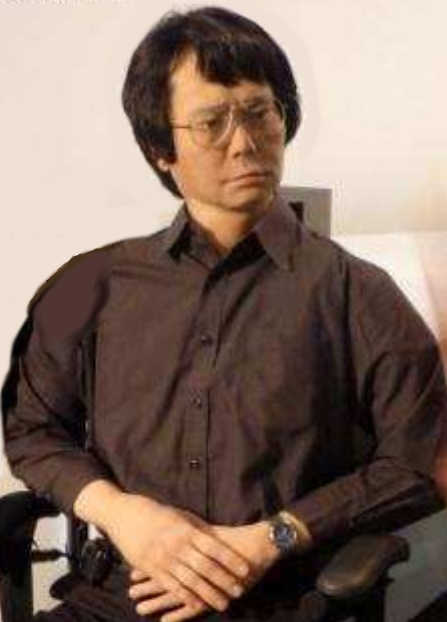}
\label{fig:geminoid}}\\
\subfloat[Leonardo "Leo"]{
\includegraphics[width=4cm,height=4cm]{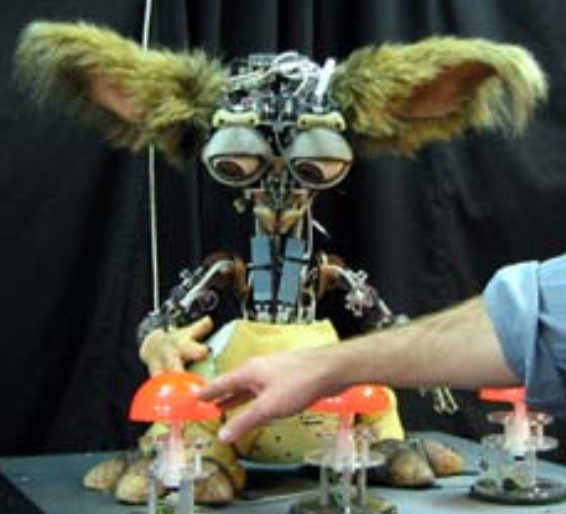}
\label{fig:leo}}
\hspace{0.5cm}
\subfloat[PeopleBot]{
\includegraphics[width=4cm,height=4cm]{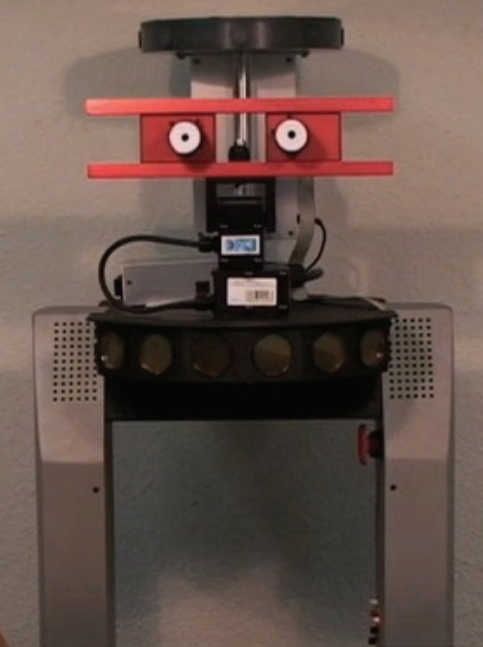}
\label{fig:peoplebot}}\\
\caption[Examples of robotic platforms with joint attention capability used in social interactions]{Examples of robotic platforms with joint attention capability used in social interactions. Source: a) \cite{breazeal2005effects} and b) \cite{ishiguro2001robovie}, c) \cite{becker2010exploring}, and d) \cite{staudte2009visual}.}
\label{fig:social_bots}
\end{figure}

Joint attention mechanisms are typically used in social robots as a means of enhancing the quality of human interaction with the robot. Robovie (\figref{fig:robovie}) \cite{ishiguro2001robovie} is one of the early robots designed for this purpose. This humanoid robot is 120 cm tall, has two arms (8 DoF), a head (3 DoF), two eyes (4 DoF) and a mobile platform. The robot's eyes have a pan-tilt capability to simulate human eye movements. In a number of experiments conducted by the creators of Robovie, it was found that human-like behaviors, such as gaze sharing, engage humans more in interaction. 

The role of shared attention and gaze control has also been investigated in more specific types of interactions. Mutlu \textit{et al.} \cite{mutlu2009nonverbal} show how gaze sharing can improve the ability of a human to play a guessing game with a robotic counterpart. The authors used two robotic platforms Robovie R-2 and Geminoid (\figref{fig:geminoid}), a human look-alike humanoid robot. The game rules are as follows: there is a table with a number of objects placed on it. The goal is for the human player to guess which object is selected by the robot by asking questions about the nature of the object. The authors observed that if the robot used very subtle eye movements towards the region where the object was placed, it would significantly improve the human player ability to correctly guess the object.  Interestingly, the human participants had a better response to the Geminoid robot owing to its human-like facial features.

Zheng \textit{et al.} \cite{zheng2015impacts} examined how gaze control can contribute to a handover activity. In this study, a PR2 robot was instructed to give a bottle of water to the human participants in three different ways: 1) while keeping its gaze constantly downward, 2) while keeping its gaze on the bottle, and 3) while maintaining eye-contact with the human participants. The authors found that in the second scenario the human participants reached for the bottle significantly earlier than in the first scenario, and in the third scenario even earlier than in the second.

Moreover, joint attention is shown to improve the efficiency of verbal communication between humans and robots. Breazeal \textit{et al.} \cite{breazeal2005effects} examine such influence using Leo (\figref{fig:leo}), a 65 DoF social robot. Through a number of interactive experiments between the robot and human participants, they show that joint attention cues such as hand gesture and eye contact help people to better understand the mental process of the robot. In this manner, the authors realized that the most effective technique is when the robot exhibits both implicit communication cues and explicit social cues during the interaction. For instance, eye contact shows that the robot is waiting for the user to say a word, and shrug indicates that it did not understand an utterance of the user. 

Imai \textit{et al.} \cite{imai2003physical} investigate the role of joint attention in a conversational context where the robot attempts to attract the attention of a human to an object. In their experiments, the robot approached the human participants and asked them to look at the poster on the wall. They noticed that when the robot moved its gaze towards the wall, the participants were more likely to follow the robot's instructions to look at the poster. A similar scenario was examined by Staudte \textit{et al.} \cite{staudte2009visual,staudte2011investigating} in which the authors instructed the robot called PeopleBot (\figref{fig:peoplebot}) to name the objects placed in front of it. In the first scenario, the robot was programmed to change its gaze towards the named objects, and in the second one it did not do so. A video of the robot's performance was recorded and displayed to a number of human participants. The authors observed the eye movements of the participants and found that the gaze changes of the robot restrict the spatial domain of the potential referents allowing the participants to disambiguate spoken references and identify the intended object faster. Similarly, Mutlu \textit{et al.} \cite{mutlu2013coordination} show that in joint action, shared attention mechanisms not only help the agents to make sense of each other's behavior, but also help them to repair conversational breakdowns as a result of failure to exchange information.

Human-like gaze following and shared attention capabilities in a robot can also make interaction more interesting and engaging \cite{yonezawa2007gaze}. Shiomi and his colleagues \cite{shiomi2006interactive} tested this hypothesis over a course of two-month trial at a science exhibition. They used various humanoid robotic platforms such as Robovie, Robovie-m (22 DoF robot with legged locomotion and bowing ability), and Robovie-ms (a stationary humanoid robot). The exhibition visitors were provided with RFID tags allowing the robots to become aware of their presence and also to retain some personal information about them to use throughout interactions. During their observations, the authors noticed that joint attention capabilities gave some sort of personality to the robots. A human-like personality makes the robots more approachable by humans and more interesting in terms of interacting with them and listening to their instructions. In other sets of experiments where the joint attention mechanisms were not present, the authors witnessed that people tended to get distracted easily by the appearance of the robots and treated them just as interesting objects, not social beings.

In addition, there are other social interactive applications that benefit from joint attention mechanism. Miyauchi \textit{et al.} \cite{miyauchi2004active} exploit eye contact cues as a means of making sense of the user's hand gestures. Intuitively speaking, the robot can distinguish between an irrelevant and an instructive hand gesture by detecting whether the user is gazing at the robot. Kim \textit{et al.} \cite{kim2015human} use joint attention mechanism in conjunction with an object detector and an action classifier to recognize the intention of the user. Joint attention, which is realized by estimating the skeleton pose of the user, helps to identify the object of interest and then associate it with the human user's activity. Monajjemi \textit{et al.} \cite{monajjemi2015uav} employ a shared attention mechanism to interface with an unmanned aerial vehicle (UAV) to attract the robot to the location of the human that needs assistance (e.g. in a search and rescue context). In this work, shared attention is realized in the form of fast hand-wave movements, which attracts the attention of the UAV from distance.

\subsubsection{Learning}

In recent years, sophisticated algorithms have been developed in which gaze direction guides the identification of objects of interest \cite{yucel2009head,yucel2009joint,yucel2013joint,gorji2016attentional,kera2016discovering}. For instance, in the context of robot learning, Yucel \textit{et al.} \cite{yucel2013joint} use a joint attention algorithm that works as follows: the instructor's face is first detected using Haar-like features. Next, the 3D pose of the face is estimated using optical flow information based on which a series of hypotheses are generated to estimate the gaze direction of the instructor. The final gaze is estimated using a tolerable error threshold which forms a cone-shaped region in front of the instructor covering a portion of the scene. In parallel with this process, a saliency map of the environment is generated using low-level features such as color and edge information to identify objects in the scene. Any of the salient objects that fall within the gaze region of the instructor can be selected as the object of interest (the object that instructor is looking at). The authors tested the proposed algorithm on a Nao humanoid robot placed on a moving platform.

The use of learning algorithms such as neural nets has also been investigated \cite{nagai2002developmental,nagai2003does,ito2004joint,nagai2005role}. In one such work \cite{nagai2005role}, the authors implicitly teach joint attention behavior to the robot. This means, given an image of the scene, the robot has to learn the motor movements that connect the gaze of the instructor to the object of interest. To do so, the robot produces a saliency map of the scene (using low-level features such as color, edges, and motion) to identify the objects. It then performs a face detection to identify the instructor's face. The detection results are then fed into a 3-layer neural net to learn the correct motor motions to follow the gaze of the instructor to the object of interest.

Besides head and eye movements, some scientists developed algorithms capable of identifying the object of interest using auditory input and pointing gestures \cite{haasch2005multi,schauerte2014look}. For instance, Haasch \textit{et al.} \cite{haasch2005multi} use skin color to identify the instructor's hand, and then, by estimating its trajectory of movement, determine what the instructor is pointing at. This helps to narrow down the attention to a small region of the scene. In addition, using speech input, the instructor specifies the characteristics of the object of interest. Using this information the system then filters out unwanted regions. For example, within the scene, if the object is blue, all regions that are not blue are filtered out. The filtered image in conjunction with the direction of pointing gesture helps the system to figure out what the intended object is.

Moreover, some scientists experimented with other means of communication and establishing joint attention \cite{billard1999experiments,dautenhahn1999studying}. For instance, in \cite{dautenhahn1999studying} the authors show that how a mobile robot that is wirelessly communicating with another robot can learn a proto-language by simply following and imitating the instructing robot's movements. In addition, in the same work, a doll-like robot is taught to imitate the hand movements of the instructor using a series of connected infrared emission sensors placed on the instructor's arm and glasses.

\subsubsection{Rehabilitation}

\begin{figure}[!t]
\centering
\captionsetup[subfigure]{}
\subfloat[Infanoid]{
\includegraphics[width=4cm,height=4cm]{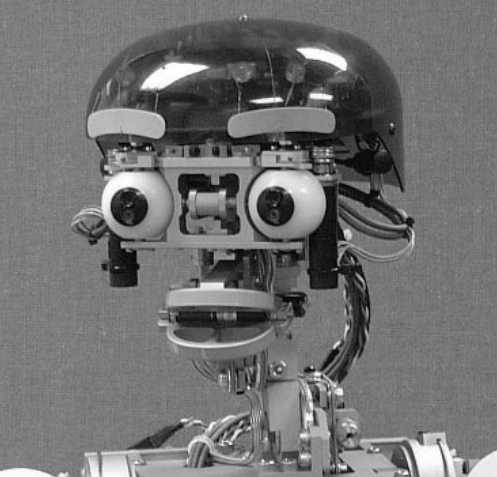}
\includegraphics[width=5cm,height=4cm]{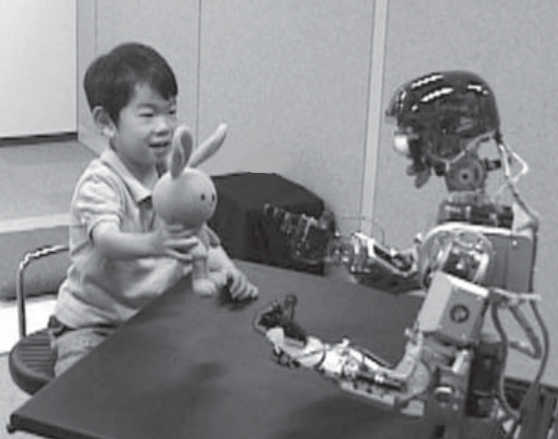}
\label{fig:infanoid}}\\
\subfloat[Keepon]{
\includegraphics[width=4cm,height=4cm]{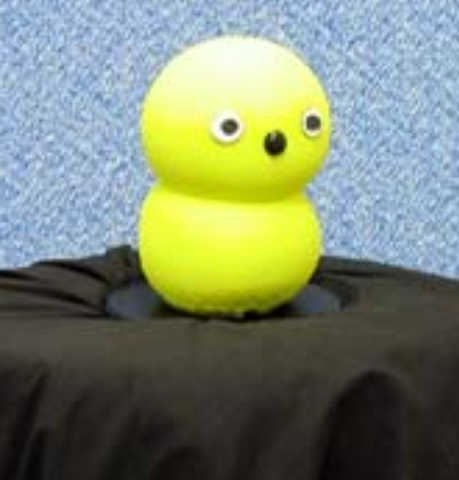}
\includegraphics[width=5cm,height=4cm]{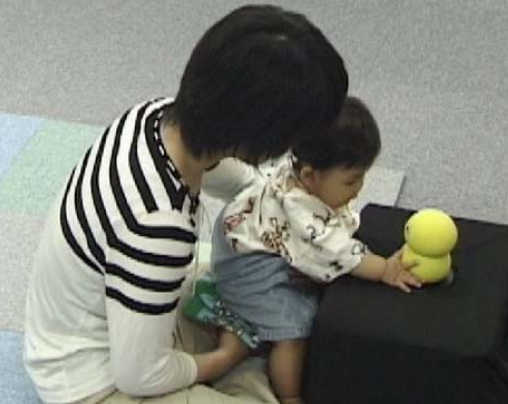}
\label{fig:keepon}}
\caption[Examples of robotic platforms with joint attention capability used for rehabilitation purposes.]{Examples of robotic platforms with joint attention capability used for rehabilitation purposes. Source: a) \cite{kozima2001robot,kozima2004can}, b) \cite{kozima2009keepon}}
\label{fig:rehab_bots}
\end{figure}
Rehabilitation robots are specifically designed for children with social disorders such as autism. Autism is a complex behavioral and cognitive disorder that causes learning problems as well as communication and interaction deficiencies \cite{dautenhahn2000issues}. Through interaction with autistic children, these robotic platforms can help to improve their communicative and social abilities.

Infanoid (\figref{fig:infanoid}) \cite{kozima1998attention,kozima2000epigenetic,kozima2001robot} is one example of rehabilitation social robots. This humanoid robot has an anthropomorphic head with a pair of eyes and servo motors that enable various eye movements. In addition, the robot has a moving neck, arms, lips, and eyebrows enabling the system to perform human-like gestures and facial expressions. The visual processing of the robot for human gaze detection is hierarchical. The robot first detects the face of the user, and then saccades to the face by centering it in the image. The zoom cameras then focus on the eye regions of the face and an algorithm detects the gaze direction. Following the direction of the gaze, the robot identifies where the object of interest is.

Through experimental evaluations, which were performed with autistics kids in the presence of their mothers,  the creators of Infanoid found that although the autistic kids were reluctant at first, proper intentional human-like eye movements could eventually engage them in close interaction and help to maintain it for a long period of time \cite{ito2004robots}. The authors observe that during interaction between the robot and the kids both dyadic and triadic (the kid, the robot, and the mother) behaviors emerged \cite{kozima2005interactive}. The authors later introduced a new robot Keepon (\figref{fig:keepon}) \cite{kozima2004can,kozima2009keepon} for interaction with infants. This small soft creature-like robot has two main functions: orienting its face and showing the emotional state, such as pleasure or excitement, by rocking its body. The use of joint attention mechanism in this new robot shows similar effect in engaging autistic children. In fact, the authors reveal that children go through three phases in order to interact with the robot: 1) neophobia (don't know how to deal with the robot), 2) exploration (through parents explore how the robot behaves), and 3) interaction. Similar studies have also been conducted on autistic kids using different robotic platforms such as a simple mobile robot \cite{dautenhahn2000issues} or LEGO robots, the result of which agree with the findings of works on Infanoid and Keepon.

\section{Hardware Means of Traffic Interaction}
Although the focus of this report is on visual scene understanding, for the sake of completeness, we will briefly discuss hardware approaches that are being (or can potentially be) used in the context of traffic interaction.

\subsection{Communication in Traffic Scenes}
Establishing communication between vehicles, and vehicles and infrastructures have been the topic of interest for the past decades. Technologies such as Vehicle to Vehicle (V2V) and Vehicle to Infrastructure (V2I), which are collectively known as V2X (or Car2X in Europe), are examples of recent developments in the field of traffic communication \cite{hobert2015enhancements,cheng2014index}. These technologies are essentially a real-time short-range wireless data exchange between the entities allowing them to share information regarding their pose, speed, and location \cite{narla2013evolution}.

Sharing this information between the vehicles and infrastructures enables the road users to detect hazards, calculate risks, issue warning signals to the drivers and take necessary measures to avoid collisions. In fact, it is estimated that V2V alone could prevent up to 76\% of the roadway crashes and that V2X technologies could lower unimpaired crashes by up to 80\% \cite{narla2013evolution}.

Recent developments extend the idea of V2X communication to connect Vehicles to Pedestrians (V2P). For instance, Honda proposes to use pedestrians' smart phones to broadcast their whereabouts as well as to receive information regarding the vehicles in their proximity. In this way, both smart vehicles and pedestrians are aware of each other's movements, and if necessary, receive warning signals when an accident is imminent \cite{Cunningham2013}.

In spite of their effectiveness in preventing accidents, V2X technologies are being criticized on two grounds. First, efficient communication between the road users is highly dependent on the functioning of all involved parties. If one device malfunctions or transmits malicious signals, it can interrupt the entire communication network. The second issue is privacy concerns. A recent study shows that one of the major problems of the road users with employing V2X technologies is sharing their personal information in the network \cite{schmidt2015v2x}.

\begin{figure}[!t]
\centering
\subfloat[]{
\includegraphics[width=0.4\textwidth, height=5cm]{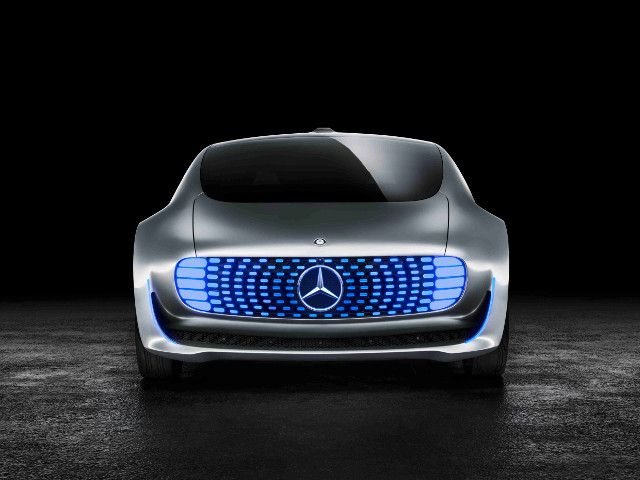}}
\hspace{0.2cm}
\subfloat[]{\includegraphics[width=0.4\textwidth, height=5cm]{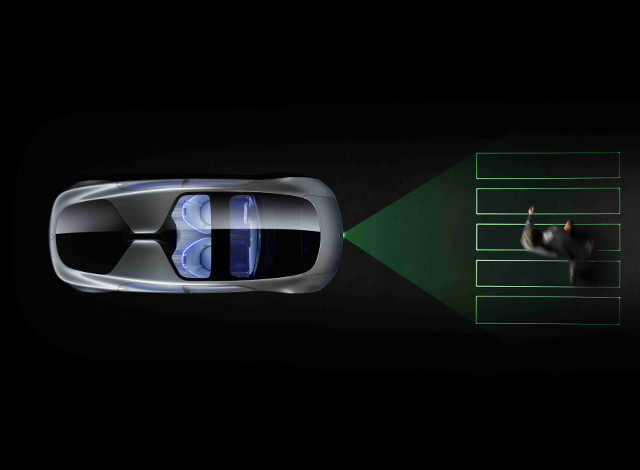}}\\
\subfloat[]{\includegraphics[width=0.4\textwidth, height=5cm]{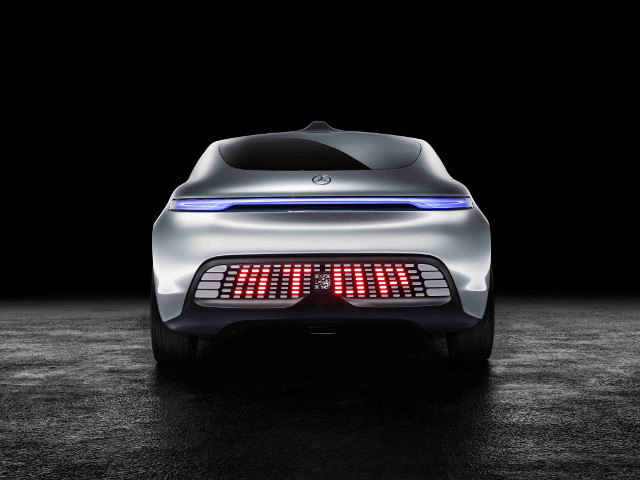}}
\hspace{0.2cm}
\subfloat[]{\includegraphics[width=0.4\textwidth, height=5cm]{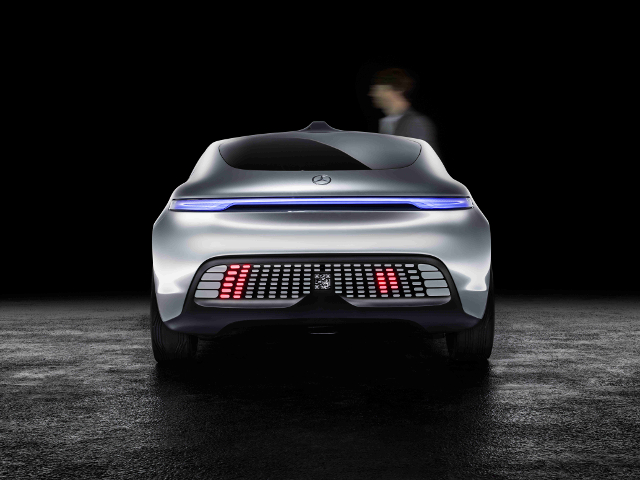}}
\caption[The concept autonomous vehicle by Mercedes Benz with communication capability.]{The concept autonomous vehicle by Mercedes Benz with communication capability: a) grill LED lights, blue indicates autonomous mode and white manual mode, b) the car is projecting a zebra crossing for the pedestrian to cross, c) rear LED lights are requesting the vehicles behind to stop, and d) rear LED lights are showing that a pedestrian is crossing in front of the vehicle. Source \cite{benzauto}.}
\label{fig:benz_auto}
\end{figure}

Since recently, car manufacturers are turning towards less intrusive forms of communication that do not require the corresponding parties to share their personal information. Techniques such as using blinking LED lights \cite{lagstrom2015avip}, color displays \cite{googledisp} or small surface projectors \cite{mitslights} to visualize the vehicle's intention have been investigated. Some vehicles also use a combination of these methods to communicate with traffic. For instance, Mercedes Benz, in their most recent concept autonomous vehicle (as illustrated in \figref{fig:benz_auto}), uses a series of LED lights at the rear end of the car to  ask other vehicles to stop/slow or inform them if a pedestrian is crossing, a set of LED fields at the front to indicate whether the vehicle is in autonomous or manual mode and a projector that can project zebra crossing on the ground for pedestrians \cite{benzauto}.

To make the communication with pedestrians more human-like, some researchers use moving-eye approach. In this method, the vehicle is able to detect the gaze of the pedestrians, and using rotatable front lights, it establishes (the feeling of) eye contact with the pedestrians and follow their gaze \cite{pennycooke2012aevita}. Some researchers also go as far as suggesting to use a humanoid robot in the driver seat so it performs human-like gestures or body movements during communication \cite{mirnig2017three}.

Aside from establishing direct communication between traffic participants, roadways can be used to transmit the intentions and whereabouts of the road users. During the recent years, the concept of smart roads has been gaining popularity in the field of intelligent driving. Smart roads are equipped with sensors and lighting equipment, which can sense various conditions such as vehicle or pedestrian crossing, changes in weather conditions or other forms of hazards that can potentially result in accidents. Through the use of visual effects, the roads then inform the road users about the potential threats \cite{siess2015hybrid}.

\begin{figure}[!t]
\centering
\subfloat[Van Gogh Path]{
\includegraphics[width=0.4\textwidth]{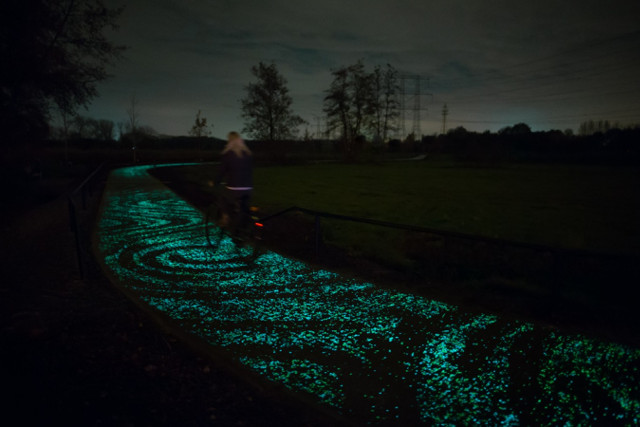}}
\hspace{0.2cm}
\subfloat[Glowing Lights]{\includegraphics[width=0.4\textwidth]{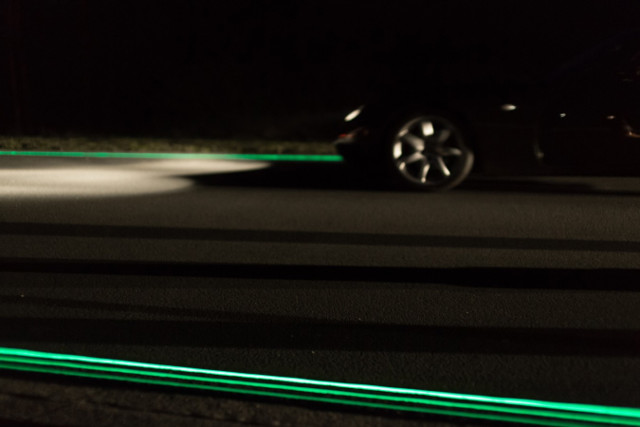}}\\
\caption[Examples of smart roads used in Netherlands.]{Examples of smart roads used in Netherlands. Source \cite{smartroad}.}
\label{fig:smartroad}
\end{figure}

In addition to the transmission of warning signals, smart roads can potentially improve the visibility of roads and attract the attention of traffic participants (especially pedestrians) from electronic devices or other distractors to the road \cite{siess2015hybrid}. Today, in some countries such as Netherlands smart roads are in use for recreational and safety purposes (see \figref{fig:smartroad}).

\subsection{Pedestrian Intention Estimation Using Sensors}

The use of various sensors for activity and event recognition has been extensively investigated  in the past decades. Although the majority of the research in this field focuses on general activity recognition \cite{tapia2004activity,kwapisz2011activity} or health monitoring systems such as fall detection in indoor environments \cite{bourke2008threshold,vo2012fall}, some works have potential application for outdoor use in the context of traffic interaction. Below, we briefly list some of these systems. 

Sensors used in activity recognition can be divided into three categories: 1) sensors mounted on the human body, 2) sensors that are in devices carried by humans or 3)  external sensors that monitor the environment. The first group are referred to as wearable sensors and can be in the form of bi-axial accelerometer \cite{lee2001recognition,bao2004activity,lee2002activity}, digital compass \cite{lee2001recognition}, bi-axial gyroscope sensor \cite{bohrer1995integrated}, inertial measurement sensors \cite{kourogi2010method,tong2009research} and Surface Electromyography (SEMG) sensors \cite{gao2016pedestrian}. The second group include sensors such as accelerometer \cite{kwapisz2011activity} or orientation sensors \cite{vo2012fall} in smartphone devices \cite{su2014activity,mehner2013location}. The third group, external sensors, ranging from RADAR \cite{stormo2014human,amin2016radar}, WiFi \cite{wang2015understanding} and motion detectors \cite{van2008accurate} to infrared light detectors \cite{lee2001recognition}, microphones \cite{toreyin2005hmm}, and state-change sensors (e.g. reed switches) \cite{van2008accurate,tapia2004activity}. The interested reader is referred to these survey papers \cite{lara2013survey,rashidi2013survey} for more information.

\section{Visual Perception: The Key to Scene Understanding}
Visual perception in autonomous driving can either be realized through the use of passive sensors such as optical cameras and infrared sensors or active sensors such as LIDAR and RADAR. Although autonomous vehicles take advantage of active sensors for range detection and depth estimation (in particular under bad weather or lighting conditions), optical cameras are still the dominant means of scene analysis and understanding. For autonomous driving optical cameras are better for a number of reasons: they provide richer information, are non-invasive, more robust to interference and less dependent on the type of objects (whereas for active sensors reflection and interference are major issues) \cite{bertozzi2000vision}. As a result, optical cameras are expected to play the main role in traffic scene analysis for years to come.

For the reasons mentioned above, we put our main focus on reviewing algorithms that use optical cameras and only briefly review some algorithms based on active sensors. Even though active sensors are fundamentally different from optical cameras, the algorithms for processing the data generated by these sensors share some commonalities.

we subdivide the materials on visual perception into three categories. Since the prerequisite to scene understanding is identifying the relevant components in the scene, we begin the discussion with algorithms for detection and classification. Then, we review pose estimation and activity recognition algorithms, which help to understand the current state of the road users in the scene. In the end, we conclude the revision by discussing the methods that predict the upcoming events in the scene, i.e. intention estimation algorithms.

\subsection{Realizing the Context: Detection and Classification}
In computer vision literature, detection and classification algorithms are either optimized for a particular class of objects or are multi-purpose and can be used for different classes of objects (generic algorithms). Here, we will review the algorithms in both categories. In the case of the object specific algorithms we put our focus on reviewing the algorithms designed for recognizing traffic scene elements, namely pedestrians, vehicles, traffic signs and roads.

Object recognition algorithms are commonly learning-based and rely on different types of datasets to train. Knowing the nature of the datasets sheds some light on the ability of the proposed algorithms in identifying various classes of objects. Hence, we begin by discussing some of the most popular datasets in the field.

\subsubsection{Datasets}
\subsubsection*{Generic Datasets}
Generic datasets, as the name implies, contain a broad range of object classes. They may include live objects such as people, animals or artificial objects such as cars, tools or furniture. A comprehensive list of the most known datasets and their properties can be found in Table \ref{table:generic_data}.

Among the available datasets, many are being widely used today. The Visual Object Classes (VOC) dataset is one of the most used and cited datasets since 2005 and contains images for object detection and classification (with bounding box annotations) and pixel-wise image segmentation. The object recognition part of the dataset contains more than 17K color images comprising 20 object classes. ImageNet is similar to VOC but in a much larger scale. Today, ImageNet contains more than 14 million images out of which over 1 million have bounding boxes for 1000 object categories. In addition, a portion of the samples in ImageNet comes with object attributes such as the type of materials, color of the objects, etc. Similar to VOC, ImageNet is used in a yearly challenge known as ImageNet Large Scale Visual Recognition Challenge (ILSVRC). 

\begin{figure}[!t]
\centering
\subfloat[ImageNet]{
\includegraphics[width=5cm,height=3.5cm]{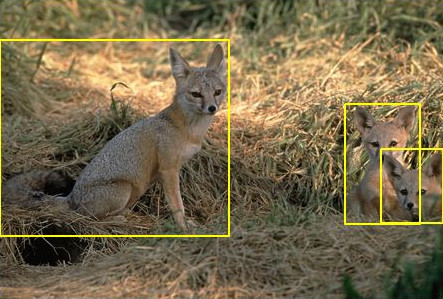}}
\subfloat[MS COCO]{
\includegraphics[width=5cm,height=3.5cm]{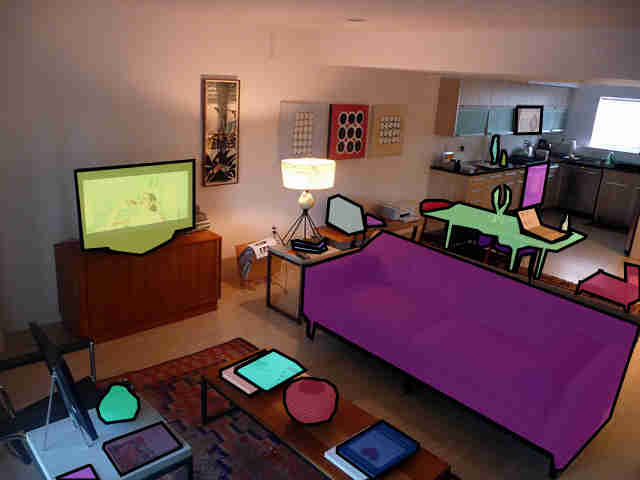}}
\subfloat[Places-205]{
\includegraphics[width=5cm,height=3.5cm]{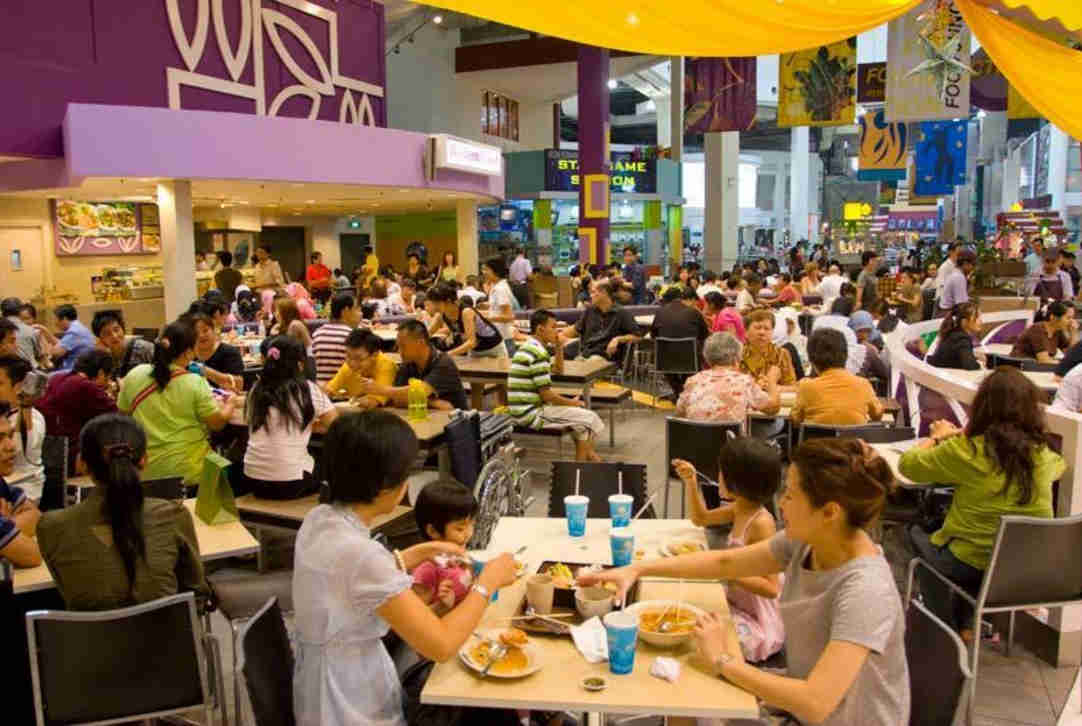}}
\caption[Sample images from three major object recognition datasets.]{Sample images from 3 major object recognition datasets.}
\label{fig:gen_data}
\end{figure}

MS COCO is another large-scale object recognition and segmentation dataset that consists of more than 300K color images with pixel-wise ground truth labels. For scene recognition, researchers at MIT have introduced the Places-205 and Places-365 datasets each with more than 2.5 and 10 million color images of various scenes respectively. The extension numbers of the Places datasets refer to the number of scene categories each dataset has (for sample images from major datasets see \figref{fig:gen_data}).

\begin{table}[!hbtp]
\caption[Generic object recognition datasets.]{Generic object recognition datasets. Abbreviations: \textit{object type:} O = Objects, D = Digits, A = Animals, AT = Attributes, S = Scenes, \textit{data type:} Gr = Greyscale, Col = Color, R = Range, CAD = 3D meshes, \textit{ground truth:} CL = Class Label, PW = Pixel-Wise, BB = Bounding Box, OB = Object Boundaries, AL = Attribute Label, P = Pose (3D).}
\centering
\resizebox{\textwidth}{!}{
\begin{tabular}{|c|c|c|c|c|c|}
\hline
Dataset & Year & Obj. Type/No. Categories & Data Type & No. Frames & Ground Truth\\
\hline \hline
COIL-20 \cite{nene1996columbia} & 1996 & O/20 & Gr/Multiviews & 2K+ & CL \\ \hline
MNIST \cite{lecun1998gradient}&1998&D/10&Gr&70K&PW \\ \hline
MSRC-21 \cite{winn2005object}&2000&O/32&Col&800&PW \\ \hline
Caltech4 \cite{fergus2003object}&2003&O/10&Col&1.4K+&BB/CL \\ \hline
Caltech 101 \cite{fei2006one}&2004&O/101&Col&32K+& OB/CL \\ \hline
VOC2006 \cite{pascal-voc-2006}&2006&O/10&Col&5200&BB/CL \\ \hline
VOC2007\cite{pascal-voc-2007}&2007&O/20&Col&5K+&BB/CL \\ \hline
VOC2008\cite{pascal-voc-2008}&2008&O/20&Col&4.3K+&BB/CL \\ \hline
SUN09 \cite{choi_cvpr10}&2009&O/200&Col&12K&BB/CL \\ \hline
AwA \cite{lampert2009learning}&2009&A/50, AT/85&Col&180K+&CL/AL \\ \hline
NUS-WIDE \cite{nus-wide-civr09}&2009&O/81&Col&269K+&CL \\ \hline
CIFAR-100 \cite{krizhevsky2009learning}&2009&O/100&Col&60K&CL \\ \hline
ImageNet \cite{ILSVRC15}&2010&O/1000, AT/21K&Col&1.2M+&OB/CL/AL \\ \hline
VOC \cite{pascal-voc-2010}&2010&O/20&Col&15K+&BB/CL \\ \hline
SUN297 \cite{xiao2010sun}&2010&O/367&Col&108K+&BB/CL \\ \hline
RGB-D \cite{lai2011large}&2011&O/300&Col and R&41K+&P/CL \\ \hline
Willow Garage \cite{aldoma2014automation}&2011&O/110&Col and R&353&CL/ PW/P \\ \hline
2D3D \cite{browatzki2011going}&2011&O/156&Col and R&11K+&- \\ \hline
Sun Attribute \cite{Patterson2012SunAttributes}&2011&AT/102&Col&14K +&CL \\ \hline
SVHN \cite{netzer2011reading}&2011&D/10&Col&600K+& OB/CL \\ \hline
NYUD2 \cite{silberman2012indoor}&2012&O/1000+&Col and R&407K+ (1449 labeled)&PW/CL \\ \hline
VOC2012\cite{pascal-voc-2012}&2012&O/20&Col&17K+&OB/CL \\ \hline
MS COCO \cite{lin2014microsoft}&2014&O/80&Col&300K+&PW/CL \\ \hline
Places-205 \cite{zhou2014learning}&2014&S/205&Col&2.5M&CL \\ \hline
ModelNet \cite{wu20153d}&2015&O660&CAD &151K+&CL \\ \hline
Places-365 \cite{zhou2017places}&2017&S/365&Col&2.5M&CL \\ \hline
\end{tabular}
}
\label{table:generic_data}
\end{table}

\subsubsection*{Traffic Scene Datasets}

\begin{figure}[!t]
\centering

\subfloat[KITTI]{
\includegraphics[width=6cm,height=4cm]{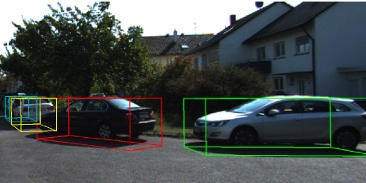}}
\subfloat[Cityscapes]{
\includegraphics[width=6cm,height=4cm]{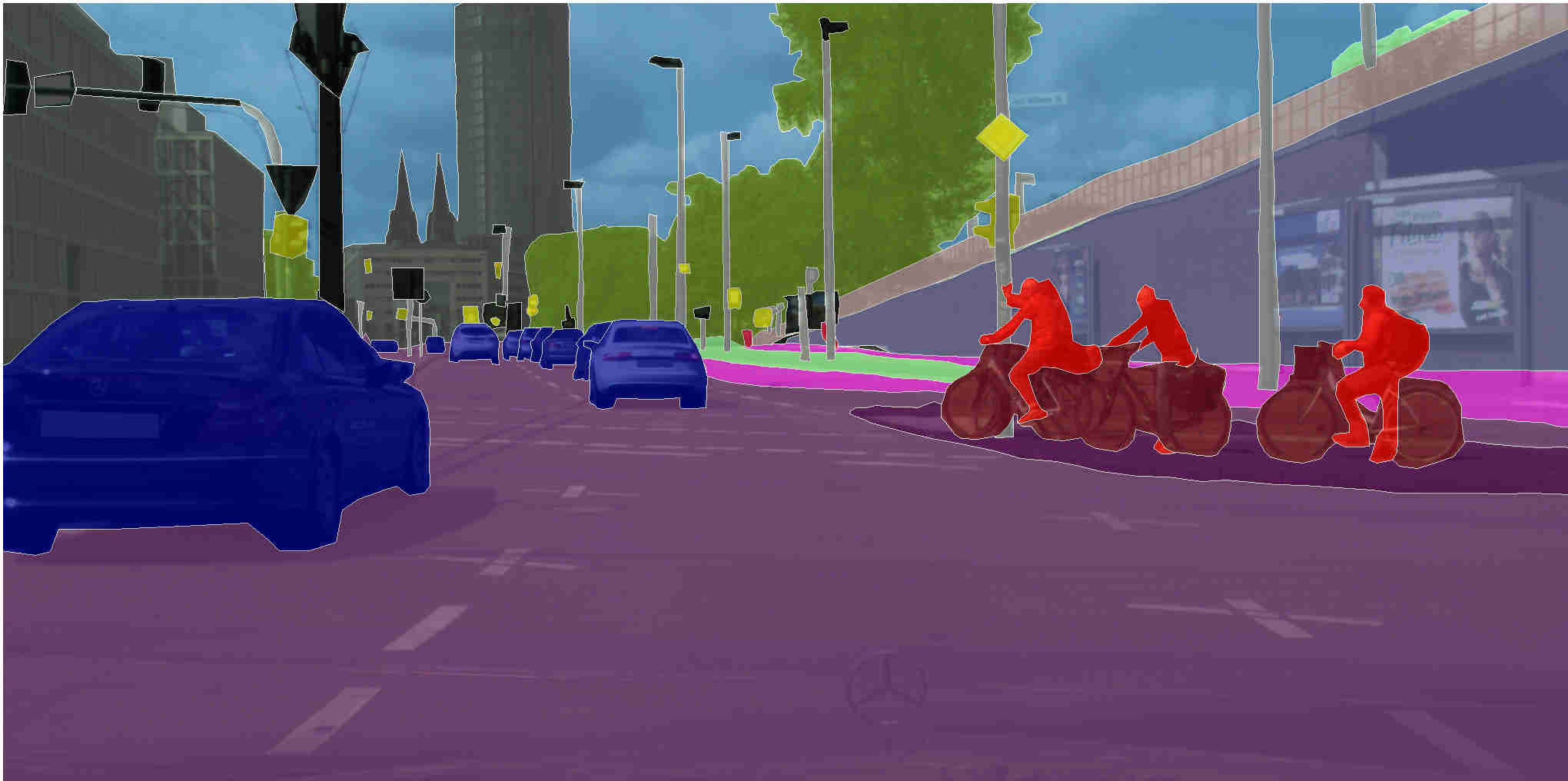}
}
\caption[Sample images from 2 major traffic scene datasets.]{Sample images from two major traffic scene datasets and their associated ground truth annotations.}
\label{fig:traffic_data}
\end{figure}

A number of datasets are particularly catered to traffic scene understanding applications. KITTI \cite{Geiger2013IJRR} is one of the best known and widely used datasets with over 30K color stereo image samples as well as associated laser point cloud data and 3D GPS information. For object recognition purposes, the dataset contains bounding box annotations for 7 classes of objects including vehicles (car, van, and truck), people (person and pedestrian), cyclists and trams. A part of the dataset is annotated with pixel-wise ground truth for road surfaces. Another major dataset, Cityscapes \cite{Cordts_2016_CVPR}, has pixel-wise ground truth annotations for other types of objects in traffic scenes including sky, signs, and trees. For sample images from KITTI and Cityscapes see \figref{fig:traffic_data}.

Table \ref{table:traffic_data} lists 7 most common datasets for traffic scene understanding. Note that these datasets are mentioned here because they contain images that are collected from street-level view via a camera placed in the car.

\begin{table}[!hbtp]
\caption[Datasets for traffic scene understanding.]{Datasets for traffic scene understanding. Abbreviations: \textit{data type:} Gr = Greyscale, Col = Color, R = Range, St = Stereo, \textit{ground truth:} CL = Class Label, PW = Pixel-Wise, BB = Bounding Box, P = Pose (3D).}
\centering
\resizebox{\textwidth}{!}{
\begin{tabular}{|c|c|c|c|c|c|}
\hline
Dataset & Year & No. Categories & Data Type & No. Frames & Ground Truth\\
\hline \hline
CBCL \cite{bileschi2006streetscenes}&2007&9&Col&3.4K&BB/PW/CL\\ \hline
CamVid \cite{BrostowSFC:ECCV08}&2008&32&Col&701&PW/CL\\ \hline
Karlsruhe \cite{Geiger2011NIPS}&2011&2&Gr&1.7K&BB/PW/CL\\ \hline
KITTI \cite{Geiger2013IJRR}&2013&7&Col, R, St, GPS&30K+&P/CL\\ \hline
Daimler-Urban \cite{scharwachter2014stixmantics}&2014&3&Col, St&5K&PW/CL\\ \hline
Cityscapes \cite{Cordts_2016_CVPR}&2016&30&Col&25K+&PW/CL\\ \hline
Inner-city \cite{yang2016exploit}&2016&4&Col&24K+&BB/CL\\ \hline
\end{tabular}
}
\label{table:traffic_data}
\end{table}

\subsubsection*{Pedestrian Detection Datasets}
Since identifying pedestrians in traffic scenes is one of the most challenging tasks, there are a number datasets that are specifically designed for training pedestrian detection algorithms. Some of the most common ones that are being used today are Caltech \cite{dollarCVPR09peds}, INRIA \cite{dalal2005histograms}, and Daimler \cite{enzweiler2009monocular}. The Caltech dataset, in particular, is one of the best known datasets for two reasons: it contains a very large number of pedestrian samples (350K) and also occlusion information in the form of a bounding box that covers only the visible portion of occluded pedestrians. Table \ref{table:ped_data} enumerates some widely used pedestrian detection datasets. One should note that since the focus of this report is on autonomous driving applications, the datasets listed in the table are the ones that contain street-level view of the pedestrians. There are, however, few exceptions that also have images from other viewpoints such as CUHK, PETA and INRIA.

\begin{table}[!hbtp]
\caption[Pedestrian detection datasets.]{Pedestrian detection datasets. Abbreviations: \textit{data type:} Gr = Greyscale, Col = Color, St = Stereo, \textit{ground truth:} CL = Class Label, PW = Pixel-Wise, BB = Bounding Box, V = Visibility, P = Pose (3D).}
\centering
\resizebox{\textwidth}{!}{
\begin{tabular}{|c|c|c|c|c|c|}
\hline
Dataset & Year & No. Ped. Samples & Data Type & No. Frames & Ground Truth\\
\hline \hline
MIT \cite{OrePap97}&1997&924&Col&924&CL\\ \hline
INRIA \cite{dalal2005histograms}&2005&1.7K+&Col&2.5K&BB\\ \hline
Daimler-Class \cite{munder2006experimental}&2006&4K&Gr&9K&CL\\ \hline
Penn-Fudan \cite{wang2007object}&2007&345&Col&170&BB/PW\\ \hline
ETHZ \cite{ess2008mobile}&2008&14.4K&Col&2.2K&BB\\ \hline
Caltech \cite{dollarCVPR09peds}&2009&350K&Col&250K+&BB,V\\ \hline
Daimler-Mono \cite{enzweiler2009monocular}&2009&15.5K+&Gr&21K+&BB\\ \hline
Daimler-Occ \cite{enzweiler2010multi}&2010&90K&Gr,St&150K+&CL\\ \hline
TUD Brussels \cite{wojek2009multi}&2010&3K+&Col&1.6K&BB\\ \hline
Daimler-Stereo \cite{keller2011new}&2011&56K+&Gr,St&28K+&BB\\ \hline
Berkley \cite{fragkiadaki2012two}&2012&2.5K&Col,St&18 Videos&PW\\ \hline
CUHK \cite{ouyang2012discriminative}&2012&-&Col&1063&BB\\ \hline
PETA \cite{deng2014pedestrian}&2014&8.7K&Col,St&19K&AL\\ \hline
Kaist \cite{hwang2015multispectral}&2015&103K+&Col,T&95K&BB,V\\ \hline
CityPerson \cite{zhang2017citypersons}&2017&35K&Col,St&5K&BB/PW\\ \hline
\end{tabular}
}
\label{table:ped_data}
\end{table}

\subsubsection*{Vehicle Recognition Datasets}
There exists a number of vehicle detection datasets that contain mainly images of passenger cars for detection and classification purposes. Some of these datasets such as Standford \cite{krause20133d} have cropped out images of vehicles for classification purposes while the others include vehicle images in street scenes (e.g. NYC3D \cite{MatzenICCV13}), making them suitable for detection tasks. A list of vehicle datasets can be found in Table \ref{table:car_data}. It should be noted that only three of these datasets (e.g. LISA \cite{sivaraman2010general}, GTI \cite{arrospide2012video} and TME \cite{TMEMotorwayDataset}) are recorded strictly from the vehicle's point of view. The rest, contain either views from different perspectives (e.g. CVLAB \cite{OzuysalLF09}) or only show side-views of the vehicles (e.g. UIUC \cite{agarwal2002learning}).

\begin{table}[!hbtp]
\caption[Vehicle detection and classification datasets.]{Vehicle detection and classification datasets. Abbreviations: \textit{data type:} Gr = Greyscale, Col = Color, R = Range, EM = Ego-Motion, \textit{ground truth:} CL = Class Label, BB = Bounding Box, RA = Rotation Angle, GL = Geographical Location, TD = Time of Day.}
\centering
\resizebox{\textwidth}{!}{
\begin{tabular}{|c|c|c|c|c|c|}
\hline
Dataset & Year & No. Vehicle Samples & Data Type & No. Frames & Ground Truth\\
\hline \hline
CMU/VASC \cite{schneiderman2000statistical}&2000&213&Gr&104&BB\\ \hline
MIT-CBCL \cite{PapPog99b}&2000&516&Col&516&CL\\ \hline
UIUC \cite{agarwal2002learning}&2002&550&Gr&1050&BB\\ \hline
CVLAB \cite{OzuysalLF09}&2009&20&Col&2K+&BB/RA\\ \hline
LISA \cite{sivaraman2010general}&2010&-&Col&3 Videos&BB\\ \hline
GTI \cite{arrospide2012video}&2012&3.4K+&Col&7K+&BB\\ \hline
TME \cite{TMEMotorwayDataset}&2012&-&Col,R,EM&30k+&BB\\ \hline
NYC3D \cite{MatzenICCV13}&2013&3.7K+&Col&5K+&3D/2D BB, CL, GL,TD\\ \hline
Stanford \cite{krause20133d}&2013&16K&Col&16K+&CL\\ \hline
\end{tabular}
}
\label{table:car_data}
\end{table}

\subsubsection*{Traffic Sign Recognition Datasets}
Among the observable elements in traffic scenes, traffic signs have one of the highest variabilities in appearances. Depending on the shape, color or descriptions on the signs, each can convey a different meaning for controlling traffic flow. Some datasets try to capture such variability by putting together cropped samples from traffic scenes (e.g. GTSRB \cite{Stallkamp-IJCNN-2011}) or showing the signs in the streets (e.g. BTSD \cite{mathias2013traffic}) accompanied by bounding box information and class labels. A number of traffic sign datasets are shown in Table \ref{table:sign_data}.

\begin{table}[!hbtp]
\caption[Sign recognition datasets.]{Sign recognition datasets. Abbreviations: \textit{data type:} Col = Color, \textit{ground truth:} CL = Class Label, BB = Bounding Box, V = Visibility, S = Size.}
\centering
\resizebox{\textwidth}{!}{
\begin{tabular}{|c|c|c|c|c|c|c|}
\hline
Dataset & Year & No. Categories & Data Type & No. Frames & Ground Truth\\
\hline \hline
UAH \cite{maldonado2007road}&2007&-&Col&200+&-\\ \hline
MASTIF \cite{vsegvic2010computer}&2009-2011&28&Col&10k&BB, S,CL\\ \hline
STSD \cite{larssonSCIA2011}&2011&-&Col&20k+&BB,V,CL\\ \hline
GTSRB \cite{Stallkamp-IJCNN-2011}&2011&43&Col&50k+&BB, S,CL\\ \hline
Lisa \cite{mogelmose2012vision}&2012&47&Col&6.6k&BB,S,V,CL\\ \hline
GTSD \cite{Houben-IJCNN-2013}&2013&42&Col&900&BB,S,V,CL\\ \hline
BTSD \cite{mathias2013traffic}&2013&62&Col&9k+&BB,S,CL\\ \hline
\end{tabular}
}
\label{table:sign_data}
\end{table}

\subsubsection*{Road Recognition Datasets}
Road detection algorithms often use traffic scene datasets such as KITTI or Cityscape. KITTI dataset, for instance, has pixel-wise annotations for different categories of road surfaces such urban unmarked (UU), and urban marked (UM), urban multiple marked lanes (UMM). There are very few datasets particularly designed for road recognition purposes. Of interest for autonomous driving applications is the Caltech-Lanes dataset \cite{aly2008real} which contains 1225 color images of streets with pixel-wise ground truth annotations of lane markings.

\subsubsection{Generic Object Recognition Algorithms}

Early works in object recognition are mainly concerned with identifying 3D shapes. These algorithms often rely on a pre-existing 3D model of an object that defines the relationship between its various aspects such as edges \cite{belongie2001shape}, vertices \cite{chakravarty1982characteristic} and/or surfaces \cite{oshima1983object}. For instance, Chakravarty and Freeman \cite{chakravarty1982characteristic} characterize a 3D object in terms of its lines (edges) and junctions (vertices), the combination of which can take one of the 5 possible junction forms (e.g. T shape). In their representation, each object, depending on what form of junctions are observable (e.g. 2 T-junctions and 1 U-junction), can have a number of unique views or as they term them characteristic-views (CVs). The task of recognition is then limited to identifying junction forms in the image and match them to those model representations in the database. For a review of similar 3D object recognition algorithms refer to \cite{besl1985three}.

The aforementioned 3D object recognition algorithms, deal with identification of rather simplistic 3D shapes such as cubes or cylinders. Later works, however, attempt to recognize more sophisticated objects using model-based techniques \cite{lowe1987three,lamdan1988object}. In a well-known algorithm by Lowe \cite{lowe1987three}, the author reduces the problem of 3D object recognition to estimating the projection parameters of the given model points into the corresponding image points by identifying correspondences between a known 3D model of the object and points observed in the scene. He argues that features used for such a purpose should have two characteristics: be viewpoint invariant and not occur accidentally in the image (due to viewpoint selection or background clutter). Based on these criteria, to form features, Lowe's algorithm perceptually groups the edges of a model based on their proximity, parallelism and collinearity. Having identifying such relationships in the image, the author then tries to match them with the object 3D model using a least square technique. Once the match occurred, the projection parameters (rotation and translation) can be calculated. Lowe points that at least three matches are required to accurately estimate the parameters.

To solve the problem of 3D object recognition, some researchers use an active approach in which the reasoning is based on the changes in the appearance of the object in the scene \cite{wilkes1992active,dickinson1994active}. Wilkes and Tsotsos \cite{wilkes1992active} perform such active recognition by using a camera mounted on a robotic arm attached to a moving platform. Here, the objective is to first find a prominent line (the longest line) in the scene, and then move the camera so the line is placed in the center. Next, the camera moves parallel to the line in 16 consecutive poses to find the view that maximizes the length of the line. In the same manner, a second prominent line is selected (that is not parallel to the first one) but this time by moving the camera perpendicularly with respect to the first line. The second viewpoint then would be the standard viewpoint of the object in which two prominent lines have the maximum length. The standard viewpoint is used in a tree search algorithm to match against the available models in the database to recognize the object.

\begin{figure}[!t]
\centering
\includegraphics[width=0.5\textwidth]{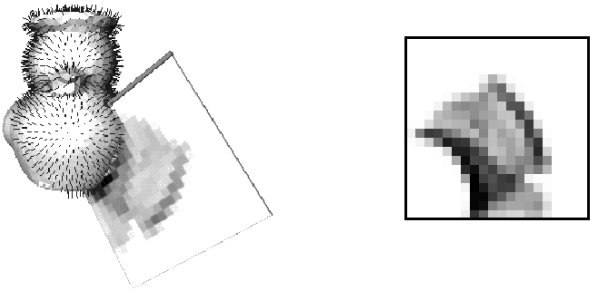}
\caption[Generation of a spin image from a 3D surface.]{Generation of a spin image from a 3D surface.}
\label{fig:spin_image}
\end{figure}

One particular drawback of the earlier 3D recognition algorithms is their dependency on identifying the edges of the objects. Therefore, they perform  best when the background clutter is minimal, for instance, the objects are placed in front of a uniform black or white background. In addition, to lower the complexity of search and reasoning, assumptions are often made to simplify the problem. These may include constraining the distance of the camera to the object, limiting the possible poses of the object and controlling the illumination of the scene. To avoid constraints, some authors use more complex 3D features to define the shape of an object. Johnson and Hebert \cite{johnson1999using} introduce the concept of spin images to characterize 3D shapes. Intuitively, spin images are 2D projections of 3D parts of an object that are generated by passing a 2D plane through the 3D shape (see \figref{fig:spin_image}). These images are generated uniformly from the 3D model of the object, hence, the relationship between different spin images is also preserved. The combination of spin images characterizes a 3D object and can be used to distinguish the object from the distractors.

Advancements in the quality of imaging sensors gave rise to the use of pixel intensity and color values for object recognition. The methods using these features either naively learn the pixel value distributions of an object \cite{swain1991color,gevers1999color}, average pixel value distributions over multiple viewpoints \cite{pontil1998support}, or generate more powerful features by finding the relationship between pixel values in local regions of the scene \cite{shotton2008semantic,torralba2004sharing}. One of the early attempts to use color for recognition is by Swain and Ballard \cite{swain1991color}. The authors propose to learn the characteristics of an object by generating a 3D color histogram of the object's appearance. Then they use the distribution to localize and classify the scene in the following way. For localization, they perform histogram backprojection which replaces the color pixel values in the scene by their distribution values from the object's 3D histogram. To decide whether the observed scene contains the object of interest, histogram intersection operation is used. Overall, although color and pixel intensities can be used to characterize an object, they are very sensitive to noise, illumination changes, and background clutter.

In the late 90's, increase in computation power allowed the researchers to develop more powerful feature descriptors for object recognition. One such method was proposed by Lowe \cite{lowe1999object,lowe2004distinctive} the Scale Invariant Feature Transform, better known as SIFT. The purpose of this method is to find key points in an image that are invariant to scale, rotation and translation. SIFT features are generated as follows: at each octave, images are successively blurred by applying a Gaussian filter and subtracted from the image in the previous scale (e.g. at the beginning first blurred image is subtracted from the original image) to generate difference of Gaussian (DOG) images. Next, in each series of images at the given octave, maxima and minima pixels are localized by comparing the pixel value with its 8 neighboring pixels in all scales. Once the key points are detected, their local gradient orientation and magnitude are calculated. A threshold is applied to pick the most dominant key points. For classification purposes, a local descriptor (for a 2x2 or 4x4 patch) around the key point is generated by calculating the gradient characteristics of the neighboring pixels. The descriptors of the object are learned by  using a classification method such as decision tree \cite{lowe1999object} or SVM \cite{bilen2014object}.

Other types of features commonly used in object recognition are Haar (a local intensity pattern) \cite{viola2001rapid,lienhart2002extended}, steerable filters \cite{torralba2003context}, and histogram of oriented gradients (HOG) \cite{felzenszwalb2010object,yao2012describing}. Depending on the application, these features achieve a different level of performance. For instance, Haar features are shown to have a discriminative representation for face detection \cite{viola2001rapid} whereas HOG features are quite effective in person detection \cite{ouyang2012discriminative}. In more recent years, due to the availability of range sensors, some researchers also take advantage of depth information for object recognition \cite{bo2014learning,gupta2014learning} to segment an object from its surroundings or to retrieve the 3D shape of an object. To achieve a better recognition performance, some scientists also experiment with combining various features for more robust representation of objects. For instance, SIFT has been used with color features \cite{uijlings2013selective}, color and HOG features \cite{deselaers2012weakly} or with HOG and local binary pattern (LBP) features \cite{chen20153d}.

Today, the field of object recognition is dominated by Convolutional Neural Networks (CNNs) both for object recognition in 2D \cite{he2015delving,krizhevsky2012imagenet,he2016deep} and 3D with the aid of range data \cite{maturana2015voxnet,wohlhart2015learning}. AlexNet, introduced by Krizhevsky \textit{et al.} \cite{krizhevsky2012imagenet}, is one of the early CNN models that achieved state-of-the-art performance in object classification (15.3\% top-5 error rate in the ImageNet dataset). AlexNet has 5 convolutional layers for learning object features and a 3-layer fully connected network to classify the learned object features into one of the 1000 classes in the ImageNet dataset. To improve the results even further, more recent CNN models propose even deeper architectures. VGG-16 \cite{simonyan2014very} is one of the widely used architectures that has 16 convolutional layers (or 19 in VGG-19 architecture) for object representation. By increasing the depth of the network, VGG-19 could achieve the top-5 error rate of 6.8\% on the ImageNet dataset.

Another widely known CNN model is GoogleNet with a 22-layer architecture \cite{szegedy2015going}. To increase the depth of the network while minimizing network parameters, the creators of GoogleNet introduce the notion of inception layers. Inception layer is based on the idea that in the earlier convolutional layers (the ones closer to the input), activations tend to be concentrated in local regions. Given that, these regions can be covered by simply applying 1x1 convolutions, which also greatly reduces the depth of the filter banks. Furthermore, to capture more spatially sparse clusters, two successive 3x3 and 5x5 convolutions are also applied. The concatenation of these three convolutions in conjunction with a max pooling operation forms the input to the next layer of the network.

Going deeper in convolutional layers, however, comes with the cost of accuracy degradation due to the saturation of the network. This is addressed by residual networks (ResNets) \cite{he2016deep}. The basic structure of the network is a series of 3x3 convolutional layers. The input to each block of 2 convolutional layers, or 3 in deeper architectures, is added directly to the output of the block. Such direct link connections are repeated throughout the layers. Using this methodology, ResNet offers an architecture as deep as 152 layers which can achieve the error rate of 5.71\% (top-5 estimation) on the ImageNet dataset.

The CNN architectures discussed above are widely used in scene recognition applications as well. Zhou \textit{et al.} \cite{zhou2014learning} train weakly supervised CNN models for scene recognition using Places-205 dataset. The networks not only perform well in scene recognition but are also surprisingly good at identifying the key attributes in a given scene. This is achieved by analyzing the activations of the last pool layer before the fully connected layers. For example, the authors show that in an indoor scene of an art gallery, the activations are concentrated on the regions with paintings. Using a similar weakly supervised method, authors in \cite{shankar2015deep} explicitly attempt to learn scene attributes from a partially labeled data. Here, in the initial stage, the network generates a series of pseudo-labels by averaging the responses of early convolutional layers. These labels are then used to train the network to recognize scene attributes.

Moreover, weakly supervised training can be used in object classification. In \cite{oquab2014weakly}, Oquab \textit{et al.} exhibit the ability of neural nets to learn object classes from images without bounding box annotations. The authors propose a multi-scale training procedure in which for each epoch of training the scale of the input images is randomized (between 0.7 to 1.4 times the original dimensions). To tackle the problem of dimension mismatch between the input size and the fully connected layers, the authors transformed the network to a fully convolutional architecture by replacing the fully connected layers with equivalent convolutional layers. The classification is then achieved by a global pooling over the final score map. At the test time, the network is run on the input image at all scales and then the results are averaged.

\begin{figure}[!t]
\centering
\includegraphics[width=0.7\textwidth]{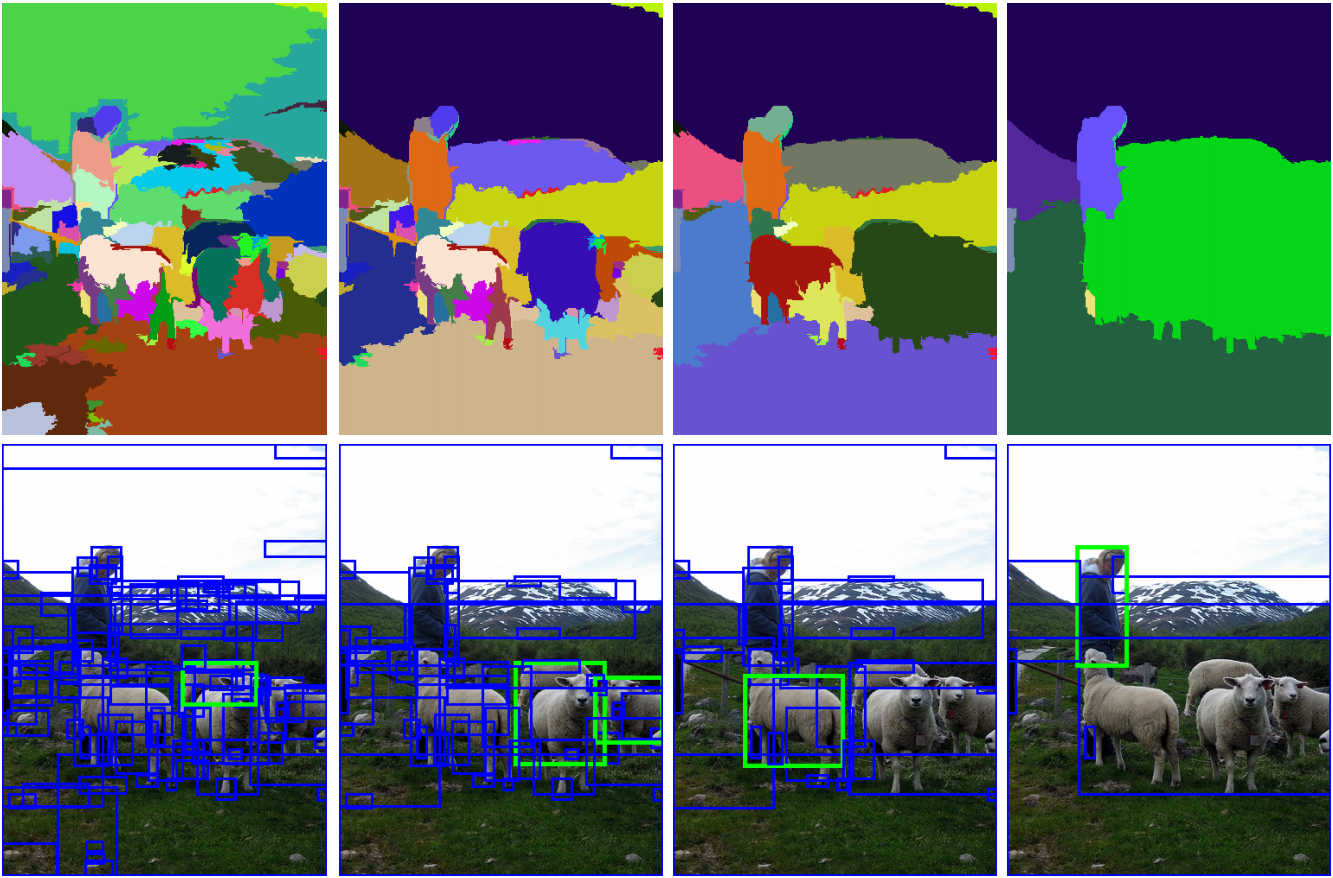}
\caption[Selective search for region proposal generation.]{Selective search for region proposal generation. Source: \cite{uijlings2013selective}.}
\label{fig:selective_search}
\end{figure}

A more challenging task in recognition is the detection and localization of objects in the scene. In the literature, there are numerous works attempting to solve this problem in a minimum amount of time \cite{erhan2014scalable,girshick2014rich,wei2014cnn,girshick2015fast,caicedo2015active,redmon2016you}. One of the well-known detection algorithms is Regions with CNN features (R-CNN) \cite{girshick2014rich} which unlike the methods that generate object specific region proposals (e.g. \cite{erhan2014scalable}) produces proposals based on the structure of the scene. R-CNN relies on the method called selective search \cite{uijlings2013selective} (see \figref{fig:selective_search}) to identify coherent regions of the scene in different spatial scales. Selective search starts by a small-scale superpixel segmentation to find regions with coherent properties. Then, in each subsequent step, it measures the similarity between the segmented regions and merges the ones that are similar. This process is repeated until a single region covers the entire image. Along with this process, at each step, proposals are generated around the identified coherent regions. Once the proposals are generated, R-CNN warps them to a fixed dimension and feeds them to a CNN (in this case AlexNet) model. The output of the CNN is a fixed feature vector that is used to identify the class of the objects in the region using a linear SVM model.

The successors of R-CNN, fast R-CNN \cite{girshick2015fast} and faster R-CNN \cite{ren2015faster}, combine a part of the region proposal module with the CNN network to speed up the process. In these methods, instead of feeding each proposal to the CNN, the entire image is first processed with the convolutional layers, and then the regions of interest (ROIs) are extracted from the convolutional feature map by a ROI pooling operation. The output of the ROI pooling layer is fed to a fully connected network with two branches: one for identifying the class of the object and the other for optimizing the positions of the bounding box. The faster R-CNN model goes one step further and generates the object proposals based on the output of the convolutional feature map using a what so called a Region Proposal Network (RPN). RPN uses a 3x3 sliding window technique to generate region proposals in the form of 256D feature vectors. These features are then fed to two fully connected networks, namely a box-regression layer and a box-classification layer. The former layer learns the position of the bounding boxes by calculating the intersection of the proposals with the ground truth information. The latter layer learns the objectness of the proposals, that is how likely the bounding box contains an object or a part of the background. In the end, the output of RPN is fed to ROI pooling layer of Fast R-CNN for final classification. 

To deal with objects of different scales, He \textit{et al.} \cite{he2014spatial} introduce spatial pooling technique in which the pooling operation is performed at different scales  (after convolutional feature map is generated), and then resulting features are concatenated to form a fixed feature vector that goes into the fully connected layers. To further speed up the proposal generation process, some algorithms such as You Only Look Once (YOLO) \cite{redmon2016you} treat the process as a regression problem. YOLO, in essence, is a FCN network. It takes as input the full resolution image, and  divides it into a SxS grid. Each cell predicts 5 bounding boxes with a confidence score indicating how likely it contains an object. Cells in turn are classified to identify which class of objects they belong to. The combination of the bounding box predictions and the predicted class of each cell forms the final detection and classification of objects in the scene.  Despite of the fact that YOLO achieves an efficient processing time of up to 45 fps (and up to 150 fps in fast version), it suffers from lower detection accuracy, inability to localize small objects and lack of generalization to unusual aspect ratios or configurations.

In more recent years, improved results have been achieved by exploiting contextual information in both detection and classification tasks. For instance, for detection purposes, in \cite{kantorov2016contextlocnet} for each generated ROI candidate, an outer region, representing the local context, is taken into account. To learn the proposals, a regression operation is performed on both addition and subtraction of ROIs and their corresponding contexts. The authors claim that this method helps generating a more discriminative representation to separate the object from its background and at the same time inferring the potential locations for the object.

Furthermore, context is shown to improve object classification. For example, Hu \textit{et al.} \cite{hu2016learning} introduce a technique for scene classification in which the relationships between the multiple objects in the scene are exploited to improve the overall classification results. In their formulation, there are 4 different levels of object annotations: coarse scene category (e.g. indoor, outdoor), scene attributes (e.g. sports field), fine-grained scene category (e.g. playground) and object-level annotation (e.g. bat, people). To capture inter-connections between different elements and their labels in the scene, a recurrent neural network (RNN) is trained. The output of this network is then used to optimize the activations of each of 4 categories prior to final classification.

A summary of algorithms discussed earlier can be found in Table \ref{table:gen_algs}.

\begin{table}[!hbtp]
\caption[A summary of generic object recognition algorithms.]{A summary of generic object recognition algorithms. Abbreviations: \textit{sensor type:} Cam = Camera, R = Range camera, L = LIDAR, K = Kinect, \textit{data type:} Gr = Greyscale, Col = Color image, D = Depth image, UB = Uniform Background, \textit{recognition:} L = Localization, C = Classification, \textit{loc. type:} 2D = 2D location, 3D = 3D Pose.}
\centering
\resizebox{\textwidth}{!}{
\begin{tabular}{|c|c|c|c|c|c|c|c|c|}
\hline
Model & Year & Features &Classification& Sensor Type &Data Type& Object Type & Recognition & Loc. Output\\ \hline \hline
CV3D \cite{chakravarty1982characteristic}&1982&Lines&-&Cam&Gr + UB&3D Objects&C&-\\ \hline
3DI \cite{oshima1983object}&1983&\makecell*{3D shapes}&-&R&D + UB&3D Objects&L,C&3D\\ \hline
2D3D \cite{lowe1987three}&1987&\makecell*{Lines}&-&Cam&Gr + UB&Razors&L,C&3D\\ \hline
AffIM \cite{lamdan1988object}&1988&\makecell*{2D models, lines}&-&Cam&Gr + UB&\makecell*{Tools}&L,C&3D\\ \hline
CI \cite{swain1991color}&1991&Color&-&Cam&Col + UB& Random Objects&L,C&2D\\ \hline
AOR \cite{wilkes1992active}&1992&\makecell*{Lines}&-&Cam&Gr + UB&Origami&C&-\\ \hline
IAVC \cite{dickinson1994active}&1994&\makecell*{Aspects}&-&Cam&Gr + UB&\makecell*{3D Objects}&L,C&3D\\ \hline
SVM3D \cite{pontil1998support}&1998&Intensitty&SVM&Cam&Col + UB&Random Objects&C&-\\ \hline
CBOR \cite{gevers1999color}&1999&Color&-&Cam&Col + UB&Random Objects&C&-\\ \hline
Spin-3D \cite{johnson1999using}&1999&3D meshes&-&R&D&Toys&L,C&3D\\ \hline
SIFT \cite{lowe1999object}&1999&SIFT&Kd-tree &Cam&Gr&Random Objects&L,C&3D\\ \hline
Shape-C \cite{belongie2001shape}&2001&Shape Context&K-NN &Cam&Gr + UB&MNIST, COIL-20&C&-\\ \hline
Boosted-C \cite{viola2001rapid}&2001&Haar&Cascade&Cam&Gr + UB&Faces&L,C&2D\\ \hline
Haar-ROD \cite{lienhart2002extended}&2002&Haar&Cascade&Cam&Gr + UB&Faces&L,C&2D\\ \hline
Context-Pl \cite{torralba2003context}&2003&\makecell*{Steerable\\ filters}&HMM&Cam&Gr&\makecell*{Places,Objects}&L,C&2D\\ \hline
MOD \cite{torralba2004sharing}&2004&Paches&Boosting&Cam&Gr&Random Objects&L,C&2D\\ \hline
STF \cite{shotton2008semantic}&2008&Color, STF &SVM&Cam&Col&\makecell*{MSRC-21,\\ VOC 2007}&L,C&2D\\ \hline
Part-Base \cite{felzenszwalb2010object}&2010&HOG&SVM &Cam&Col&VOC 2006-08&L,C&2D\\ \hline
WSL \cite{deselaers2012weakly}&2012&\makecell*{HOG, GIST,\\ SURF, Color}&CRF, SVM &Cam&Col&\makecell*{VOC 2006-07,\\Caltech 4}&L,C&2D\\ \hline
AlexNet \cite{krizhevsky2012imagenet}&2012&Conv&Neural Net&Cam&Col&ImageNet&C&-\\ \hline
origMSRC \cite{yao2012describing}&2012&HOG&SVM &Cam&Col&\makecell*{MSRC 21,\\ VOC 2010}&L,C&2D\\ \hline
SSHG \cite{uijlings2013selective}&2013&Color, SIFT &SVM&Cam&Col&VOC 2007&L,C&2D\\ \hline
Deep-Scene \cite{zhou2014learning}&2014&Conv&Neural Net&Cam&Col&\makecell*{ImageNet,\\Places-205}&C&-\\ \hline
Adapt-R \cite{bilen2014object}&2014&SIFT&SVM&Cam&Col&VOC 2007&L,C&2D\\ \hline
HMP-D \cite{bo2014learning}&2014&\makecell*{SPC, Color\\Depth}&SVM&K&Col + D&\makecell*{RGBD,Willow,2D3D}&L,C&2D\\ \hline
R-CNN \cite{girshick2014rich}&2014&Conv&Neural Net&Cam&Col&VOC 2007,10-12&L,C&2D\\ \hline
MultiBox \cite{erhan2014scalable}&2014&Conv&Neural Net&Cam&Col&ILSVRC 2012&L,C&2D\\ \hline
LRF-D \cite{gupta2014learning}&2014&\makecell*{Conv,Depth}&SVM&R&Col&NYUD2 &L,C&2D\\ \hline
SPP-net \cite{he2014spatial}&2014&Conv&Neural Net&Cam&Col&\makecell*{ILSVRC 2012,\\VOC 2007,\\Caltech101}&C&-\\ \hline
VGG \cite{simonyan2014very}&2014&Conv&Neural Net&Cam&Col&\makecell*{ILSVRC 2014,\\  VOC 2012}&L,C&2D\\ \hline
WSO-CNN \cite{oquab2014weakly}&2014&Conv&Neural Net&Cam&Col&VOC 2012&C&-\\ \hline
HCP \cite{wei2014cnn}&2014&Conv&Neural Net&Cam&Col&VOC 2007/12&L,C&-\\ \hline
AAR \cite{caicedo2015active}&2015&Conv&Neural Net&Cam&Col&VOC 2007&L,C&2D\\ \hline
CtxSVM-AMM \cite{chen2015contextualizing}&2015&\makecell*{SIFT, HOG,\\ LBP}&SVM&Cam&Col&\makecell*{VOC 2007/10,\\ SUN09}&L,C&2D\\ \hline
Fast R-CNN \cite{girshick2015fast}&2015&Conv&Neural Net&Cam&Col&VOC 2007/10/12&L,C&2D\\ \hline
PReLU \cite{he2015delving}&2015&Conv&Neural Net&Cam&Col&\makecell*{VOC 2007,\\ ILSVRC 2012}&L,C&2D\\ \hline
RCNN \cite{liang2015recurrent}&2015&Conv&Neural Net&Cam&Col&\makecell*{CIFAR-10, CIFAR-100, \\MNIST, SVHN}&C&2D\\ \hline
VoxNet \cite{maturana2015voxnet}&2015&Conv&Neural Net&L&D&NYUD2, ModelNet 40&C&2D\\ \hline
Faster R-CNN \cite{ren2015faster}&2015&Conv&Neural Net&Cam&Col&VOC 2007/12&L,C&2D\\ \hline
DEEP-CARVING \cite{shankar2015deep}&2015&Conv&Neural Net&Cam&Col& \makecell*{CAMIT – NSAD,\\ SUN attributes}&C&-\\ \hline
GoogleNet \cite{szegedy2015going}&2015&Conv&Neural Net&Cam&Col&ImageNet 2014&L,C&2D\\ \hline
3DP-CNN \cite{wohlhart2015learning}&2015&Conv&K-NN &K&Col + D&\makecell*{Syntehthic,Depth}&L,C&3D\\ \hline
ResNet \cite{he2016deep}&2016&Conv&Neural Net&Cam&Col&\makecell*{CIFAR-10,COCO\\ ILSVRC 2014,\\VOC2007/12}&L,C&2D\\ \hline
SINN \cite{hu2016learning}&2016&Conv&Neural Net&Cam&Col&\makecell*{AwA, NUS-WIDE,\\ SUN397}&C&-\\ \hline
ContextLocNet \cite{kantorov2016contextlocnet}&2016&Conv&Neural Net&Cam&Col&VOC 2007/12&L,C&2D\\ \hline
YOLO \cite{redmon2016you}&2016&Conv&Neural Net&Cam&Col&VOC 2007/12&L,C&2D\\ \hline

\end{tabular}  
}
\label{table:gen_algs}
\end{table}

\subsubsection{Algorithms for Traffic Scene Understanding}
The nature of algorithms for traffic scene understanding, both in terms of detection and classification, is very similar to the generic object recognition algorithms discussed earlier. As a result, here, we will only focus on the parts of the algorithms that deal with unique characteristics of objects in traffic scenes.

A number of algorithms try to make sense of traffic scenes by identifying and localizing multiple objects or structures either in  2D image plane \cite{gavrila1999real,papageorgiou2000trainable,ess2009segmentation} or 3D world coordinates \cite{wojek2013monocular,chen2016monocular} (see Table \ref{table:traffic_algs} for a short summary). For instance, in one of the early works \cite{gavrila1999real} a hierarchical representation of 2D models is used to detect humans and recognize signs. The hierarchical representation allows one to efficiently search through a large number of available models to find the best match with the candidate object in the scene. Using this approach, for example, sign models form a tree structure with branches corresponding to different color, shape, or type of the sign.

Recent approaches use more powerful techniques for traffic object detection such as CNNs. Chen \textit{et al.} \cite{chen2016monocular} introduce the Mono-3D algorithm in which the network estimates the 3D pose of the objects in the scene using only the 2D image plane information. To achieve this, the authors first use the prior knowledge of the road with respect to the image plane (the distance of the camera from the ground is known and the road is assumed to be orthogonal to the image plane). This knowledge is used to generate initial 3D region proposals for objects. Using these proposals in conjunction with pixel-wise semantic information allows the algorithm to select the regions that contain an object. Note that the dimension and orientation of the 3D proposal boxes are learned based on the data and are not recovered from the scene, i.e. for all three classes of objects (pedestrians, cyclists, and cars), the authors estimated three possible sizes and two possible orientations ([0,90] degrees) for 3D boxes.

\begin{table}[!hbtp]
\caption[Examples of traffic scene understanding algorithms.]{Examples of traffic scene understanding algorithms. Abbreviations: \textit{sensor type:} Cam = Camera, IMU = Inertial Measurement Unit, \textit{data type:} Col = Color, Gr = Greyscale, \textit{recognition:} L = Localization, C = Classification, \textit{loc. type:} 2D = 2D location, 3D = 3D Pose, Pix = Pixel-wise.}
\centering
\resizebox{\textwidth}{!}{
\begin{tabular}{|c|c|c|c|c|c|c|c|c|}
\hline
Model & Year & Features &Classification& Sensor Type & Data Type & Object Type & Recognition& Loc. Output\\
\hline \hline
RT-Smart \cite{gavrila1999real}&1999& 2D models&-&Cam&Gr&Pedestrians, Signs&L,C& 2D \\ \hline
TSOD \cite{papageorgiou2000trainable} & 2000& Haar& SVM & Cam & Gr & \makecell*{Pedestrians,Cars,\\Faces} & L,C & 2D \\ \hline
SUTSU \cite{ess2009segmentation} & 2009 & Color & Adaboost & Cam & Col & \makecell*{13 Class\\ e.g. Signs,Cars} & C & Pix \\ \hline
Tracklet \cite{wojek2013monocular} & 2013 & HOG &SVM & Cam + IMU & Col & Cars, Pedestrians & L,C & 3D\\ \hline
Mono-3D \cite{chen2016monocular} & 2016 & Conv & Neural Net & Cam & Col & \makecell*{Cars, Pedestrians,\\ Cyclists, Roads}&L,C & 3D \\ \hline
\end{tabular}
}
\label{table:traffic_algs}
\end{table}

\subsubsection{Pedestrian Detection}
Pedestrian detection, perhaps, is one of the most challenging tasks in traffic scene understanding. This is due to the fact that human body shape is often confused with various elements in the scene such as trees, poles, mailboxes, etc. Given the challenging nature of pedestrian detection, a large number of attempts have been made in the past decades trying to develop robust representations of human body that would be discriminative enough to separate them from other objects in the scene.

To deal with these ambiguities, intelligent transportation systems often rely on different types of sensors such as LIDAR \cite{fuerstenberg2002pedestrian,kidono2011pedestrian,broggi2009new} to segment pedestrians from the background, Ultrasonic or RADAR sensors \cite{beckwith1998passive} to detect pedestrians by analyzing motion information or infrared \cite{nanda2002probabilistic,bertozzi2003shape,bertozzi2005stereo} to recognize humans by detecting body heat. The models that only detect motion are generally susceptible to noise and  can easily be distracted by any kind of moving objects. That is why they are being used for monitoring places that are specifically dedicated to pedestrians (e.g. zebra crossings) \cite{beckwith1998passive}. The depth-based approaches that rely only on point clouds or depth information are vulnerable to occlusion or background clutter. These methods use stacks of planar laser scans by which they extract the shape of pedestrians \cite{kidono2011pedestrian}. So if the pedestrian is standing next to another object, the extracted shape can be distorted.

Among all non-optical sensors, infrared is proven to be very useful when lighting conditions are not favorable, e.g. at night. Infrared sensors identify the heat emitted by the human body and generate a dense greyscale representation of the environment. Naive algorithms simply threshold the captured heat map by a value close to expected human body temperature to identify pedestrians \cite{bertozzi2003shape}. Since heat source alone can be sensitive to noise, especially in hot weather,  some algorithms try to identify heat patterns similar to human body shape. For instance, Nanda and Davis \cite{nanda2002probabilistic} use a probabilistic template of the human body to match against the detected heat patterns. Bertozzi \textit{et al.} \cite{bertozzi2005stereo} use a stereo infrared camera to generate a depth map whereby the human is segmented from other heat sources in the scene.

Some early camera-based pedestrian detection models use techniques similar to those used by infrared-based models. For instance, in the cases where the camera position is fixed, motion is often used to detect human subjects. The detected motion pattern is then either identified as a pedestrian \cite{richards1995vision} or is compared with pre-learned 2D human body models to confirm the presence of the human in the scene \cite{rohr1993incremental}. Stereo techniques, on the other hand, generate a depth map to identify the potential regions where human might be \cite{wohler1998time,broggi2000shape}. For instance, Broggi \textit{et al.} \cite{broggi2000shape} use a similar technique but further refine the identified regions by discarding the ones with low symmetry. Then, the authors use a human head template and compare it to the final regions to identify the ones that correspond to a pedestrian. Of course, one major drawback of this technique is that it does not detect pedestrians from side-view.

A better detection rate can be achieved using more descriptive features including but not limited to Haar \cite{oren1997trainable}, color \cite{dollar2009integral}, and HOG \cite{dalal2006human,ouyang2012discriminative,yan2013robust}. Among these features, HOG,  originally designed for human detection \cite{dalal2005histograms} have been one of the most common techniques until recently. For detection purposes, these features are used in various forms. For instance, some algorithms use them in isolation \cite{yan2013robust} whereas others combine them with other types of features such as color \cite{dollar2009integral} or optical flow \cite{dalal2006human} to build more powerful image descriptors.

The most recent state-of-the-art techniques are based on convolutional neural networks \cite{tian2015deep,tian2015pedestrian,schlosser2016fusing,hu2017pushing}. The feature generation and classification of pedestrian detection algorithms are similar to those of the generic object recognition algorithms described earlier. However, CNNs designed for pedestrian detection pay special attention to challenges related to distinguishing pedestrians from outliers. For instance, to deal with the problem of occlusion in the scene, Tian \textit{et al.} \cite{tian2015deep} detect pedestrians based on their body parts. The authors construct a pool of pedestrian parts comprising 45 different types of human body parts under various forms of occlusion. These parts are used to train a series of CNNs which are used at the test time to detect humans. To reduce the number of false positives and to increase true positives, in \cite{tian2015pedestrian}, the authors explicitly train a classifier on objects that are most often confused with pedestrians. For instance, they identify hard negative samples (e.g. trees, poles) and explicitly learn them as separate classes in the network.  In addition, they improve the detection of the pedestrians by identifying relevant attributes such as gender, their angle with respect to the camera or even what they carry (e.g. handbag). A more recent approach also achieves a similar objective by combining pedestrian detection with semantic segmentation of the scene to explicitly identify the element that may be confused with pedestrians \cite{brazil2017illuminating}. A summary of pedestrian detection algorithms discussed in this section can be found in Table \ref{table:ped_algs}.

In the literature, there are a number of benchmark studies \cite{dollar2009pedestrian,dollar2012pedestrian,benenson2014ten} that give a good overview of recent algorithms and their performance on various datasets. In addition, there are a number of survey papers that provide a comprehensive overview of pedestrian detection algorithms, for example \cite{enzweiler2009monocular}, a survey of monocular camera-based pedestrian detection algorithms, and \cite{geronimo2010survey}, which summarizes models that use different types of sensors.

\begin{table}[!hbtp]
\caption[A summary of pedestrian detection algorithms.]{A summary of pedestrian detection algorithms. Abbreviations: \textit{sensor type:} Cam = Camera, R = RADAR, U = Ultrasonic, I = Infrared, St = Stereo, L = LIDAR, \textit{sensor pose:} FiP = Fixed Position, SiV = Side-View, StL = Street-Level, BeV = Bird's eye View, MuV = Multiple-Views, \textit{data type:} Col = Color, Gr = Greyscale, Vid = Video, Ht = Heat map, PC = Point Cloud, D  = Depth map, M = Movies, \textit{loc. type:} 2D = 2D location, 3D = 3D pose.}
\centering
\resizebox{\textwidth}{!}{
\begin{tabular}{|c|c|c|c|c|c|c|c|}
\hline
Model & Year & Features &Classification& Sensor Type & Sensor Pose & Data Type & Loc. Output\\
\hline \hline
IRP \cite{rohr1993incremental}&1993&\makecell*{Motion,\\ 2D model}&Matching &Cam&FiP +BeV +SiV&Vid+Gr&3D \\ \hline
VBI \cite{richards1995vision}&1995&Motion&-&Cam&FiP +BeV&Vid+Gr&2D\\ \hline
TSPD \cite{oren1997trainable}&1997&Haar &SVM&Cam&StL&Col&2D\\ \hline
PPD \cite{beckwith1998passive}&1998&Motion&-&R + U + I&FiP+BeV&Ht&-\\ \hline
UTA \cite{wohler1998time}&1998&Depth&Neural Net&StCam&StL&Vid+Gr&2D\\ \hline
ARGO \cite{broggi2000shape}&2000&\makecell*{Edges\\Head model}&-&StCam&StL&Gr&2D\\ \hline
PRUT \cite{fuerstenberg2002pedestrian}&2002&Depth&Parameter-based&L&StL&PC&2D\\ \hline
PTB \cite{nanda2002probabilistic}&2002&\makecell*{Intensity,\\ Template}&-&I&StL&Ht&2D\\ \hline
PDII \cite{bertozzi2003shape}&2003&Intensity&-&I&StL&Ht&2D\\ \hline
SVB \cite{bertozzi2005stereo}&2005&\makecell*{Intensity, Edges,\\ Depth}&-&SI&StL&Ht&2D\\ \hline
HOG \cite{dalal2006human}&2006&\makecell*{Optical flow,\\ HOG}&SVM&Cam&MuV&M&2D\\ \hline
ACF \cite{dollar2009integral}&2009&Color, HOG&AdaBoost&Cam&MuV&Caltech, INRIA&2D\\ \hline
PDAB \cite{broggi2009new}&2009&Haar&AdaBoost&Cam + L&StL&Gr&2D\\ \hline
PR-L \cite{kidono2011pedestrian}&2011&Depth&SVM&L&StL&PC&2D\\ \hline
DPM \cite{ouyang2012discriminative}&2013&HOG&SVM&Cam&StL&\makecell*{TUD-Brussels,\\ Caltech,INRIA}&2D\\ \hline
MT-DPM \cite{yan2013robust}&2013&HOG&SVM&Cam&StL&Caltech&2D\\ \hline
CCF \cite{yang2015convolutional}&2015&Conv&Decision forest&Cam&StL&Caltech&2D\\ \hline
Deep-Parts \cite{tian2015deep}&2015&Conv&Neural Net, SVM &Cam&StL&Caltech&2D\\ \hline
TA-CNN \cite{tian2015pedestrian}&2015&Conv&Neural Net&Cam&StL&Caltech, ETHZ&2D\\ \hline
LIDAR-CNN \cite{schlosser2016fusing}&2016&Conv&Neural Net&Cam + L&StL&KITTI&2D\\ \hline
RPN-BF \cite{zhang2016faster}&2016&Conv&Neural Net&Cam&StL&\makecell*{Caltech, KITTI,\\ INRIA, ETH}&2D\\ \hline
Fast−CFM \cite{hu2017pushing}&2017&Conv& Decision forest&Cam&StL&\makecell*{Caltech, KITTI,\\ INRIA}&2D\\ \hline
SDS-RCNN \cite{brazil2017illuminating}&2017&Conv&Neural Net&Cam&StL&Caltech, KITTI&2D\\ \hline
\end{tabular}
}
\label{table:ped_algs}
\end{table}

\subsubsection{Vehicle Recognition}

Vehicles are relatively easier to detect in traffic scenes. Given their rigid body property, early detection methods often rely on simple shape detectors, e.g. circular shapes for wheels \cite{radford1989vehicle} or rectangular shapes for vehicle bodies \cite{betke1996multiple,bertozzi1997real}. Vehicles also have symmetric appearance when they are seen from the front or the back, which is often the case with detecting vehicles on the road. Some authors take advantage of this property and identify symmetric patterns in the scene to detect vehicles \cite{bertozzi2000stereo,alessandretti2007vehicle}. For instance, in \cite{bertozzi2000stereo}, a stereo camera is used to segment the road edges. Then, a vertical line moves across the image, and at each step the regions on the right and left sides of the line are compared to measure their symmetry. The areas that yield the highest symmetry measure (above a certain threshold) are identified as vehicles.

Similar to pedestrian detection, if the position and view of the camera are fixed, motion can be used to detect potential regions that may correspond to vehicles \cite{bullock1993neural,gupte2002detection}. If the type of the vehicle (e.g. truck or sedan) is of interest, the nominated regions can be further processed, for example by measuring the aspect ratio of the bounding boxes to classify the vehicles \cite{gupte2002detection}.

Given the lack of robustness of simple features such as edges, corners or motion, more complex 2D features are widely used in vehicle detection. Some popular examples are Haar features \cite{papageorgiou1999trainable}, Gabor filters \cite{sun2005road}, HOG features \cite{niknejad2011vehicle,sivaraman2013looking} and color \cite{lopez2008nighttime}. Compared to generic detection algorithms, vehicle detection methods use these features differently. For instance, Sivaraman and Trivedi \cite{sivaraman2013looking}, instead of generating HOG descriptors for vehicles as a whole, learn the back and front parts of the vehicles separately. Then, at the test time, once each individual component is found, they fit a bounding box that covers both parts and consequently localize the entire vehicle. The authors argue that using this technique helps to identify vehicles despite occlusion. In another work \cite{niknejad2011vehicle} a part-based learning approach using HOG features is employed to detect vehicles at night. To improve the detection results, the authors modify the contributing weight of each part differently. For example, they increase the weight of back lights which are easily observable at night. Lopez \textit{et al.} \cite{lopez2008nighttime} use the intensity and the color of vehicle lights at nighttime to determine the distance to the vehicle and its direction of motion

Today, the field of vehicle detection algorithms is dominated by CNNs in such applications as satellite view traffic control \cite{cao2016robust} or autonomous driving \cite{chabot2017deep}. The techniques used here are fairly similar to those used in generic algorithms. For instance, in \cite{yang2016exploit} a multi-scale ROI pooling algorithm is used to detect vehicles in various scales. The authors use a VGG-16 architecture and connect three ROI pooling layers after convolutional layers 3, 4 and 5. The output of each ROI pooling layer is connected to a separate fully connected network to perform classification at different scales. The motivation for this architecture is that different scale object features emerge in different levels of the neural network, i.e. earlier layers are generally better at detecting smaller objects (due to their larger field of view) and the later ones are better at finding the large objects.

In autonomous driving,  different sensor modalities can be used to improve the performance of detection. For instance, Lange \textit{et al.} \cite{lange2016online}, to save processing time, use LIDAR readings to discard the ROI candidates that belong to cars far away from the vehicle (and focus only on up to 10 closest vehicles). Chen \textit{et al.} \cite{chen2017multi} combine LIDAR readings from two different points of view (bird's eye view and street level) with RGB images to improve the detection results. Here, the final classification takes place using the output of 3 ROI pooling sources from each input sensor reading. In this algorithm, the variations in viewpoints can disambiguate challenging situations caused by occlusion or illumination conditions.

The algorithms discussed in this section are summarized in Table \ref{table:car_algs}. In addition, there are a number of survey papers that give a good overview of early \cite{sun2006road} and more recent \cite{sivaraman2013looking} vehicle detection algorithms.

\begin{table}[!hbtp]
\caption[A summary of vehicle recognition algorithms.]{A summary of vehicle recognition algorithms. Abbreviations: \textit{sensor type:} Cam = Camera, R = RADAR, I = Infrared, St = Stereo, L = LIDAR, \textit{sensor pose:} FiP = Fixed Position, SiV = Side-View, StL = Street-Level, BeV = Bird's eye View, \textit{data type:} Col = Color, Gr = Greyscale, Ht = Heat map, PC = Point Cloud, \textit{loc. type:} 2D = 2D location, 3D = 3D Pose, P = Presence only.}
\centering
\resizebox{\textwidth}{!}{
\begin{tabular}{|c|c|c|c|c|c|c|c|}
\hline
Model & Year & Features &Classification& Sensor Type & Sensor Pose & Data Type & Loc. Output\\
\hline \hline
OWS-HT \cite{radford1989vehicle}&1989&Lines&-&Cam&SiV&Gr&2D\\ \hline
WADS \cite{bullock1993neural}&1993&Intensity&Neural Net&Cam&FiP + BeV&Gr&P\\ \hline
MVD \cite{betke1996multiple}&1996&\makecell*{Lines, Corners,\\ Template}&-&Cam&StL&Gr&2D\\ \hline
PAPRICA \cite{bertozzi1997real}&1997&Lines, Corners&-&Cam&StL&Gr&2D\\ \hline
Haar-CD \cite{papageorgiou1999trainable}&1999&Haar&SVM&Cam&StL&Col&-\\ \hline
SVB \cite{bertozzi2000stereo}&2000&Lines, Corners&-&StCam&StL&Gr&2D\\ \hline
DCV \cite{gupte2002detection}&2002&Motion, Lines&-&Cam&FiP + BeV&Gr&2D\\ \hline
EGFO \cite{sun2005road}&2005&Gabor&SVM&Cam&StL&Gr&2D\\ \hline
VGRD \cite{alessandretti2007vehicle}&2007&Lines&-&Cam + R&StL&Gr&2D\\ \hline
NVD-IHC \cite{lopez2008nighttime}&2008&Color&AdaBoost&Cam&StL&Col&2D\\ \hline
NUAD \cite{niknejad2011vehicle}&2011&HOG&AdaBoost, SVM&I&StL&Ht&2D\\ \hline
VDIP \cite{sivaraman2013vehicle}&2013&HOG&AdaBoost, SVM&Cam&StL&Gr&2D\\ \hline
On-DNN \cite{cao2016robust}&2016&Conv&SVM&Cam&Sat&Col&2D\\ \hline
LID-CNN \cite{lange2016online}&2016&Conv&Neural Net&Cam + L + R&StL&Gr + PC&3D\\ \hline
SDP \cite{yang2016exploit}&2016&Conv&\makecell*{Neural Net,\\ AdaBoost}&Cam&StL&\makecell*{KITTI,\\ VOC 2007,\\ Inner-city}&2D\\ \hline
Deep-MANTA \cite{chabot2017deep}&2017&Conv&Neural Net&Cam&StL&KITTI&3D\\ \hline
Multi-3D \cite{chen2017multi}&2017&Conv&Neural Net&Cam + L&StL&KITTI&3D\\ \hline
\end{tabular}
}
\label{table:car_algs}
\end{table}

\subsubsection{Traffic Sign Recognition}
Color and shape are perhaps the two most distinctive features that separate signs from the background and at the same time, define their meaning (e.g. regulatory, informative or warning). Relying on these features alone, signs can be segmented from the background \cite{priese1993traffic,kehtarnavaz1995traffic,de1997road,arlicot2009circular} and classified \cite{seo2012recognizing,balali2015multi}. The algorithms that use color for sign detection, rely on a similar approach in which the image color values are thresholded by color values of the sought sign. For this purpose, different color spaces are investigated such as YIQ \cite{kehtarnavaz1995traffic}, HSI \cite{de1997road,de2003traffic} and HSV \cite{arlicot2009circular}. The classification methods are similar to those widely used in object recognition and may include neural nets \cite{kehtarnavaz1995traffic,de1997road,de2003traffic}, Parzen window \cite{paclik2000road} and SVM \cite{balali2015multi}.

Besides shape and color, the position of traffic signs can be used to improve localization. For instance, Zhu \textit{et al.} \cite{zhu2016traffic} use AlexNet architecture to detect and classify traffic signs in the scene. Here, instead of generating object proposals on the entire input image, the position of the signs, which are often located on the both sides of the roads, are used as a prior knowledge to generate proposals for classification.

A summary of papers reviewed in this section can be found in Table \ref{table:sign_algs}. In addition, a comprehensive survey of sign recognition algorithms can be found in \cite{mogelmose2012vision}.

\begin{table}[!hbtp]
\caption[A summary of traffic sign recognition algorithms.]{A summary of traffic sign recognition algorithms. Abbreviations: \textit{sensor type:} Cam = Camera \textit{sensor pose:} StL = Street-Level, MuV = Multi View, \textit{data type:} Col = Color, Gr = Greyscale, \textit{class type:} All = All types of signs, Cir = Circular signs, Tri = Triangular signs, Work = Work zone signs, \textit{loc. type:} 2D = 2D location, 3D = 3D Pose, P = Presence only.}
\centering
\resizebox{\textwidth}{!}{
\begin{tabular}{|c|c|c|c|c|c|c|c|c|}
\hline
Model & Year & Features &Classification& Sensor Type & Sensor Pose & Data Type & Class Type & Loc. Output\\
\hline \hline
ART2 \cite{priese1993traffic}&1994&Color, Shape&-&Cam&StL&Col&Cir, Tri&2D\\ \hline
NOS \cite{kehtarnavaz1995traffic}&1995&Color, Edges&Neural Net&Cam&MuV&Col&All&-\\ \hline
RCE \cite{de1997road}&1997&Color&Neural Net&Cam&StL&Col&Cir, Tri&2D\\ \hline
HSFM \cite{paclik2000road}&2000&Intensity&Parzen Window&Cam&StL&Gr&Cir&-\\ \hline
TSRA \cite{de2003traffic}&2003&Color, Shape&Neural Net&Cam&StL&Col&Cir, Tri&2D\\ \hline
CRSE \cite{arlicot2009circular}&2009&Color, Shape&-&Cam&MuV&Col&Cir&2D\\ \hline
RTSC \cite{seo2012recognizing}&2012&Color&SVM&Cam&MuV&Col&Work&2D\\ \hline
GSV \cite{balali2015multi}&2015&Color, HOG&SVM&Cam&StL&Col&All&2D\\ \hline
FCN \cite{zhu2016traffic}&2016&Conv&Neural Net&Cam&StL&Col&STSD&2D\\ \hline
\end{tabular}
}
\label{table:sign_algs}
\end{table}

\subsubsection{Road Detection}

Road detection algorithms are useful in two ways: they help us to understand the structure of the streets, and at the same time can improve the localization of objects such as cars, pedestrians or signs (e.g. as in \cite{chen2016monocular}). The early works on road detection mainly use edge features to identify the boundaries of the road. The detected boundaries are often compared with a pre-learned model to estimate the structure of the road \cite{crisman1991unscarf,broggi1995vision,he2004color,dahlkamp2006self}. Depending on the type of the road, edge segmentation can be achieved by lane markings (in structured roads) \cite{broggi1995vision,he2004color} or the color of the surface (in unstructured roads) \cite{crisman1991unscarf,dahlkamp2006self}. Some algorithms such as the one used in Stanley at DARPA 2005 \cite{dahlkamp2006self} use LIDAR readings to localize the road surface prior to road boundary detection.

In contrast to traditional learning techniques, neural networks learn an explicit model of the road surfaces through successive generation and classification of convolutional features \cite{mohan2014deep,laddha2016map,mendes2016exploiting,oliveira2016efficient}. However, generating enough annotated sample data for training is a daunting task because it requires pixel-wise ground truth binary masks. To deal with this problem, Laddha \textit{et al.} \cite{laddha2016map} propose a technique that automatically generates ground truth annotations based on the information of the vehicle position, detailed map information (including the position of rigid objects) and the camera's calibration parameters. Based on this knowledge, a 3D scene around the vehicle is constructed and is used for identifying a rough estimate of the road surface. The result is further refined using the color information in the scene by forcing the road segments to contain only colors that are coherent with its characteristic.

Another important factor in road detection, especially in autonomous driving applications, is the speed of processing. A number of approaches are proposed to reduce the computational load, such as classifying patches instead of pixels \cite{mendes2016exploiting} or using smaller convolutional filter sizes and employing fully convolutional architectures \cite{oliveira2016efficient}.

Table \ref{table:road_algs} gives an overview of algorithms discussed in this section. A more in depth review of road detection algorithms can be found in \cite{hillel2014recent}.

\begin{table}[!hbtp]
\caption[A summary of road detection algorithms.]{A summary of road detection algorithms. Abbreviations: \textit{sensor type:} Cam = Camera, L = LIDAR, \textit{data type:} Col = Color, Gr = Greyscale, PC = Point Cloud, \textit{loc. type:} BD = Boundaries, SF = Surface.}
\centering
\resizebox{\textwidth}{!}{
\begin{tabular}{|c|c|c|c|c|c|c|c|}
\hline
Model & Year & Road Type &Features &Classification& Sensor Type & Data Type & Loc. Output\\
\hline \hline
UNSCARF \cite{crisman1991unscarf}&1991&Unstructured roads&\makecell*{Color, Edge,\\ Road model}&-&Cam&Col&BD\\ \hline
VBRD \cite{broggi1995vision}&1995&Streets&Edge&-&Cam&Gr&BD\\ \hline
CBRD \cite{he2004color}&2004&Streets&Color, Edge&-&Cam&Col&BD + SF\\ \hline
Stan \cite{dahlkamp2006self}&2006&Unstructured roads&Color, Edge&K-means&Cam + L&Col + PC&BD\\ \hline
DNN-SP \cite{mohan2014deep}&2014&Streets&Conv&Neural Net&Cam&\makecell*{Stanford Background,\\ SIFT Flow,\\ CamVid, KITTI}&SF\\ \hline
MSRD \cite{laddha2016map}&2016&Streets&Conv&Neural Net&Cam&KITTI&SF\\ \hline
NiN \cite{mendes2016exploiting}&2016&Streets&Conv&Neural Net&Cam&KITTI&SF\\ \hline
Deep-MRS \cite{oliveira2016efficient}&2016&Streets&Conv&Neural Net&Cam&KITTI&SF\\ \hline
\end{tabular}
}
\label{table:road_algs}
\end{table}

\subsection{What the Pedestrian is Doing: Pose Estimation and Activity Recognition}
In this section, we review two topics: pose estimation and activity recognition. In the context of pedestrian behavior understanding, pose plays an important role. On its own pose may imply the state of the pedestrian. For example, head pose or body posture may indicate that the pedestrian is intending to cross. In some applications, changes in pose are also used to understand one's activity, i.e. it serves as a prerequisite to activity recognition. Activity recognition is also important for obvious reasons. For instance, it helps identifying someone's walking direction towards the street or handwave to send a signal.

We start the discussion by listing some datasets publicly available for studying pose estimation and activity recognition. Then, for each topic, we discuss some of the known algorithms. As before, we try to cover a broad range of methods used in different applications, with a particular focus on the ones that can be potentially used in the context of pedestrian behavior understanding.
\subsubsection{Datasets}

\subsubsection*{Pose Datasets}
Pose estimation datasets are collected in different ways. Some are gathered for sports scene analysis \cite{Johnson10,kazemi2013multi} while the others are catered to a wider range of applications \cite{andriluka14cvpr,zhang2014facial}. The data type and the availability of ground truth annotations varies from one dataset to another. For instance, datasets such as Buffy \cite{Ferrari08} and Human Pose Evaluator Dataset (HPED) \cite{Jammalamadaka12a} are collected from TV shows and movies and only contain upper body pose information. The ones for sport or general scene understanding often come from various sport broadcasts videos \cite{kazemi2013multi}, videos collected from the web \cite{Charles13} or generated by the researchers for special purposes \cite{gasparrini2014depth}.

The ground truth annotations that come with pose datasets are often in the form of joint locations and their connections \cite{Ferrari08,modec13}. A few datasets include depth information and 3D joint positions \cite{kazemi2013multi} or body parts tags \cite{Shafaei16}. Table \ref{table:pose_data} shows some of the most common datasets for pose estimation.

\begin{table}[!hbtp]
\caption[Pose estimation datasets.]{Pose estimation datasets. Abbreviations: \textit{categories:} P = Pose, A = Activity, \textit{action Type:} M = Movies, Sp = Sport, B = Basics (e.g. walking, sitting), \textit{parts:} UT = Upper Torso, Full = Full body, \textit{camera Pose:} F = Front view, Mul = Multi view, Sky = Sky view, CU = Close Up, \textit{data Type:} Img = Image, Col = Color, D = Depth, Vid = Video, Syn = Synthetic, St = Stereo, \textit{ground Truth:} J = Joints, AL = Activity Label, SM = Stickmen, PL = Pose Label, FP = Facial Points, BP = Body Pose}
\centering
\resizebox{\textwidth}{!}{
\begin{tabular}{|c|c|c|c|c|c|c|c|c|c|}
\hline
Dataset&Year&Cat.&Act. type&Parts&No. Class&Cam. Pose&Data Type&No. Frames&GT \\
\hline
\hline
Buffy \cite{Ferrari08}&2008&P&M&UT&-&F&Img+Col&748&J\\ \hline
Buffy-Pose \cite{ferrari20092d}&2009&P&M&UT&3&F&Img + Col&245&BB,PL\\ \hline
LSP \cite{Johnson10}&2010&P&Sp&Full&-&F&Img+Col&2K&J\\ \hline
VideoPose \cite{sapp2011parsing}&2011&P&M&UT&-&Mul&Vid+Col&44&J \\ \hline
CAD-60 \cite{sung2012unstructured}&2012&P,A&B&Full&12&F&D+Vid+Col&60&J,AL\\ \hline
KTHI \cite{kazemi2012using}&2012&P&Sp&Full&-&Mul&Img+Col&771&J\\ \hline
HPED \cite{Jammalamadaka12a}&2012&P&M&UT&-&F&Vid+Col&6.5K&SM\\ \hline
CAD-120 \cite{koppula2013learning}&2013&P,A&B&Full&20&F&D+Vid+Col&120&J,AL\\ \hline
FashionPose \cite{dantone2013human}&2013&P&B&Full&-&F&Img+Col&7.5K&J\\ \hline
KTHII \cite{kazemi2013multi}&2013&P&Sp&Full&-&Mul&Vid+Img+Col&800+5.9K&3DJ\\ \hline
VGG \cite{Charles13}&2013&P&Mix&UT&-&F&D+Vid+Col&172&J\\ \hline
FLIC \cite{modec13}&2013&P&M&UT&-&F&Img+Col&5K&J\\ \hline
APE \cite{yu2013unconstrained}&2013&P,A&B&Full&7&F&D+Vid+Col&245&3DJ,AL\\ \hline
ChaLearn \cite{escalera2013multi}&2013&P,A&B&Full&20&F&D+Vid+Col&23Hr&J,AL\\ \hline
PennAction \cite{zhang2013actemes}&2013&P,A&Mix&Full&15&F&Vid+Col&2.3K&J,AL\\ \hline
Human3.6M \cite{h36m_pami}&2014&P,A&B&Full&17&Mul&D+Vid+Col&3.6M&3DJ,AL\\ \hline
PARSE \cite{Antol2014}&2014&P,A&Sp&Full&14&F&Syn+Img+Col&1.5K+305&J,AL\\ \hline
TST-Fallv1 \cite{gasparrini2014depth}&2014&P&B&Full&2&Sky&D+Vid&20&PL\\ \hline
FLD \cite{zhang2014facial}&2014&P&Mix&Face&-&CU&Img+Col&33K&FP\\ \hline
MHPE \cite{belagiannis20143d}&2014&p&B&Full&-&Mul&Vid+Col&2&J\\ \hline
PiW \cite{Cherian14}&2014&P&M&UT&-&F&Vid+Col&30&J\\ \hline
MPII \cite{andriluka14cvpr}&2014&P,A&Mix&Full&491&F&Img+Col&40K&J,AL\\ \hline
TST-TUG \cite{cippitelli2015time}&2015&P&B&Full&-&F&D+Vid+Col&20&J\\ \hline
HandNet \cite{WetzlerBMVC15}&2015&P&B&Hand&-&F&D+Vid&100K&J\\ \hline
SHPED \cite{lopezquintero2015mvap}&2015&P&Mix&UT&-&F&St+Img+Col&630K&SM\\ \hline
VI-3DHP \cite{haque2016viewpoint}&2016&P,A&B&Full&15&Mul&D+Img&100K&J,AL\\ \hline
UBC3V \cite{Shafaei16}&2016&P&B&Full&-&F&D&210K&BP\\ \hline
TST-Fallv2 \cite{gasparrini2016proposal}&2016&P,A&B&Full&264&F&D+Vid&264&J,AL\\ \hline
\end{tabular}
}
\label{table:pose_data}
\end{table}

\subsubsection*{Activity Datasets}
Activity recognition datasets often comprise temporal sequences and, similarly to pose datasets, are extracted from different sources including sport videos \cite{rodriguez2008action}, movies \cite{marszalek09} or are made for a particular application \cite{schuldt2004recognizing}. The ground truth annotations of these datasets are often in the form of activity labels with \cite{haque2016viewpoint} or without temporal correspondence \cite{KarpathyCVPR14}. In addition, some datasets have explicit pose information as joint positions \cite{sung2012unstructured,yu2013unconstrained}, which makes them suitable for both pose estimation and activity recognition. Table \ref{table:activity_data} lists some common datasets for activity recognition. For a comprehensive list of activity recognition datasets including non-vision-based ones refer to \cite{chaquet2013survey}.

\begin{table}[!hbtp]
\caption[Activity recognition datasets.]{Activity recognition datasets. Abbreviations: \textit{categories:} P = Pose, A = Activity, \textit{action Type:} M = Movies, Sp = Sport, B = Basics (e.g. walking, sitting), C = Cooking, Int = interaction, G = Gait, Fall = Fall detection, Grp = Group \textit{parts:} H = Hand, UT = Upper Torso, Full = Full body, \textit{camera Pose:} F = Front view, Mul = Multi view, Sky = Sky view, CU = Close Up, FPer = First Person, BeV = Bird's eye View, \textit{data Type:} Img = Image, Col = Color, D = Depth, Vid = Video, Syn = Synthetic, St = Stereo, Gr = Grey, \textit{ground Truth:} J = Joints, AL = Activity Label, PL = Pose Label, FP = Facial Points, BP = Body Pose, BB = Bounding Box, Dir = Direction, PB = Pixel-wise Binary, TL = Temporal Localization.}
\centering
\resizebox{\textwidth}{!}{
\begin{tabular}{|c|c|c|c|c|c|c|c|c|c|}
\hline
Dataset&Year&Cat.&Act. type&Parts&No. Class&Cam. Pose&Data Type&No. Frames&GT \\
\hline
\hline
CASIA Giat \cite{wang2003silhouette}&2001&A&G&Full&3&F&Vid+Col&12&Dir\\ \hline
KTH \cite{schuldt2004recognizing}&2004&A&B&Full&6&F&Vid+Gr&2.3K+&AL\\ \hline
ASTS \cite{ActionsAsSpaceTimeShapes_iccv05}&2005&A&B&Full&10&F&Vid+Col&90&AL\\ \hline
IXMAS \cite{weinland2006free}&2006&A&B&Mix&13&Mul&Vid+Col&36&AL\\ \hline
Cam-HGD \cite{kim2007tensor}&2007&A&B&H&9&F&Vid+Col&900&PL\\ \hline
CASIA-Act \cite{wang2007human}&2007&A&B,Int&Full&8,7&Mul&Vid+Col&1446&BB,AL\\ \hline
UCF \cite{rodriguez2008action}&2008&A&Sp&Full&10&F&ViD+Col&150&AL,BB\\ \hline
UCF-Crowd \cite{ali2008floor}&2008&A&Grp&Full&2&BeV&Vid+Col&3&BB,Dir\\ \hline
LHA \cite{laptev:08}&2008&A&M&Mix&8&Mul&Vid+Col&32&AL\\ \hline
MAR-ML \cite{tran2008human}&2008&A&Sp&Full&14&BeV&Vid+Col&541&BB,AL\\ \hline
UCF-Aerial \cite{UCF-Aerial}&2009&A&B&Full&9&Sky&Vid+Col&7&BB,AL\\ \hline
MSR \cite{MSRData}&2009&A&B&Full&20&F&D+Vid+Col&20&AL\\ \hline
Col-AD \cite{choi2009they}&2009&A&Grp&Full&5&F&Vid+Col&44&BB,AL,Dir\\ \hline
PETS-Flow \cite{PETS2009}&2009&A&Grp&Full&6&BeV&Vid+Col&9&AL\\ \hline
Buffy-Pose \cite{ferrari20092d}&2009&P&M&UT&3&F&Img+Col&245&BB,PL\\ \hline
Holly2 \cite{marszalek09}&2009&A&M&Mix&12&Mul&Vid+Col&3.6K+&AL\\ \hline
UCF11 \cite{liu2009recognizing}&2009&A&Sp&Mix&11&Mul&Vid+Col&1.1K+&AL\\ \hline
OSD \cite{niebles2010modeling}&2010&A&Sp&Mix&16&Mul&Vid+Col&800&AL\\ \hline
MCFA \cite{auvinet2010multiple}&2010&A&Fall&Full&24&Mul&Vid+Col&24&0\\ \hline
BEHAVE \cite{blunsden2010behave}&2010&A&Int&Full&10&BeV&Vid+Col&4&BB,AL\\ \hline
TV-HID \cite{patron2010high}&2010&A&M,Int&Mix&4&Mul&Vid+Col&300&AL\\ \hline
SDHA \cite{UT-Interaction-Data}&2010&A&Int&Full&6&BeV&Vid+Col&20&AL\\ \hline
SDHA-Air \cite{UT-Tower-Data}&2010&A&B&Full&9&BeV&Vid+Col&108&BB,AL\\ \hline
Videoweb \cite{denina2011videoweb}&2010&A&Int&Mix&9&Mul&Vid+Col&2.5Hr&AL\\ \hline
Willow-Act \cite{delaitre2010recognizing}&2010&A&B&Mix&7&Mul&Img+Col&968&AL\\ \hline
MUHAVI \cite{singh2010muhavi}&2010&A&B&Full&17&Mul&Vid+Col&17&AL,PB\\ \hline
VIRAT \cite{oh2011large}&2011&A&B&Full&12&BeV&Vid+Col&8,5Hr&BB,AL\\ \hline
UCF-ARG \cite{UCF-Arg}&2011&A&B&Mix&10&Mul&Vid+Col&480&BB,AL\\ \hline
HMDB \cite{Kuehne11}&2011&A&Mix&Mix&51&Mul&Vid+Col&6.8K+&AL\\ \hline
UCF50 \cite{reddy2013recognizing}&2012&A&Mix&Mix&50&Mul&Vid+Col&6.6K+&AL\\ \hline
CAD-60 \cite{sung2012unstructured}&2012&P,A&B&Full&12&F&D+Vid+Col&60&J,AL\\ \hline
ChaLearn \cite{escalera2013multi}&2013&P,A&B&Full&20&F&D+Vid+Col&23Hr&J,AL\\ \hline
CAD-120 \cite{koppula2013learning}&2013&P,A&B&Full&20&F&D+Vid+Col&120&J,AL\\ \hline
JPL-FPI \cite{ryoo2013first}&2013&A&Int&Full&7&FP&Vid+Col&57&AL\\ \hline
UCF101 \cite{soomro2012ucf101}&2013&A&Mix&Mix&101&Mul&Vid+Col&13K+&AL\\ \hline
APE \cite{yu2013unconstrained}&2013&P,A&B&Full&7&F&D+Vid+Col&245&3DJ,AL\\ \hline
ChaLearn \cite{escalera2013multi}&2013&P,A&B&Full&20&F&D+Vid+Col&23Hr&J,AL\\ \hline
MPII \cite{andriluka14cvpr}&2014&P,A&Mix&Full&491&F&Img+Col&40K&J,AL\\ \hline
VAD \cite{waltner14a}&2014&A&Sp&Full&7&F&Vid+Col&6&AL\\ \hline
Sports-1M \cite{KarpathyCVPR14}&2014&A&Mix&Mix&487&Mul&Vid+Col&1.1M&AL\\ \hline
PARSE \cite{Antol2014}&2014&P,A&Sp&Full&14&F&Syn+Img+Col&1.5K+,305&J,AL\\ \hline
Human3.6M \cite{h36m_pami}&2014&P,A&B&Full&17&Mul&D+Vid+Col&3.6M&3DJ,AL\\ \hline
Crepe \cite{lee2015stare}&2015&A&C&Full&9&F&Vid+Col&6.5K&AL,BB\\ \hline
ActivityNet \cite{caba2015activitynet}&2015&A&B&Mix&203&Mul&Vid+Col&27K&Al,TL\\ \hline
TST-Fallv2 \cite{gasparrini2016proposal}&2016&P,A&B&Full&264&F&D+Vid&264&J,AL\\ \hline
VI-3DHP \cite{haque2016viewpoint}&2016&P,A&B&Full&15&Mul&D+Img&100K&J,AL\\ \hline
\end{tabular}
}
\label{table:activity_data}
\end{table}

\subsubsection{Pose Estimation}

Pose estimation algorithms can be divided into two main groups: exemplar-based and part-based models \cite{wang2011learning}. The former approaches try to identify the pose as a whole whereas part-based models use local body part appearances and the connections between them to estimate the overall pose.

Early pose estimation methods are predominantly exemplar-based and rely on body templates to estimate the pose. 
Templates can be created using different techniques such as simple silhouette models of human body \cite{ohya1994human}, real image samples pre-processed by applying some form of filters (e.g Gabor filters) \cite{mckenna1998real} or 3D models of human body \cite{ueda2003hand}.

More recent algorithms are learning-based and estimate body poses by learning from examples. For instance, Ng and Gong \cite{ng2002composite} use a large dataset of greyscale head pose images, each corresponding to a specific head orientation. These images are then normalized and used to train an SVM model to learn the correspondences between the 2D images and 3D head poses. Likewise, Agrawal and Triggs \cite{agarwal20043d} use a regression technique to learn the full 3D human body poses from a series of image silhouettes. The input to regression model is a 100D feature vector generated from the image silhouettes and the output is a 55D vector that estimates 3 angles for each of 18 body joints.

The exemplar-based models, however, suffer from a major drawback, that is they require a good match between the templates and proposals. This is problematic because it is often the case that a complete model of the body (or its component) is not retrievable from the image due to occlusion or background clutter. More importantly, even when a complete model of the body is available, if the ratios or the poses of the body components do not match entirely to those of the training images (e.g. legs are similar to some training images, but the arms are similar to the others), the algorithm will fail to estimate the pose.

Part-based models, as explained earlier, consist of two stages: learning human body part appearances, and determining the relationship between them. The body parts can be represented using simple edge maps and color distributions \cite{ramanan2007learning} or more complex descriptors such as HOG \cite{johnson2010pose}.

There are two ways to learn the relationships between body parts- explicit or implicit. The explicit techniques generally use tree-structured graphical models, such as CRF \cite{ramanan2007learning,wang2011learning}, to learn the interdependencies between the connected parts. Here, the learning is in the form of a spatial prior that describes the relative arrangement of the parts both in terms of the location and orientation of the parts. Implicit techniques, on the other hand, learn the relationships between the parts directly from the data. Regression techniques such as SVM \cite{johnson2010pose} or neural nets \cite{ouyang2014multi} are examples of such methods.

Part-based models also have a disadvantage that they only learn the relationships between parts locally and lack a global knowledge of how the overall human body, in a given pose, should look like. There are a number of solutions proposed to overcome this problem. Wang \textit{et al.} \cite{wang2011learning} introduce a hierarchical part representation, which, in addition to 10 common body parts (e.g hands, torso, head), uses 10 intermediate part representations formed by combining basic human parts together, such as torso with right arm and hand, left arm with the left hand, all the way up to full human body. The appearance of every part representation is learned and used at the test time to improve the result. The authors claim that this form of representation is a bridge between the pure exemplar- and part-based methods.

To improve pose estimation even further, other types of body part representations are also investigated. For instance, Dantone \textit{et al.} \cite{dantone2013human}, in addition to human limbs, learn the joint models that connect the limbs together. The tree structure that describes the overall pose is based on the position of the joints starting from the nose position all the way down to the ankles. Cherian \textit{et al.} \cite{cherian2014mixing}, in addition to modeling the relationships between body joints, learn the temporal correspondence of the same joint across a number of frames. The potential position of a joint from one frame to another is measured via calculating the optical flow.

Pishchulin \textit{et al.} \cite{pishchulin2013strong} combine the methods of \cite{wang2011learning} and \cite{dantone2013human} and enhance the robustness of body part representations by learning both their absolute rotation appearance and their relative rotation appearance with respect to the overall body pose. To achieve the former, body part training data is divided into 16 bins (each corresponding to 22.5 degrees of rotation) and a detector is trained for each bin. The second component also performs a similar training procedure with the difference of normalizing the body part rotation with respect to the overall body pose.

Not surprisingly, the majority of the recent pose estimation models are based on CNNs in which the body part representations, as well as underlying relations between them, are all learned together under one framework. These models achieve state-of-the-art performance in both 2D \cite{wei2016cpm} and 3D \cite{zhou2016sparseness} pose estimation. In one of the early CNN-based techniques, DeepPose \cite{toshev2014deeppose}, the authors first train a network to localize the positions of the joints on the full body image. Then, in the next stage, a series of sub-images are cropped around the predicted joints from the first stage and are fed to another regressor for each joint. Using this method, the subsequent regressors see higher resolution images, and as a result, learn features at finer
scales and ultimately achieve a higher precision. Employing a similar technique, in \cite{fan2015combining}, instead of a two-stage learning, the authors use a two-stream network. In one stream the close-up views of body parts are learned and another stream captures a holistic view of the human body. The features learned from each stream are concatenated before the fully connected layers.

To better capture the spatial context around each joint, Wei \textit{et al.} \cite{wei2016cpm} use a multi-stage training scheme. In the first stage, the network only predicts part beliefs from local image evidence. The next stage accepts as input the image features and the features computed for each of the parts in the previous stage and applies convolutional filters with larger receptive fields. The increase in the receptive field of filters allows the learning of potentially complex and long-range correlations between parts.

\begin{figure}[!t]
\centering
\includegraphics[width=1\textwidth]{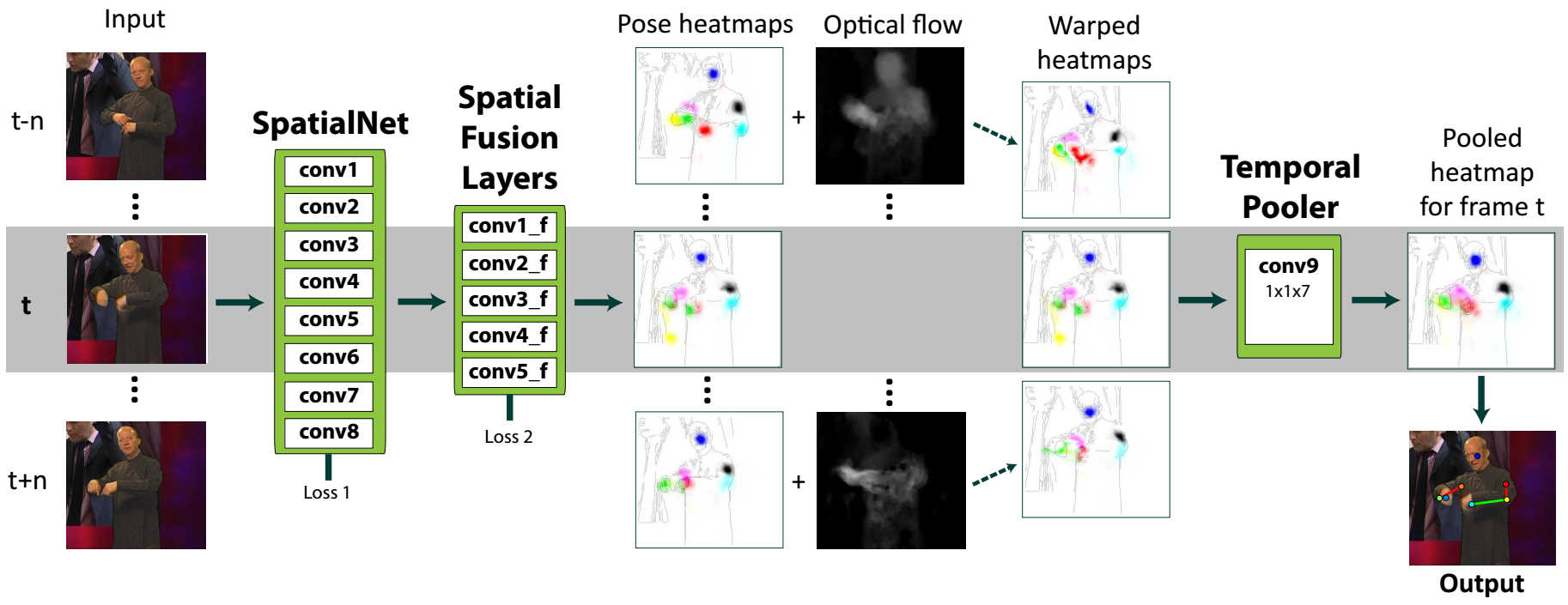}
\caption[An example of a pose estimation architecture.]{The pose estimation architecture proposed in \cite{pfister2015flowing}.}
\label{fig:pose_Est}
\end{figure}

Pfister \textit{et al.} \cite{pfister2015flowing} (see \figref{fig:pose_Est}) take advantage of temporal correspondences in videos to improve pose estimation. At each step, the poses in N frames before and after the given frame are estimated. Then, the spatial relationships between the joints are learned using a second network that takes as input the responses from the 3rd and 7th convolutional layers from the first network. Once the joint locations are optimized on all frames, the results are warped into the reference frame by calculating their optical flow. Next, the warped predictions (from multiple frames) are pooled with another convolutional layer (a temporal pooler) that learns how to weigh the warped predictions from nearby frames. The final joint locations are selected as the maximum of the pooled predictions.

In one of the most recent works, a deep learning method is introduced that is capable of robust estimation of the poses of multiple people in the scene \cite{cao2017realtime}. In this work, the authors extend the method in \cite{wei2016cpm} to explicitly learn the associations between different parts. They use a descriptor known as Part Affinity Fields (PAFs), which is a set of 2D vector fields that encode the location and orientation of limbs over the image domain. These fields help to disambiguate the relationships between various detected parts which might belong to different people.

A summary of described models can be found in Table \ref{table:pose_est}. For a more in depth review of pose estimation algorithms please refer to \cite{andriluka20142d,liu2015survey}.

\begin{table}[!hbtp]
\caption[A summary of pose estimation algorithms.]{A summary of pose estimation algorithms. Abbreviations: \textit{Parts:} H = Hand, UT = Upper Torso, Full = Full body, Hd = Head, \textit{Data Type:} Img = Image, Col = Color, Syn = Synthetic, Gr = Grey. }
\centering
\resizebox{\textwidth}{!}{
\begin{tabular}{|c|c|c|c|c|c|}
\hline
Model&Year&Features&Classif.&Parts&Data Type \\
\hline
\hline
HPE-GA \cite{ohya1994human}&1994&Templates&-&UT&Img+Syn\\ \hline
RA-FPE \cite{mckenna1998real}&1998&Gabor&-&Hd&Img+Gr\\ \hline
SVM-FACE \cite{ng2002composite}&2002&Intensity&SVM&Hd&Img+Gr\\ \hline
Hand-PE \cite{ueda2003hand}&2003&3D Model&-&H&Mul+Img+Gr\\ \hline
3D-HPE \cite{agarwal20043d}&2004&Silhouette&RVM&Full&Img+Col\\ \hline
Parse-AB \cite{ramanan2007learning}&2007&Color, Edge&CRF&Full&Weizmann\\ \hline
Cluster-P \cite{johnson2010pose}&2010&HOG&SVM&Full&LSP\\ \hline
Poslet \cite{wang2011learning}&2011&HOG&SVM&Full&UIUC,Sport\\ \hline
Body-Parts \cite{dantone2013human}&2013&HOG,Color&Dec. Forest&Full&FashionPose\\ \hline
ESM \cite{pishchulin2013strong}&2013&HOG&SVM&Full&Parse\\ \hline
MBP \cite{cherian2014mixing}&2014&HOG,Optical Flow&SVM&UT&VideoPose,MPII,PiW\\ \hline
MDL \cite{ouyang2014multi}&2014&HOG&SVM, Neural Net&Full&LSP, UIUC,PARSE\\ \hline
DeepPose \cite{toshev2014deeppose}&2014&Conv&Neural Net&Full&FLIC, LSP\\ \hline
DS-CNN \cite{fan2015combining}&2015&Conv&Neural Net&Full&FLIC, LSP\\ \hline
Flow-CNN \cite{pfister2015flowing}&2015&Conv&Neural Net&Full&FLIC, ChaLearn, PiW, BBC\\ \hline
CPM \cite{wei2016cpm}&2016&Conv&Neural Net&Full&MPII, LSP, FLIC\\ \hline
SMD \cite{zhou2016sparseness}&2016&Conv&Neural Net&Full&Human3.6M, PennAction\\ \hline
PFA \cite{cao2017realtime}&2017&Conv&Neural Net&Full&MPII,COCO\\ \hline
\end{tabular}
}
\label{table:pose_est}
\end{table}

\subsubsection{Action Recognition}
Activity recognition is perhaps one of the most studied fields in computer vision. The amount of work done in this field is apparent from a long list of survey papers published throughout the past decades, such as surveys on hand gesture recognition \cite{rautaray2015vision}, general vision-based action recognition \cite{poppe2010survey,weinland2011survey}, video surveillance \cite{vishwakarma2013survey}, action recognition using 3D data \cite{aggarwal2014human}, action recognition in still images \cite{guo2014survey}, semantic action recognition \cite{ziaeefard2015semantic}, action recognition using deep networks \cite{herath2017going} or hand-crafted feature learning \cite{sargano2017comprehensive}. Some works also present extensive evaluation of the state-of-the-art on popular datasets in the field \cite{caba2015activitynet}.

Depending on the difficulty of the task, activity recognition algorithms come in 4 types (in the order of difficulty from easiest to hardest): \textit{gesture recognition}, \textit{action recognition}, \textit{interaction recognition} and \textit{group activity}.

Gesture recognition is commonly used in close encounters for applications such as Human Computer Interaction (HCI). The process often involves the recognition of the hand, and identifying its transformation in terms of pose and velocity in temporal domain \cite{campbell1996invariant}. Given that gesture recognition is very application specific and the fact that activity recognition algorithms capture the gesture changes to some degree, we will focus our discussion mainly on other three categories of activity recognition.

Action recognition algorithms, in their simplest form, often rely on some form of template matching to identify a particular action pattern. Template matching can be done in spatial domain by identifying certain body forms (e.g. pedestrian legs) to infer a type of action (e.g. walking) \cite{curio2000walking}. More sophisticated algorithms use templates generated in spatiotemporal domain \cite{niyogi1994analyzing,bobick2001recognition,ke2007event}. For instance, Niyogi and Adelson \cite{niyogi1994analyzing} detect a pedestrian in the image sequence using background subtraction. They then cut through the temporal domain and generate a 2D image, which reflects the temporal changes (at a given height of the pedestrian). This 2D image is compared with a pre-learned template to realize the gait of the pedestrian.

Bobick and Davis \cite{bobick2001recognition} introduce two types of features, motion-energy image (MEI) and motion-history image (MHI), to characterize human activities through time. MEI is a binary image that shows instantaneous motion pattern in the sequence. This image is generated by simply aggregating the motion patterns through time in a single 2D image. MHI, on the other hand, is a scalar-valued image in which the value of each pixel is a function of time where pixels corresponding to more recent movements have higher intensity values compared to the rest. The combination of these two images forms a feature vector, which in turn can be compared with pre-learned features to identify the action. An extension of this approach for more realistic scenes is employed in \cite{ke2007event}, where, in addition to temporal changes, color features are used to separate the human body from the background. To deal with occlusion, the templates are divided into parts (e.g. upper body and legs) and are identified separately at test time.

The major drawback of the template-based approaches is that they often make simplistic assumptions about the environment setting. For example, in some works the height or velocity of the pedestrian is assumed to be fixed \cite{niyogi1994analyzing}, or the camera position is considered fixed with minimal background motion \cite{bobick2001recognition}. In practice, such assumptions can significantly constrain the applicability of these approaches to complex and cluttered scenes.

Learning algorithms are also very popular tools for action recognition both in still images \cite{ikizler2008recognizing,yao2010modeling} and videos \cite{feichtenhofer2015dynamically,yue2015beyond}. For instance, in still images, human pose is often used to recognize actions \cite{ikizler2008recognizing,yao2010modeling,yang2010recognizing}. Ikizler \textit{et al.} \cite{ikizler2008recognizing} use the pose detection algorithm in \cite{ramanan2007learning} to identify body parts. The orientation of body parts is measured and quantized in a histogram to form a descriptor of the image. The descriptors for each activity is then learned using a linear SVM algorithm. In addition to the pose, in \cite{yao2010modeling}, the authors learn the relationship between the pose and certain objects (e.g. tennis racquet) via a graphical model. Besides human pose, Delaitre \textit{et al.} \cite{delaitre2010recognizing} investigate the role of 2D features, e.g. SIFT, in conjunction with popular regression techniques such as SVM. The authors show that for simple action recognition tasks, using 2D features can result in state-of-the-art performance.

Similar to action recognition in still images, video-based models take advantage of various learning techniques such graphical models (e.g. Bayesian networks) and regression algorithms (e.g. SVM). For instance, Yamato \textit{et al.} \cite{yamato1992recognizing} use a Hidden Markov Model (HMM) for inference of action types in videos. The authors train a separate HMM for each action category (6 common actions in tennis). The regression based models often use features that are generated from optical flow information \cite{efros2003recognizing} or extracted from spatiotemporal domain \cite{schuldt2004recognizing,blank2005actions}. For example, Schuldt \textit{et al.} \cite{schuldt2004recognizing} use local space-time features to characterize the critical moments in an observed action. These features are generated by computing second-moment matrix using spatiotemporal image gradients within a Gaussian neighborhood around each point and selecting positions with local maxima. Blank \textit{et al.} \cite{blank2005actions} employ a similar approach to \cite{bobick2001recognition} by characterizing the entire space-time volume resulting from the action. Here, every internal space-time point is assigned a value that reflects its relative position within the space-time shape (the boundaries of the volume).

Today, action recognition algorithms are moving towards neural networks. In its simplest form, a CNN, using 3D convolutional filters, can infer actions from a stacked image sequence \cite{ji20133d}. This method can be further improved by performing the convolutions at multiple resolution levels to extract both fine and coarse features. The authors of \cite{karpathy2014large} investigate such a technique by using a two-stream network. One stream receives as input the image in its original resolution (context-stream) and the other a zoomed cropped portion of the image from the center (fovea-stream). The output of these two networks is then fused for final classification.

To achieve a better representation, more recent algorithms learn spatial and temporal features separately \cite{simonyan2014two,feichtenhofer2016convolutional,yue2015beyond,wang2016actions}. These models typically use a two-stream architecture, in which one stream receives a color image at time $t$ and another stream takes a stack of temporal representations,  often in the form of optical flow measurements for $n$ frames before and after the image at time $t$. The outputs of the streams are then fused prior to final classification. The fusion process might be at the early (e.g. after 3rd convolutional layer) or late (e.g. right before fully connected layers) stages.

Some CNN-based algorithms implement graphical reasoning using Recurrent Neural Nets (RNNs) or its variants such as popular Long Short Term Memory (LSTM) models \cite{yue2015beyond,ma2016learning}. LSTM networks are equipped with a form of internal memory, which allows them to learn the dependencies between the consecutive frames. The internal control of LSTMs is through the use of gates, whereby the contribution of new information and past learning for final inference is determined. For instance, Yue \textit{et al.} \cite{yue2015beyond} propose a method for learning spatial and temporal features using two common CNN models (e.g. GoogleNet). The outputs of the networks prior to the loss layers are fused and fed to a 5-stack LSTM network for final inference.

\begin{figure}[!t]
\centering
\includegraphics[width=1\textwidth]{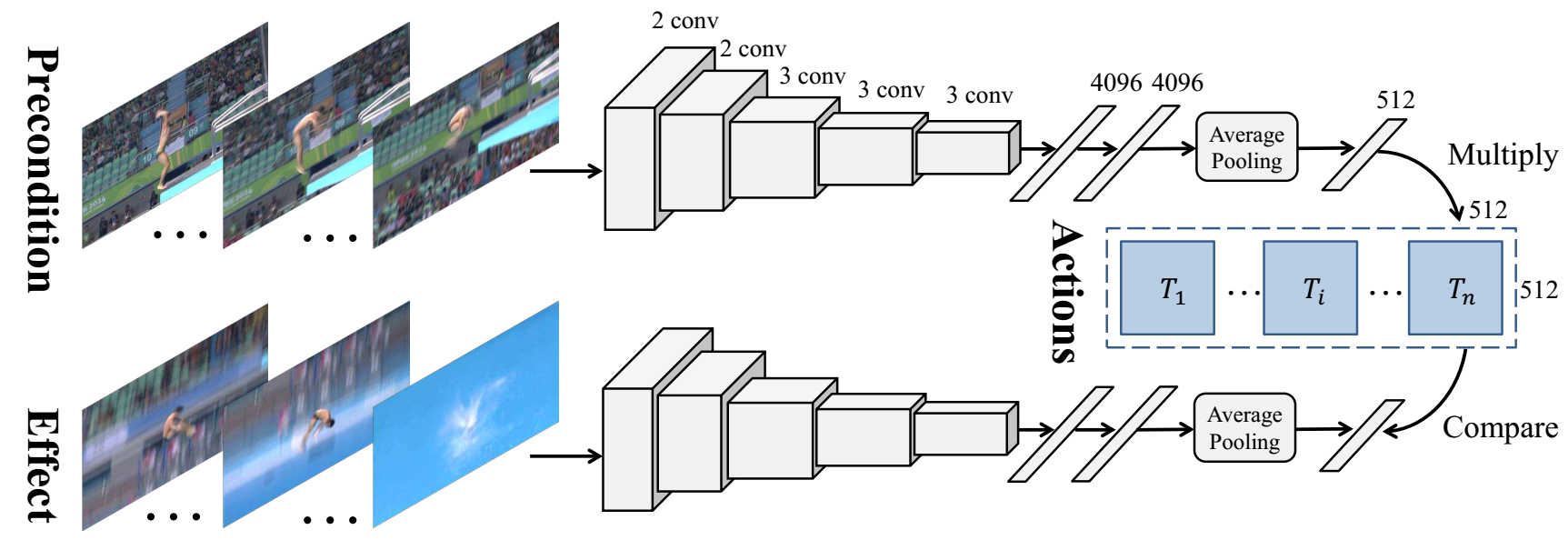}
\caption[Actions as transformation architecture.]{The transformation framework proposed in \cite{wang2016actions}.}
\label{fig:activity_rec}
\end{figure}

Other methods of inference using neural nets have also been investigated \cite{wang2016actions,kar2016adascan}. For example, Wang \textit{et al.} \cite{wang2016actions} (see \figref{fig:activity_rec}) treat the action recognition problem as learning the transformation between the precondition frames (typically from the beginning of the video) and the effects (a portion of the frames illustrating the consequence of the action). The authors use a Siamese network architecture which receives as input two streams of frames (i.e. the precondition and effect frames), learns their corresponding characteristics by applying a series of convolutional and fully connected layers, applies a linear transformation to the output of precondition stream, and finally measures the similarity in terms of distance between the transformed output of the precondition stream with the effect one. Here, the objective of learning is to find a transformation matrix that minimizes the distance between the two streams of the network. Using this approach, the authors argue that the overfitting problem that exists in other approaches due to scene context learning can be avoided because the networks are explicitly forced to encode the change in the environment, i.e. the inherent reason that convinced the agent to perform the action.

Interaction recognition algorithms \cite{park2003recognition,ryoo2009spatio} often infer the type of activity by directly estimating the relationship between the subjects' body parts. For instance, Park \textit{et al.} \cite{park2003recognition} first identify the body parts of the humans in the scene, and then, using a graphical model, infer their interaction using the distance between each person's body parts and the corresponding one in another person as well as the overall orientation of their entire bodies.

Similar direct approaches have been used in group activity recognition. Choi \textit{et al.} \cite{choi2009they}, for instance, use so called spatiotemporal local (STL) features to characterize individuals in the scene. This descriptor, in essence, is a histogram that records the following factors with respect to the reference person: the number of people surrounding the person, their relative pose and distance to that person. These descriptors are generated for each frame and concatenated for the entire video. At the end, using a linear classifier such as SVM, the descriptors are classified into different group activities.

A number of algorithms use two-stage inference where, first, the actions of individuals are recognized, and then their spatial formation and dynamic changes are used to recognize the overall activity \cite{lan2012social,deng2015deep}. For example, in \cite{deng2015deep} a neural net architecture is proposed. This network, by applying a series of ConvNets, first estimates the pose and action of each individual as well as the class of the scene (e.g. fall event). Next, the output of these networks is passed to a message passing network that learns the semantic dependencies between the inputs. At the end, the output of the message passing network facilitates the refinement of estimations for individual activities. For instance, a person standing in a queue might be identified as standing, but after taking into account the entire scene, it would be relabeled as queuing. For a summary of algorithms reviewed in the section please refer to Table \ref{table:act_rec}.

\begin{table}[!hbtp]
\caption[A summary of action recognition algorithms.]{A summary of action recognition algorithms. Abbreviations: \textit{Features:} ST - Spatiotemporal features, \textit{Act. Type:} Gest = Gesture, Act = Action, Int = Interaction, Grp = Group, \textit{Data Type:} Img = Image, Col = Color, Vid = Video, Gr = Grey, FiP = Fixed Position. }
\centering
\resizebox{\textwidth}{!}{
\begin{tabular}{|c|c|c|c|c|c|c|}
\hline
Model&Year&Features&Classif.&Act. Type&Data Type&No. Class \\
\hline
\hline
Mesh-HMM \cite{yamato1992recognizing}&1992&Mesh features&HMM&Act&Vid+Gr+FiP &6\\ \hline
XYT \cite{niyogi1994analyzing}&1994&Motion,Template&-&Act&Vid+Gr+FiP &1\\ \hline
3D-Gesture \cite{campbell1996invariant}&1996&Position, Velocity&HMM&Gest&St+Vid+Gr+FiP&19\\ \hline
Walk-Ped \cite{curio2000walking}&2000&Template, Edges&-&Act&Vid+Gr&1\\ \hline
Aerobics \cite{bobick2001recognition}&2001&Template, Motion&-&Act&Vid+Gr+FiP&18\\ \hline
Act-Dist \cite{efros2003recognizing}&2003&Optical flow&KNN&Act&Vid+Col&30\\ \hline
Two-Int \cite{schuldt2004recognizing}&2003&Pose&BN, HMM&Int&Vid+Col+FiP&9\\ \hline
Local-SVM \cite{schuldt2004recognizing}&2004&ST&SVM&Act&RHA&6\\ \hline
Space-time \cite{blank2005actions}&2005&ST&KNN&Act&ASTS&9\\ \hline
EDCV \cite{ke2007event}&2007&Template, Optical flow&-&Act&Vid+Col&6\\ \hline
RA-Still \cite{ikizler2008recognizing}&2008&Pose&SVM&Act&Img+Col&6\\ \hline
CAC-STP \cite{choi2009they}&2009&Pose, ST&SVM&Grp&Vid+Col&5\\ \hline
STRM \cite{ryoo2009spatio}&2009&ST&K-means&Int&Vid+Col+FiP&6\\ \hline
MMC-OHP \cite{yao2010modeling}&2010&Shape context&Graphical&Act&Img+Col&6\\ \hline
Bag-AR \cite{delaitre2010recognizing}&2010&SIFT&SVM&Act&Willow-Act&7\\ \hline
RHA-SI \cite{yang2010recognizing}&2010&Pose&SVM&Act&Img+Col&5\\ \hline
Social \cite{lan2012social}&2012&HOG&SVM&Grp&Vid+Col&3\\ \hline
3D-Conv \cite{ji20133d}&2013&Conv&Neural Net&Act&KTH&6\\ \hline
LSVC \cite{karpathy2014large}&2014&Conv&Neural Net&Act&Sports-1M,UCF101&588\\ \hline
Two-Stream \cite{simonyan2014two}&2014&Conv&Neural Net&Act&UCF101, HMDB&152\\ \hline
Deep-Struct \cite{deng2015deep}&2015&Conv&Neural Net&Grp&Col-AD&5\\ \hline
Deep-Snippets \cite{yue2015beyond}&2015&Conv&Neural Net&Act&Sports-1M,UCF101&487\\ \hline
SIM \cite{deng2016structure}&2016&Conv&Neural Net&Grp&Col-AD&5\\ \hline
Early-Det \cite{ma2016learning}&2016&Conv&Neural Net&Act&ActivityNet&203\\ \hline
Act-Transform \cite{wang2016actions}&2016&Conv&Neural Net&Act&ACT&43\\ \hline
Two-Stream-3D \cite{feichtenhofer2016convolutional}&2016&Conv&Neural Net&Act&UCF101, HMDB&152\\ \hline
AdaScan \cite{kar2016adascan}&2017&Conv&Neural Net&Act&HMDB,UCF101&152\\ \hline
\end{tabular}
}
\label{table:act_rec}
\end{table}

\subsection{What the Pedestrian is Going to Do: Intention Estimation}

\subsubsection{Datasets}
In the past few decades, a number of large-scale datasets have been collected in different geographical locations to study pedestrian and driver behavior in traffic scenes. Some of the most well-known datasets are 100-car Naturalistic Study \cite{neale2005overview}, UDRIVE \cite{eenink2014udrive}, SHRP2 \cite{richard2015shrp}, and many more \cite{habibovic2013driver,dozza2014introducing}. Although these datasets provide invaluable statistics about various factors that influence the behavior of road users, the raw videos or any information that can be used for visual modeling is not publicly available.

A number of existing vision datasets can be potentially used for intention estimation such as activity recognition and object tracking datasets \cite{CAVIAR,PETS2009} or pedestrian detection datasets that have temporal correspondences \cite{dollarCVPR09peds,Geiger2013IJRR}. The drawback of these datasets is that besides bounding box information, they only contain a limited (if any) number of annotations for contextual elements such as street structure, activities, group size, signals, etc. Therefore they are mainly suitable for predicting behavior based on pedestrians' dynamics.

The number of datasets that are specifically tailored for intention estimation applications is very small.
One of the main datasets suitable for behavioral studies with visual information is Joint Attention in Autonomous Driving (JAAD) \cite{kotseruba2016joint}. This dataset comprises  346 high resolution videos with bounding box information for pedestrians. A subset of the pedestrians (over 600) are annotated with behavioral data, such as whether the pedestrians are looking, walking, crossing or waving hands, as well as those pedestrians' demographic information. In addition each frame in the video sequences is annotated with contextual information such as street delineation, street width, weather, etc. Daimler-Path \cite{schneider2013pedestrian} is a dataset designed for pedestrian path prediction. It contains 68 greyscale video samples of pedestrians from a moving car perspective. The data is collected using a stereo camera and is annotated with vehicle speed, bounding box information and 4 pedestrian motion types including crossing, stopping, starting to walk and bending in.

Another dataset for pedestrian intention estimation is Daimler-Intend \cite{kooij2014context} that comes with 58 greyscale stereo video sequences annotated with bounding box information, the degree of head orientation, pedestrian distance to the curb and the vehicle as well as the vehicle's speed.

\subsubsection{Pedestrian Intention Estimation}
As discussed earlier, intention estimation is a broad topic and can be applied in various applications. In intelligent driving systems alone, intention estimation techniques have been widely used for predicting the behavior of the drivers \cite{ohn2014head,molchanov2015multi}, other drivers \cite{laugier2011probabilistic,li2016recognizing}, pedestrians \cite{kooij2014analysis,kohler2012early} or combinations of any of these three \cite{phan2013estimating,bahram2016combined} (for a more detailed list of these techniques see \cite{ohn2016looking}). In this section, however, we particularly discuss pedestrian intention estimation methods in the context of intelligent transportation systems with a few mentions of some of the techniques used in mobile robotics.

Essentially, intention estimation algorithms are very similar to object tracking systems. One's intention can be estimated by looking at their past and current behavior including their dynamics, current activity and context. In autonomous driving, for example, we want to estimate the next location where the pedestrian might be appearing, realize whether they attempt to cross, or predict the changes in their dynamics.

There are a number of works that are purely data-driven, meaning that they attempt to model pedestrian walking direction with the assumption that all relevant information is known to the system. These models either base their estimation merely on dynamic information such as the position and velocity of pedestrians \cite{schulz2015controlled}, or in addition, take into account the contextual information of the scene such as pedestrian signal state, whether the pedestrian is walking alone or in a group, and their distance to the curb \cite{hashimoto2015probability}. In a work by Brouwer \textit{et al.} \cite{brouwer2016comparison}, the authors investigate the role of different types of information in collision estimation. More specifically, they consider the following four factors: \textit{dynamics} (directions pedestrian can potentially move and time to collision), \textit{physiological elements} (pedestrian's moving direction and distance to the car and velocity), \textit{awareness} (in terms of head orientation towards the vehicle), and \textit{obstacles} (the position of obstacles in the scene and sidewalks). The authors show that, in isolation, second and third factors are the best predictors of collision. They also show that by fusing the information from all four factors, the best prediction results can be achieved.

The vision-based intention estimation algorithms often treat the problem as tracking a dynamic object by simply taking into account the changes in the position, velocity and orientation of the pedestrian \cite{goldhammer2014analysis,trivedi2015trajectory} or by considering the changes in their 3D pose \cite{quintero2014pedestrian}. For instance, Kooij \textit{et al.} \cite{kooij2014analysis} employ a dynamical Bayesian model, which takes as input the current position of the pedestrian and, based on their motion history, infers in which direction the pedestrian might move next. In addition to pedestrian position, Volz \textit{et al.} \cite{volz2016data} use information regarding the pedestrian's distance to the curb and the car as well as the pedestrian's velocity at the time. This information is fed to a LSTM network to infer whether the pedestrian is going to cross the street or not.

In robotics, intention prediction algorithms are used as a means of improving trajectory selection and navigation. Besides dynamic information, these techniques assume a potential goal for pedestrians based on which their trajectories are predicted \cite{bandyopadhyay2013intention,bai2015intention}. The drawback of these algorithms is that they assign to pedestrians a predefined goal location, which is based on the current orientation of the pedestrian  towards one of the potential goal locations.

\begin{figure}[!t]
\centering
\includegraphics[width=0.8\textwidth]{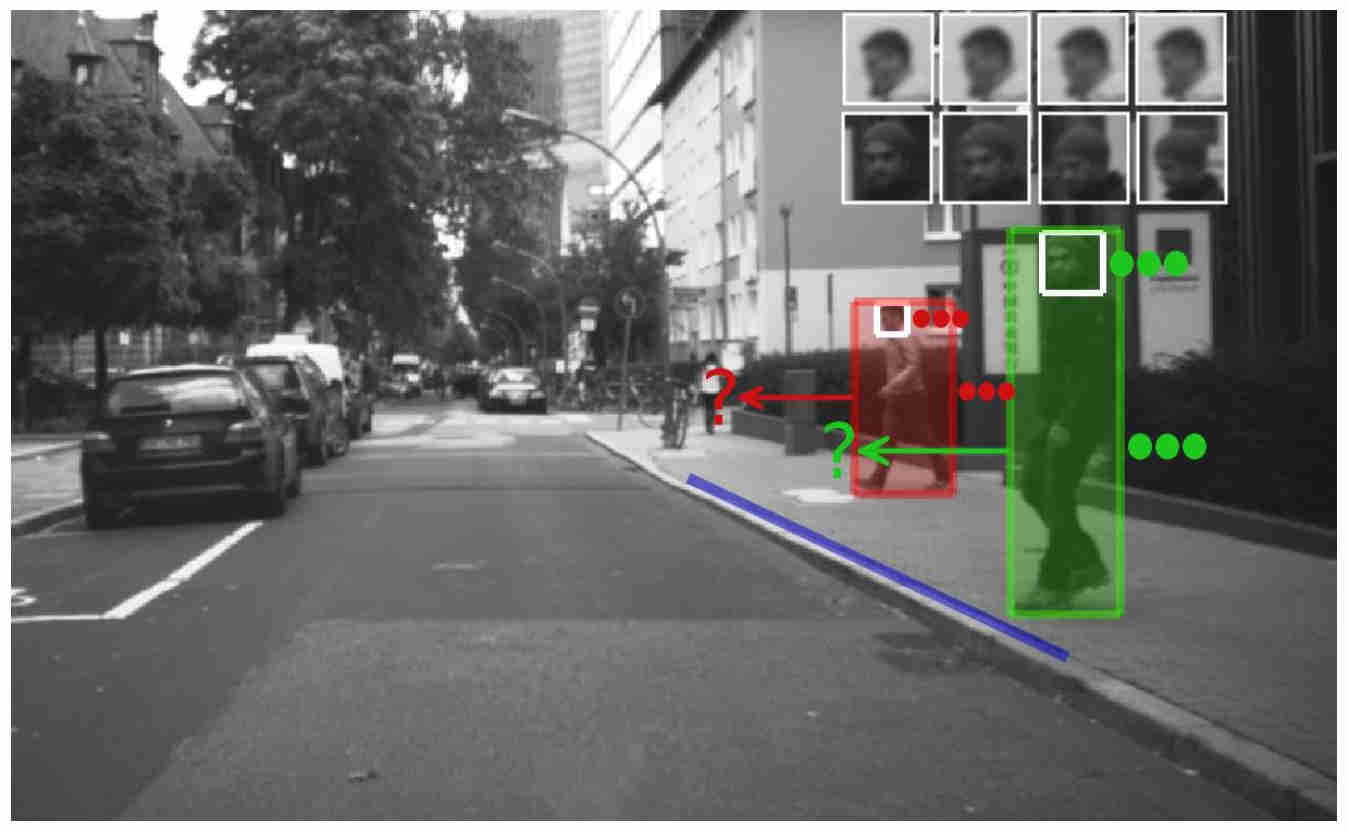}
\caption[Pedestrian's head orientation as the sign of awareness.]{Pedestrian's head orientation as the sign of awareness. Source: \cite{kooij2014context}.}
\label{fig:intent_est}
\end{figure}

In some recent works, social context is exploited to estimate intention. For instance, pedestrian awareness is measured by the head orientation relative to the vehicle (see \figref{fig:intent_est})  \cite{kooij2014context,hariyono2016estimation,kwak2017pedestrian}. Kooij \textit{et al.} \cite{kooij2014context} employ a graphical model that takes into account factors such as pedestrian trajectory, distance to the curb and awareness. Here, they argue that the pedestrian looking towards the car is a sign that he noticed the car and is less likely to cross the street. This model, however, is based on a scripted data which means that the participants were instructed to perform certain actions, and the videos were only recorded in a narrow non-signalized street.

For intention estimation, some scholars also consider social forces, which refer to people's tendency to maintain a certain level of distance from one another. In their simplest form, social forces can be treated as a dynamic navigation problem in which pedestrians choose the path that minimizes the likelihood of colliding with others \cite{pellegrini2009you}. Social forces also reflect the relationship between pedestrians, which in turn can be used to predict their future behavior. For instance, Madrigal \textit{et al.} \cite{madrigal2014intention} define two types of social forces: repulsion and attraction. In this interpretation, for example, if two pedestrians are walking close to one another for a period of time, it is more likely that they are interacting, therefore the tracker estimates their future states close together.

Apart from the explicit tracking of pedestrian behavior, a number of works try to solve the intention estimation problem using various classification approaches. Kohler \textit{et al.} \cite{kohler2012early}, via using a SVM algorithm, classify pedestrian posture as about to cross or not crossing. The postures are extracted in the form of silhouette body models from motion images which are generated by background subtraction. In the extensions of this work \cite{kohler2013autonomous,kohler2015stereo}, the authors use a HOG-based detection algorithm to first localize the pedestrian, and then using stereo information, they extract the body silhouette from the scene. To account for the previous action, they perform the same process for N consecutive frames and superimpose all silhouettes into a single image. The final image is used to classify whether the pedestrian is going to cross.

Rangesh \etal \cite{rangesh2018vehicles} estimates the pose of pedestrians in the scene, and identifies whether they are holding cellphones. The combination of the pedestrians' pose and the presence of a cellphone is used to estimate the level of pedestrians engagement in their devices. In \cite{rasouli2017they}, the authors use various contextual information such as characteristics of the road, the presence of traffic signals and zebra crossing lines, in conjunction with pedestrians' state to estimate whether they are going to cross. In this method, two neural network architectures are used. One network is responsible for detecting contextual elements in the scene and the other identifying whether the pedestrian is walking/standing and looking/not-looking. The scores from both networks are then fed to a linear SVM to classify the intention of the pedestrians. The authors report that by taking into account the context, intention estimation accuracy can be improved by upto 23\%.   

In addition to appearance-based models, Volz \textit{et al.} \cite{volz2015feature} use pedestrian velocity, and  distance to the curb, crosswalk and the car. Schneemann \textit{et al.} \cite{schneemann2016context} consider the structure of the street as a factor influencing crossing behavior. The authors generate an image descriptor in the form of a grid which contains the following information: \textit{street-zones} in the scene including ego-zone (the vehicle's lane), non-ego lanes (other street lanes), sidewalks, and mixed-zones (places where cars may park), \textit{crosswalk occupancy} (the position of scene elements with respect to the current position of the pedestrians), and \textit{waiting area occupancy} (occupancy of waiting areas such as bus stops with respect to the pedestrian's orientation and position). Such descriptors are generated for a number of consecutive frames and concatenated to form the final descriptor. At the end, an SVM algorithm is used to decide how likely the pedestrian is to cross. Despite its sophistication for exploiting various contextual elements, this algorithm does not perform any perceptual tasks to identify the aforementioned elements and simply assumes they are all known in advance.

In the context of robotic navigation, Park \textit{et al.} \cite{park2016hi} classify observed trajectories to measure the imminence of collisions. The authors recorded over 2.5 hours of videos of the pedestrians who were instructed to engage in various activities with the robot (e.g. approaching the robot for interaction or simply blocking its way). Using a Gaussian process, the trajectories were then classified into blocking and non-blocking groups.

Table \ref{table:intention_est} summarizes the papers discussed in this section.

\begin{table}[!hbtp]
\caption[A summary of intention estimation algorithms.]{A summary of intention estimation algorithms. Abbreviations: \textit{Factors:} PP = Pedestrian Position, PV = Pedestrian Velocity, SC = Social Context, PPs = Pedestrian Posture, SS = Street Structure, MH = Motion History, HO = Head Orientation, G = Goal, GS = Group Size, Si = Signal, DC = Distance to curb, DCr = Distance to Crosswalk, DV = Distance Velocity, Vehicle Dynamics = VD, \textit{Inference:} GD = Gradient Decent, PF = Particle Filter, GP = Gaussian Process, \textit{Pred. Type:} Traj = Trajectory, Cross = Crossing, \textit{Data Type:} Img = Image, Col = Color, Vid = Video, Gr = Greyn, St = Stereo, I = Infrared \textit{Cam Pose:} F = Front view, BeV = Bird's Eye View, Mult = Multiple view, FiP = Fixed Position. }
\centering
\resizebox{\textwidth}{!}{
\begin{tabular}{|c|c|c|c|c|c|c|}
\hline
Model&Year&Factors&Inference&Pred. Type&Data Type&Cam Pose\\
\hline
\hline
LTA \cite{pellegrini2009you}&2009&PP, PV,G, SC&GD&Traj&Vid+Col&BeV+FiP\\ \hline
Early-et \cite{kohler2012early}&2012&PPs&SVM&Cross&Img+Col&F+FiP\\ \hline
IAPA \cite{bandyopadhyay2013intention}&2013&PP,PV,G&MDP&Traj,Cross&Vid+Col&F\\ \hline
Evasive \cite{kohler2013autonomous}&2013&PPs,MH&SVM&Cross&St+Vid+Gr&F+FiP\\ \hline
Early-Pred \cite{goldhammer2013early}&2014&PP,PV&SVM&Traj&Vid+Gr&Mult+FiP\\ \hline
Veh-Pers \cite{kooij2014analysis}&2014&PP,PV,VD&BN&Traj&Vid+Gr&F\\ \hline
Context-Based \cite{kooij2014context}&2014&SS,HO,PP,VD&BN&Cross&Daimler-Intend&F\\ \hline
Intent-Aware \cite{madrigal2014intention}&2014&PP,PV,SC&BN&Traj&PETS&F\\ \hline
Path-Predict \cite{quintero2014pedestrian}&2014&PP,PV,Pose&BN&Traj, Pose&Vid+Col&F\\ \hline
MMF \cite{schulz2015controlled}&2015&PP,PV,SC&CRF&Traj&Daimler-Path&F\\ \hline
Intend-MDP \cite{bai2015intention}&2015&PP,PV,G,VD&MDP&Traj&Vid+Col+L&F\\ \hline
SVB \cite{kohler2015stereo}&2015&MH,PPs&SVM&Cross&St+Vid+Gr&F+FiP\\ \hline
PE-PC \cite{hashimoto2015probability}&2015&GS,PP,PV,Si&BN&Traj, Cross&Vid+Gr&F\\ \hline
Traj-Pred \cite{trivedi2015trajectory}&2015&PP,PV&PF&Traj&Vid+Col&F\\ \hline
FRE \cite{volz2015feature}&2015&PV,DC,DCr,DV,VD&SVM&Cross&Vid+L&F\\ \hline
Eval-PMM \cite{brouwer2016comparison}&2016&PP,PV,HO&BN&Traj&Vid+Col&F\\ \hline
ECR \cite{hariyono2016estimation}&2016&PP,PV,HO &BN&Collision&Caltech, Daimler-Mono&F\\ \hline
HI-Robot \cite{park2016hi}&2016&MH&GP&Collision&Vid+L&F\\ \hline
CBD \cite{schneemann2016context}&2016& PP,SS,MH&SVM&Cross&Vid+Col&F\\ \hline
DDA \cite{volz2016data}&2016&DC,DCr,DV,VD&Neural Net&Cross&Vid+L&F\\ \hline
DFA \cite{kwak2017pedestrian}&2017&PP,PV,MH&DFA&Cross&Vid+I&F\\ \hline
Cross-Intent \cite{rasouli2017they}&2017&PPs,SS,HO&\makecell*{Neural Net,\\SVM}&Cross&JAAD&F\\ \hline
Ped-Phones \cite{rangesh2018vehicles}&2018&PPs&SVM,BN&Pose&Vid+Col&F\\ \hline
\end{tabular}
}
\label{table:intention_est}
\end{table}

\section{Reasoning and Decision Making: How Pedestrians Analyze the Information}

Discussing reasoning as a separate topic is difficult because it is involved in every aspect of an intelligent system. For instance, as was shown in the previous sections, visual perception algorithms employ various types of reasoning whether for data acquisition (e.g. active recognition) or processing the sensory input (e.g. the use of time series for activity recognition). As a result, in this section, we will discuss reasoning at a macro level and review a number of control architectures used by intelligent robots. These architectures include all the subtasks required for accomplishing a task, ranging from perception to actuation of the plans.

Finding a common definition for all types of reasoning used in practical systems is a daunting task. Primarily, this is due to the fact that the terminology and grouping of reasoning approaches vary significantly in different fields of computer science such as AI, robotics or cognitive systems. Therefore, we leave the detailed definitions of different AI systems to AI experts and instead will briefly discuss a number of control and planning architectures designed for actual autonomous robots or vehicles.

\subsection{Decision Making in Autonomous Vehicles}

\begin{figure}[!t]
\centering
\includegraphics[width=0.6\textwidth]{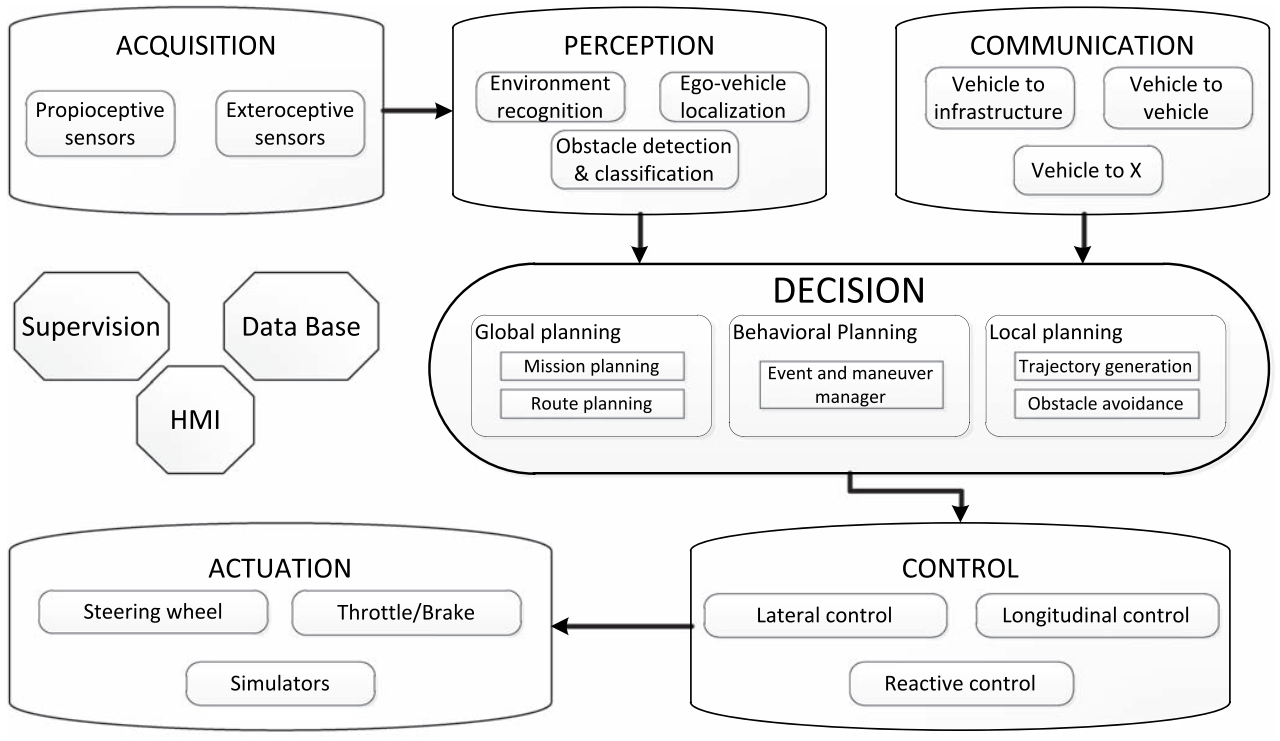}
\caption[A general control architecture for autonomous vehicles.]{A general control architecture for autonomous vehicles. Source: \cite{gonzalez2016review}}
\label{fig:action_planning}
\end{figure}

At the very least a practical intelligent robot must have three sub-modules including perception, decision-making (planning)\footnote{The term may vary in different literature and terms such as decision-making, planning or action selection are used interchangeably.} and control \cite{fraichard2001fuzzy}. Note that some architectures such as Subsumption \cite{brooks1990elephants}, that merely rely on bottom-up sensory input, have achieved intelligent behavior, such as moving or navigating in an environment, to some extent. Here, our focus is rather on intelligent vehicles and the submodules necessary for their proper functioning.

In addition to the three main modules mentioned above, some systems include a monitoring component that oversees the performance of other modules and intervenes (e.g. by changing operating constraints) if a failure is detected \cite{mitchell1987planning}. Given the inherent complexity of the autonomous driving task, more recent architectures further divide these components into sub-modules \cite{gonzalez2016review} (see \figref{fig:action_planning}). For instance, data acquisition can be decoupled from perception and actuation can be separated from control.

% In this interpretation, there is also a communication module responsible for sharing information with other road users or infrastructure.

The heart of a robotic control architecture is the decision-making module, which is responsible for reasoning about what control commands to produce given the perceptual input and the mission's objectives. Decision-making has several sub-components including \textit{global planning, behavioral planning}, and \textit{local planning}. Global planning can be seen as strategic planning in which the abstract mission goals are translated into geographical goals (mission planning) and then, by taking into account the path constraints, a route is specified (route planning) \cite{mitchell1987planning}. The behavioral planning module (sometimes called behavioral execution module) is responsible for guaranteeing the system adherence to various rules of the road, especially those concerning structured interactions with other traffic and road blockages. Examples include selecting an appropriate lane when driving in an urban setting or handling intersections by observing other vehicles and determining who has the right of way \cite{ferguson2008reasoning}. Last but not least is the local planning module which is in charge of executing the current goal. This module generates trajectories and deals with dynamic factors in the scene (obstacles avoidance) \cite{gonzalez2016review}.

\subsection{Knowledge-based Systems}

Now the question is, how can reasoning in each level of planning be performed? Classical models use knowledge-based systems (KBS) to deal with reasoning tasks. KBS typically have two major components, a knowledge-base and an inference engine \cite{davis1980meta}. Here, knowledge is represented in the form of rules and facts about the world. The inference engine task is then to use the knowledge base to solve a given problem often by repeating selection and application of the knowledge to the problem. There are many instances of successful applications of KBS to practical systems such as traffic monitoring and control \cite{cucchiara2000image} and autonomous driving \cite{ferguson2008reasoning} (e.g. for traffic rules).

Knowledge-based systems, however, have a number of limitations. One of the major ones is in the knowledge elicitation process in which expert knowledge is transformed into rules. This process can be challenging because it requires special skills and often takes many man-years to accomplish. In addition, KBS require an explicit model of the world, which in practice makes them less flexible when dealing with new problems \cite{watson1994case}. Such limitations gave rise to case-based systems (CBS) that learn new knowledge as they experience new phenomena. CBS, on their own or in conjunction with rule-based systems \cite{golding1991improving}, show more practically feasible performance in complex systems.

The core of CBS is a library of cases which is acquired by the system through experience. In this context, a case comprises three components \cite{watson1994case}: \textit{the problem} that describes the state of the world, \textit{the solution} to that problem, and/or \textit{the outcome} which describes the state of the world after the case has occurred. CBS involve four processes: \textit{retrieve} the most similar case, \textit{reuse} the case to attempt to solve the problem, \textit{revise} the proposed solution (typically if outcome is seen), and \textit{retain} the new solution as a part of a new case.

\begin{figure}[!t]
\centering
\subfloat[]{
\includegraphics[width=7cm,height=5cm]{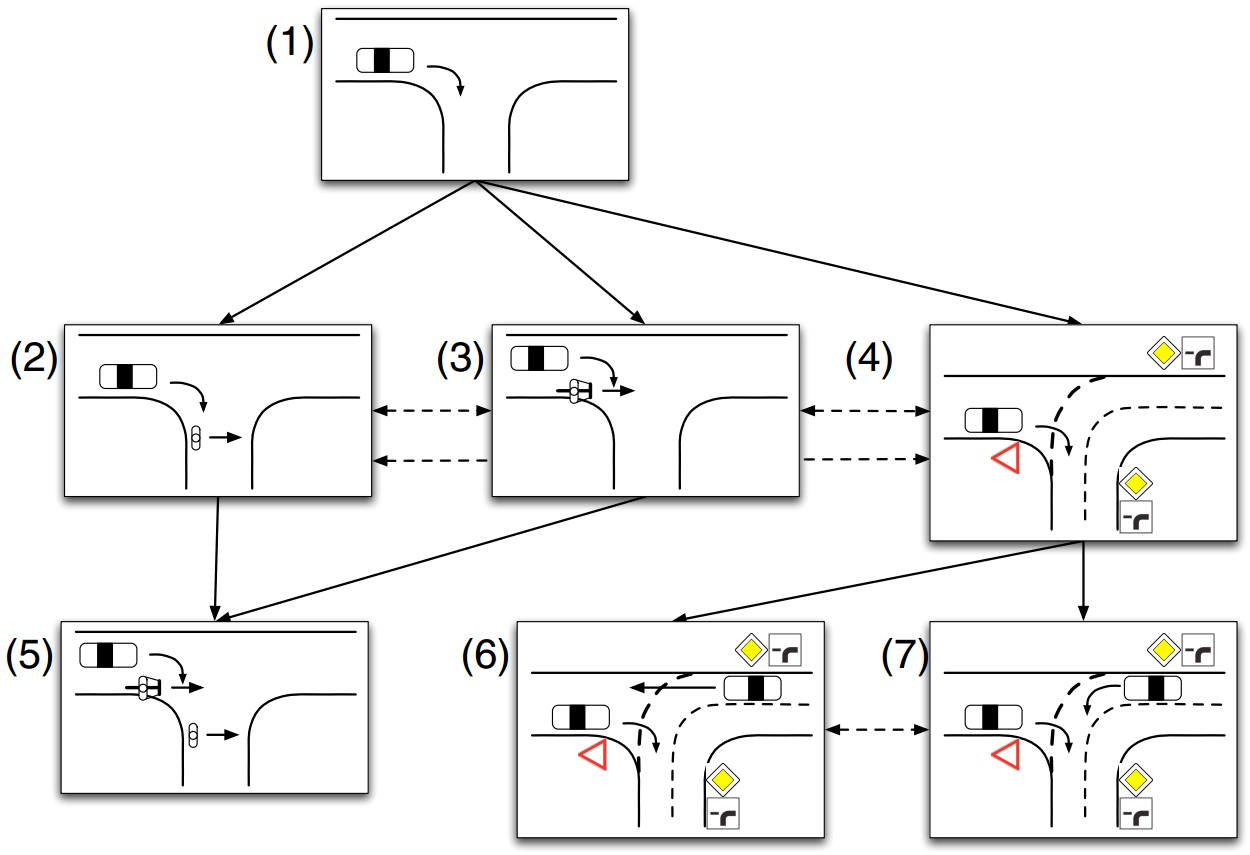}
\label{fig:CSB1}}
\hspace{1cm}
\subfloat[]{
\includegraphics[width=7cm,height=5cm]{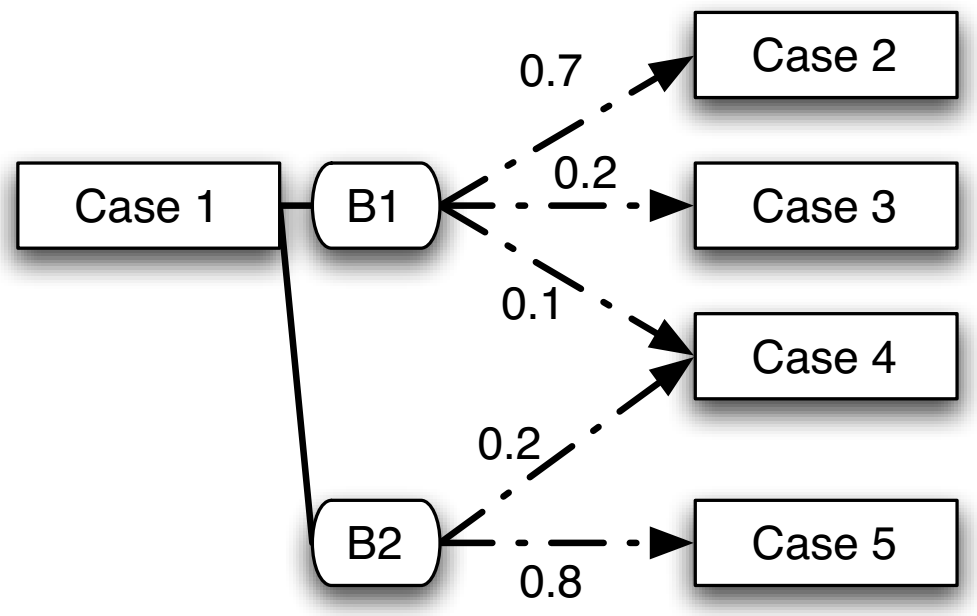}
\label{fig:CSB2}
}
\caption[Case-based reasoning for autonomous vehicle control.]{a) The hierarchical structure of the case base. Solid arrows denote specialization between the cases, and dashed lines link the cases at the same level of specialization. b) the temporal linkage between the cases.}
\label{fig:CSB}
\end{figure}

CBS have been used in robotic applications such as navigation and obstacle avoidance \cite{moorman1992case,urdiales2006purely} and autonomous driving \cite{vacek2007using}. For instance, Vacek \textit{et al.} \cite{vacek2007using} propose a hierarchical structure where each case is built by producing a description of the scene. To do so, first, the perception module identifies the features (e.g. signs, road users, etc.) in the scene. Then, depending on the criticality of each feature, a value between 0 and 1 is associated with it. For example, a person standing on the sidewalk is not as critical as an approaching car from the opposite direction. The combination of all features forms the final description of the situation which in turn characterizes the case. The resulting cases are then stored hierarchically in the memory. Depending on their specialization level, cases are connected either vertically or horizontally (see \figref{fig:CSB1}). In addition, there are temporal linkages between cases that highlight the evolution of a case throughout the time (\figref{fig:CSB2}).

\subsection{Probabilistic Reasoning Systems}

One particular challenge for reasoning in autonomous robotic applications is dealing with uncertainties. This problem arises when the belief of the system about the world is incomplete which can be caused by numerous factors such as noise in sensory input. To tackle this problem, probabilistic approaches such as Bayesian inference \cite{rasouli2014visual} and Markov decision processes (MDPs) \cite{bandyopadhyay2013intention,bai2015intention} are widely used. In robotic control applications, MDPs and their variants such as partially observable MDPs (POMDP), are widely used. An MDP represents a domain by a finite set of states, a finite set of actions, and an action model, which specifies the probability of performing each action in the given state that results in a transition to the next state \cite{hertzberg2008ai}. For instance, in \cite{bai2015intention}, a POMDP is used to control the speed of the vehicle along the planned path while taking into account the behavior of pedestrians. In this formulation, each state contains the position, orientation and instantaneous speed of the vehicle as well as the pedestrian's position, goal and instantaneous speed. The actions to choose from are discretized into \textit{ACCELERATE, MAINTAIN} and \textit{DECELERATE}.

\section{Visual Attention: Focusing on Relevance}

The role of attention in optimizing visual processing tasks such as visual search is undeniable \cite{tsotsos1990analyzing, tsotsos1995toward}. Practical systems are no exception and are shown to benefit from attention mechanisms which reduce the processing complexity of perceptual tasks, and as a result save computational resources. This, in particular, is important for practical systems such as robots or autonomous cars which are dealing with complex real-world scenarios and have limited resources \cite{potapova2017survey}. 

Given the importance of visual attention in machine vision, during the past decades a great deal of works have been done on developing computational attention models for applications such as object detection and recognition \cite{ba2014multiple}, image captioning \cite{xu2015show} and robotics \cite{rasouli2017integrating}. 

There are a large number of computational attention models (mainly on visual saliency), the majority of which are generic in a sense that they can be used in different applications simply as they are or with some minor tuning for particular tasks. Going into details and discussing the body of work on computational attention models is beyond the scope of this report. As a result,  we limit our discussion to first, describing the categories of visual saliency models and some of the common techniques to develop them, and second, reviewing some of the techniques designed for intelligent driving systems. At the end we briefly  argue whether the current approaches to visual attention suffice for autonomous driving applications.

\subsection{Computational Models of Visual Attention}

In computer vision community, the majority of the work on attention have been realized in the form of generating visual saliency maps. Depending on the objective of saliency and the way the image is processed, visual saliency algorithms can be divided into two categories, \textit{bottom-up} and \textit{top-down} algorithms \cite{bylinskii2015towards}. 

As the name implies, bottom-up algorithms are data driven and measure saliency based on detecting the regions of the image that stand out in comparison to the rest of the scene. The saliency generated by these maps can be either  object-based (identifying the dominant objects) or fixation-based (predicting human eye fixations).

To generate bottom-up saliency maps, input images are commonly decomposed into features, and then the distribution of these features is measured, either locally or globally, to identify the uniqueness of a given region in the image. The types of features used for this purpose include superpixels \cite{Chang2011fusing,Jiang2011salobj}, low-level features such as color and intensity \cite{Itti1998rapidscene}, higher level learned features such as Independent Component Analysis (ICA) \cite{Bruce2007aim}, or in more recent works convolutional features using deep learning techniques \cite{Li2015cnnsal,Zhao2015deepsal}.

In contrast to bottom-up algorithms, top-down saliency models identify saliency as the regions with similar properties to a specific object or task \cite{Cave1999top}. These algorithms often use features such as SIFT in conjunction with learning techniques such as SVM \cite{Zhu2014context}, conditional random fields (CRF) \cite{JYang2012tpdown} or neural networks \cite{He2016deeptop} to determine the presence of the object of interest based on a combination of pre-learned features. 

Some top-down algorithms take into account the nature of the task to generate saliency maps. Such works define the task as a search problem in which the objective is to find regions with similar properties to the object of interest. To achieve this some techniques use neural network architectures that are trained on different object classes. At the run time, the task (i.e. object class) is provided to the network, for instance in the form of object exemplars \cite{He2016deeptop} or labels \cite{zhang2016top}, and in return the network highlights areas that have similar properties to the object.

Besides static images, a number of models also estimate saliency in videos. Similar to the previous techniques, these models are either data driven in which, for example, optical flow generated based on image sequences is used to characterize the scene \cite{wang2018video} or are task driven in a sense that they attempt to learn the human gaze patterns in various scenarios \cite{leifman2016learning}.  

For further details on general saliency algorithms please refer to \cite{Borji2013quantanalysis, bruce2015computational,filipe2013human} and for 3D visual saliency in robotics refer to \cite{potapova2017survey}.

\subsection{Attention in Intelligent Driving Systems}

The use of attention mechanisms in autonomous driving goes as far back as the early 90s in the work of Ernst Dickmanns and his colleagues \cite{dickmanns1990integrated}. Their autonomous van, VaMoRs, was equipped with a dual-focal length camera mounted on a pan-tilt unit. Using this system, the wide angle image was used to analyze global features such as the road boundaries, whereas, the enlarged image was used for focusing on objects and obstacles ahead. The pan-tilt unit allowed the vehicle to maintain its focus of attention on the center of the road in the cases when it was turning or entering a tight curve. The primary purpose of this attention mechanism was to deal with the limitations of camera sensors (e.g. narrow viewing angle) at the time. Today, the new technological advancements in designing wide view angle cameras make the use of such mechanisms obsolete. 

In more recent years, computational attention models have found their way into intelligent assistive driving systems. These works commonly rely on some form of inside looking cameras to monitor the changes in the drivers attention, and if necessary, alert the driver or control the vehicle in the case of inattention. The types of sensors used in such systems may vary ranging from a single looking camera \cite{doshi2010attention} to a distributed network of cameras for high resolution 3D face reconstruction and tracking \cite{tawari2014robust, tawari2014looking}. 

Some assistive driving systems go one step further and match the focus of driver's attention to objects surrounding the vehicle to make better sense of the situation \cite{tanishige2014prediction,tawari2014attention}. For example, Tawari \etal \cite{tawari2014attention} use a head mounted camera and a Google Glass (which records eye movements) to simultaneously detect the object of interest (in this case a pedestrian) and determine whether it is in the center of the driver's attention. 

More recent works attempt to simulate human attention patterns in autonomous driving systems. The focus of these works is on developing saliency models to imitate human fixation patterns, and consequently identify regions of interest for further processing \cite{deng2016does,palazzi2017learning,tawari2017computational}.

In \cite{deng2016does}, the authors recorded the fixation patterns of 40 subjects with various driving background by showing them 100 still traffic images. Based on their observation, they concluded that the majority of the subjects fixated on the vanishing point of the road. Using this knowledge, they propose a saliency algorithm consisting of a bottom-up algorithm with a top-down bias on the vanishing point of the road. 

Since still images are not proper representatives of driving task, more recent algorithms rely on video datasets for estimating saliency.  One such dataset is Dr(Eye)ve \cite{alletto2016dr}, which is a collection of 74 video sequences of 5 minute long, each recorded from 8 drivers during actual driving experience. The dataset contains 550K frames and is accompanied with fixations of the drivers as well as GPS information, car speed and car course. In a subsequent work \cite{palazzi2017learning}, the scholars behind Dr(Eye)ve show how using an off-the-shelf CNN algorithm trained on their data can predict human fixation patterns in traffic scenes. Tawari and Kang \cite{tawari2017computational} further improve fixation predictions on Dr(Eye)ve by taking into account the task of driving. In their algorithm, the authors used yaw rate, which is an indicative of events such as turning, as a prior to narrow the predictions to more relevant areas.

\subsection{Are the Vehicles Attentive Enough?}

Since attention plays a key role in optimizing visual perception in the machine, in this section we briefly discuss the limitations of computational attention models in autonomous driving and point out possible future directions. 

\subsubsection*{Data collection limitations}

One problem with collecting fixation data is the reproducibility of driving conditions for the drivers, given the dynamic nature of the environment. As a result, driving fixation datasets such as Dr(Eye)ve \cite{alletto2016dr} provide only one instance of driving on a given road. 

Having only one fixation sample is problematic and is not representative of general human fixation patterns. As discussed earlier in Section \ref{the_role_of_context} the allocation of attention is highly subjective, meaning that it depends on the characteristics of the driver (e.g. novice vs expert \cite{underwood2003visual} or culture \cite{amer2017cultural}).

Another potential issue is with the way fixations are recorded These datasets often record the eye movements of drivers which is a form of overt visual attention. However, going back to psychological studies on human attention, we see that attentional fixations can be covert meaning that humans often internally focus on a particular region of a scene without any explicit eye movement \cite{wright1998visual,bruce2015computational}. This fact was also evident in the studies of visual attention in driving where it was argued that expert drivers tend to use their peripheral vision to focus on objects, reducing the need for eye movements \cite{underwood2003visual}. As one would expect, measuring covert fixation patterns is extremely difficult and not recorded in current datasets.
 
Last but not least, the content of the datasets is not representative of complex driving scenarios. Dr(Eye)ve \cite{alletto2016dr}, which to the best of our knowledge is the only publicly available driving fixation dataset, mainly contains empty rural roads and lacks the presence of road users or other environmental factors. Given the nature of the dataset, the fixation patterns are highly biased towards vanishing point of roads.

\subsubsection*{The limitations of computational models}
\begin{figure}[!t]
\centering
%\captionsetup[subfigure]{labelformat=empty}
\subfloat[]{
\includegraphics[width=0.7\textwidth]{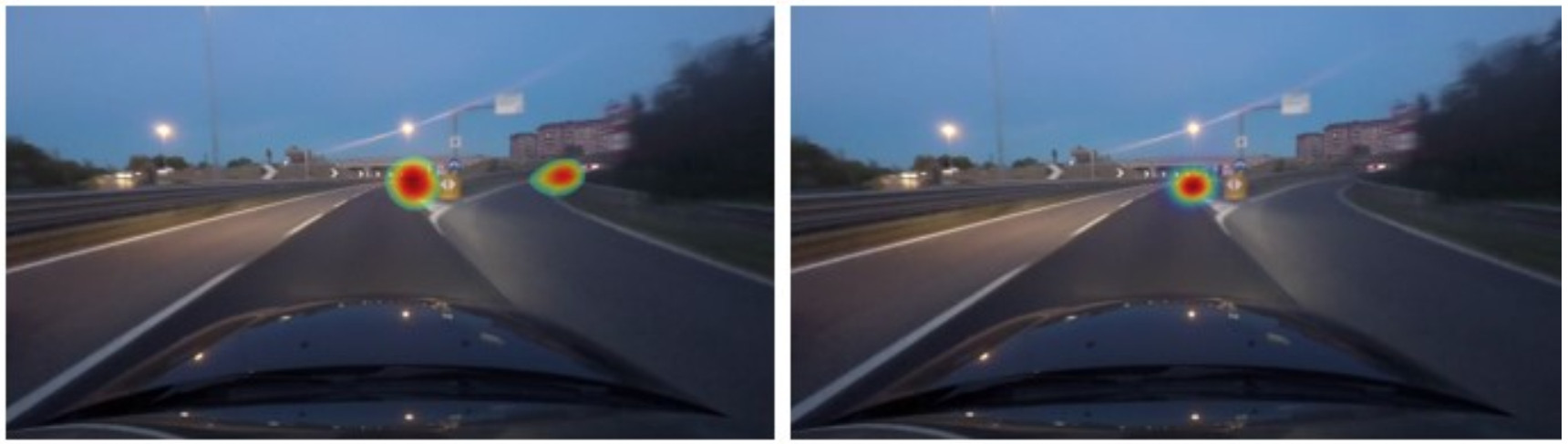}
 }\\
\subfloat[]{
\includegraphics[width=0.7\textwidth]{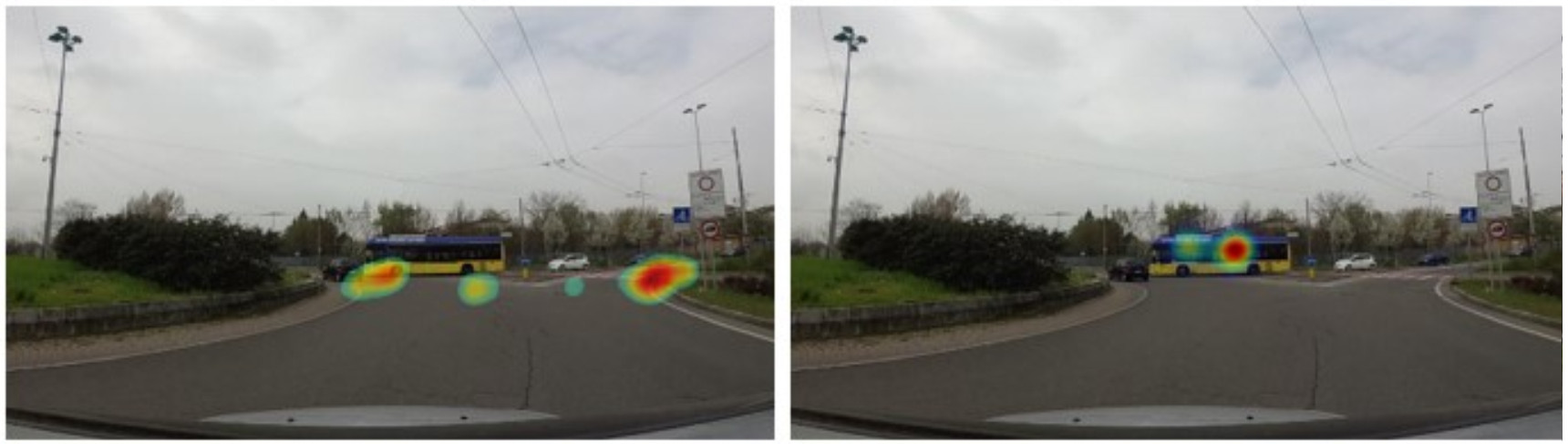}
\label{fig:task_driving_attend_bus} }
\caption[Examples of false attention prediction in driving scene 1]{Examples of fixation predictions of the algorithm in \cite{tawari2017computational} (\textit{left}) and human ground truth data (\textit{right}). }
\label{fig:task_driving_attend}
\end{figure}

As we discussed earlier (see Section \ref{attenion_role}), there is strong evidence that attention is mainly biased by the task. The nature of the task can influence where the human looks, the way his visual receptors are optimized, and how the features are represented and processed internally. 

Unfortunately, in the context of autonomous driving, the role of task in attention has been largely ignored. Some works such as \cite{tawari2017computational} attempted to capture the influence of task on the focus of attention, but they failed in more complex scenarios. \figref{fig:task_driving_attend} shows two examples from \cite{tawari2017computational} indicating why the task is important in fixation prediction. As can be seen, in both of these instances the vehicle is approaching a branching road. In these cases the car is maintaining its path on the main road, hence the driver's attention is focused on the road ahead. The algorithm, on the other hand, falsely predicts fixations on the vanishing points of both branches. This means the algorithm does not take into account the driver's task.

Another implication of \figref{fig:task_driving_attend} is that the data (as mentioned earlier) is highly biased towards vanishing points of the road. For example, in \figref{fig:task_driving_attend_bus} the driver is clearly focusing on the yellow bus as it is turning in the path of the vehicle whereas the algorithm focuses on the vanishing points instead.

\begin{figure}[!t]
\centering
%\captionsetup[subfigure]{labelformat=empty}
\subfloat[]{
\includegraphics[width=0.7\textwidth]{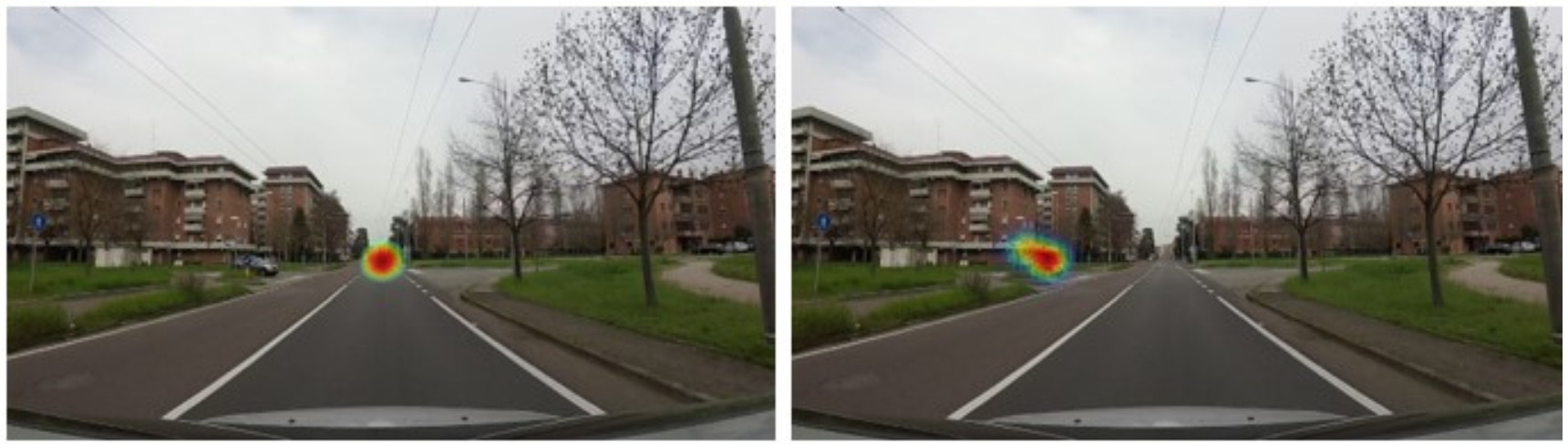}
 \label{fig:task_driving_attend_car}}\\
\subfloat[]{
\includegraphics[width=0.7\textwidth]{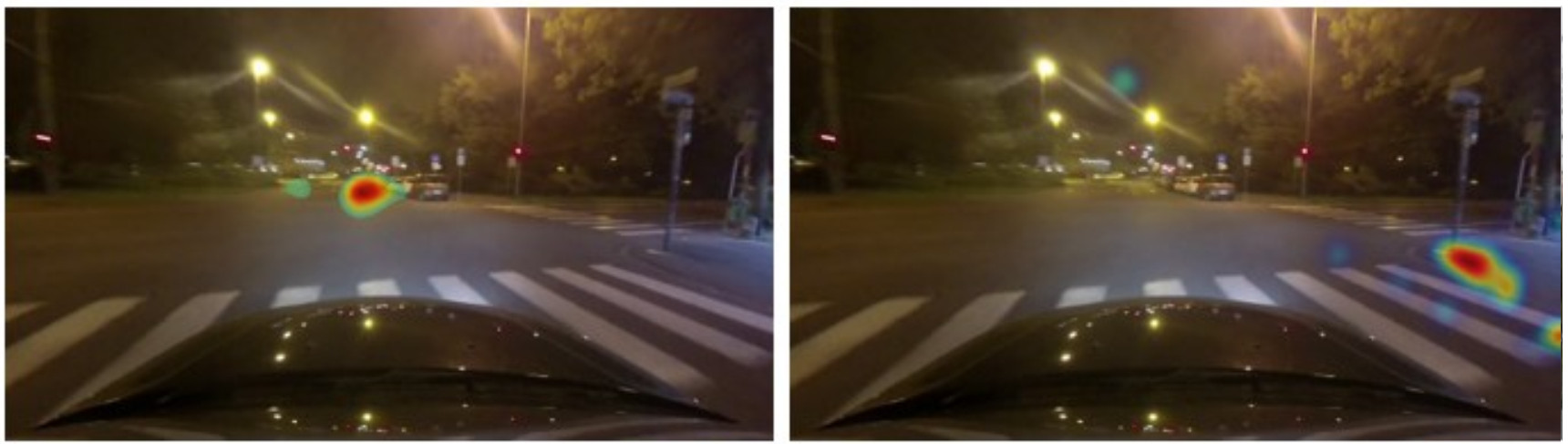}
\label{fig:task_driving_attend_sign} }
\caption[Examples of false attention prediction in driving scene 2]{Examples of fixation predictions of the algorithm in \cite{tawari2017computational} (\textit{left}) and human ground truth data (\textit{right}). }
\label{fig:task_driving_attend_context}
\end{figure}

In the psychology literature, it is also argued that context can be a good predictor of human focus of attention \cite{bruce2015computational}. In driving, as discussed before, context comprises other road users, road structure or traffic signs. Referring again to \cite{tawari2017computational}, we can see how the lack of considerations for context can result in misprediction of human fixations. For instance, in \figref{fig:task_driving_attend_car} we can see that a car is approaching the road, hence, the driver is focusing on that car. Similarly, in \figref{fig:task_driving_attend_sign} the driver is paying attention to the pedestrian crosswalk below the traffic sign. In both cases, however, the algorithm falsely picks the vanishing points as the focus of attention.

\section{Joint Attention, Traffic Interaction, and More...: Case Studies}

Thus far, we have reviewed various theoretical and technical aspects of pedestrian behavior understanding in terms of what we have to look for in the scene (e.g. contextual elements), where to look for them (e.g. attention algorithms) and how to identify them (e.g. visual detection algorithms).

In this section, with the aid of two case studies, we will relate the previous findings to the problem of driver-pedestrian interaction. More specifically, we will analyze the problem to identify what has to be done, and how (un)suitable the current state of the art algorithms are for solving different aspects of the problem. 

\subsection{Driver-pedestrian interaction in a complex traffic scene}

Before discussing the case studies, it is worth formulating the problem of driver-pedestrian interaction in a typical traffic scenario. Overall, we can divide the interaction problem into three phases:

\begin{enumerate}
\item Detection: It is important to reliably identify various traffic scene elements that are potentially relevant to the task of interaction. The detection process can be generic or directed (attention). In a generic approach, we identify all the traffic elements in the scene whereas in a directed approach we perform detection on the parts of the scene that are relevant to the task. 

\item Selection: The detected elements have to be evaluated to determine which ones are relevant, even if a directed detection approach is used. For instance, we might use a directed approach to detect pedestrians that are in close proximity to the vehicle or its path. Next, we need to determine which of these pedestrians are potentially going to interact with the driver based on, for example, their direction of motion, the state of the traffic signal or head orientation.

\item Inference and negotiation: Once we selected the potential candidates and relevant contextual elements, we need to determine what the candidates are going to do next. This is where joint attention plays a significant role. It helps to identify the center of focus of the pedestrian (e.g. by following their head/eye movements) and interpret their actions according to their objectives and bodily movements. It is also involved in negotiating  the right of way with the pedestrian, by using explicit (e.g. hand gesture) or implicit (e.g. change of pace) signals. The end result is an agreement (or a contract) between the driver and the pedestrian, which is demonstrated by changing or maintaining the current behavior of the involved parties.
\end{enumerate}

The following case studies will focus on different aspects of the problem as described above. Case 1 mainly addresses the problem of detection and selection and case 2 focuses on the problem of joint attention.

\subsection{Case 1: Making a right turn at a four-way intersection}
\begin{figure}[!t]
\centering
%\captionsetup[subfigure]{labelformat=empty}
\subfloat[A view of the scene]{
\includegraphics[width=0.5\textwidth]{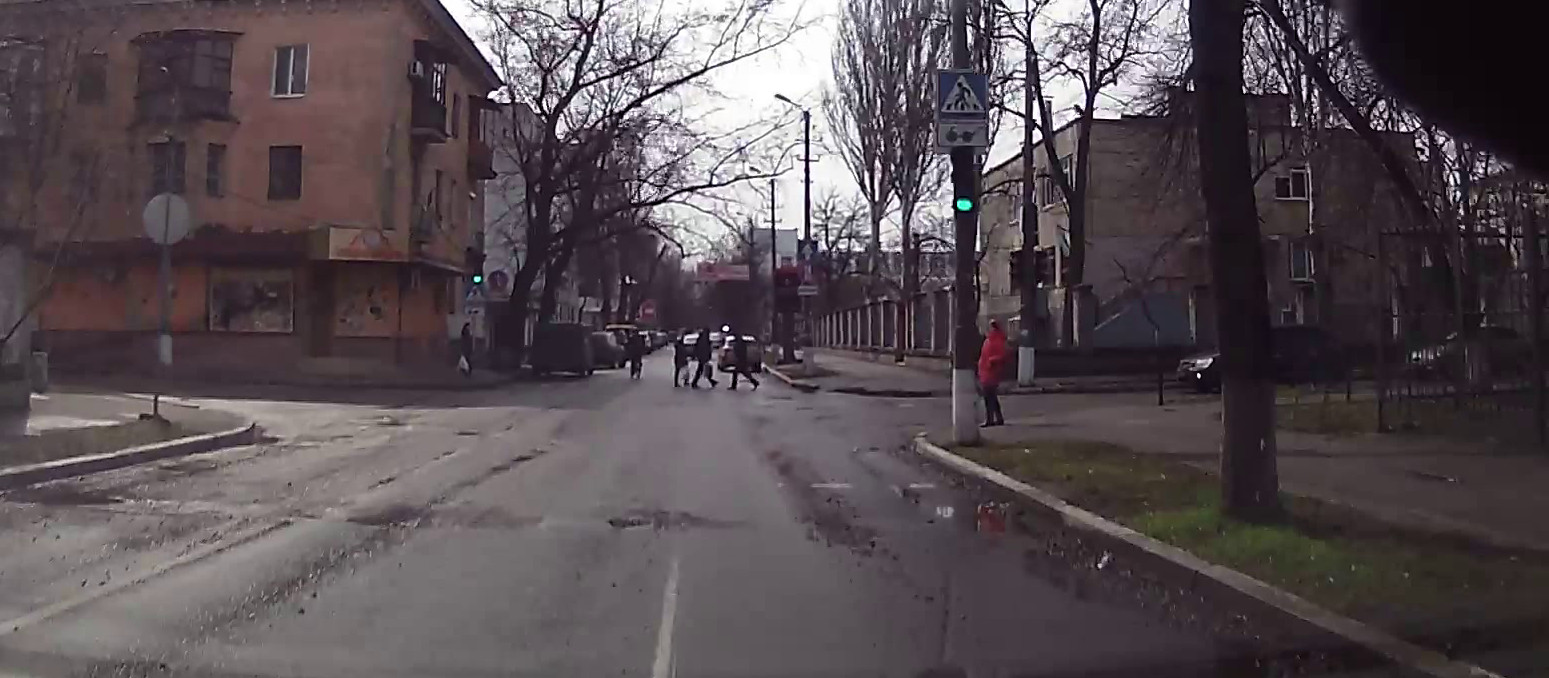}
 \label{fig:case1}}
\subfloat[Potentially relevant regions]{
\includegraphics[width=0.5\textwidth]{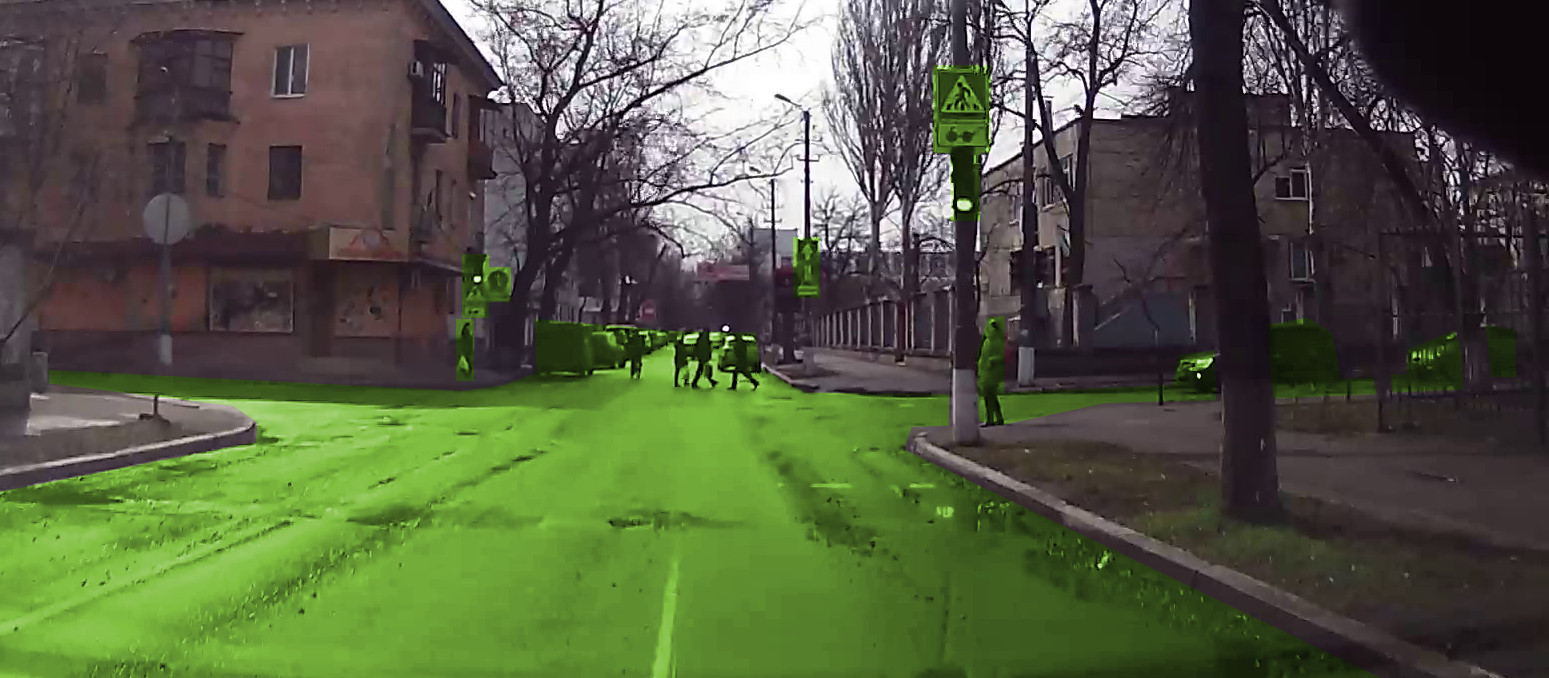}
\label{fig:case1_all_context} }\\
\subfloat[Actual relevant regions]{
\includegraphics[width=0.5\textwidth]{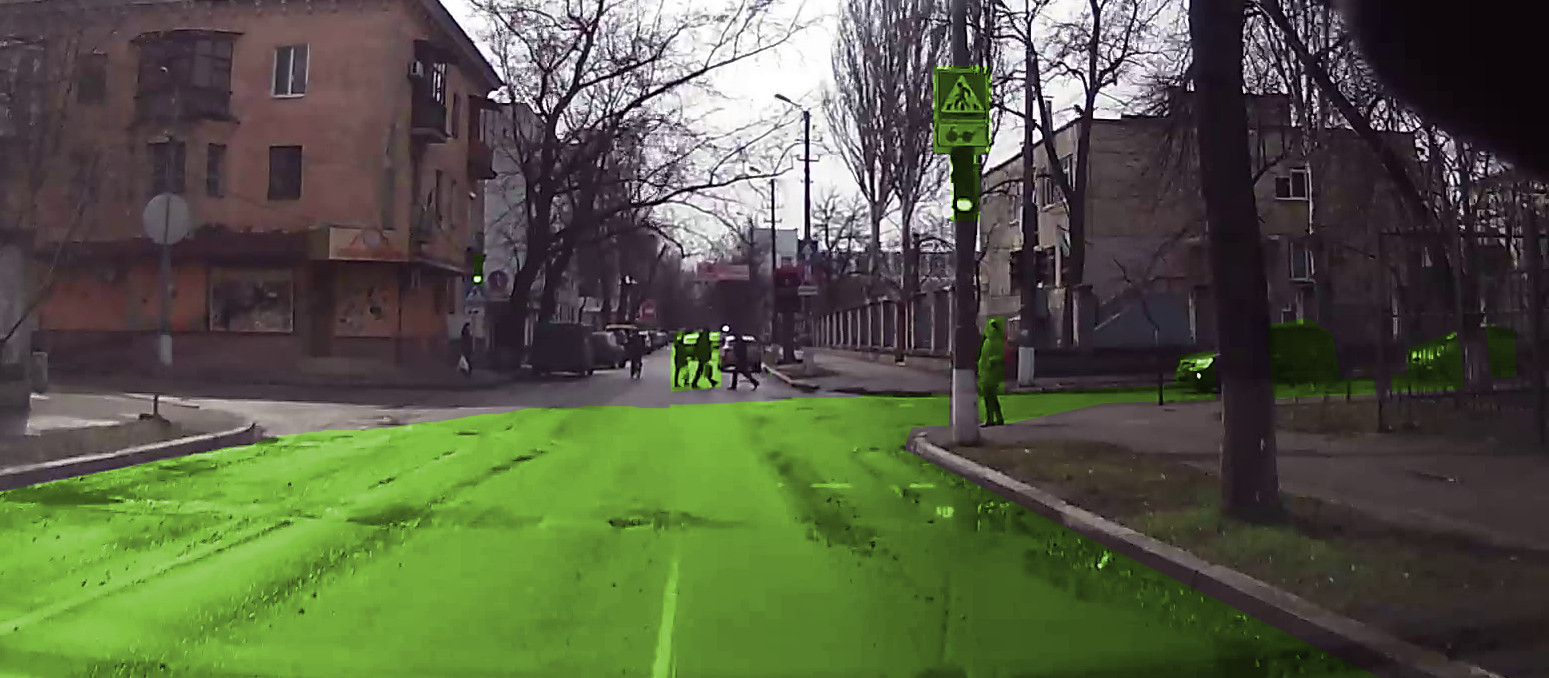}
\label{fig:case1_relevant_context} }
\subfloat[Priority of attention]{
\includegraphics[width=0.5\textwidth]{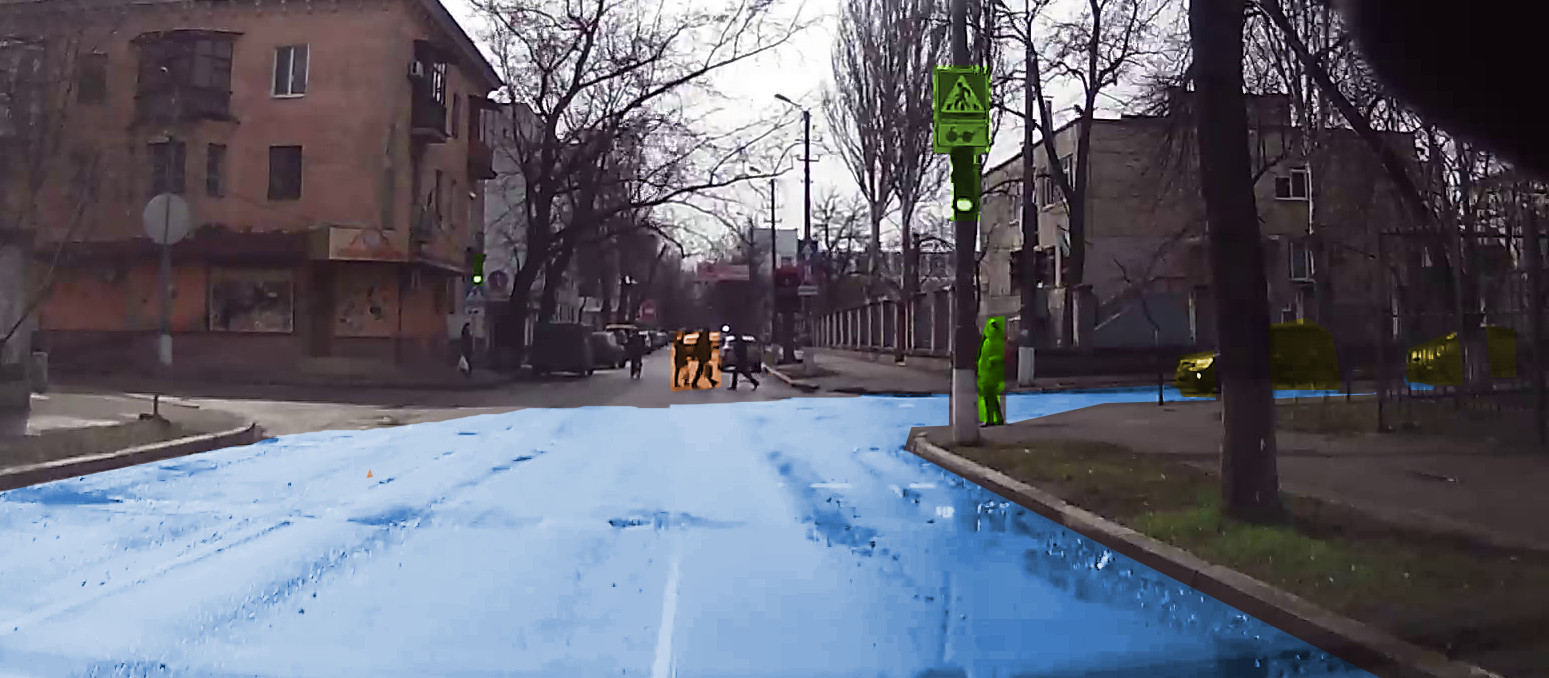}
\label{fig:case1_t1_relevant_context} }\\
\subfloat[Priority of attention after one second]{
\includegraphics[width=0.5\textwidth]{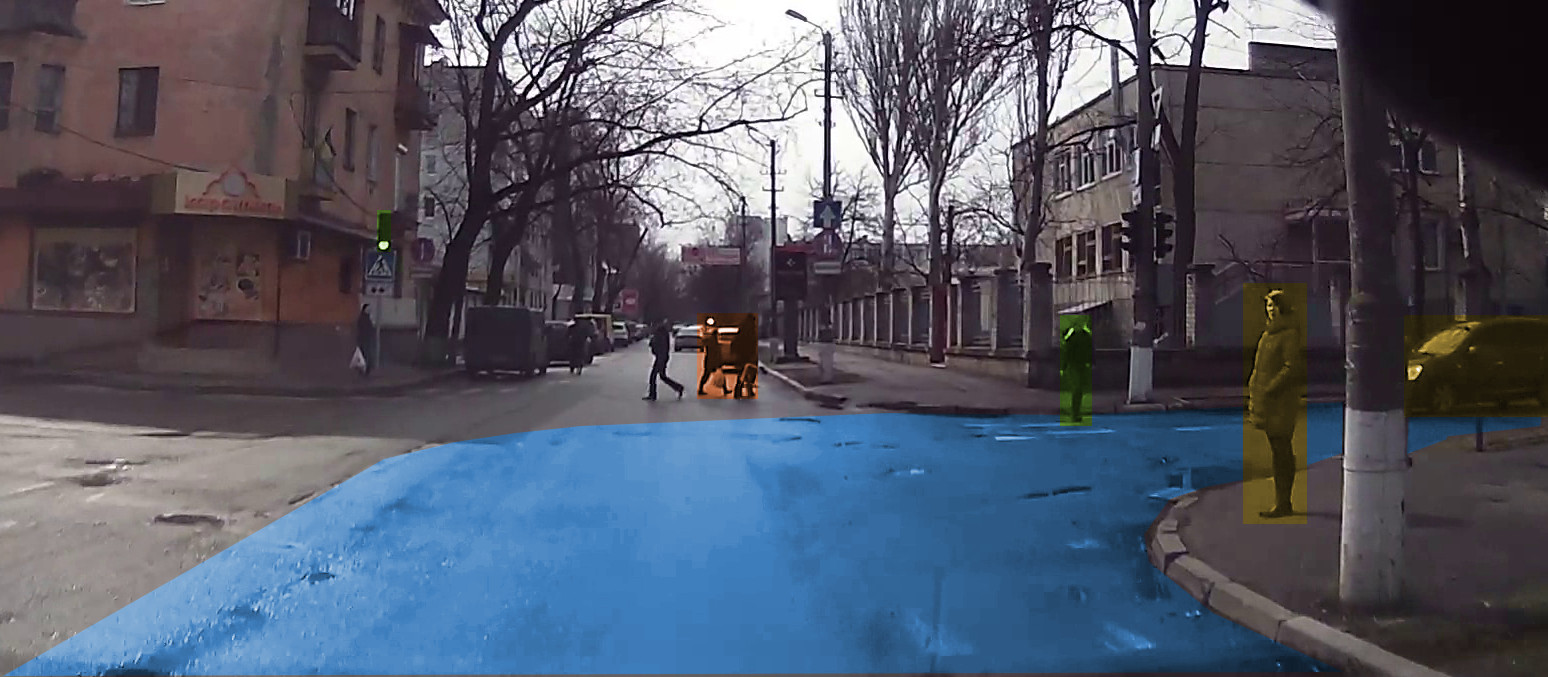}
\label{fig:case1_t2_relevant_context} }
\subfloat[Priority of attention after two seconds]{
\includegraphics[width=0.5\textwidth]{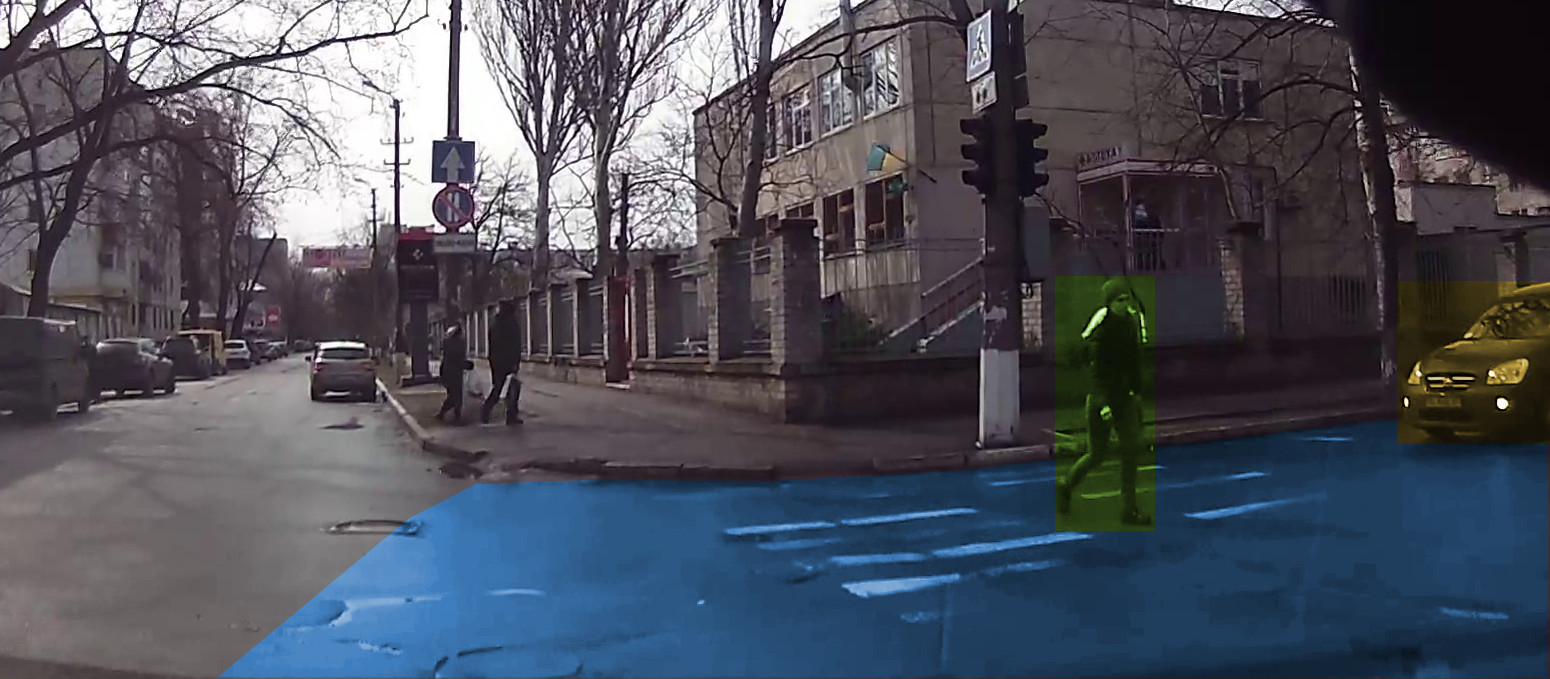}
\label{fig:case1_t3_relevant} }

\caption[Attention allocation in a traffic scene - case1.]{Attention allocation and behavioral understanding in a traffic scene. In sub-figures \textit{d-f} colors indicate the priority of attention, from the highest to the lowest:  {\color{green}\textit{green}}, {\color{blue}\textit{blue}}, {\color{yellow}\textit{yellow}} and {\color{red}\textit{red}}}.
\label{fig:case1_driving_scenario}
\end{figure}

We begin with a typical interaction scenario where the vehicle is driving towards a four-way intersection (see \figref{fig:case1}) and a pedestrian is standing near the curb waiting to cross the street. We can assume that the primary objective of the vehicle is safety, that is avoiding collisions with other road users or the infrastructure.

\subsubsection{Perceiving the scene}
\begin{figure}[!ht]
\centering
\subfloat[Rainy]{\includegraphics[width=0.3\textwidth]{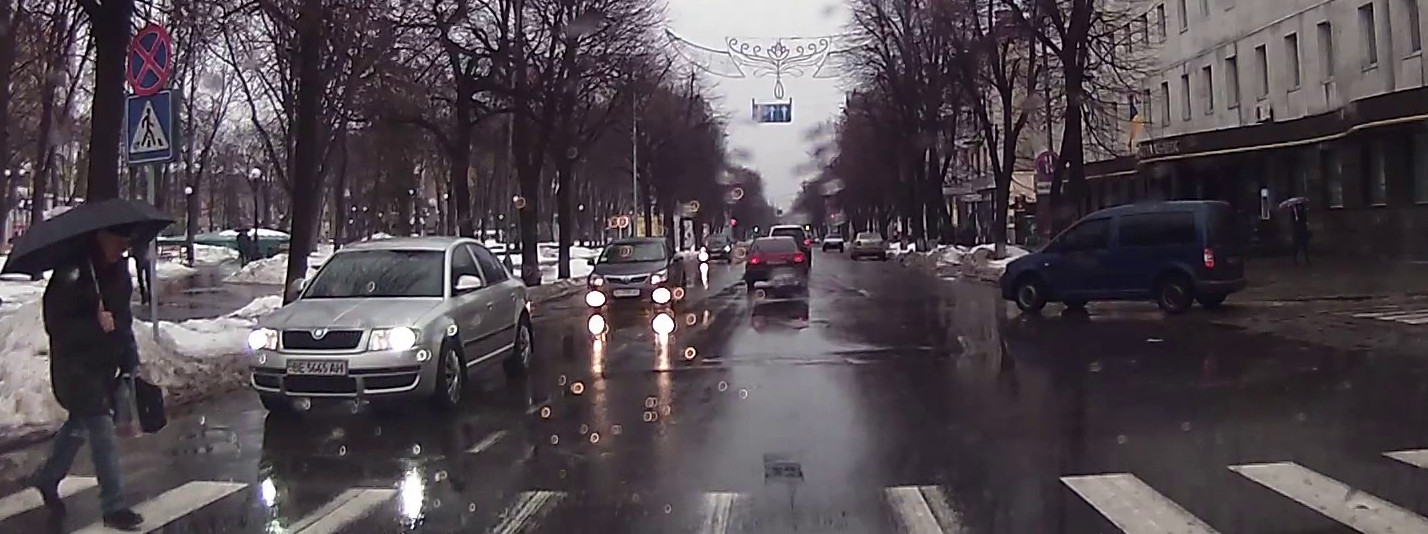}
\label{fig:rainy}}
\subfloat[Snowy]{\includegraphics[width=0.3\textwidth]{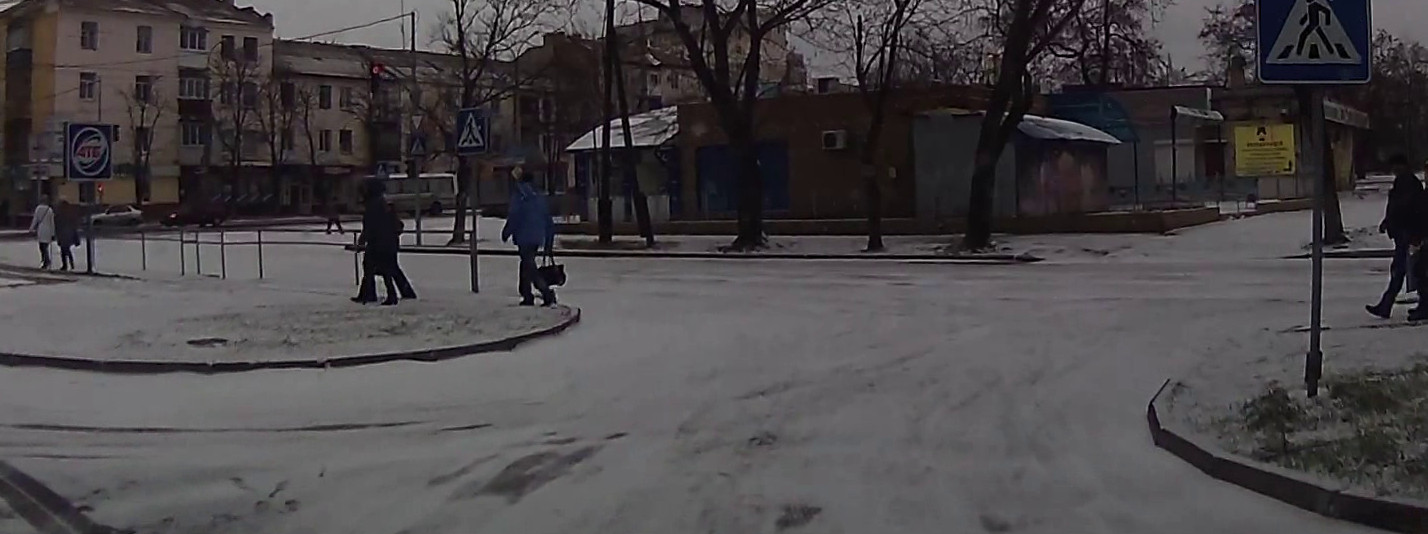}
\label{fig:snowy}}
\subfloat[Sunny]{\includegraphics[width=0.3\textwidth]{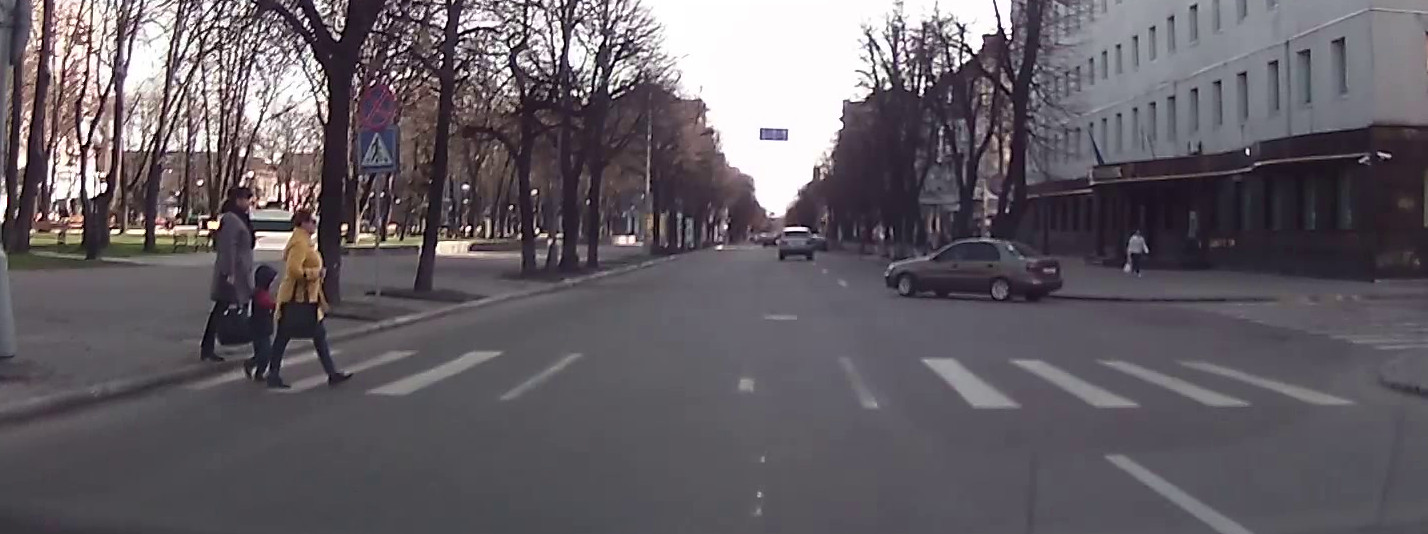}
\label{fig:sunny}}
\caption[Traffic scenes with various weather conditions]{Traffic scenes with various weather conditions.}
\label{fig:weather}
\end{figure} 

\figref{fig:case1_all_context} highlights the potentially relevant contextual elements in the scene. The current state-of-the-art visual perception algorithms are capable of detecting \cite{girshick2015fast,redmon2016you, chen2016monocular} and recognizing \cite{krizhevsky2012imagenet,he2016deep} traffic scene elements such as pedestrians, vehicles and signs with a reasonable margin of error. 

However, there are still challenges that need to be addressed in traffic scene understanding research. The object detection algorithms for traffic scenes are commonly trained and evaluated using datasets, such as KITTI \cite{Geiger2013IJRR} or Cityscapes \cite{Cordts_2016_CVPR} or, in the case of pedestrian detection, Caltech \cite{dollarCVPR09peds}, all of which are collected in daylight under clear sky conditions. Such lack of variability in weather and lighting conditions raises the concern of how reliable these methods would be in real traffic scenarios.  As illustrated in \figref{fig:weather}, we can see, for example, the strong reflection of light on the road surface in a rainy weather (\figref{fig:rainy}) or appearance changes of pedestrians in a snowy day (\figref{fig:snowy}), none of which exists in a typical sunny day (\figref{fig:sunny}). Therefore, it is important to train object detection algorithms using datasets that capture the variability of traffic scenes.

\subsubsection{Where to attend}

It is easy to see that identifying all the contextual elements can be computationally expensive. Such a holistic view of the scene can increase the chance of error in the detection task and complicates reasoning processes about the behavior of the agents present in the scene.

To limit the focus of attention, we need to identify the contextual elements that are \textit{relevant} \cite{sperber1987precis}. The relevance is defined based on the nature of the task, which, in our case, is the vehicle making a right turn. One can see that, simply by taking into account the task, a big portion of initially identified elements such as pedestrians walking on the other side of the street, the road straight ahead or to  the left, and the parked vehicles are discarded as they are deemed irrelevant (see \figref{fig:case1_relevant_context}).

The role of task on attention allocation has been widely studied in behavioral psychology, but unfortunately very few computational models of attention take task into consideration. For instance, in \cite{zhu2016traffic} the authors use a spatial prior for sign detection to limit the detection to regions on the borders of the road. As for top-down models, they focus on highlighting regions with properties coherent with the object of interest but fail to prioritize the instances of the same class according to the task in hand \cite{bylinskii2016should}. In the case of models of attention for driving, we also saw that the models which simply learn the driver's gaze behavior \cite{palazzi2017learning} fail to predict regions of interest in more complex  scenarios, where the vehicle is not driving on a straight road. In addition, the same issue was apparent in the case of the model that attempts to consider the task \cite{tawari2017computational} in the form of yaw rate and the structure of the street. We saw in the examples earlier (see Figures \ref{fig:task_driving_attend}-\ref{fig:task_driving_attend_context}) that using such simple cues is  certainly not enough to identify regions of importance. 

\subsubsection{Prioritizing the attention}

Although discarding all irrelevant elements speeds up the analysis of the scene, it does not suffice for a fast dynamic task such as driving. For instance, as illustrated in \figref{fig:case1_relevant_context}, the relevant contexts include three pedestrians, 2 vehicles and the traffic lights. Tracking and analyzing all these elements can be expensive, thus might not be possible in a timely manner to avoid collision.

A solution to the above problem is to prioritize the objects from the highest to the lowest importance. Consider the elements illustrated in \figref{fig:case1_t1_relevant_context}. The vehicle is approaching the intersection to turn right. So the immediate relevant contextual elements are traffic signs and lights indicating whether the vehicle is allowed to pass or not. Of course the signs on the other side of the intersection are irrelevant as they concern drivers who are driving straight ahead.

In our example the road users possess different levels of priority. The first participant important to the driver is the pedestrian wearing a red coat and standing near the traffic light pole. She is intending to cross in front of the vehicle, therefore the primary focus should be on what she might do next. Other road users are less important. For instance, the cars stopped the red light at the intersection are driving in a different direction and might not pose any threat unless they perform a reckless behavior. The pedestrians already crossing the intersection on the road ahead have the lowest priority.  They are moving to the right side of the street (\figref{fig:case1_t1_relevant_context} and can potentially cross the street again, however, their speed is much slower than the vehicle and there is a very low chance that they'll be crossing the street by the time the vehicle is turning. 

Last but not least is the street. Generally speaking, the road on which the vehicle is driving or intending to drive has a high priority. However, in the given context the road's state mainly depends on dynamic objects that are in the driver's center of attention, hence compared to the traffic light or the pedestrian closest to the vehicle, it has a lower priority.

The problem of attending to the objects according to their degree of priority has extensively been  studied in human behavioral studies \cite{wright1998visual,bruce2015computational}. Bottom-up computational models define priority in terms of how distinctive various features in the scene are \cite{Bruce2007aim,Li2015cnnsal}. Top-down algorithms \cite{JYang2012tpdown,He2016deeptop,zhang2016top}, on the other hand, define priority in terms of similarity of the image regions to the object of interest. These models, however, define the task very narrowly as visual search based on a template or a label. These algorithms lack the the capability of understanding the relationship between various objects in the scene and the influence of temporal changes on the task (e.g. changes in traffic light color or the state of pedestrians).

\subsubsection{Evolving the attention}
 
In a dynamic task such as driving, the focus of attention has to evolve according to the changes in the objective of the task. Going back to the example in \figref{fig:case1_driving_scenario}, we can see that the pedestrian wearing a red coat is more likely to cross the street when the car is far away (\figref{fig:case1_t1_relevant_context}) than when the car is in a close proximity (\figref{fig:case1_t2_relevant_context}). Therefore, a lower priority is assigned to the pedestrian in the second time step. The same is true about the pedestrians at the back in \figref{fig:case1_t1_relevant_context} who are not relevant anymore in \figref{fig:case1_t3_relevant} as they are clearly not going to a cross the vehicle's path. At the same time we see that a new pedestrian is appearing in the scene (\figref{fig:case1_t2_relevant_context}) who is crossing the path of the car,and thus has the highest priority at this stage.

Video saliency algorithms \cite{wang2018video,leifman2016learning}, to some extent, deal with temporal evolution of attention by taking into account optical flow or fixation data from human subjects. The bottom-up algorithms, of course, are not suitable for such a task as they are only capable of identifying certain patterns in the image, e.g. moving objects or activities, without any connection to the context within which they occurred. 

As we saw earlier, the algorithms learning human fixation patterns \cite{tawari2017computational,palazzi2017learning} also fail to highlight what is important in a given driving context due to a number of reasons. The datasets used in these techniques are unbalanced and the occurrence of traffic interactions such as crossing intersections are rarer compared to driving on regular road in which the driver looks straight ahead. Furthermore, there is a high variability in driving tasks, for example, different types of intersections, different numbers of pedestrians or cars present in the scene or even weather conditions. Collecting data for all of these scenarios is a daunting task. Finally, these algorithms do not account for the nature of the task (some only primitively in the form of route \cite{tawari2017computational}) and context when predicting the focus of attention.

\subsubsection{Foreseeing the future}

The dynamic nature of driving makes the problem of attention and scene understanding quite complex. Given that driving is a temporal task the relevant context goes beyond the static configuration of the elements observed in the scene. As we discussed earlier in Section \ref{the_role_of_context}, when identifying the context in a dynamic interactive scenario, one should consider the intentions and consequences of the actions performed by the parties involved, in this case the other road users. In some cases actions have to be predicted even in the absence of traffic participants. To explain this further, lets use the example in \figref{fig:case1_driving_scenario}.

We begin with the pedestrian wearing a red coat.  She is standing at the traffic light and looking towards the ego-vehicle. At the current situation (\figref{fig:case1_t1_relevant_context}) she is the first immediate traffic participant that is relevant to the task. Although the traffic light is green (red for the pedestrian) there is still a chance that she will start crossing (similar to the people at the back who are crossing while the light is green for the cars). The pedestrian is looking towards the vehicle, the gap between the pedestrian and the vehicle is sufficiently big and no other vehicle is driving in the opposite direction. In addition, the pedestrian's posture indicates that she is not in a fully static position. Taking all these factors into consideration, she should be the main focus of attention.

After one second, the car is much closer to the pedestrian, she is in a fully static condition and the traffic light remains green for the ego-vehicle. This significantly lowers the chance of crossing, therefore the priority of attending to the pedestrian wearing a red coat is reduced.

Earlier, we saw how intention estimation algorithms \cite{hashimoto2015probability,kooij2014analysis,kooij2014context} predict pedestrian behavior, albeit in a limited context. These algorithms mainly rely on the dynamics of the vehicle and the pedestrian \cite{hashimoto2015probability,kooij2014analysis} and some also take into account pedestrians' awareness by observing their head orientation \cite{kooij2014context}. Although these algorithms can be effective in a number of cases, they have some drawbacks. Mainly relying on the trajectory of the pedestrian means that if there is a discontinuity in the pedestrian's motion or the pedestrian is motionless (which is the case in our example), these algorithms fail to predict upcoming movements.

\begin{figure}[!t]
\centering
\subfloat[]{\includegraphics[width=0.1\textwidth]{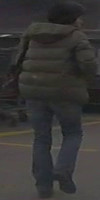}
\label{fig:looking:minor}}
\hspace{0.4cm}
\subfloat[]{\includegraphics[width=0.1\textwidth]{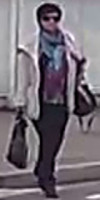}}
\hspace{0.4cm}
\subfloat[]{\includegraphics[width=0.1\textwidth]{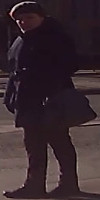}}
\hspace{0.4cm}
\subfloat[]{\includegraphics[width=0.1\textwidth]{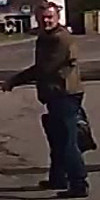}
\label{fig:looking:extreme}}\\
\caption[Examples of pedestrians looking at the traffic.]{Examples of pedestrians looking at the traffic.}

\label{fig:looking}
\end{figure}

Measuring head orientation as a sign of awareness may also be tricky in some scenarios. The variability of head orientation and looking actions by pedestrians is very high (see \figref{fig:looking}) and depends on their positioning or the structure of the scene. In some cases, e.g. \figref{fig:looking:minor}, pedestrians may not explicitly turn their heads towards the vehicle, and instead notice the vehicle with their peripheral vision. Pedestrians' head in some cases may not even be visible enough for performing recognition (see \figref{fig:case1_t1_relevant_context}).

\begin{figure}[!t]
\centering
%\captionsetup[subfigure]{labelformat=empty}
\subfloat[]{
\includegraphics[width=0.5\textwidth]{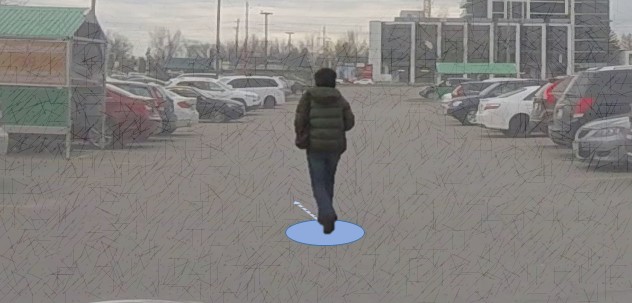}
 \label{fig:case2-1-1}}
\subfloat[]{
\includegraphics[width=0.5\textwidth]{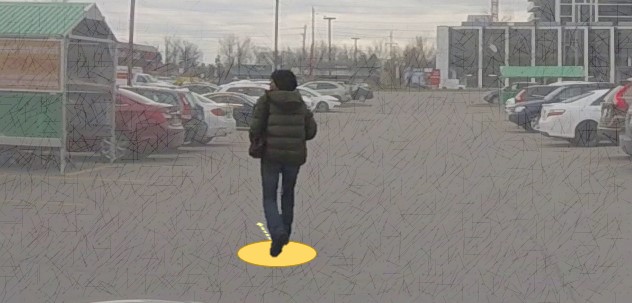}
\label{fig:case2-1-2} }\\
\subfloat[]{
\includegraphics[width=0.5\textwidth]{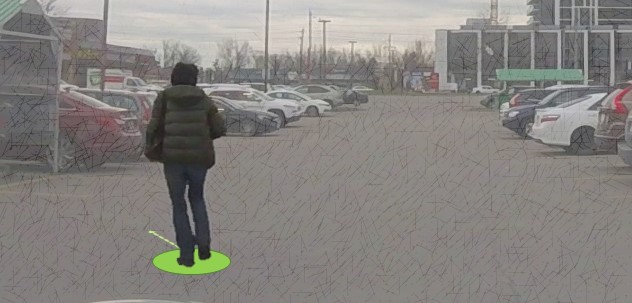}
\label{fig:case2-1-3} }
\subfloat[]{
\includegraphics[width=0.5\textwidth]{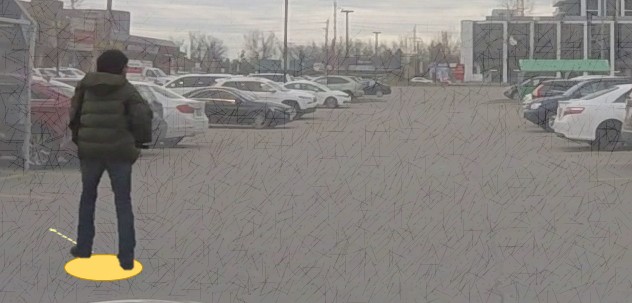}
\label{fig:case2-1-4} }\\
\subfloat[]{
\includegraphics[width=0.5\textwidth]{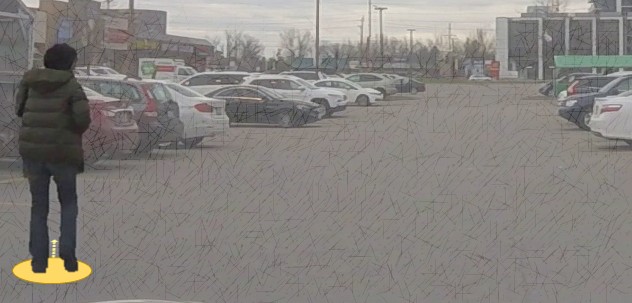}
\label{fig:case2-1-5} }
\subfloat[]{
\includegraphics[width=0.5\textwidth]{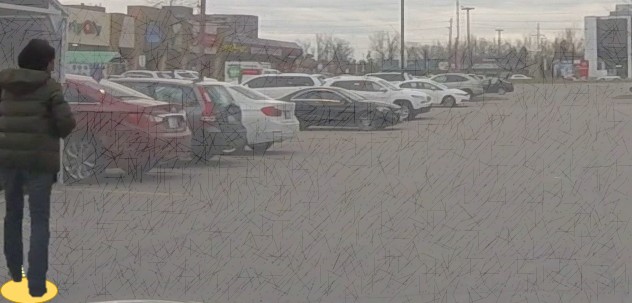}
\label{fig:case2-1-6} }\\
\subfloat[]{
\includegraphics[width=0.5\textwidth]{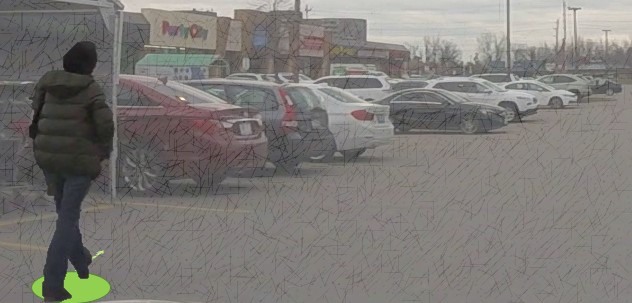}
\label{fig:case2-1-7} }
\subfloat[]{
\includegraphics[width=0.5\textwidth]{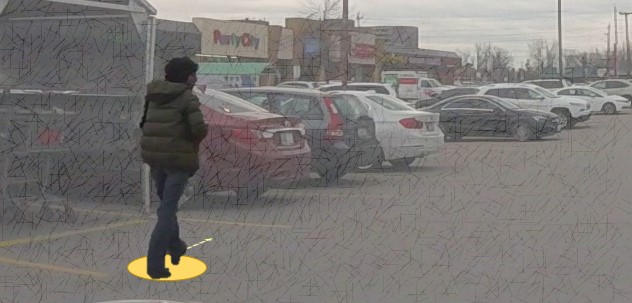}
\label{fig:case2-1-8} }

\caption[An example of pedestrian noticing the vehicle and reacting.]{An example of a pedestrian subtly looking at the vehicle and changing her path. Arrows show the direction of motion and colors indicate whether the pedestrian is moving without noticing the car {\color{blue}\textit{blue}},  the pedestrian is looking at the car while moving {\color{yellow}\textit{yellow}}, and the pedestrian is clearing the path in response to the car {\color{green}\textit{green}}}.
\label{fig:case2-1-subtle-looking}
\vspace{0.4cm}
\end{figure}

To remedy the problem of measuring pedestrian awareness, besides head orientation, one should rely on more implicit clues. For instance, the activity of the pedestrian, e.g. stopping or clearing path, can be an indicator that they noticed or are aware of the vehicles presence (see \figref{fig:case2-1-subtle-looking}). In cases where no obvious changes in the pedestrian's state are observable, we can rely on other clues to measure their awareness. For example, in \figref{fig:case1_t1_relevant_context} the posture of the pedestrian wearing a red coat is a clear indicator  that she is looking towards the vehicle, even though her head or face is not fully visible.

Another drawback of practical intention estimation algorithms is their lack of consideration for contextual information beyond pedestrians' or vehicles' states. Some simulation-based works exploit higher level contextual information such as social forces \cite{hashimoto2015probability} or street structure \cite{brouwer2016comparison} but they do not provide any algorithmic solution for solving the perception problem, assuming everything is detected and also do not address the complex interactions between the traffic participants.

As we mentioned earlier, in some cases, behavior prediction should be performed in the absence of other road users. In road traffic, it is often the case that various elements in the scene are not immediately visible to drivers. For instance, the view of a pedestrian or a car might be blocked by another vehicle, or as in our example (\figref{fig:case1_t2_relevant_context}) by a traffic pole. In such situations, an experienced driver takes note of obstructing elements and adjusts his driving style accordingly.

In assisstive driving context, some scholars address the issue of obstruction in traffic scenes by identifying, for example, the gap between vehicles on the road side and directing the system's attention to those regions as the potential areas where pedestrians might appear \cite{broggi2009new}. For autonomous driving, however, an extension of such approaches is needed to identify any forms of potential obstruction in the scene that might affect the view of the scene.  

\subsection{Case 2: Interaction with pedestrians in a parking lot}

\begin{figure}[!t]
\centering
%\captionsetup[subfigure]{labelformat=empty}
\subfloat[A view of the scene]{
\includegraphics[width=0.5\textwidth]{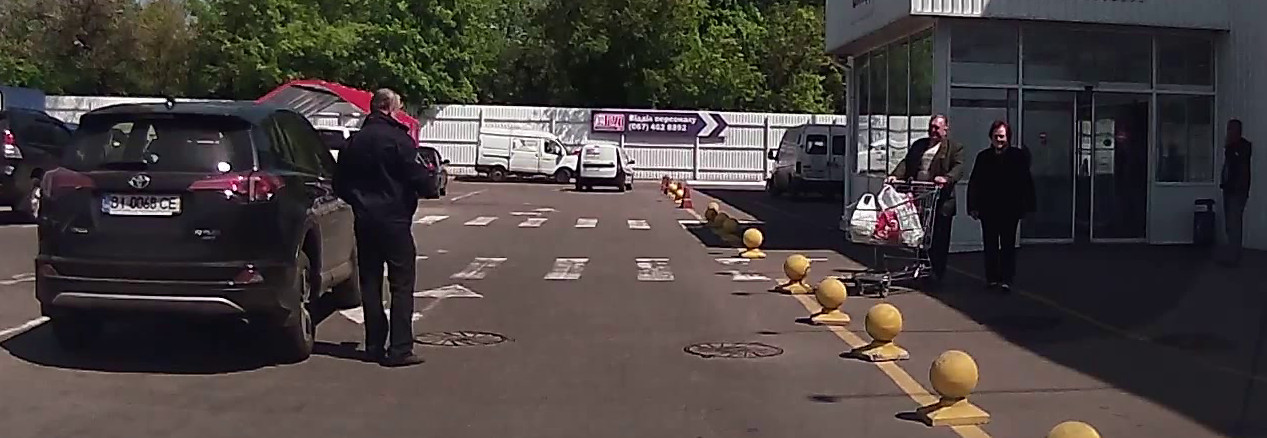}
 \label{fig:case2}}
\subfloat[Potential relevant elements]{
\includegraphics[width=0.5\textwidth]{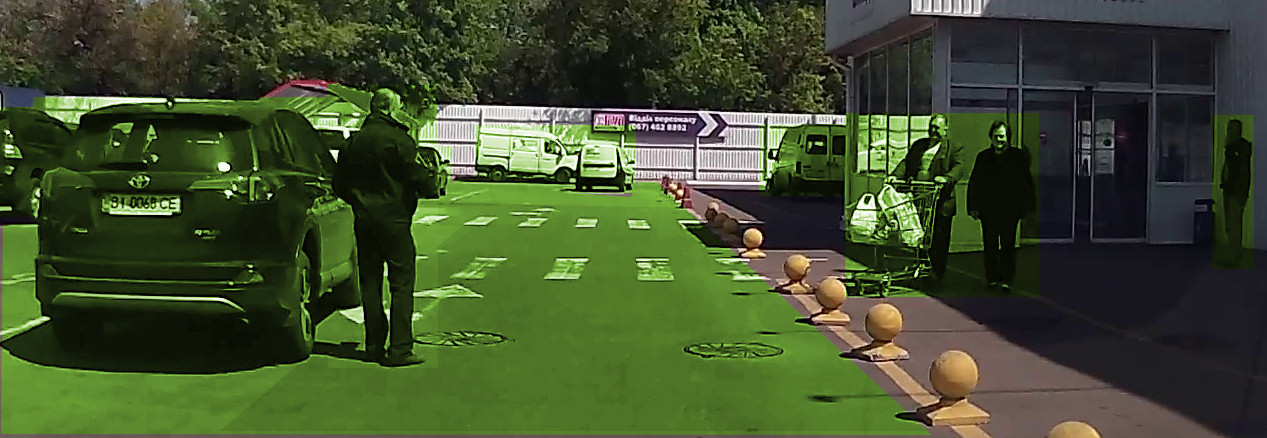}
\label{fig:case2_all_context} }\\
\subfloat[Actual relevant elements]{
\includegraphics[width=0.5\textwidth]{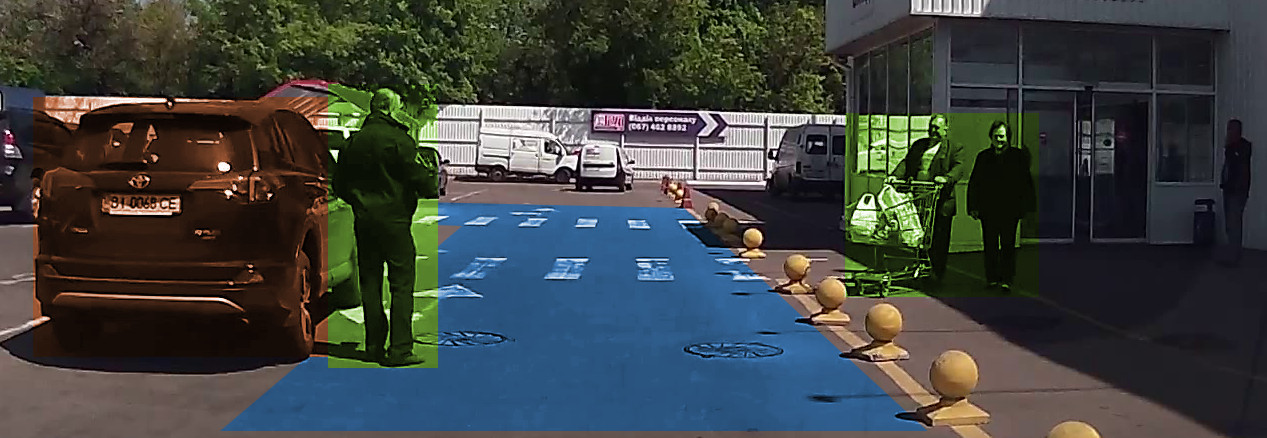}
\label{fig:case2_relevant_context} }
\subfloat[Interacting pedestrians]{
\includegraphics[width=0.5\textwidth]{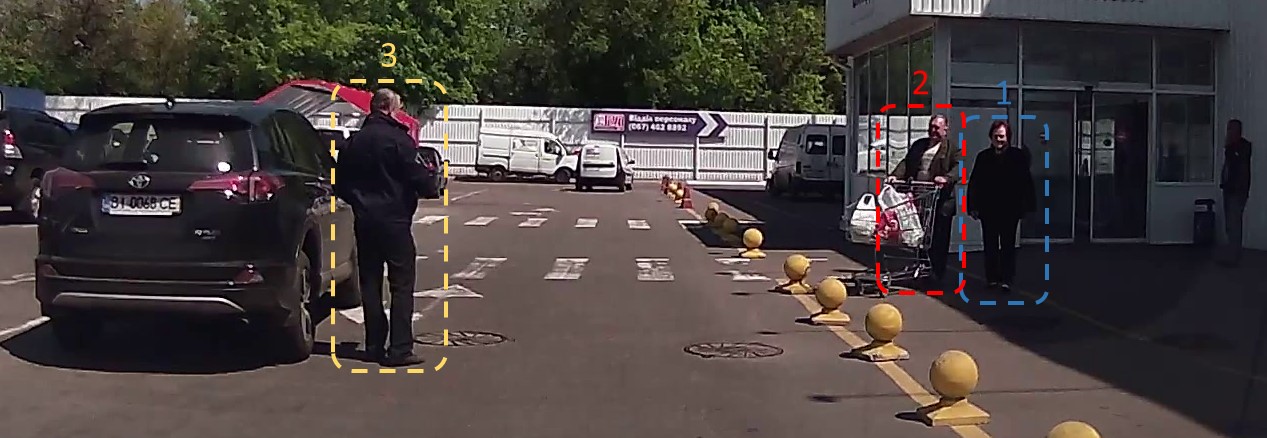}
\label{fig:case2_t1_peds} }\\
\subfloat[Pedestrians 2 and 3 are engaged in eye contact]{
\includegraphics[width=0.5\textwidth]{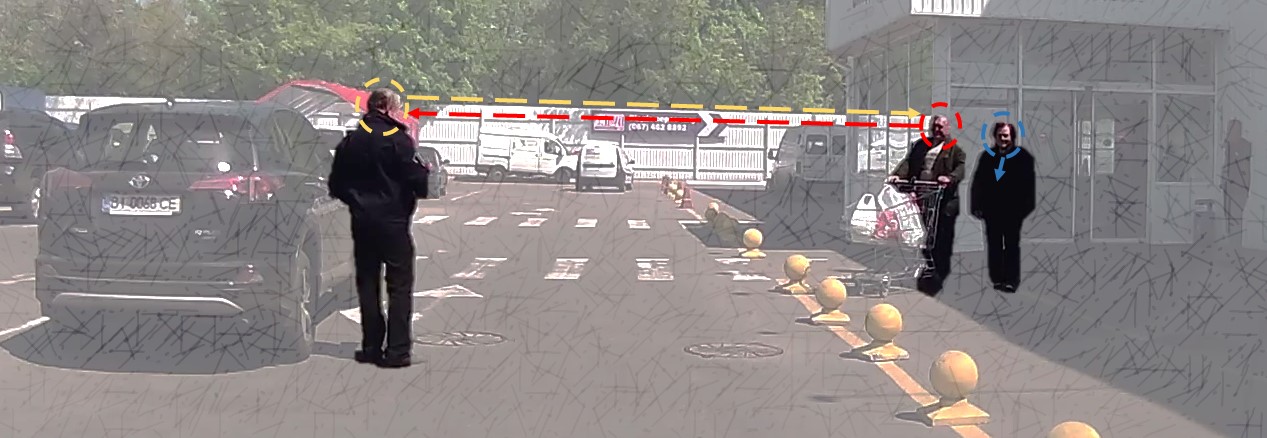}
\label{fig:case2_t1_view} }
\subfloat[Pedestrian 1 engages in eye contact with pedestrian 3]{
\includegraphics[width=0.5\textwidth]{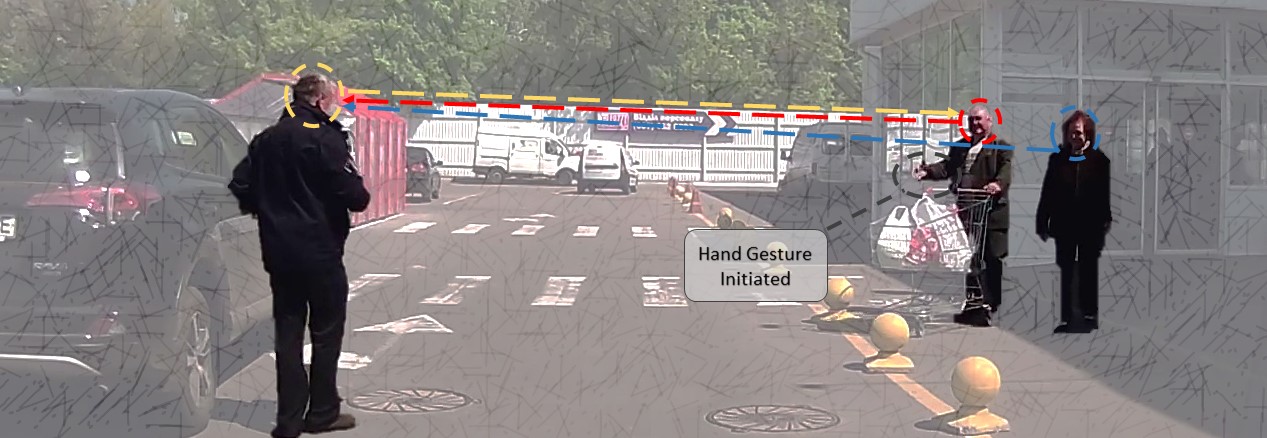}
\label{fig:case2_t2} }\\
\subfloat[Pedestrian 2 is looking at the vehicle]{
\includegraphics[width=0.5\textwidth]{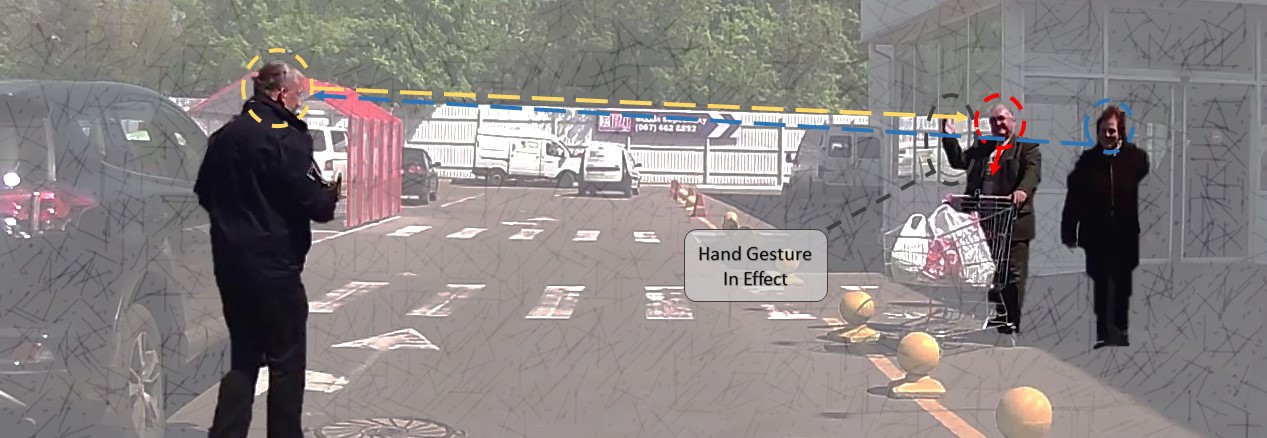}
\label{fig:case2_t3} }
\subfloat[Pedestrians 1 and 3 start turning towards the car]{
\includegraphics[width=0.5\textwidth]{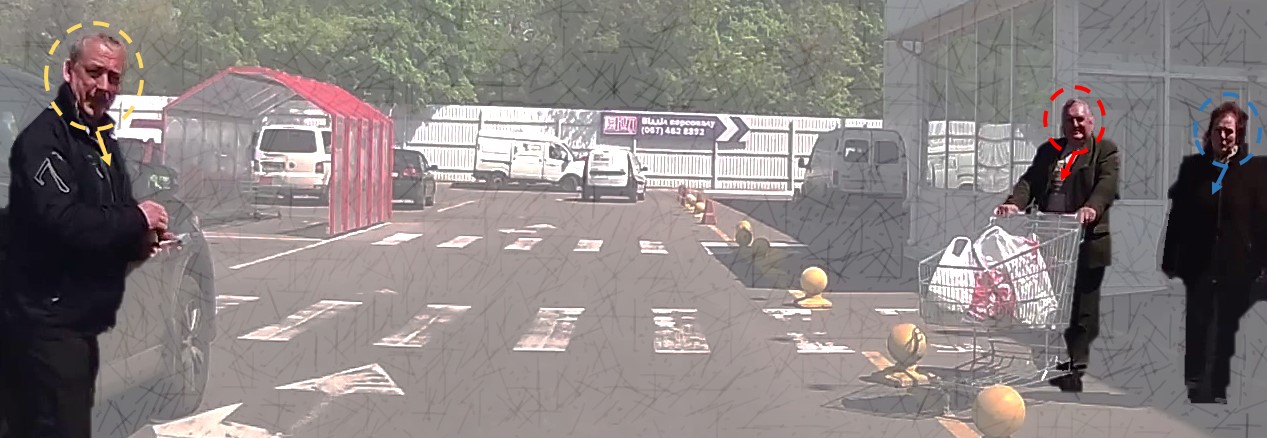}
\label{fig:case2_t4} }

\caption[Joint attention and nonverbal communication in a traffic scene - case2.]{Joint attention and nonverbal communication in a traffic scene. In sub-figure \textit{c} colors indicate the priority of attention, from highest to lowest: {\color{green}\textit{green}}, {\color{blue}\textit{blue}}, {\color{yellow}\textit{yellow}} and {\color{red}\textit{red}}.}
\label{fig:case2_driving_scenario}
\end{figure}

The second example is a traffic scenario occurring in a parking lot (see \figref{fig:case2_driving_scenario}) where two pedestrians are intending to cross while interacting with another pedestrian on the other side of the road. Similar to case 1, we can first identify all potentially relevant elements in the scene (\figref{fig:case2_all_context}) and then specify which ones  are actually relevant to the current task (\figref{fig:case2_relevant_context}). Given that we extensively discussed the problem of attention in the previous case, here we mainly focus on the interaction between the pedestrians and the driver.

\subsubsection{Joint attention and nonverbal communication}
Pedestrians 1 (p1) and 2 (p2) (as illustrated in \figref{fig:case2_t1_peds})  are clearly intending to cross as they are approaching the road whereas the intention of pedestrian 3 (p3) is unknown as he is looking away from the vehicle and standing next to a parked car. In order to make sense of their actions, we need to identify and  analyze instances of joint attention and interaction. In this case, we can observe five instances of joint attention:

\begin{enumerate}

\item \textbf{P1 and the driver:} In \figref{fig:case2_t1_view}, p1 is making eye contact with the driver indicating her intention of crossing and assessing the position and velocity of the vehicle. She then switches her gaze at p3 (\figref{fig:case2_t2})  and looks back at the driver again. This is a classical instance of joint attention in which we can simply by following the gaze of the pedestrian, similar to the works presented earlier \cite{kim2015human,staudte2011investigating,kera2016discovering}, identify her focus of attention. However, this case is different from the social robotic applications. Instead of one task, the pedestrian is involved in two tasks, crossing the road and communicating with another pedestrian. This fact makes the reasoning particularly tricky because we have to identify which task is the main focus of the pedestrian's attention, which in turn will influence the way she is going to behave. Such multi-tasking behavior in the context of joint attention has not previously been addressed.

\item \textbf{P1 and P3:} P1 makes  eye contact with p3 (\figref{fig:case2_t2}) and turns her head towards the ego-vehicle (\figref{fig:case2_t4}) influencing p3 to also follow her gaze towards the same point. 

\item \textbf{P2 and P3:} P2 is initially looking at p3 (\figref{fig:case2_t1_view}). While engaged in communication, he initiates a hand gesture at p3 (\figref{fig:case2_t2}) and at the same time begins turning his head towards the vehicle. In \figref{fig:case2_t3} p2's hand gesture is still in effect while he is looking towards the vehicle. Once again state-of-the-art action recognition \cite{yue2015beyond,deng2016structure,ma2016learning} can detect these behaviors. 

Besides perception issues, it is important to make sense of the communication cue exhibited by p2 since it can potentially influence the behavior of p3. By the time the hand gesture is observed (\figref{fig:case2_t3}) p3 has not yet looked towards the vehicle, and as a result, we do not know whether he is aware of the approaching vehicle and what he is going to do next. For instance, the hand gesture might induce p3 to cross or signal him to stop. 

Understanding nonverbal communication cues have been extensively studied in HCI applications \cite{kim2015human,schauerte2014look,miyauchi2004active,campbell1996invariant} and to some extent in general scene understanding tasks \cite{monajjemi2015uav}. However, these algorithms do not usually consider the semantics of nonverbal signals and specifically do not link them to possible actions (or intentions) of the recipient of the signal.

\item \textbf{P2 and the driver:} The instance of joint attention in this case is particularly  interesting for a number of reasons:

\begin{enumerate}

\item P2 looks at the driver right after he engages in communication with p3 (\figref{fig:case2_t3}). This means in order to realize p2's primary focus of attention, we have to follow his gaze prior to being engaged in eye contact with him. This is in contrast to widely used approaches in joint attention applications where the gaze following and identifying focus of attention takes place after eye contact.

\item P2 makes a hand gesture which is clearly intended for p3 while looking at the driver (\figref{fig:case2_t3}). In joint attention applications, two techniques are common to identify the intention of a nonverbal cue. The communication is performed after an eye contact \cite{zheng2015impacts,mutlu2013coordination}, so the recipient would be the last person who made eye contact with the communicator (e.g. the robot). Another technique associates the cue (e.g. hand gesture) with the eye contact. As a result,  whoever the communicator (e.g. the person interacting with the machine) is looking towards at the time of communicating is the intended recipient \cite{miyauchi2004active}. In our case, however, both these techniques fail because the action performed for someone else is co-occurring with looking at the driver. To solve this issue one has to track the behavior of the agents and be aware of what has happen prior to the eye contact.

\item The communication between p2 and p3 certainly can shed light on understanding the intention of p2  when he makes eye contact with the driver. The current intention estimation algorithms are not addressing the role of communication on pedestrian behavior understanding. Some components of this problem, however, can be solved using various visual perception algorithms such as interaction \cite{park2003recognition,ryoo2009spatio} and group activity recognition \cite{choi2009they,deng2015deep} algorithms.

\end{enumerate}  

\item \textbf{P3 and the driver:} This instance also has some similarities with the previous one in which the primary focus of p3 precedes his eye contact with the driver. Here, once the eye contact is observed, we have more certainty about p3's upcoming crossing decision. 

\end{enumerate}

In this section, we introduced two cases of attention and interaction in driving to highlight various challenges in pedestrian behavior understanding. Although parts of these examples might be seen as very rare and exceptional, the components involved in understanding them are still relevant in typical situations. For instance, joint attention is a very common phenomenon and is involved in almost every interactive traffic scenarios. Nonverbal communication is very common as well, not only between pedestrians and drivers, but also between different drivers or sometimes even authorities and road users, e.g. police force and vehicles.

\section{Summary and Future Work}
Autonomous driving is a complex task and requires the cooperation between many branches of artificial intelligence to become a reality. In this report some of the open problems in the field, in particular, the interaction between autonomous vehicles and other road users, were discussed. It was shown that the interaction between traffic participants can guarantee the flow of traffic, prevent accidents and reduce the likelihood of malicious actions that might happen against the vehicles. 

We showed that at the center of social interactions is joint attention, or the ability to share intention regarding a common object or an event. Joint attention often signals the intention of two or more parties to coordinate in order to accomplish a task. During the coordination, humans observe each other's behavior and try to predict what comes next.

To make the prediction accurate, humans often communicate their intentions and analyze the context in which the behavior is observed. In traffic scenarios, communication is commonly nonverbal, i.e. through the use of cues such as eye contact, hand gestures or head movements to transmit the intention of crossing, showing gratitude or asking for the right of way. As for the context, there is a broad set of factors that potentially impact the behavior of pedestrians and drivers. Some of these factors are: dynamics of the scene (e.g. vehicles' speed and distance), social norms, demographics, social forces (e.g. group size, social status), and physical context (e.g. signals, road geometry).

Realizing the ability of social interaction in practice is challenging and involves a vast number of research areas ranging from detection and identification of elements (e.g. pedestrians, signs, roads) in the scene to reasoning about locations, poses and the types of activities observed.

Today's state-of-the-art computer vision algorithms attempt to resolve the perceptual tasks involved in traffic scene understanding. Despite their advancements and successes in a number of contexts, these algorithms are still far from ideal to be used in practice. For instance, pedestrian detection algorithms, which are only tested on datasets with favorable conditions, such as scenes in broad daylight or clear weather conditions, have not yet achieved a performance level close to humans. Activity recognition algorithms are mainly focusing on either video classification problems where the entire length of the activity is known at the detection time or applications where the outliers or background clutter is minimal , e.g. sport scenes. In addition, given that the majority of vision algorithms are designed for off-line applications, their processing times usually far exceed the real-time requirement for driving tasks.     

In recent years, a number of attempts have been made to design algorithms for estimating the behavior of pedestrians at crosswalks. These works combine various contextual cues such as the dynamics of the vehicle and the pedestrian, pedestrian head orientation or the structure of the street to estimate whether the pedestrian will cross or not. However, these algorithms either lack visual perception methods necessary for analyzing the scenes, or are only applied to very limited contexts, e.g. narrow streets with no signal.  

Given the broadness of the social interaction task in autonomous driving, there are many open problems in this field. 

A great deal of studies on pedestrian behavior are dated back to a few decades ago. During this time we have witnessed major technological and socioeconomic changes, which means the behavioral studies have to be revisited in order to account for modern times. In particular, there is a major shortage of studies addressing the interaction issues between pedestrians and autonomous vehicles. 

The majority of the practical joint attention systems are designed for tasks such as learning, rehabilitation or imitation. There is a very few attempts to build a system
capable of joint attention in complex cooperative tasks, and none in the context of autonomous driving.

As it was presented earlier, the state of the art visual
attention algorithms are heavily data-driven and are unsuitable for driving tasks. There is a need for a
task-driven visual attention model that can dynamically focus on regions of interest during the course
of driving.

Algorithms capable of understanding non-verbal communication are mainly designed for human-computer interaction where the proximity of the user to the
machine is small, the background clutter is minimal, and the task often involves direct understanding
of various bodily movements without considering the context. There are only few attempts on under-
standing communication in traffic scenes, none of which infer the meaning of the observed cues, but
rather identify the type of the cues.

The current intention estimation algorithms are
very limited in context, and often are not accompanied with necessary visual perception algorithms
to analyze the scenes. In addition, the data used in these algorithms is either not naturalistically
obtained (e.g. participants are scripted), or not sufficiently diverse to include various traffic scenarios.
This points to the need for implementing a system that can, first, identify the relevant elements in the
scene, second, reason about the interconnections between these elements, and third, infer the upcoming
actions of the road users. The algorithm should also be universal in a sense that it can be used in
various traffic scenarios with different street structures, traffic signals, crosswalk configurations, etc.

\bibliographystyle{IEEEtran}
\bibliography{../references/beh_ref,../references/auto_ref,../references/intention_ref,../references/joint_comm,../references/cog_reasoning,../references/hardware_ref,../references/jt_robot_ref,../references/gen_data_ref,../references/gen_detection_ref,../references/car_data_ref,../references/car_detection_ref,../references/ped_data_ref,../references/ped_detection_ref,../references/road_detection_ref,../references/sign_data_ref,../references/sign_detection_ref,../references/traffic_data_ref,../references/traffic_detection_ref,../references/activity_data_ref,../references/pose_ref,../references/activity_ref,../references/attention,../references/ours_ref}

\begin{thebibliography}{100}
\providecommand{\url}[1]{#1}
\csname url@samestyle\endcsname
\providecommand{\newblock}{\relax}
\providecommand{\bibinfo}[2]{#2}
\providecommand{\BIBentrySTDinterwordspacing}{\spaceskip=0pt\relax}
\providecommand{\BIBentryALTinterwordstretchfactor}{4}
\providecommand{\BIBentryALTinterwordspacing}{\spaceskip=\fontdimen2\font plus
\BIBentryALTinterwordstretchfactor\fontdimen3\font minus
  \fontdimen4\font\relax}
\providecommand{\BIBforeignlanguage}[2]{{%
\expandafter\ifx\csname l@#1\endcsname\relax
\typeout{** WARNING: IEEEtran.bst: No hyphenation pattern has been}%
\typeout{** loaded for the language `#1'. Using the pattern for}%
\typeout{** the default language instead.}%
\else
\language=\csname l@#1\endcsname
\fi
#2}}
\providecommand{\BIBdecl}{\relax}
\BIBdecl

\bibitem{kroger2016automated}
F.~Kr{\"o}ger, ``Automated driving in its social, historical and cultural
  contexts,'' in \emph{Autonomous Driving}.\hskip 1em plus 0.5em minus
  0.4em\relax Springer, 2016, pp. 41--68.

\bibitem{winkle2016safety}
T.~Winkle, ``Safety benefits of automated vehicles: Extended findings from
  accident research for development, validation and testing,'' in
  \emph{Autonomous Driving}.\hskip 1em plus 0.5em minus 0.4em\relax Springer,
  2016, pp. 335--364.

\bibitem{litman2014autonomous}
T.~Litman, ``Autonomous vehicle implementation predictions,'' \emph{Victoria
  Transport Policy Institute}, vol.~28, 2014.

\bibitem{dickmanns1987curvature}
E.~D. Dickmanns and A.~Zapp, ``A curvature-based scheme for improving road
  vehicle guidance by computer vision,'' in \emph{Cambridge
  Symposium\_Intelligent Robotics Systems}.\hskip 1em plus 0.5em minus
  0.4em\relax International Society for Optics and Photonics, 1987, pp.
  161--168.

\bibitem{darms2008multisensor}
M.~Darms, P.~E. Rybski, and C.~Urmson, ``A multisensor multiobject tracking
  system for an autonomous vehicle driving in an urban environment,'' in
  \emph{9th International Symposium on Advanced Vehicle Control (AVEC)}, 2008.

\bibitem{bishop2000intelligent}
R.~Bishop, ``Intelligent vehicle applications worldwide,'' \emph{IEEE
  Intelligent Systems and Their Applications}, vol.~15, no.~1, pp. 78--81,
  2000.

\bibitem{Radovanovic2016}
\BIBentryALTinterwordspacing
D.~Radovanovic and D.~Muoio, ``This is what the evolution of self-driving cars
  looks like,'' Online, 2017-05-28. [Online]. Available:
  \url{http://www.businessinsider.com/what-are-the-different-levels-of-driverless-cars-2016-10/#-1}
\BIBentrySTDinterwordspacing

\bibitem{sae2017}
\BIBentryALTinterwordspacing
``Automated driving levels of driving automation are defined in new {SAE}
  international standard j3016,'' Online, 2017-05-28. [Online]. Available:
  \url{https://www.sae.org/misc/pdfs/automated_driving.pdf}
\BIBentrySTDinterwordspacing

\bibitem{Nguyen2017}
\BIBentryALTinterwordspacing
V.~Nguyen, ``2019 audi a8 level 3 autonomy first-drive: Chasing the perfect
  ‘jam’,'' Online, 2017-11-10. [Online]. Available:
  \url{https://www.slashgear.com/2019-audi-a8-level-3-autonomy-first-drive-chasing-the-perfect-jam-11499082/}
\BIBentrySTDinterwordspacing

\bibitem{cart1921}
\BIBentryALTinterwordspacing
``Unmanned ground vehicle,'' Online, 2017-05-28. [Online]. Available:
  \url{http://www.wikiwand.com/en/Unmanned_ground_vehicle}
\BIBentrySTDinterwordspacing

\bibitem{Oagana2016}
\BIBentryALTinterwordspacing
A.~Oagana, ``A short history of mercedes-benz autonomous driving technology,''
  Online, 2017-05-28. [Online]. Available:
  \url{https://www.autoevolution.com/news/a-short-history-of-mercedes-benz-autonomous-driving-technology-68148.html}
\BIBentrySTDinterwordspacing

\bibitem{broggi1999argo}
A.~Broggi, M.~Bertozzi, A.~Fascioli, C.~G.~L. Bianco, and A.~Piazzi, ``The argo
  autonomous vehicle’s vision and control systems,'' \emph{International
  Journal of Intelligent Control and Systems}, vol.~3, no.~4, pp. 409--441,
  1999.

\bibitem{stanley2005}
\BIBentryALTinterwordspacing
``Self driving car,'' Online, 2017-05-28. [Online]. Available:
  \url{http://stanford.edu/~cpiech/cs221/apps/driverlessCar.html}
\BIBentrySTDinterwordspacing

\bibitem{boss2007}
\BIBentryALTinterwordspacing
``De 1977 à nos jours, beaucoup de progrès !'' Online, 2017-05-28. [Online].
  Available:
  \url{http://voitureautonome-2014.kazeo.com/de-1977-a-nos-jours-beaucoup-de-progres-a124503004}
\BIBentrySTDinterwordspacing

\bibitem{vislab2010}
\BIBentryALTinterwordspacing
``Vislab intercontinental autonomous challenge: Inaugural ceremony – milan,
  italy,'' Online, 2017-05-28. [Online]. Available:
  \url{http://manonthemove.com/2010/07/21/vislab-intercontinental-autonomous-challenge-inaugural-ceremony-milan-italy/}
\BIBentrySTDinterwordspacing

\bibitem{shelley2012}
\BIBentryALTinterwordspacing
``Watch {Stanford’s} self-driving vehicle hit 120mph: {Autonomous} {Audi}
  proves to be just as good as a race car driver,'' Online, 2017-05-28.
  [Online]. Available:
  \url{http://www.dailymail.co.uk/sciencetech/article-3472223/Watch-Stanford-s-self-driving-vehicle-hit-120mph-Autonomous-Audi-proves-just-good-race-car-driver.html}
\BIBentrySTDinterwordspacing

\bibitem{Davieg2016}
\BIBentryALTinterwordspacing
A.~Davieg, ``We take a ride in the self-driving {Uber} now roaming
  {Pittsburgh},'' Online, 2017-05-28. [Online]. Available:
  \url{https://www.wired.com/2016/09/self-driving-autonomous-uber-pittsburgh/#slide-8}
\BIBentrySTDinterwordspacing

\bibitem{walter1948image}
\BIBentryALTinterwordspacing
``W. {Grey Walter’s Tortoises} – {Self-recognition} and narcissism,''
  Online, 2017-05-26. [Online]. Available:
  \url{http://cyberneticzoo.com/cyberneticanimals/w-grey-walter-tortoises-picture-gallery-2/attachment/la-scienza-illustrata-1950_10-walter-tortoise-2-x640/}
\BIBentrySTDinterwordspacing

\bibitem{moravecimage}
\BIBentryALTinterwordspacing
``1960 – {Stanford Cart – (American)},'' Online, 2017-05-26. [Online].
  Available:
  \url{http://cyberneticzoo.com/cyberneticanimals/1960-stanford-cart-american/}
\BIBentrySTDinterwordspacing

\bibitem{walter1948}
\BIBentryALTinterwordspacing
``Elsie (electro-mechanical robot, light sensitive with internal and external
  stability,'' Online, 2017-05-26. [Online]. Available:
  \url{http://cyberneticzoo.com/cyberneticanimals/elsie-cyberneticanimals/elsie/}
\BIBentrySTDinterwordspacing

\bibitem{moravec1980obstacle}
H.~Moravec, ``Obstacle avoidance and navigation in the real world by a seeing
  robot rover.'' DTIC Document, Tech. Rep., 1980.

\bibitem{moravec1983stanford}
H.~P. Moravec, ``The {Stanford Cart} and the {CMU} rover,'' \emph{Proceedings
  of the IEEE}, vol.~71, no.~7, pp. 872--884, 1983.

\bibitem{mysliwetz1987distributed}
B.~D. Mysliwetz and E.~Dickmanns, ``Distributed scene analysis for autonomous
  road vehicle guidance,'' in \emph{Robotics and IECON'87 Conferences}.\hskip
  1em plus 0.5em minus 0.4em\relax International Society for Optics and
  Photonics, 1987, pp. 72--79.

\bibitem{dickmanns1990integrated}
E.~D. Dickmanns, B.~Mysliwetz, and T.~Christians, ``An integrated
  spatio-temporal approach to automatic visual guidance of autonomous
  vehicles,'' \emph{IEEE Transactions on Systems, Man, and Cybernetics},
  vol.~20, no.~6, pp. 1273--1284, 1990.

\bibitem{Mueller2017}
\BIBentryALTinterwordspacing
M.~Mueller-Freitag, ``Germany asleep at the wheel?'' Online, 2017-05-28.
  [Online]. Available:
  \url{https://medium.com/twentybn/germany-asleep-at-the-wheel-d800445d6da2}
\BIBentrySTDinterwordspacing

\bibitem{pomerleau1991combining}
D.~A. Pomerleau, J.~Gowdy, and C.~E. Thorpe, ``Combining artificial neural
  networks and symbolic processing for autonomous robot guidance,''
  \emph{Engineering Applications of Artificial Intelligence}, vol.~4, no.~4,
  pp. 279--285, 1991.

\bibitem{pomerleau1992progress}
D.~Pomerleau, ``Progress in neural network-based vision for autonomous robot
  driving,'' in \emph{Intelligent VehiclesSymposium (IV)}.\hskip 1em plus 0.5em
  minus 0.4em\relax IEEE, 1992, pp. 391--396.

\bibitem{pomerleau1996neural}
D.~A. Pomerleau, ``Neural network vision for robot driving,'' in \emph{The
  Handbook of Brain Theory and Neural Networks}.\hskip 1em plus 0.5em minus
  0.4em\relax Citeseer, 1996.

\bibitem{baluja1996evolution}
S.~Baluja, ``Evolution of an artificial neural network based autonomous land
  vehicle controller,'' \emph{IEEE Transactions on Systems, Man, and
  Cybernetics, Part B (Cybernetics)}, vol.~26, no.~3, pp. 450--463, 1996.

\bibitem{thorpe1991toward}
C.~Thorpe, M.~Herbert, T.~Kanade, and S.~Shafer, ``Toward autonomous driving:
  the {CMU Navlab}. i. perception,'' \emph{IEEE expert}, vol.~6, no.~4, pp.
  31--42, 1991.

\bibitem{jochem1995vision}
T.~M. Jochem, D.~A. Pomerleau, and C.~E. Thorpe, ``Vision-based neural network
  road and intersection detection and traversal,'' in \emph{IROS},
  vol.~3.\hskip 1em plus 0.5em minus 0.4em\relax IEEE, 1995, pp. 344--349.

\bibitem{yim2003three}
Y.~U. Yim and S.-Y. Oh, ``Three-feature based automatic lane detection
  algorithm ({TFALDA}) for autonomous driving,'' \emph{IEEE Transactions on
  Intelligent Transportation Systems}, vol.~4, no.~4, pp. 219--225, 2003.

\bibitem{kluge1994extracting}
K.~Kluge, ``Extracting road curvature and orientation from image edge points
  without perceptual grouping into features,'' in \emph{Intelligent Vehicles
  Symposium (IV)}.\hskip 1em plus 0.5em minus 0.4em\relax IEEE, 1994, pp.
  109--114.

\bibitem{dickmanns1994seeing}
E.~D. Dickmanns, R.~Behringer, D.~Dickmanns, T.~Hildebrandt, M.~Maurer,
  F.~Thomanek, and J.~Schiehlen, ``The seeing passenger car {'VaMoRs-P'},'' in
  \emph{Intelligent Vehicles Symposium (IV)}.\hskip 1em plus 0.5em minus
  0.4em\relax IEEE, 1994, pp. 68--73.

\bibitem{franke1994daimler}
U.~Franke, S.~Mehring, A.~Suissa, and S.~Hahn, ``The {Daimler-Benz} steering
  assistant: a spin-off from autonomous driving,'' in \emph{Intelligent
  Vehicles Symposium (IV)}.\hskip 1em plus 0.5em minus 0.4em\relax IEEE, 1994,
  pp. 120--124.

\bibitem{bohrer1995integrated}
S.~Bohrer, T.~Zielke, and V.~Freiburg, ``An integrated obstacle detection
  framework for intelligent cruise control on motorways,'' in \emph{Intelligent
  Vehicles Symposium (IV)}.\hskip 1em plus 0.5em minus 0.4em\relax IEEE, 1995,
  pp. 276--281.

\bibitem{hong2000intelligent}
T.~Hong, M.~Abrams, T.~Chang, and M.~Shneier, ``An intelligent world model for
  autonomous off-road driving,'' \emph{CVIU}, 2000.

\bibitem{behringer2004darpa}
R.~Behringer, S.~Sundareswaran, B.~Gregory, R.~Elsley, B.~Addison,
  W.~Guthmiller, R.~Daily, and D.~Bevly, ``The {DARPA} grand
  challenge-development of an autonomous vehicle,'' in \emph{Intelligent
  Vehicles Symposium (IV)}.\hskip 1em plus 0.5em minus 0.4em\relax IEEE, 2004,
  pp. 226--231.

\bibitem{dang2006path}
T.~Dang, S.~Kammel, C.~Duchow, B.~Hummel, and C.~Stiller, ``Path planning for
  autonomous driving based on stereoscopic and monoscopic vision cues,'' in
  \emph{IEEE International Conference on Multisensor Fusion and Integration for
  Intelligent Systems}.\hskip 1em plus 0.5em minus 0.4em\relax IEEE, 2006, pp.
  191--196.

\bibitem{thrun2006stanley}
S.~Thrun, M.~Montemerlo, H.~Dahlkamp, D.~Stavens, A.~Aron, J.~Diebel, P.~Fong,
  J.~Gale, M.~Halpenny, G.~Hoffmann \emph{et~al.}, ``Stanley: The robot that
  won the {DARPA} grand challenge,'' \emph{Journal of field Robotics}, vol.~23,
  no.~9, pp. 661--692, 2006.

\bibitem{thrun2006probabilistic}
S.~Thrun, M.~Montemerlo, and A.~Aron, ``Probabilistic terrain analysis for
  high-speed desert driving.'' in \emph{Robotics: Science and Systems}, 2006,
  pp. 16--19.

\bibitem{hoffmann2007autonomous}
G.~M. Hoffmann, C.~J. Tomlin, M.~Montemerlo, and S.~Thrun, ``Autonomous
  automobile trajectory tracking for off-road driving: Controller design,
  experimental validation and racing,'' in \emph{American Control Conference
  (ACC)}.\hskip 1em plus 0.5em minus 0.4em\relax IEEE, 2007, pp. 2296--2301.

\bibitem{dolgov2008practical}
D.~Dolgov, S.~Thrun, M.~Montemerlo, and J.~Diebel, ``Practical search
  techniques in path planning for autonomous driving,'' \emph{Ann Arbor}, vol.
  1001, p. 48105, 2008.

\bibitem{urmson2008autonomous}
C.~Urmson, J.~Anhalt, D.~Bagnell, C.~Baker, R.~Bittner, M.~Clark, J.~Dolan,
  D.~Duggins, T.~Galatali, C.~Geyer \emph{et~al.}, ``Autonomous driving in
  urban environments: Boss and the urban challenge,'' \emph{Journal of Field
  Robotics}, vol.~25, no.~8, pp. 425--466, 2008.

\bibitem{dolgov2009autonomous}
D.~Dolgov and S.~Thrun, ``Autonomous driving in semi-structured environments:
  Mapping and planning,'' in \emph{ICRA}.\hskip 1em plus 0.5em minus
  0.4em\relax IEEE, 2009, pp. 3407--3414.

\bibitem{kummerle2009autonomous}
R.~Kummerle, D.~Hahnel, D.~Dolgov, S.~Thrun, and W.~Burgard, ``Autonomous
  driving in a multi-level parking structure,'' in \emph{ICRA}.\hskip 1em plus
  0.5em minus 0.4em\relax IEEE, 2009, pp. 3395--3400.

\bibitem{wei2009robust}
J.~Wei and J.~M. Dolan, ``A robust autonomous freeway driving algorithm,'' in
  \emph{Intelligent Vehicles Symposium (IV)}.\hskip 1em plus 0.5em minus
  0.4em\relax IEEE, 2009, pp. 1015--1020.

\bibitem{wei2013towards}
J.~Wei, J.~M. Snider, J.~Kim, J.~M. Dolan, R.~Rajkumar, and B.~Litkouhi,
  ``Towards a viable autonomous driving research platform,'' in
  \emph{Intelligent Vehicles Symposium (IV)}.\hskip 1em plus 0.5em minus
  0.4em\relax IEEE, 2013, pp. 763--770.

\bibitem{brechtel2014probabilistic}
S.~Brechtel, T.~Gindele, and R.~Dillmann, ``Probabilistic decision-making under
  uncertainty for autonomous driving using continuous pomdps,'' in
  \emph{Intelligent Transportation Systems (ITSC)}.\hskip 1em plus 0.5em minus
  0.4em\relax IEEE, 2014, pp. 392--399.

\bibitem{bertozzi2010vislab}
M.~Bertozzi, L.~Bombini, A.~Broggi, M.~Buzzoni, E.~Cardarelli, S.~Cattani,
  P.~Cerri, S.~Debattisti, R.~Fedriga, M.~Felisa \emph{et~al.}, ``The {VISLAB}
  intercontinental autonomous challenge: 13,000 km, 3 months, no driver,'' in
  \emph{17th World Congress on ITS}, 2010.

\bibitem{shelleydef}
\BIBentryALTinterwordspacing
C.~SQUATRIGLIA, ``{Audi's} robotic car climbs pikes peak,'' Online, 2017-05-28.
  [Online]. Available:
  \url{https://www.wired.com/2010/11/audis-robotic-car-climbs-pikes-peak/}
\BIBentrySTDinterwordspacing

\bibitem{waymo}
\BIBentryALTinterwordspacing
``Waymo,'' Online, 2017-05-30. [Online]. Available: \url{https://waymo.com/}
\BIBentrySTDinterwordspacing

\bibitem{baidu}
\BIBentryALTinterwordspacing
S.~Millward, ``Baidu's driverless cars on china’s roads by 2020,'' Online,
  2017-05-30. [Online]. Available:
  \url{https://www.techinasia.com/baidu-autonomous-car-sales-2020}
\BIBentrySTDinterwordspacing

\bibitem{toyota}
\BIBentryALTinterwordspacing
a.~English, ``Toyota's driveless car,'' Online, 2017-05-30. [Online].
  Available:
  \url{http://www.telegraph.co.uk/motoring/car-manufacturers/toyota/10404575/Toyotas-driverless-car.html}
\BIBentrySTDinterwordspacing

\bibitem{companiesauto}
\BIBentryALTinterwordspacing
``44 corporations working on autonomous vehicles,'' Online, 2017-05-30.
  [Online]. Available:
  \url{https://www.cbinsights.com/blog/autonomous-driverless-vehicles-corporations-list/}
\BIBentrySTDinterwordspacing

\bibitem{tesla}
\BIBentryALTinterwordspacing
``Full self-driving hardware on all cars,'' Online, 2017-05-30. [Online].
  Available: \url{https://www.tesla.com/en_CA/autopilot?redirect=no}
\BIBentrySTDinterwordspacing

\bibitem{nica2016}
\BIBentryALTinterwordspacing
G.~Nica, ``{BMW CEO} wants autonomous driving cars within five years,'' Online,
  2017-05-28. [Online]. Available:
  \url{http://www.bmwblog.com/2016/08/02/bmw-ceo-wants-autonomous-driving-cars-within-five-years/}
\BIBentrySTDinterwordspacing

\bibitem{ziegler2014making}
J.~Ziegler, P.~Bender, M.~Schreiber, H.~Lategahn, T.~Strauss, C.~Stiller,
  T.~Dang, U.~Franke, N.~Appenrodt, C.~G. Keller \emph{et~al.}, ``Making
  {Bertha} drive-{An} autonomous journey on a historic route,'' \emph{IEEE
  Intelligent Transportation Systems Magazine}, vol.~6, no.~2, pp. 8--20, 2014.

\bibitem{truck}
\BIBentryALTinterwordspacing
A.~Davieg, ``Uber’s self-driving truck makes its first delivery: 50,000
  beers,'' Online, 2017-05-30. [Online]. Available:
  \url{https://www.wired.com/2016/10/ubers-self-driving-truck-makes-first-delivery-50000-beers/}
\BIBentrySTDinterwordspacing

\bibitem{bus}
\BIBentryALTinterwordspacing
A.~Marshal, ``Don’t look now, but even buses are going autonomous,'' Online,
  2017-05-30. [Online]. Available:
  \url{https://www.wired.com/2017/05/reno-nevada-autonomous-bus/}
\BIBentrySTDinterwordspacing

\bibitem{ship}
\BIBentryALTinterwordspacing
O.~Levander, ``Forget autonomous cars-autonomous ships are almost here,''
  Online, 2017-05-30. [Online]. Available:
  \url{https://www.wired.com/2016/10/ubers-self-driving-truck-makes-first-delivery-50000-beers/}
\BIBentrySTDinterwordspacing

\bibitem{teslafull}
\BIBentryALTinterwordspacing
F.~Lambert, ``Elon {Musk} clarifies tesla’s plan for level 5 fully autonomous
  driving: 2 years away from sleeping in the car,'' Online, 2017-05-30.
  [Online]. Available:
  \url{https://electrek.co/2017/04/29/elon-musk-tesla-plan-level-5-full-autonomous-driving/}
\BIBentrySTDinterwordspacing

\bibitem{toyotaauto}
\BIBentryALTinterwordspacing
E.~Ackerman, ``{Toyota's Gill Pratt} on self-driving cars and the reality of
  full autonomy,'' Online, 2017-05-30. [Online]. Available:
  \url{http://spectrum.ieee.org/cars-that-think/transportation/self-driving/toyota-gill-pratt-on-the-reality-of-full-autonomy}
\BIBentrySTDinterwordspacing

\bibitem{Friedrich2016}
B.~Friedrich, ``The effect of autonomous vehicles on traffic,''
  \emph{Autonomous Driving}, pp. 317--334, 2016.

\bibitem{Gasser2016}
T.~M. Gasser, ``Fundamental and special legal questions for autonomous
  vehicles,'' \emph{Autonomous Driving}, pp. 523--551, 2016.

\bibitem{muoio2016}
\BIBentryALTinterwordspacing
D.~Muoio, ``6 scenarios self-driving cars still can't handle,'' Online,
  2017-05-30. [Online]. Available:
  \url{http://www.businessinsider.com/autonomous-car-limitations-2016-8/#1-driverless-cars-struggle-going-over-bridges-1}
\BIBentrySTDinterwordspacing

\bibitem{ron2016}
\BIBentryALTinterwordspacing
R.~Tussy, ``The challenges facing autonomous vehicles,'' Online, 2017-05-30.
  [Online]. Available:
  \url{http://auto-sens.com/the-challenges-facing-autonomous-vehicles/}
\BIBentrySTDinterwordspacing

\bibitem{lambert2016}
\BIBentryALTinterwordspacing
F.~Lambert, ``Tesla {Model S} driver crashes into a van while on autopilot
  [video],'' Online, 2017-05-30. [Online]. Available:
  \url{https://electrek.co/2016/05/26/tesla-model-s-crash-autopilot-video/}
\BIBentrySTDinterwordspacing

\bibitem{teslapolice2017}
\BIBentryALTinterwordspacing
``Tesla on autopilot hits police motorcycle,'' Online, 2017-05-30. [Online].
  Available:
  \url{http://www.government-fleet.com/channel/safety-accident-management/news/story/2017/03/tesla-on-autopilot-hits-police-motorcycle.aspx}
\BIBentrySTDinterwordspacing

\bibitem{uberaccident2017}
\BIBentryALTinterwordspacing
``Uber suspends self-driving fleet after {Ariz}. crash,'' Online, 2017-05-30.
  [Online]. Available:
  \url{http://www.automotive-fleet.com/news/story/2017/03/uber-self-driving-car-struck-in-ariz-crash.aspx}
\BIBentrySTDinterwordspacing

\bibitem{teslafatalt1}
\BIBentryALTinterwordspacing
``Tesla driver dies in first fatal crash while using autopilot mode,'' Online,
  2017-05-30. [Online]. Available:
  \url{https://www.theguardian.com/technology/2016/jun/30/tesla-autopilot-death-self-driving-car-elon-musk}
\BIBentrySTDinterwordspacing

\bibitem{teslafatalt2}
\BIBentryALTinterwordspacing
``Another fatal tesla crash reportedly on autopilot emerges, {Model S} hits a
  streetsweeper truck – caught on dashcam,'' Online, 2017-05-30. [Online].
  Available:
  \url{https://electrek.co/2016/09/14/another-fatal-tesla-autopilot-crash-emerges-model-s-hits-a-streetsweeper-truck-caught-on-dashcam/}
\BIBentrySTDinterwordspacing

\bibitem{wolf2016interaction}
I.~Wolf, ``The interaction between humans and autonomous agents,'' in
  \emph{Autonomous Driving}.\hskip 1em plus 0.5em minus 0.4em\relax Springer,
  2016, pp. 103--124.

\bibitem{farber2016communication}
B.~F{\"a}rber, ``Communication and communication problems between autonomous
  vehicles and human drivers,'' in \emph{Autonomous Driving}.\hskip 1em plus
  0.5em minus 0.4em\relax Springer, 2016, pp. 125--144.

\bibitem{richtel2016}
\BIBentryALTinterwordspacing
M.~Richtel, ``Google's driverless cars run into problem: Cars with drivers,''
  Online, 2017-05-30. [Online]. Available:
  \url{https://www.nytimes.com/2015/09/02/technology/personaltech/google-says-its-not-the-driverless-cars-fault-its-other-drivers.html?_r=2}
\BIBentrySTDinterwordspacing

\bibitem{anthony2016}
\BIBentryALTinterwordspacing
S.~E. Anthony, ``The trollable self-driving car,'' Online, 2017-05-30.
  [Online]. Available:
  \url{http://www.slate.com/articles/technology/future_tense/2016/03/google_self_driving_cars_lack_a_human_s_intuition_for_what_other_drivers.html}
\BIBentrySTDinterwordspacing

\bibitem{gough2016}
\BIBentryALTinterwordspacing
M.~Gough, ``Machine smarts: how will pedestrians negotiate with driverless
  cars?'' Online, 2017-05-30. [Online]. Available:
  \url{https://www.theguardian.com/sustainable-business/2016/sep/09/machine-smarts-how-will-pedestrians-negotiate-with-driverless-cars}
\BIBentrySTDinterwordspacing

\bibitem{sun2002modeling}
D.~Sun, S.~Ukkusuri, R.~F. Benekohal, and S.~T. Waller, ``Modeling of
  motorist-pedestrian interaction at uncontrolled mid-block crosswalks,''
  \emph{Urbana}, vol.~51, p. 61801, 2002.

\bibitem{parkes1995potential}
A.~M. Parkes, N.~J. Ward, and L.~Bossi, ``The potential of vision enhancement
  systems to improve driver safety,'' \emph{Le Travail Humain}, vol.~58, no.~2,
  p. 151, 1995.

\bibitem{bully}
\BIBentryALTinterwordspacing
M.~McFarland, ``Robots hit the streets -- and the streets hit back,'' Online,
  2017-05-30. [Online]. Available:
  \url{http://money.cnn.com/2017/04/28/technology/robot-bullying/}
\BIBentrySTDinterwordspacing

\bibitem{kumashiro2003natural}
M.~Kumashiro, H.~Ishibashi, Y.~Uchiyama, S.~Itakura, A.~Murata, and A.~Iriki,
  ``Natural imitation induced by joint attention in {Japanese} monkeys,''
  \emph{International Journal of Psychophysiology}, vol.~50, no.~1, pp. 81--99,
  2003.

\bibitem{kidwell2007joint}
M.~Kidwell and D.~H. Zimmerman, ``Joint attention as action,'' \emph{Journal of
  Pragmatics}, vol.~39, no.~3, pp. 592--611, 2007.

\bibitem{butterworth1980towards}
G.~Butterworth and E.~Cochran, ``Towards a mechanism of joint visual attention
  in human infancy,'' \emph{International Journal of Behavioral Development},
  vol.~3, no.~3, pp. 253--272, 1980.

\bibitem{scaife1975capacity}
M.~Scaife and J.~S. Bruner, ``The capacity for joint visual attention in the
  infant.'' \emph{Nature}, 1975.

\bibitem{tomasello1983joint}
M.~Tomasello and J.~Todd, ``Joint attention and lexical acquisition style,''
  \emph{First language}, vol.~4, no.~12, pp. 197--211, 1983.

\bibitem{botero2016tactless}
M.~Botero, ``Tactless scientists: Ignoring touch in the study of joint
  attention,'' \emph{Philosophical Psychology}, vol.~29, no.~8, pp. 1200--1214,
  2016.

\bibitem{sheldrake2016joint}
R.~Sheldrake and A.~Beeharee, ``Is joint attention detectable at a distance?
  {Three} automated, internet-based tests,'' \emph{Explore: The Journal of
  Science and Healing}, vol.~12, no.~1, pp. 34--41, 2016.

\bibitem{moore1997role}
C.~Moore, M.~Angelopoulos, and P.~Bennett, ``The role of movement in the
  development of joint visual attention,'' \emph{Infant Behavior and
  Development}, vol.~20, no.~1, pp. 83--92, 1997.

\bibitem{mundy1997joint}
P.~Mundy and M.~Crowson, ``Joint attention and early social communication:
  Implications for research on intervention with autism,'' \emph{Journal of
  Autism and Developmental disorders}, vol.~27, no.~6, pp. 653--676, 1997.

\bibitem{dube2004toward}
W.~V. Dube, R.~P. MacDonald, R.~C. Mansfield, W.~L. Holcomb, and W.~H. Ahearn,
  ``Toward a behavioral analysis of joint attention,'' \emph{The Behavior
  Analyst}, vol.~27, no.~2, p. 197, 2004.

\bibitem{holth2005operant}
P.~Holth, ``An operant analysis of joint attention skills.'' \emph{Journal of
  Early and Intensive Behavior Intervention}, vol.~2, no.~3, p. 160, 2005.

\bibitem{tomasello2007shared}
M.~Tomasello and M.~Carpenter, ``Shared intentionality,'' \emph{Developmental
  science}, vol.~10, no.~1, pp. 121--125, 2007.

\bibitem{charman2000testing}
T.~Charman, S.~Baron-Cohen, J.~Swettenham, G.~Baird, A.~Cox, and A.~Drew,
  ``Testing joint attention, imitation, and play as infancy precursors to
  language and theory of mind,'' \emph{Cognitive development}, vol.~15, no.~4,
  pp. 481--498, 2000.

\bibitem{carpenter1998social}
M.~Carpenter, K.~Nagell, M.~Tomasello, G.~Butterworth, and C.~Moore, ``Social
  cognition, joint attention, and communicative competence from 9 to 15 months
  of age,'' \emph{Monographs of the society for research in child development},
  pp. i--174, 1998.

\bibitem{mundy1998individual}
P.~Mundy and A.~Gomes, ``Individual differences in joint attention skill
  development in the second year,'' \emph{Infant behavior and development},
  vol.~21, no.~3, pp. 469--482, 1998.

\bibitem{macdonald2006behavioral}
R.~MacDonald, J.~Anderson, W.~V. Dube, A.~Geckeler, G.~Green, W.~Holcomb,
  R.~Mansfield, and J.~Sanchez, ``Behavioral assessment of joint attention: A
  methodological report,'' \emph{Research in Developmental Disabilities},
  vol.~27, no.~2, pp. 138--150, 2006.

\bibitem{deroche2016joint}
T.~Deroche, C.~Castanier, A.~Perrot, and A.~Hartley, ``Joint attention is
  slowed in older adults,'' \emph{Experimental aging research}, vol.~42, no.~2,
  pp. 144--150, 2016.

\bibitem{goffman1978presentation}
E.~Goffman \emph{et~al.}, \emph{The presentation of self in everyday
  life}.\hskip 1em plus 0.5em minus 0.4em\relax Harmondsworth, 1978.

\bibitem{badis1979}
P.~D. Bardis, ``Social interaction and social processes,'' \emph{Social
  Science}, vol.~54, no.~3, pp. 147--167, 1979.

\bibitem{sebanz2006joint}
N.~Sebanz, H.~Bekkering, and G.~Knoblich, ``Joint action: {Bodies} and minds
  moving together,'' \emph{Trends in cognitive sciences}, vol.~10, no.~2, pp.
  70--76, 2006.

\bibitem{fiebich2013joint}
A.~Fiebich and S.~Gallagher, ``Joint attention in joint action,''
  \emph{Philosophical Psychology}, vol.~26, no.~4, pp. 571--587, 2013.

\bibitem{nuku2008joint}
P.~Nuku and H.~Bekkering, ``Joint attention: Inferring what others perceive
  (and don’t perceive),'' \emph{Consciousness and Cognition}, vol.~17, no.~1,
  pp. 339--349, 2008.

\bibitem{sucha2017pedestrian}
M.~Sucha, D.~Dostal, and R.~Risser, ``Pedestrian-driver communication and
  decision strategies at marked crossings,'' \emph{Accident Analysis \&
  Prevention}, vol. 102, pp. 41--50, 2017.

\bibitem{fogassi2005parietal}
L.~Fogassi, P.~F. Ferrari, B.~Gesierich, S.~Rozzi, F.~Chersi, and
  G.~Rizzolatti, ``Parietal lobe: from action organization to intention
  understanding,'' \emph{Science}, vol. 308, no. 5722, pp. 662--667, 2005.

\bibitem{umilta2001know}
M.~A. Umilta, E.~Kohler, V.~Gallese, L.~Fogassi, L.~Fadiga, C.~Keysers, and
  G.~Rizzolatti, ``I know what you are doing: A neurophysiological study,''
  \emph{Neuron}, vol.~31, no.~1, pp. 155--165, 2001.

\bibitem{verfaillie2002representing}
K.~Verfaillie and A.~Daems, ``Representing and anticipating human actions in
  vision,'' \emph{Visual Cognition}, vol.~9, no. 1-2, pp. 217--232, 2002.

\bibitem{flanagan2003action}
J.~R. Flanagan and R.~S. Johansson, ``Action plans used in action
  observation,'' \emph{Nature}, vol. 424, no. 6950, p. 769, 2003.

\bibitem{dennett1981brainstorms}
D.~C. Dennett, \emph{Brainstorms: Philosophical essays on mind and
  psychology}.\hskip 1em plus 0.5em minus 0.4em\relax MIT press, 1981.

\bibitem{baron1995mindblindness}
S.~Baron-Cohen, ``Mindblindness: An essay on autism and theory of mind.
  cambridge, ma: Bradford,'' 1995.

\bibitem{humphrey1984consciousness}
N.~Humphrey, \emph{Consciousness regained: Chapters in the development of
  mind}.\hskip 1em plus 0.5em minus 0.4em\relax Nicholas Humphrey, 1984.

\bibitem{sperber1987precis}
D.~Sperber and D.~Wilson, ``Precis of relevance: Communication and cognition,''
  \emph{Behavioral and brain sciences}, vol.~10, no.~4, pp. 697--710, 1987.

\bibitem{briton1995beliefs}
N.~J. Briton and J.~A. Hall, ``Beliefs about female and male nonverbal
  communication,'' \emph{Sex Roles}, vol.~32, no.~1, pp. 79--90, 1995.

\bibitem{hecht1999nonverbal}
M.~A. Hecht and N.~Ambady, ``Nonverbal communication and psychology: Past and
  future,'' \emph{Atlantic Journal of Communication}, vol.~7, no.~2, pp.
  156--170, 1999.

\bibitem{buck2002verbal}
R.~Buck and C.~A. VanLear, ``Verbal and nonverbal communication: Distinguishing
  symbolic, spontaneous, and pseudo-spontaneous nonverbal behavior,''
  \emph{Journal of Communication}, vol.~52, no.~3, pp. 522--541, 2002.

\bibitem{darwin1998expression}
C.~Darwin, \emph{The expression of the emotions in man and animals}.\hskip 1em
  plus 0.5em minus 0.4em\relax Oxford University Press, USA, 1998.

\bibitem{krauss1996nonverbal}
R.~M. Krauss, Y.~Chen, and P.~Chawla, ``Nonverbal behavior and nonverbal
  communication: What do conversational hand gestures tell us?'' \emph{Advances
  in experimental social psychology}, vol.~28, pp. 389--450, 1996.

\bibitem{mehrabian1976public}
A.~Mehrabian, \emph{Public places and private spaces: the psychology of work,
  play, and living environments}.\hskip 1em plus 0.5em minus 0.4em\relax Basic
  Books New York, 1976.

\bibitem{birdwhistell2010kinesics}
R.~L. Birdwhistell, \emph{Kinesics and context: Essays on body motion
  communication}.\hskip 1em plus 0.5em minus 0.4em\relax University of
  Pennsylvania press, 2010.

\bibitem{dimatteo1980predicting}
M.~R. DiMatteo, A.~Taranta, H.~S. Friedman, and L.~M. Prince, ``Predicting
  patient satisfaction from physicians' nonverbal communication skills,''
  \emph{Medical care}, pp. 376--387, 1980.

\bibitem{nowicki1994individual}
S.~Nowicki and M.~P. Duke, ``Individual differences in the nonverbal
  communication of affect: The diagnostic analysis of nonverbal accuracy
  scale,'' \emph{Journal of Nonverbal behavior}, vol.~18, no.~1, pp. 9--35,
  1994.

\bibitem{argyle1965eye}
M.~Argyle and J.~Dean, ``Eye-contact, distance and affiliation,''
  \emph{Sociometry}, pp. 289--304, 1965.

\bibitem{senju2009eye}
A.~Senju and M.~H. Johnson, ``The eye contact effect: mechanisms and
  development,'' \emph{Trends in cognitive sciences}, vol.~13, no.~3, pp.
  127--134, 2009.

\bibitem{rothenbucher2016ghost}
D.~Rothenb{\"u}cher, J.~Li, D.~Sirkin, B.~Mok, and W.~Ju, ``Ghost driver: A
  field study investigating the interaction between pedestrians and driverless
  vehicles,'' in \emph{International Symposium on Robot and Human Interactive
  Communication (RO-MAN)}.\hskip 1em plus 0.5em minus 0.4em\relax IEEE, 2016,
  pp. 795--802.

\bibitem{gueguen2015pedestrian}
N.~Gu{\'e}guen, S.~Meineri, and C.~Eyssartier, ``A pedestrian’s stare and
  drivers’ stopping behavior: A field experiment at the pedestrian
  crossing,'' \emph{Safety science}, vol.~75, pp. 87--89, 2015.

\bibitem{risser1985behavior}
R.~Risser, ``Behavior in traffic conflict situations,'' \emph{Accident Analysis
  \& Prevention}, vol.~17, no.~2, pp. 179--197, 1985.

\bibitem{scheflen1964significance}
A.~E. Scheflen, ``The significance of posture in communication systems,''
  \emph{Psychiatry}, vol.~27, no.~4, pp. 316--331, 1964.

\bibitem{wilde1980immediate}
G.~Wilde, ``Immediate and delayed social interaction in road user behaviour,''
  \emph{Applied Psychology}, vol.~29, no.~4, pp. 439--460, 1980.

\bibitem{clay1995driver}
D.~Clay, ``Driver attitude and attribution: implications for accident
  prevention,'' Ph.D. dissertation, Cranfield University, 1995.

\bibitem{rasouliunderstanding}
A.~Rasouli, I.~Kotseruba, and J.~K. Tsotsos, ``Understanding pedestrian
  behavior in complex traffic scenes,'' \emph{IEEE Transactions on Intelligent
  Vehicles}, vol.~PP, no.~99, pp. 1--1, 2017.

\bibitem{varhelyi1998drivers}
A.~Varhelyi, ``Drivers' speed behaviour at a zebra crossing: a case study,''
  \emph{Accident Analysis \& Prevention}, vol.~30, no.~6, pp. 731--743, 1998.

\bibitem{klienke1977compliance}
C.~Klienke, ``Compliance to requests made by gazing and touching,'' \emph{J.
  exp. soc. Psychol.}, vol.~13, pp. 218--223, 1977.

\bibitem{walker2007drivers}
I.~Walker and M.~Brosnan, ``Drivers’ gaze fixations during judgements about a
  bicyclist’s intentions,'' \emph{Transportation research part F: traffic
  psychology and behaviour}, vol.~10, no.~2, pp. 90--98, 2007.

\bibitem{hamlet1984eye}
C.~C. Hamlet, S.~Axelrod, and S.~Kuerschner, ``Eye contact as an antecedent to
  compliant behavior,'' \emph{Journal of Applied Behavior Analysis}, vol.~17,
  no.~4, pp. 553--557, 1984.

\bibitem{price2000relationship}
J.~M. Price and S.~J. Glynn, ``The relationship between crash rates and
  drivers' hazard assessments using the connecticut photolog,'' in
  \emph{Proceedings of the Human Factors and Ergonomics Society Annual
  Meeting}, vol.~44, no.~20.\hskip 1em plus 0.5em minus 0.4em\relax SAGE
  Publications, 2000, pp. 3--263.

\bibitem{crundall1999driving}
D.~Crundall, ``Driving experience and the acquisition of visual information,''
  Ph.D. dissertation, University of Nottingham, 1999.

\bibitem{sullivan2011differences}
J.~M. Sullivan and M.~J. Flannagan, ``Differences in geometry of pedestrian
  crashes in daylight and darkness,'' \emph{Journal of safety research},
  vol.~42, no.~1, pp. 33--37, 2011.

\bibitem{tom2011gender}
A.~Tom and M.-A. Grani{\'e}, ``Gender differences in pedestrian rule compliance
  and visual search at signalized and unsignalized crossroads,'' \emph{Accident
  Analysis \& Prevention}, vol.~43, no.~5, pp. 1794--1801, 2011.

\bibitem{reed2008intersection}
M.~Reed, ``Intersection kinematics: a pilot study of driver turning behavior
  with application to pedestrian obscuration by a-pillars,'' University of
  Michigan, Tech. Rep., 2008.

\bibitem{tawari2014attention}
A.~Tawari, A.~M{\o}gelmose, S.~Martin, T.~B. Moeslund, and M.~M. Trivedi,
  ``Attention estimation by simultaneous analysis of viewer and view,'' in
  \emph{Intelligent Transportation Systems Conference (ITSC)}.\hskip 1em plus
  0.5em minus 0.4em\relax IEEE, 2014, pp. 1381--1387.

\bibitem{heimstra1969experimental}
N.~W. Heimstra, J.~Nichols, and G.~Martin, ``An experimental methodology for
  analysis of child pedestrian behavior,'' \emph{Pediatrics}, vol.~44, no.~5,
  pp. 832--838, 1969.

\bibitem{neale2005overview}
V.~L. Neale, T.~A. Dingus, S.~G. Klauer, J.~Sudweeks, and M.~Goodman, ``An
  overview of the 100-car naturalistic study and findings,'' \emph{National
  Highway Traffic Safety Administration, Paper}, no. 05-0400, 2005.

\bibitem{eenink2014udrive}
R.~Eenink, Y.~Barnard, M.~Baumann, X.~Augros, and F.~Utesch, ``{UDRIVE}: the
  european naturalistic driving study,'' in \emph{Proceedings of Transport
  Research Arena}.\hskip 1em plus 0.5em minus 0.4em\relax IFSTTAR, 2014.

\bibitem{wang2010study}
T.~Wang, J.~Wu, P.~Zheng, and M.~McDonald, ``Study of pedestrians' gap
  acceptance behavior when they jaywalk outside crossing facilities,'' in
  \emph{Intelligent Transportation Systems (ITSC)}.\hskip 1em plus 0.5em minus
  0.4em\relax IEEE, 2010, pp. 1295--1300.

\bibitem{rosenbloom2009crossing}
T.~Rosenbloom, ``Crossing at a red light: Behaviour of individuals and
  groups,'' \emph{Transportation research part F: traffic psychology and
  behaviour}, vol.~12, no.~5, pp. 389--394, 2009.

\bibitem{ishaque2008behavioural}
M.~M. Ishaque and R.~B. Noland, ``Behavioural issues in pedestrian speed choice
  and street crossing behaviour: a review,'' \emph{Transport Reviews}, vol.~28,
  no.~1, pp. 61--85, 2008.

\bibitem{johnston1973road}
D.~Johnston, ``Road accident casuality: A critique of the literature and an
  illustrative case,'' \emph{Ontario: Grand Rounds. Department of Psy chiatry,
  Hotel Dieu Hospital}, 1973.

\bibitem{gheri1963blickverhalten}
M.~Gheri, ``{\"U}ber das blickverhalten von kraftfahrern an kreuzungen,''
  \emph{Kuratorium f{\"u}r Verkehrssicherheit, Kleine Fachbuchreihe Bd},
  vol.~5, 1963.

\bibitem{vsucha2014road}
M.~{\v{S}}ucha, ``Road users’ strategies and communication: driver-pedestrian
  interaction,'' \emph{Transport Research Arena (TRA)}, 2014.

\bibitem{yagil2000beliefs}
D.~Yagil, ``Beliefs, motives and situational factors related to pedestrians’
  self-reported behavior at signal-controlled crossings,'' \emph{Transportation
  Research Part F: Traffic Psychology and Behaviour}, vol.~3, no.~1, pp. 1--13,
  2000.

\bibitem{lefkowitz1955status}
M.~Lefkowitz, R.~R. Blake, and J.~S. Mouton, ``Status factors in pedestrian
  violation of traffic signals.'' \emph{The Journal of Abnormal and Social
  Psychology}, vol.~51, no.~3, p. 704, 1955.

\bibitem{papadimitriou2009critical}
E.~Papadimitriou, G.~Yannis, and J.~Golias, ``A critical assessment of
  pedestrian behaviour models,'' \emph{Transportation research part F: traffic
  psychology and behaviour}, vol.~12, no.~3, pp. 242--255, 2009.

\bibitem{twisk2012understanding}
D.~Twisk, N.~Van~Nes, and J.~Haupt, ``Understanding safety critical
  interactions between bicycles and motor vehicles in europe by means of
  naturalistic driving techniques,'' in \emph{Proceedings of the first
  international cycling safety conference}, 2012.

\bibitem{goldhammer2014analysis}
M.~Goldhammer, A.~Hubert, S.~Koehler, K.~Zindler, U.~Brunsmann, K.~Doll, and
  B.~Sick, ``Analysis on termination of pedestrians' gait at urban
  intersections,'' in \emph{Intelligent Transportation Systems (ITSC)}.\hskip
  1em plus 0.5em minus 0.4em\relax IEEE, 2014, pp. 1758--1763.

\bibitem{waizman2015micro}
G.~Waizman, S.~Shoval, and I.~Benenson, ``Micro-simulation model for assessing
  the risk of vehicle--pedestrian road accidents,'' \emph{Journal of
  Intelligent Transportation Systems}, vol.~19, no.~1, pp. 63--77, 2015.

\bibitem{rasouliagree}
A.~Rasouli, I.~Kotseruba, and J.~K. Tsotsos, ``Agreeing to cross: How drivers
  and pedestrians communicate,'' in \emph{Intelligent Vehicles Symposium (IV)},
  June 2017, pp. 264--269.

\bibitem{oudejans1996cross}
R.~R. Oudejans, C.~F. Michaels, B.~van Dort, and E.~J. Frissen, ``To cross or
  not to cross: The effect of locomotion on street-crossing behavior,''
  \emph{Ecological psychology}, vol.~8, no.~3, pp. 259--267, 1996.

\bibitem{crossing2010}
\BIBentryALTinterwordspacing
D.~Cottingham, ``Pedestrian crossings and islands,'' Online, 2017-06-3.
  [Online]. Available:
  \url{http://mocktheorytest.com/resources/pedestrian-crossings-and-islands/}
\BIBentrySTDinterwordspacing

\bibitem{tian2013pilot}
R.~Tian, E.~Y. Du, K.~Yang, P.~Jiang, F.~Jiang, Y.~Chen, R.~Sherony, and
  H.~Takahashi, ``Pilot study on pedestrian step frequency in naturalistic
  driving environment,'' in \emph{Intelligent Vehicles Symposium (IV)}.\hskip
  1em plus 0.5em minus 0.4em\relax IEEE, 2013, pp. 1215--1220.

\bibitem{crompton1979pedestrian}
D.~Crompton, ``Pedestrian delay, annoyance and risk: preliminary results from a
  2 years study,'' in \emph{Proceedings of PTRC Summer Annual Meeting}, 1979,
  pp. 275--299.

\bibitem{underwood2003visual}
G.~Underwood, P.~Chapman, N.~Brocklehurst, J.~Underwood, and D.~Crundall,
  ``Visual attention while driving: {Sequences} of eye fixations made by
  experienced and novice drivers,'' \emph{Ergonomics}, vol.~46, no.~6, pp.
  629--646, 2003.

\bibitem{klauer2005driver}
S.~G. Klauer, V.~L. Neale, T.~A. Dingus, D.~Ramsey, and J.~Sudweeks, ``Driver
  inattention: A contributing factor to crashes and near-crashes,'' in
  \emph{Proceedings of the Human Factors and Ergonomics Society Annual
  Meeting}, vol.~49, no.~22.\hskip 1em plus 0.5em minus 0.4em\relax SAGE
  Publications Sage CA: Los Angeles, CA, 2005, pp. 1922--1926.

\bibitem{underwood2007visual}
G.~Underwood, ``Visual attention and the transition from novice to advanced
  driver,'' \emph{Ergonomics}, vol.~50, no.~8, pp. 1235--1249, 2007.

\bibitem{barnard2016study}
Y.~Barnard, F.~Utesch, N.~Nes, R.~Eenink, and M.~Baumann, ``The study design of
  {UDRIVE}: the naturalistic driving study across europe for cars, trucks and
  scooters,'' \emph{European Transport Research Review}, vol.~8, no.~2, pp.
  1--10, 2016.

\bibitem{sabey1975interacting}
B.~E. Sabey and G.~Staughton, ``Interacting roles of road environment vehicle
  and road user in accidents,'' \emph{Ceste I Mostovi}, 1975.

\bibitem{nasar2008mobile}
J.~Nasar, P.~Hecht, and R.~Wener, ``Mobile telephones, distracted attention,
  and pedestrian safety,'' \emph{Accident analysis \& prevention}, vol.~40,
  no.~1, pp. 69--75, 2008.

\bibitem{schwebel2012distraction}
D.~C. Schwebel, D.~Stavrinos, K.~W. Byington, T.~Davis, E.~E. O’Neal, and
  D.~De~Jong, ``Distraction and pedestrian safety: {How} talking on the phone,
  texting, and listening to music impact crossing the street,'' \emph{Accident
  Analysis \& Prevention}, vol.~45, pp. 266--271, 2012.

\bibitem{hyman2010did}
I.~E. Hyman, S.~M. Boss, B.~M. Wise, K.~E. McKenzie, and J.~M. Caggiano, ``Did
  you see the unicycling clown? {Inattentional} blindness while walking and
  talking on a cell phone,'' \emph{Applied Cognitive Psychology}, vol.~24,
  no.~5, pp. 597--607, 2010.

\bibitem{schmidt2009pedestrians}
S.~Schmidt and B.~F{\"a}rber, ``Pedestrians at the kerb--recognising the action
  intentions of humans,'' \emph{Transportation research part F: traffic
  psychology and behaviour}, vol.~12, no.~4, pp. 300--310, 2009.

\bibitem{lindgren2008requirements}
A.~Lindgren, F.~Chen, P.~W. Jordan, and H.~Zhang, ``Requirements for the design
  of advanced driver assistance systems-the differences between {Swedish} and
  {Chinese} drivers,'' \emph{International Journal of Design}, vol.~2, no.~2,
  2008.

\bibitem{bjorklund2005driver}
G.~M. Bj{\"o}rklund and L.~{\AA}berg, ``Driver behaviour in intersections:
  Formal and informal traffic rules,'' \emph{Transportation Research Part F:
  Traffic Psychology and Behaviour}, vol.~8, no.~3, pp. 239--253, 2005.

\bibitem{rosenbloom2004heaven}
T.~Rosenbloom, H.~Barkan, and D.~Nemrodov, ``For heaven’s sake keep the
  rules: Pedestrians’ behavior at intersections in ultra-orthodox and secular
  cities,'' \emph{Transportation Research Part F}, vol.~7, pp. 395--404, 2004.

\bibitem{sun2015estimation}
R.~Sun, X.~Zhuang, C.~Wu, G.~Zhao, and K.~Zhang, ``The estimation of vehicle
  speed and stopping distance by pedestrians crossing streets in a naturalistic
  traffic environment,'' \emph{Transportation research part F: traffic
  psychology and behaviour}, vol.~30, pp. 97--106, 2015.

\bibitem{harrell1991factors}
W.~A. Harrell, ``Factors influencing pedestrian cautiousness in crossing
  streets,'' \emph{The Journal of Social Psychology}, vol. 131, no.~3, pp.
  367--372, 1991.

\bibitem{lin2016impact}
P.-S. Lin, Z.~Wang, and R.~Guo, ``Impact of connected vehicles and autonomous
  vehicles on future transportation,'' \emph{Bridging the East and West},
  p.~46, 2016.

\bibitem{yangpedestrian}
\BIBentryALTinterwordspacing
E.~CYingzi~Du, K.~Yang, F.~Jiang, P.~Jiang, R.~Tian, M.~Luzetski, Y.~Chen,
  R.~Sherony, and H.~Takahashi, ``Pedestrian behavior analysis using 110-car
  naturalistic driving data in {USA},'' Online, 2017-06-3. [Online]. Available:
  \url{https://www-nrd.nhtsa.dot.gov/pdf/Esv/esv23/23ESV-000291.pdf}
\BIBentrySTDinterwordspacing

\bibitem{caird1994perception}
J.~Caird and P.~Hancock, ``The perception of arrival time for different
  oncoming vehicles at an intersection,'' \emph{Ecological Psychology}, vol.~6,
  no.~2, pp. 83--109, 1994.

\bibitem{edwards1954theory}
W.~Edwards, ``The theory of decision making.'' \emph{Psychological bulletin},
  vol.~51, no.~4, p. 380, 1954.

\bibitem{legrenzi1993focussing}
P.~Legrenzi, V.~Girotto, and P.~N. Johnson-Laird, ``Focussing in reasoning and
  decision making,'' \emph{Cognition}, vol.~49, no.~1, pp. 37--66, 1993.

\bibitem{simon1959theories}
H.~A. Simon, ``Theories of decision-making in economics and behavioral
  science,'' \emph{The American economic review}, vol.~49, no.~3, pp. 253--283,
  1959.

\bibitem{carroll1993human}
J.~B. Carroll, \emph{Human cognitive abilities: A survey of factor-analytic
  studies}.\hskip 1em plus 0.5em minus 0.4em\relax Cambridge University Press,
  1993.

\bibitem{sternberg1978toward}
R.~J. Sternberg, ``Toward a unified componential theory of human reasoning.''
  DTIC Document, Tech. Rep., 1978.

\bibitem{barlow1974inductive}
H.~Barlow, ``Inductive inference, coding, perception, and language,''
  \emph{Perception}, vol.~3, no.~2, pp. 123--134, 1974.

\bibitem{johnson2015logic}
P.~Johnson-Laird, S.~S. Khemlani, and G.~P. Goodwin, ``Logic, probability, and
  human reasoning,'' \emph{Trends in cognitive sciences}, vol.~19, no.~4, pp.
  201--214, 2015.

\bibitem{oaksford2001probabilistic}
M.~Oaksford and N.~Chater, ``The probabilistic approach to human reasoning,''
  \emph{Trends in cognitive sciences}, vol.~5, no.~8, pp. 349--357, 2001.

\bibitem{aliseda2006abductive}
A.~Aliseda, \emph{Abductive reasoning}.\hskip 1em plus 0.5em minus 0.4em\relax
  Springer, 2006, vol. 330.

\bibitem{fischer2001abductive}
H.~R. Fischer, ``Abductive reasoning as a way of worldmaking,''
  \emph{Foundations of Science}, vol.~6, no.~4, pp. 361--383, 2001.

\bibitem{typesreason}
\BIBentryALTinterwordspacing
``Types of reasoning,'' Online, 2017-06-25. [Online]. Available:
  \url{http://changingminds.org/disciplines/argument/types_reasoning/types_reasoning.htm}
\BIBentrySTDinterwordspacing

\bibitem{zadeh1975fuzzy}
L.~A. Zadeh, ``Fuzzy logic and approximate reasoning,'' \emph{Synthese},
  vol.~30, no.~3, pp. 407--428, 1975.

\bibitem{cosmides1989logic}
L.~Cosmides, ``The logic of social exchange: Has natural selection shaped how
  humans reason? studies with the {Wason} selection task,'' \emph{Cognition},
  vol.~31, no.~3, pp. 187--276, 1989.

\bibitem{byrne1989spatial}
R.~M. Byrne and P.~N. Johnson-Laird, ``Spatial reasoning,'' \emph{Journal of
  memory and language}, vol.~28, no.~5, pp. 564--575, 1989.

\bibitem{frank1992qualitative}
A.~U. Frank, ``Qualitative spatial reasoning about distances and directions in
  geographic space,'' \emph{Journal of Visual Languages \& Computing}, vol.~3,
  no.~4, pp. 343--371, 1992.

\bibitem{vila1994survey}
L.~Vila, ``A survey on temporal reasoning in artificial intelligence,''
  \emph{AI Communications}, vol.~7, no.~1, pp. 4--28, 1994.

\bibitem{pani2001temporal}
A.~Pani and G.~Bhattacharjee, ``Temporal representation and reasoning in
  artificial intelligence: A review,'' \emph{Mathematical and Computer
  Modelling}, vol.~34, no. 1-2, pp. 55--80, 2001.

\bibitem{klenk2005solving}
M.~Klenk, K.~D. Forbus, E.~Tomai, H.~Kim, and B.~Kyckelhahn, ``Solving everyday
  physical reasoning problems by analogy using sketches,'' in \emph{National
  Conference on Artificial Intelligence}, vol.~20, no.~1, 2005, p. 209.

\bibitem{baillargeon1995physical}
R.~Baillargeon, ``Physical reasoning in infancy,'' \emph{The cognitive
  neurosciences}, pp. 181--204, 1995.

\bibitem{cohn1997qualitative}
A.~G. Cohn, ``Qualitative spatial representation and reasoning techniques,'' in
  \emph{Annual Conference on Artificial Intelligence}.\hskip 1em plus 0.5em
  minus 0.4em\relax Springer, 1997, pp. 1--30.

\bibitem{davis2008}
E.~Davis, ``Physical reasoning,'' \emph{Handbook of knowledge representation},
  vol.~1, pp. 597--620, 2008.

\bibitem{jaeger2016artificial}
H.~Jaeger, ``Artificial intelligence: Deep neural reasoning,'' \emph{Nature},
  vol. 538, no. 7626, pp. 467--468, 2016.

\bibitem{peng2015towards}
B.~Peng, Z.~Lu, H.~Li, and K.-F. Wong, ``Towards neural network-based
  reasoning,'' \emph{arXiv preprint arXiv:1508.05508}, 2015.

\bibitem{socher2013reasoning}
R.~Socher, D.~Chen, C.~D. Manning, and A.~Ng, ``Reasoning with neural tensor
  networks for knowledge base completion,'' in \emph{Advances in neural
  information processing systems}, 2013, pp. 926--934.

\bibitem{graves2016hybrid}
A.~Graves, G.~Wayne, M.~Reynolds, T.~Harley, I.~Danihelka,
  A.~Grabska-Barwi{\'n}ska, S.~G. Colmenarejo, E.~Grefenstette, T.~Ramalho,
  J.~Agapiou \emph{et~al.}, ``Hybrid computing using a neural network with
  dynamic external memory,'' \emph{Nature}, vol. 538, no. 7626, pp. 471--476,
  2016.

\bibitem{sutskever2014sequence}
I.~Sutskever, O.~Vinyals, and Q.~V. Le, ``Sequence to sequence learning with
  neural networks,'' in \emph{Advances in neural information processing
  systems}, 2014, pp. 3104--3112.

\bibitem{graves2014neural}
A.~Graves, G.~Wayne, and I.~Danihelka, ``Neural {Turing} machines,''
  \emph{arXiv preprint arXiv:1410.5401}, 2014.

\bibitem{sichman1998social}
J.~S. Sichman, R.~Conte, Y.~Demazeau, and C.~Castelfranchi, ``A social
  reasoning mechanism based on dependence networks,'' in \emph{Proceedings of
  11th European Conference on Artificial Intelligence}, 1998, pp. 416--420.

\bibitem{de1965social}
C.~B. De~Soto, M.~London, and S.~Handel, ``Social reasoning and spatial
  paralogic.'' \emph{Journal of Personality and Social Psychology}, vol.~2,
  no.~4, p. 513, 1965.

\bibitem{wright1998visual}
R.~D. Wright, \emph{Visual attention}.\hskip 1em plus 0.5em minus 0.4em\relax
  Oxford University Press, 1998.

\bibitem{tsotsos1995toward}
J.~K. Tsotsos, ``Toward a computational model of visual attention,'' in
  \emph{Early vision and beyond}.\hskip 1em plus 0.5em minus 0.4em\relax MIT
  Press, Cambridge, MA, 1995, pp. 207--218.

\bibitem{bruce2015computational}
N.~D. Bruce, C.~Wloka, N.~Frosst, S.~Rahman, and J.~K. Tsotsos, ``On
  computational modeling of visual saliency: Examining what’s right, and
  what’s left,'' \emph{Vision research}, vol. 116, pp. 95--112, 2015.

\bibitem{lee2009interaction}
Y.-C. Lee, J.~D. Lee, and L.~Ng~Boyle, ``The interaction of cognitive load and
  attention-directing cues in driving,'' \emph{Human factors}, vol.~51, no.~3,
  pp. 271--280, 2009.

\bibitem{klauer2006impact}
S.~G. Klauer, T.~A. Dingus, V.~L. Neale, J.~D. Sudweeks, D.~J. Ramsey
  \emph{et~al.}, ``The impact of driver inattention on near-crash/crash risk:
  An analysis using the 100-car naturalistic driving study data,'' 2006.

\bibitem{llaneras2013human}
R.~E. Llaneras, J.~Salinger, and C.~A. Green, ``Human factors issues associated
  with limited ability autonomous driving systems: Drivers’ allocation of
  visual attention to the forward roadway,'' in \emph{Proceedings of the 7th
  International Driving Symposium on Human Factors in Driver Assessment,
  Training and Vehicle Design}.\hskip 1em plus 0.5em minus 0.4em\relax Public
  Policy Center, University of Iowa Iowa City, 2013, pp. 92--98.

\bibitem{takeda2016electrophysiological}
Y.~Takeda, T.~Sato, K.~Kimura, H.~Komine, M.~Akamatsu, and J.~Sato,
  ``Electrophysiological evaluation of attention in drivers and passengers:
  Toward an understanding of drivers’ attentional state in autonomous
  vehicles,'' \emph{Transportation research part F: traffic psychology and
  behaviour}, vol.~42, pp. 140--150, 2016.

\bibitem{scassellati1996mechanisms}
B.~Scassellati, ``Mechanisms of shared attention for a humanoid robot,'' in
  \emph{Embodied Cognition and Action: Papers from the 1996 AAAI Fall
  Symposium}, vol.~4, no.~9, 1996, p.~21.

\bibitem{scassellati1999imitation}
{B. Scassellati}, ``Imitation and mechanisms of joint attention: A
  developmental structure for building social skills on a humanoid robot,'' pp.
  176--195, 1999.

\bibitem{breazeal2008social}
C.~Breazeal, A.~Takanishi, and T.~Kobayashi, ``Social robots that interact with
  people,'' in \emph{Springer handbook of robotics}.\hskip 1em plus 0.5em minus
  0.4em\relax Springer, 2008, pp. 1349--1369.

\bibitem{fong2003survey}
T.~Fong, I.~Nourbakhsh, and K.~Dautenhahn, ``A survey of socially interactive
  robots,'' \emph{Robotics and autonomous systems}, vol.~42, no.~3, pp.
  143--166, 2003.

\bibitem{scassellati1999knowing}
B.~Scassellati, ``Knowing what to imitate and knowing when you succeed,'' in
  \emph{Proceedings of the AISB’99 Symposium on Imitation in Animals and
  Artifacts}, 1999, pp. 105--113.

\bibitem{breazeal2005robot}
C.~Breazeal and R.~Brooks, ``Robot emotion: A functional perspective,''
  \emph{Who needs emotions}, pp. 271--310, 2005.

\bibitem{breazeal1999robot}
C.~Breazeal and J.~Velasquez, ``Robot in society: Friend or appliance,'' in
  \emph{Proceedings of the 1999 Autonomous Agents Workshop on Emotion-Based
  Agent Architectures}, 1999, pp. 18--26.

\bibitem{brooks1999cog}
R.~A. Brooks, C.~Breazeal, M.~Marjanovi{\'c}, B.~Scassellati, and M.~M.
  Williamson, ``The {Cog} project: Building a humanoid robot,'' in
  \emph{Computation for metaphors, analogy, and agents}.\hskip 1em plus 0.5em
  minus 0.4em\relax Springer, 1999, pp. 52--87.

\bibitem{breazeal2005effects}
C.~Breazeal, C.~D. Kidd, A.~L. Thomaz, G.~Hoffman, and M.~Berlin, ``Effects of
  nonverbal communication on efficiency and robustness in human-robot
  teamwork,'' in \emph{IROS}.\hskip 1em plus 0.5em minus 0.4em\relax IEEE,
  2005, pp. 708--713.

\bibitem{ishiguro2001robovie}
H.~Ishiguro, T.~Ono, M.~Imai, T.~Maeda, T.~Kanda, and R.~Nakatsu, ``Robovie: An
  interactive humanoid robot,'' \emph{Industrial robot: An international
  journal}, vol.~28, no.~6, pp. 498--504, 2001.

\bibitem{becker2010exploring}
C.~Becker-Asano, K.~Ogawa, S.~Nishio, and H.~Ishiguro, ``Exploring the uncanny
  valley with {Geminoid HI-1} in a real-world application,'' in
  \emph{Proceedings of IADIS International conference interfaces and human
  computer interaction}, 2010, pp. 121--128.

\bibitem{staudte2009visual}
M.~Staudte and M.~W. Crocker, ``Visual attention in spoken human-robot
  interaction,'' in \emph{the 4th ACM/IEEE international conference on Human
  robot interaction}.\hskip 1em plus 0.5em minus 0.4em\relax ACM, 2009, pp.
  77--84.

\bibitem{mutlu2009nonverbal}
B.~Mutlu, F.~Yamaoka, T.~Kanda, H.~Ishiguro, and N.~Hagita, ``Nonverbal leakage
  in robots: Communication of intentions through seemingly unintentional
  behavior,'' in \emph{the 4th ACM/IEEE international conference on Human robot
  interaction (HRI)}.\hskip 1em plus 0.5em minus 0.4em\relax ACM, 2009, pp.
  69--76.

\bibitem{zheng2015impacts}
M.~Zheng, A.~Moon, E.~A. Croft, and M.~Q.-H. Meng, ``Impacts of robot head gaze
  on robot-to-human handovers,'' \emph{International Journal of Social
  Robotics}, vol.~7, no.~5, pp. 783--798, 2015.

\bibitem{imai2003physical}
M.~Imai, T.~Ono, and H.~Ishiguro, ``Physical relation and expression: Joint
  attention for human-robot interaction,'' \emph{IEEE Transactions on
  Industrial Electronics}, vol.~50, no.~4, pp. 636--643, 2003.

\bibitem{staudte2011investigating}
M.~Staudte and M.~W. Crocker, ``Investigating joint attention mechanisms
  through spoken human--robot interaction,'' \emph{Cognition}, vol. 120, no.~2,
  pp. 268--291, 2011.

\bibitem{mutlu2013coordination}
B.~Mutlu, A.~Terrell, and C.-M. Huang, ``Coordination mechanisms in human-robot
  collaboration,'' in \emph{Proceedings of the Workshop on Collaborative
  Manipulation, 8th ACM/IEEE International Conference on Human-Robot
  Interaction (HRI)}, 2013.

\bibitem{yonezawa2007gaze}
T.~Yonezawa, H.~Yamazoe, A.~Utsumi, and S.~Abe, ``Gaze-communicative behavior
  of stuffed-toy robot with joint attention and eye contact based on ambient
  gaze-tracking,'' in \emph{Proceedings of the 9th international conference on
  Multimodal interfaces}.\hskip 1em plus 0.5em minus 0.4em\relax ACM, 2007, pp.
  140--145.

\bibitem{shiomi2006interactive}
M.~Shiomi, T.~Kanda, H.~Ishiguro, and N.~Hagita, ``Interactive humanoid robots
  for a science museum,'' in \emph{Proceedings of the 1st ACM SIGCHI/SIGART
  conference on Human-robot interaction (HRI)}.\hskip 1em plus 0.5em minus
  0.4em\relax ACM, 2006, pp. 305--312.

\bibitem{miyauchi2004active}
D.~Miyauchi, A.~Sakurai, A.~Nakamura, and Y.~Kuno, ``Active eye contact for
  human-robot communication,'' in \emph{CHI'04 Extended Abstracts on Human
  Factors in Computing Systems}.\hskip 1em plus 0.5em minus 0.4em\relax ACM,
  2004, pp. 1099--1102.

\bibitem{kim2015human}
S.~Kim, Z.~Yu, J.~Kim, A.~Ojha, and M.~Lee, ``Human-robot interaction using
  intention recognition,'' in \emph{Proceedings of the 3rd International
  Conference on Human-Agent Interaction}.\hskip 1em plus 0.5em minus
  0.4em\relax ACM, 2015, pp. 299--302.

\bibitem{monajjemi2015uav}
M.~Monajjemi, J.~Bruce, S.~A. Sadat, J.~Wawerla, and R.~Vaughan, ``{UAV}, do
  you see me? {Establishing} mutual attention between an uninstrumented human
  and an outdoor {UAV} in flight,'' in \emph{IROS}.\hskip 1em plus 0.5em minus
  0.4em\relax IEEE, 2015, pp. 3614--3620.

\bibitem{yucel2009head}
Z.~Y{\"u}cel and A.~A. Salah, ``Head pose and neural network based gaze
  direction estimation for joint attention modeling in embodied agents,'' in
  \emph{Annual Meeting of Cognitive Science Society}, 2009.

\bibitem{yucel2009joint}
Z.~Yucel, A.~A. Salah, C.~Meri{\c{c}}li, and T.~Meri{\c{c}}li, ``Joint visual
  attention modeling for naturally interacting robotic agents,'' in
  \emph{International Symposium on Computer and Information Sciences
  (ISCIS)}.\hskip 1em plus 0.5em minus 0.4em\relax IEEE, 2009, pp. 242--247.

\bibitem{yucel2013joint}
Z.~Y{\"u}cel, A.~A. Salah, {\c{C}}.~Meri{\c{c}}li, T.~Meri{\c{c}}li,
  R.~Valenti, and T.~Gevers, ``Joint attention by gaze interpolation and
  saliency,'' \emph{IEEE Transactions on cybernetics}, vol.~43, no.~3, pp.
  829--842, 2013.

\bibitem{gorji2016attentional}
S.~Gorji and J.~J. Clark, ``Attentional push: Augmenting salience with shared
  attention modeling,'' \emph{arXiv preprint arXiv:1609.00072}, 2016.

\bibitem{kera2016discovering}
H.~Kera, R.~Yonetani, K.~Higuchi, and Y.~Sato, ``Discovering objects of joint
  attention via first-person sensing,'' in \emph{Conference on Computer Vision
  and Pattern Recognition Workshops (CVPRW)}, 2016, pp. 7--15.

\bibitem{nagai2002developmental}
Y.~Nagai, M.~Asada, and K.~Hosoda, ``Developmental learning model for joint
  attention,'' in \emph{IROS}, vol.~1.\hskip 1em plus 0.5em minus 0.4em\relax
  IEEE, 2002, pp. 932--937.

\bibitem{nagai2003does}
Y.~Nagai, K.~Hosoda, and M.~Asada, ``How does an infant acquire the ability of
  joint attention?: A constructive approach,'' 2003.

\bibitem{ito2004joint}
M.~Ito and J.~Tani, ``Joint attention between a humanoid robot and users in
  imitation game,'' in \emph{the International Conference on Development and
  Learning (ICDL)}, 2004.

\bibitem{nagai2005role}
Y.~Nagai, ``The role of motion information in learning human-robot joint
  attention,'' in \emph{ICRA}.\hskip 1em plus 0.5em minus 0.4em\relax IEEE,
  2005, pp. 2069--2074.

\bibitem{haasch2005multi}
A.~Haasch, N.~Hofemann, J.~Fritsch, and G.~Sagerer, ``A multi-modal object
  attention system for a mobile robot,'' in \emph{IROS}.\hskip 1em plus 0.5em
  minus 0.4em\relax IEEE, 2005, pp. 2712--2717.

\bibitem{schauerte2014look}
B.~Schauerte and R.~Stiefelhagen, ``“{Look} at this!” learning to guide
  visual saliency in human-robot interaction,'' in \emph{IROS}.\hskip 1em plus
  0.5em minus 0.4em\relax IEEE, 2014, pp. 995--1002.

\bibitem{billard1999experiments}
A.~Billard and K.~Dautenhahn, ``Experiments in learning by imitation-grounding
  and use of communication in robotic agents,'' \emph{Adaptive behavior},
  vol.~7, no. 3-4, pp. 415--438, 1999.

\bibitem{dautenhahn1999studying}
K.~Dautenhahn and A.~Billard, ``Studying robot social cognition within a
  developmental psychology framework,'' in \emph{Advanced Mobile Robots,
  1999.(Eurobot'99) 1999 Third European Workshop on}.\hskip 1em plus 0.5em
  minus 0.4em\relax IEEE, 1999, pp. 187--194.

\bibitem{kozima2001robot}
H.~Kozima and H.~Yano, ``A robot that learns to communicate with human
  caregivers,'' in \emph{Proceedings of the First International Workshop on
  Epigenetic Robotics}, 2001, pp. 47--52.

\bibitem{kozima2004can}
H.~Kozima, C.~Nakagawa, and H.~Yano, ``Can a robot empathize with people?''
  \emph{Artificial life and robotics}, vol.~8, no.~1, pp. 83--88, 2004.

\bibitem{kozima2009keepon}
H.~Kozima, M.~P. Michalowski, and C.~Nakagawa, ``Keepon,'' \emph{International
  Journal of Social Robotics}, vol.~1, no.~1, pp. 3--18, 2009.

\bibitem{dautenhahn2000issues}
K.~Dautenhahn and I.~Werry, ``Issues of robot-human interaction dynamics in the
  rehabilitation of children with autism,'' \emph{Proceedings From animals to
  animats}, vol.~6, pp. 519--528, 2000.

\bibitem{kozima1998attention}
H.~Kozima and A.~Ito, ``An attention-based approach to symbol acquisition,'' in
  \emph{Intelligent Control (ISIC) Held jointly with IEEE International
  Symposium on Computational Intelligence in Robotics and Automation (CIRA),
  Intelligent Systems and Semiotics (ISAS)}.\hskip 1em plus 0.5em minus
  0.4em\relax IEEE, 1998, pp. 852--856.

\bibitem{kozima2000epigenetic}
H.~Kozima and J.~Zlatev, ``An epigenetic approach to human-robot
  communication,'' in \emph{9th IEEE International Workshop on Robot and Human
  Interactive Communication}.\hskip 1em plus 0.5em minus 0.4em\relax IEEE,
  2000, pp. 346--351.

\bibitem{ito2004robots}
A.~Ito, S.~Hayakawa, and T.~Terada, ``Why robots need body for mind
  communication-an attempt of eye-contact between human and robot,'' in
  \emph{IEEE International Workshop on Robot and Human Interactive
  Communication}.\hskip 1em plus 0.5em minus 0.4em\relax IEEE, 2004, pp.
  473--478.

\bibitem{kozima2005interactive}
H.~Kozima, C.~Nakagawa, and Y.~Yasuda, ``Interactive robots for
  communication-care: A case-study in autism therapy,'' in \emph{IEEE
  International Workshop on Robot and Human Interactive Communication}.\hskip
  1em plus 0.5em minus 0.4em\relax IEEE, 2005, pp. 341--346.

\bibitem{hobert2015enhancements}
L.~Hobert, A.~Festag, I.~Llatser, L.~Altomare, F.~Visintainer, and A.~Kovacs,
  ``Enhancements of v2x communication in support of cooperative autonomous
  driving,'' \emph{IEEE Communications Magazine}, vol.~53, no.~12, pp. 64--70,
  2015.

\bibitem{cheng2014index}
X.~Cheng, M.~Wen, L.~Yang, and Y.~Li, ``Index modulated {OFDM} with interleaved
  grouping for {V2X} communications,'' in \emph{Intelligent Transportation
  Systems (ITSC)}.\hskip 1em plus 0.5em minus 0.4em\relax IEEE, 2014, pp.
  1097--1104.

\bibitem{narla2013evolution}
S.~R. Narla, ``The evolution of connected vehicle technology: From smart
  drivers to smart cars to... self-driving cars,'' \emph{Institute of
  Transportation Engineers. ITE Journal}, vol.~83, no.~7, p.~22, 2013.

\bibitem{Cunningham2013}
\BIBentryALTinterwordspacing
W.~Cunningham, ``Honda tech warns drivers of pedestrian presence,'' Online,
  2017-06-30. [Online]. Available:
  \url{https://www.cnet.com/roadshow/news/honda-tech-warns-drivers-of-pedestrian-presence/}
\BIBentrySTDinterwordspacing

\bibitem{schmidt2015v2x}
T.~Schmidt, R.~Philipsen, and M.~Ziefle, ``From v2x to control2trust,'' in
  \emph{Proceedings of the Third International Conference on Human Aspects of
  Information Security, Privacy, and Trust}.\hskip 1em plus 0.5em minus
  0.4em\relax Springer-Verlag New York, Inc., 2015, pp. 570--581.

\bibitem{lagstrom2015avip}
T.~Lagstrom and V.~M. Lundgren, ``{AVIP}-autonomous vehicles interaction with
  pedestrians,'' Master's thesis, Chalmers {U}niversity of {T}echnology,
  Gothenborg, {S}weden, 2015.

\bibitem{googledisp}
C.~P. Urmson, I.~J. Mahon, D.~A. Dolgov, and J.~Zhu, ``Pedestrian
  notifications,'' US Patent US 9\,196\,164B1, 11 24, 2015.

\bibitem{mitslights}
\BIBentryALTinterwordspacing
``Mitsubishi electric introduces road-illuminating directional indicators,''
  Online, 2017-06-30. [Online]. Available:
  \url{http://www.mitsubishielectric.com/news/2015/1023.html?cid=rss}
\BIBentrySTDinterwordspacing

\bibitem{benzauto}
\BIBentryALTinterwordspacing
``Overview: {Mercedes-Benz F} 015 luxury in motion,'' Online, 2017-06-30.
  [Online]. Available:
  \url{http://media.daimler.com/marsMediaSite/en/instance/ko/Overview-Mercedes-Benz-F-015-Luxury-in-Motion.xhtml?oid=9904624}
\BIBentrySTDinterwordspacing

\bibitem{pennycooke2012aevita}
N.~Pennycooke, ``{AEVITA}: {Designing} biomimetic vehicle-to-pedestrian
  communication protocols for autonomously operating \& parking on-road
  electric vehicles,'' Master's thesis, Massachusetts Institute of Technology,
  2012.

\bibitem{mirnig2017three}
N.~Mirnig, N.~Perterer, G.~Stollnberger, and M.~Tscheligi, ``Three strategies
  for autonomous car-to-pedestrian communication: A survival guide,'' in
  \emph{ACM/IEEE International Conference on Human-Robot Interaction}.\hskip
  1em plus 0.5em minus 0.4em\relax ACM, 2017, pp. 209--210.

\bibitem{siess2015hybrid}
A.~Sie{\ss}, K.~H{\"u}bel, D.~Hepperle, A.~Dronov, C.~Hufnagel, J.~Aktun, and
  M.~W{\"o}lfel, ``Hybrid city lighting-improving pedestrians' safety through
  proactive street lighting,'' in \emph{International Conference on Cyberworlds
  (CW)}.\hskip 1em plus 0.5em minus 0.4em\relax IEEE, 2015, pp. 46--49.

\bibitem{smartroad}
\BIBentryALTinterwordspacing
``Smart highway,'' Online, 2017-06-30. [Online]. Available:
  \url{https://www.studioroosegaarde.net/projects/#smog-free-project}
\BIBentrySTDinterwordspacing

\bibitem{tapia2004activity}
E.~M. Tapia, S.~S. Intille, and K.~Larson, ``Activity recognition in the home
  using simple and ubiquitous sensors,'' in \emph{Pervasive}, vol.~4.\hskip 1em
  plus 0.5em minus 0.4em\relax Springer, 2004, pp. 158--175.

\bibitem{kwapisz2011activity}
J.~R. Kwapisz, G.~M. Weiss, and S.~A. Moore, ``Activity recognition using cell
  phone accelerometers,'' \emph{ACM SigKDD Explorations Newsletter}, vol.~12,
  no.~2, pp. 74--82, 2011.

\bibitem{bourke2008threshold}
A.~K. Bourke and G.~M. Lyons, ``A threshold-based fall-detection algorithm
  using a bi-axial gyroscope sensor,'' \emph{Medical engineering \& physics},
  vol.~30, no.~1, pp. 84--90, 2008.

\bibitem{vo2012fall}
Q.~V. Vo, G.~Lee, and D.~Choi, ``Fall detection based on movement and smart
  phone technology,'' in \emph{International Conference on Computing and
  Communication Technologies, Research, Innovation, and Vision for the Future
  (RIVF)}.\hskip 1em plus 0.5em minus 0.4em\relax IEEE, 2012, pp. 1--4.

\bibitem{lee2001recognition}
S.-W. Lee and K.~Mase, ``Recognition of walking behaviors for pedestrian
  navigation,'' in \emph{International Conference on Control
  Applications}.\hskip 1em plus 0.5em minus 0.4em\relax IEEE, 2001, pp.
  1152--1155.

\bibitem{bao2004activity}
L.~Bao and S.~Intille, ``Activity recognition from user-annotated acceleration
  data,'' \emph{Pervasive computing}, pp. 1--17, 2004.

\bibitem{lee2002activity}
S.~W. Lee and K.~Mase, ``Activity and location recognition using wearable
  sensors,'' \emph{IEEE pervasive computing}, vol.~1, no.~3, pp. 24--32, 2002.

\bibitem{kourogi2010method}
M.~Kourogi, T.~Ishikawa, and T.~Kurata, ``A method of pedestrian dead reckoning
  using action recognition,'' in \emph{Position Location and Navigation
  Symposium (PLANS)}.\hskip 1em plus 0.5em minus 0.4em\relax IEEE, 2010, pp.
  85--89.

\bibitem{tong2009research}
L.~Tong, W.~Chen, Q.~Song, and Y.~Ge, ``A research on automatic human fall
  detection method based on wearable inertial force information acquisition
  system,'' in \emph{International Conference on Robotics and Biomimetics
  (ROBIO)}.\hskip 1em plus 0.5em minus 0.4em\relax IEEE, 2009, pp. 949--953.

\bibitem{gao2016pedestrian}
N.~Gao and L.~Zhao, ``A pedestrian dead reckoning system using {SEMG} based on
  activities recognition,'' in \emph{Chinese Guidance, Navigation and Control
  Conference (CGNCC)}.\hskip 1em plus 0.5em minus 0.4em\relax IEEE, 2016, pp.
  2361--2365.

\bibitem{su2014activity}
X.~Su, H.~Tong, and P.~Ji, ``Activity recognition with smartphone sensors,''
  \emph{Tsinghua Science and Technology}, vol.~19, no.~3, pp. 235--249, 2014.

\bibitem{mehner2013location}
S.~Mehner, R.~Klauck, and H.~Koenig, ``Location-independent fall detection with
  smartphone,'' in \emph{Proceedings of the 6th International Conference on
  PErvasive Technologies Related to Assistive Environments}.\hskip 1em plus
  0.5em minus 0.4em\relax ACM, 2013, p.~11.

\bibitem{stormo2014human}
K.~Stormo, ``Human fall detection using distributed monostatic {UWB} radars,''
  Master's thesis, Institutt for teknisk kybernetikk, 2014.

\bibitem{amin2016radar}
M.~G. Amin, Y.~D. Zhang, F.~Ahmad, and K.~D. Ho, ``Radar signal processing for
  elderly fall detection: The future for in-home monitoring,'' \emph{IEEE
  Signal Processing Magazine}, vol.~33, no.~2, pp. 71--80, 2016.

\bibitem{wang2015understanding}
W.~Wang, A.~X. Liu, M.~Shahzad, K.~Ling, and S.~Lu, ``Understanding and
  modeling of wifi signal based human activity recognition,'' in
  \emph{Proceedings of the 21st annual international conference on mobile
  computing and networking}.\hskip 1em plus 0.5em minus 0.4em\relax ACM, 2015,
  pp. 65--76.

\bibitem{van2008accurate}
T.~Van~Kasteren, A.~Noulas, G.~Englebienne, and B.~Kr{\"o}se, ``Accurate
  activity recognition in a home setting,'' in \emph{Proceedings of the 10th
  international conference on Ubiquitous computing}.\hskip 1em plus 0.5em minus
  0.4em\relax ACM, 2008, pp. 1--9.

\bibitem{toreyin2005hmm}
B.~U. Toreyin, Y.~Dedeoglu, and A.~E. {\c{C}}etin, ``{HMM} based falling person
  detection using both audio and video,'' \emph{Lecture Notes in Computer
  Science}, vol. 3766, p. 211, 2005.

\bibitem{lara2013survey}
O.~D. Lara and M.~A. Labrador, ``A survey on human activity recognition using
  wearable sensors.'' \emph{IEEE Communications Surveys and Tutorials},
  vol.~15, no.~3, pp. 1192--1209, 2013.

\bibitem{rashidi2013survey}
P.~Rashidi and A.~Mihailidis, ``A survey on ambient-assisted living tools for
  older adults,'' \emph{IEEE journal of biomedical and health informatics},
  vol.~17, no.~3, pp. 579--590, 2013.

\bibitem{bertozzi2000vision}
M.~Bertozzi, A.~Broggi, and A.~Fascioli, ``Vision-based intelligent vehicles:
  State of the art and perspectives,'' \emph{Robotics and Autonomous systems},
  vol.~32, no.~1, pp. 1--16, 2000.

\bibitem{nene1996columbia}
S.~A. Nene, S.~K. Nayar, H.~Murase \emph{et~al.}, ``Columbia object image
  library ({COIL}-20),'' 1996.

\bibitem{lecun1998gradient}
Y.~LeCun, L.~Bottou, Y.~Bengio, and P.~Haffner, ``Gradient-based learning
  applied to document recognition,'' \emph{Proceedings of the IEEE}, vol.~86,
  no.~11, pp. 2278--2324, 1998.

\bibitem{winn2005object}
J.~Winn, A.~Criminisi, and T.~Minka, ``Object categorization by learned
  universal visual dictionary,'' in \emph{ICCV}, vol.~2.\hskip 1em plus 0.5em
  minus 0.4em\relax IEEE, 2005, pp. 1800--1807.

\bibitem{fergus2003object}
R.~Fergus, P.~Perona, and A.~Zisserman, ``Object class recognition by
  unsupervised scale-invariant learning,'' in \emph{CVPR}, vol.~2.\hskip 1em
  plus 0.5em minus 0.4em\relax IEEE, 2003, pp. II--II.

\bibitem{fei2006one}
L.~Fei-Fei, R.~Fergus, and P.~Perona, ``One-shot learning of object
  categories,'' \emph{PAMI}, vol.~28, no.~4, pp. 594--611, 2006.

\bibitem{pascal-voc-2006}
E.~M., A.~Zisserman, C.~K.~I. Williams, and L.~Van~Gool, ``The {PASCAL}
  {V}isual {O}bject {C}lasses {C}hallenge 2006 {(VOC2006)} {R}esults,''
  http://www.pascal-network.org/challenges/VOC/voc2006/results.pdf.

\bibitem{pascal-voc-2007}
M.~Everingham, L.~Van~Gool, C.~K.~I. Williams, J.~Winn, and A.~Zisserman, ``The
  {PASCAL} {V}isual {O}bject {C}lasses {C}hallenge 2007 {(VOC2007)}
  {R}esults,''
  http://www.pascal-network.org/challenges/VOC/voc2007/workshop/index.html.

\bibitem{pascal-voc-2008}
M.~Everingham, L.~Van~Gool, C.~K. Williams, J.~Winn, and A.~Zisserman, ``The
  {PASCAL} {V}isual {O}bject {C}lasses {C}hallenge 2008 {(VOC2008)}
  {R}esults,''
  http://www.pascal-network.org/challenges/VOC/voc2008/workshop/index.html.

\bibitem{choi_cvpr10}
M.~J. Choi, J.~J. Lim, A.~Torralba, and A.~S. Willsky, ``Exploiting
  hierarchical context on a large database of object categories,'' in
  \emph{CVPR}, 2010.

\bibitem{lampert2009learning}
C.~H. Lampert, H.~Nickisch, and S.~Harmeling, ``Learning to detect unseen
  object classes by between-class attribute transfer,'' in \emph{CVPR}.\hskip
  1em plus 0.5em minus 0.4em\relax IEEE, 2009, pp. 951--958.

\bibitem{nus-wide-civr09}
T.-S. Chua, J.~Tang, R.~Hong, H.~Li, Z.~Luo, and Y.-T. Zheng, ``{NUS-WIDE}: A
  real-world web image database from {National University of Singapore},'' in
  \emph{ACM Conference on Image and Video Retrieval (CIVR)}, Santorini,
  Greece., 2009.

\bibitem{krizhevsky2009learning}
A.~Krizhevsky and G.~Hinton, ``Learning multiple layers of features from tiny
  images,'' 2009.

\bibitem{ILSVRC15}
O.~Russakovsky, J.~Deng, H.~Su, J.~Krause, S.~Satheesh, S.~Ma, Z.~Huang,
  A.~Karpathy, A.~Khosla, M.~Bernstein, A.~C. Berg, and L.~Fei-Fei, ``{ImageNet
  Large Scale Visual Recognition Challenge},'' \emph{IJCV}, vol. 115, no.~3,
  pp. 211--252, 2015.

\bibitem{pascal-voc-2010}
M.~Everingham, L.~Van~Gool, C.~K.~I. Williams, J.~Winn, and A.~Zisserman, ``The
  {PASCAL} {V}isual {O}bject {C}lasses {C}hallenge 2010 {(VOC2010)}
  {R}esults,''
  http://www.pascal-network.org/challenges/VOC/voc2010/workshop/index.html.

\bibitem{xiao2010sun}
J.~Xiao, J.~Hays, K.~A. Ehinger, A.~Oliva, and A.~Torralba, ``{SUN} database:
  Large-scale scene recognition from abbey to zoo,'' in \emph{CVPR}.\hskip 1em
  plus 0.5em minus 0.4em\relax IEEE, 2010, pp. 3485--3492.

\bibitem{lai2011large}
K.~Lai, L.~Bo, X.~Ren, and D.~Fox, ``A large-scale hierarchical multi-view
  {RGB-D} object dataset,'' in \emph{ICRA}.\hskip 1em plus 0.5em minus
  0.4em\relax IEEE, 2011, pp. 1817--1824.

\bibitem{aldoma2014automation}
A.~Aldoma, T.~F{\"a}ulhammer, and M.~Vincze, ``Automation of "ground truth"
  annotation for multi-view {RGB-D} object instance recognition datasets,'' in
  \emph{IROS}.\hskip 1em plus 0.5em minus 0.4em\relax IEEE, 2014, pp.
  5016--5023.

\bibitem{browatzki2011going}
B.~Browatzki, J.~Fischer, B.~Graf, H.~H. B{\"u}lthoff, and C.~Wallraven,
  ``Going into depth: Evaluating {2D} and {3D} cues for object classification
  on a new, large-scale object dataset,'' in \emph{ICCVW}.\hskip 1em plus 0.5em
  minus 0.4em\relax IEEE, 2011, pp. 1189--1195.

\bibitem{Patterson2012SunAttributes}
G.~Patterson and J.~Hays, ``{SUN} attribute database: Discovering, annotating,
  and recognizing scene attributes,'' in \emph{CVPR}, 2012.

\bibitem{netzer2011reading}
Y.~Netzer, T.~Wang, A.~Coates, A.~Bissacco, B.~Wu, and A.~Y. Ng, ``Reading
  digits in natural images with unsupervised feature learning,'' in
  \emph{NIPSW}, vol. 2011, no.~2, 2011, p.~5.

\bibitem{silberman2012indoor}
N.~Silberman, D.~Hoiem, P.~Kohli, and R.~Fergus, ``Indoor segmentation and
  support inference from {RGBD} images,'' in \emph{ECCV}.\hskip 1em plus 0.5em
  minus 0.4em\relax Springer, 2012, pp. 746--760.

\bibitem{pascal-voc-2012}
M.~Everingham, L.~Van~Gool, C.~K.~I. Williams, J.~Winn, and A.~Zisserman, ``The
  {PASCAL} {V}isual {O}bject {C}lasses {C}hallenge 2012 {(VOC2012)}
  {R}esults,''
  http://www.pascal-network.org/challenges/VOC/voc2012/workshop/index.html.

\bibitem{lin2014microsoft}
T.-Y. Lin, M.~Maire, S.~Belongie, J.~Hays, P.~Perona, D.~Ramanan,
  P.~Doll{\'a}r, and C.~L. Zitnick, ``Microsoft {COCO}: Common objects in
  context,'' in \emph{ECCV}.\hskip 1em plus 0.5em minus 0.4em\relax Springer,
  2014, pp. 740--755.

\bibitem{zhou2014learning}
B.~Zhou, A.~Lapedriza, J.~Xiao, A.~Torralba, and A.~Oliva, ``Learning deep
  features for scene recognition using places database,'' in \emph{Advances in
  neural information processing systems}, 2014, pp. 487--495.

\bibitem{wu20153d}
Z.~Wu, S.~Song, A.~Khosla, F.~Yu, L.~Zhang, X.~Tang, and J.~Xiao, ``{3D}
  shapenets: A deep representation for volumetric shapes,'' in \emph{CVPR},
  2015, pp. 1912--1920.

\bibitem{zhou2017places}
B.~Zhou, A.~Lapedriza, A.~Khosla, A.~Oliva, and A.~Torralba, ``Places: A 10
  million image database for scene recognition,'' \emph{PAMI}, 2017.

\bibitem{Geiger2013IJRR}
A.~Geiger, P.~Lenz, C.~Stiller, and R.~Urtasun, ``Vision meets robotics: The
  {KITTI} dataset,'' \emph{International Journal of Robotics Research (IJRR)},
  2013.

\bibitem{Cordts_2016_CVPR}
M.~Cordts, M.~Omran, S.~Ramos, T.~Rehfeld, M.~Enzweiler, R.~Benenson,
  U.~Franke, S.~Roth, and B.~Schiele, ``The {Cityscapes} dataset for semantic
  urban scene understanding,'' in \emph{CVPR}, June 2016.

\bibitem{bileschi2006streetscenes}
S.~M. Bileschi, ``Streetscenes: Towards scene understanding in still images,''
  MASSACHUSETTS INST OF TECH CAMBRIDGE, Tech. Rep., 2006.

\bibitem{BrostowSFC:ECCV08}
G.~J. Brostow, J.~Shotton, J.~Fauqueur, and R.~Cipolla, ``Segmentation and
  recognition using structure from motion point clouds,'' in \emph{ECCV}, 2008,
  pp. 44--57.

\bibitem{Geiger2011NIPS}
A.~Geiger, C.~Wojek, and R.~Urtasun, ``Joint {3D} estimation of objects and
  scene layout,'' in \emph{NIPS}, 2011.

\bibitem{scharwachter2014stixmantics}
T.~Scharw{\"a}chter, M.~Enzweiler, U.~Franke, and S.~Roth, ``Stixmantics: A
  medium-level model for real-time semantic scene understanding,'' in
  \emph{ECCV}.\hskip 1em plus 0.5em minus 0.4em\relax Springer, 2014, pp.
  533--548.

\bibitem{yang2016exploit}
F.~Yang, W.~Choi, and Y.~Lin, ``Exploit all the layers: Fast and accurate cnn
  object detector with scale dependent pooling and cascaded rejection
  classifiers,'' in \emph{CVPR}, 2016, pp. 2129--2137.

\bibitem{dollarCVPR09peds}
P.~Doll\'ar, C.~Wojek, B.~Schiele, and P.~Perona, ``Pedestrian detection: A
  benchmark,'' in \emph{CVPR}, June 2009.

\bibitem{dalal2005histograms}
N.~Dalal and B.~Triggs, ``Histograms of oriented gradients for human
  detection,'' in \emph{CVPR}, vol.~1.\hskip 1em plus 0.5em minus 0.4em\relax
  IEEE, 2005, pp. 886--893.

\bibitem{enzweiler2009monocular}
M.~Enzweiler and D.~M. Gavrila, ``Monocular pedestrian detection: Survey and
  experiments,'' \emph{PAMI}, vol.~31, no.~12, pp. 2179--2195, 2009.

\bibitem{OrePap97}
M.~Oren, C.~Papageorgiou, P.~Sinha, E.~Osuna, and T.~Poggio, ``Pedestrian
  detection using wavelet templates,'' in \emph{CVPR}, 1997, pp. 193--99.

\bibitem{munder2006experimental}
S.~Munder and D.~M. Gavrila, ``An experimental study on pedestrian
  classification,'' \emph{IEEE transactions on pattern analysis and machine
  intelligence}, vol.~28, no.~11, pp. 1863--1868, 2006.

\bibitem{wang2007object}
L.~Wang, J.~Shi, G.~Song, and I.-F. Shen, ``Object detection combining
  recognition and segmentation,'' in \emph{ACCV}.\hskip 1em plus 0.5em minus
  0.4em\relax Springer, 2007, pp. 189--199.

\bibitem{ess2008mobile}
A.~Ess, B.~Leibe, K.~Schindler, and L.~Van~Gool, ``A mobile vision system for
  robust multi-person tracking,'' in \emph{CVPR}.\hskip 1em plus 0.5em minus
  0.4em\relax IEEE, 2008, pp. 1--8.

\bibitem{enzweiler2010multi}
M.~Enzweiler, A.~Eigenstetter, B.~Schiele, and D.~M. Gavrila, ``Multi-cue
  pedestrian classification with partial occlusion handling,'' in
  \emph{CVPR}.\hskip 1em plus 0.5em minus 0.4em\relax IEEE, 2010, pp. 990--997.

\bibitem{wojek2009multi}
C.~Wojek, S.~Walk, and B.~Schiele, ``Multi-cue onboard pedestrian detection,''
  in \emph{CVPR}.\hskip 1em plus 0.5em minus 0.4em\relax IEEE, 2009, pp.
  794--801.

\bibitem{keller2011new}
C.~G. Keller, M.~Enzweiler, and D.~M. Gavrila, ``A new benchmark for
  stereo-based pedestrian detection,'' in \emph{Intelligent Vehicles Symposium
  (IV)}.\hskip 1em plus 0.5em minus 0.4em\relax IEEE, 2011, pp. 691--696.

\bibitem{fragkiadaki2012two}
K.~Fragkiadaki, W.~Zhang, G.~Zhang, and J.~Shi, ``Two-granularity tracking:
  Mediating trajectory and detection graphs for tracking under occlusions,'' in
  \emph{ECCV}.\hskip 1em plus 0.5em minus 0.4em\relax Springer, 2012, pp.
  552--565.

\bibitem{ouyang2012discriminative}
W.~Ouyang and X.~Wang, ``A discriminative deep model for pedestrian detection
  with occlusion handling,'' in \emph{CVPR}.\hskip 1em plus 0.5em minus
  0.4em\relax IEEE, 2012, pp. 3258--3265.

\bibitem{deng2014pedestrian}
Y.~Deng, P.~Luo, C.~C. Loy, and X.~Tang, ``Pedestrian attribute recognition at
  far distance,'' in \emph{Proceedings of the 22nd ACM international conference
  on Multimedia}.\hskip 1em plus 0.5em minus 0.4em\relax ACM, 2014, pp.
  789--792.

\bibitem{hwang2015multispectral}
S.~Hwang, J.~Park, N.~Kim, Y.~Choi, and I.~So~Kweon, ``Multispectral pedestrian
  detection: Benchmark dataset and baseline,'' in \emph{CVPR}, 2015, pp.
  1037--1045.

\bibitem{zhang2017citypersons}
S.~Zhang, R.~Benenson, and B.~Schiele, ``Citypersons: A diverse dataset for
  pedestrian detection,'' \emph{arXiv preprint arXiv:1702.05693}, 2017.

\bibitem{krause20133d}
J.~Krause, M.~Stark, J.~Deng, and L.~Fei-Fei, ``{3D} object representations for
  fine-grained categorization,'' in \emph{ICCVW}, 2013, pp. 554--561.

\bibitem{MatzenICCV13}
K.~Matzen and N.~Snavely, ``{NYC3DCars}: A dataset of {3D} vehicles in
  geographic context,'' in \emph{ICCV}, 2013.

\bibitem{sivaraman2010general}
S.~Sivaraman and M.~M. Trivedi, ``A general active-learning framework for
  on-road vehicle recognition and tracking,'' \emph{IEEE Transactions on
  Intelligent Transportation Systems}, vol.~11, no.~2, pp. 267--276, 2010.

\bibitem{arrospide2012video}
J.~Arr{\'o}spide, L.~Salgado, and M.~Nieto, ``Video analysis-based vehicle
  detection and tracking using an mcmc sampling framework,'' \emph{EURASIP
  Journal on Advances in Signal Processing}, vol. 2012, no.~1, p.~2, 2012.

\bibitem{TMEMotorwayDataset}
C.~Caraffi, T.~Vojir, J.~Trefny, J.~Sochman, and J.~Matas, ``{A System for
  Real-time Detection and Tracking of Vehicles from a Single Car-mounted
  Camera},'' in \emph{ITSC}, Sep. 2012, pp. 975--982.

\bibitem{OzuysalLF09}
M.~Ozuysal, V.~Lepetit, and P.Fua, ``Pose estimation for category specific
  multiview object localization,'' in \emph{CVPR}, Miami, FL, June 2009.

\bibitem{agarwal2002learning}
S.~Agarwal and D.~Roth, ``Learning a sparse representation for object
  detection,'' in \emph{ECCV}.\hskip 1em plus 0.5em minus 0.4em\relax Springer,
  2002, pp. 113--127.

\bibitem{schneiderman2000statistical}
H.~Schneiderman and T.~Kanade, ``A statistical method for 3d object detection
  applied to faces and cars,'' in \emph{CVPR}, vol.~1.\hskip 1em plus 0.5em
  minus 0.4em\relax IEEE, 2000, pp. 746--751.

\bibitem{PapPog99b}
C.~Papageorgiou and T.~Poggio, ``A trainable object detection system: Car
  detection in static images,'' mitai, Tech. Rep. 1673, Oct. 1999.

\bibitem{Stallkamp-IJCNN-2011}
J.~Stallkamp, M.~Schlipsing, J.~Salmen, and C.~Igel, ``The {G}erman {T}raffic
  {S}ign {R}ecognition {B}enchmark: A multi-class classification competition,''
  in \emph{IEEE International Joint Conference on Neural Networks}, 2011, pp.
  1453--1460.

\bibitem{mathias2013traffic}
M.~Mathias, R.~Timofte, R.~Benenson, and L.~Van~Gool, ``Traffic sign
  recognition-how far are we from the solution?'' in \emph{The International
  Joint Conference on Neural Networks (IJCNN)}.\hskip 1em plus 0.5em minus
  0.4em\relax IEEE, 2013, pp. 1--8.

\bibitem{maldonado2007road}
S.~Maldonado-Bascon, S.~Lafuente-Arroyo, P.~Gil-Jimenez, H.~Gomez-Moreno, and
  F.~L{\'o}pez-Ferreras, ``Road-sign detection and recognition based on support
  vector machines,'' \emph{IEEE Transactions on Intelligent Transportation
  Systems}, vol.~8, no.~2, pp. 264--278, 2007.

\bibitem{vsegvic2010computer}
S.~{\v{S}}egvic, K.~Brki{\'c}, Z.~Kalafati{\'c}, V.~Stanisavljevi{\'c},
  M.~{\v{S}}evrovi{\'c}, D.~Budimir, and I.~Dadi{\'c}, ``A computer vision
  assisted geoinformation inventory for traffic infrastructure,'' in
  \emph{Intelligent Transportation Systems (ITSC)}.\hskip 1em plus 0.5em minus
  0.4em\relax IEEE, 2010, pp. 66--73.

\bibitem{larssonSCIA2011}
F.~Larsson, M.~Felsberg, and P.-E. Forssen, ``{Correlating Fourier descriptors
  of local patches for road sign recognition},'' \emph{IET Computer Vision},
  vol.~5, no.~4, pp. 244--254, 2011.

\bibitem{mogelmose2012vision}
A.~Mogelmose, M.~M. Trivedi, and T.~B. Moeslund, ``Vision-based traffic sign
  detection and analysis for intelligent driver assistance systems:
  Perspectives and survey,'' \emph{IEEE Transactions on Intelligent
  Transportation Systems}, vol.~13, no.~4, pp. 1484--1497, 2012.

\bibitem{Houben-IJCNN-2013}
S.~Houben, J.~Stallkamp, J.~Salmen, M.~Schlipsing, and C.~Igel, ``Detection of
  traffic signs in real-world images: The {G}erman {T}raffic {S}ign {D}etection
  {B}enchmark,'' in \emph{International Joint Conference on Neural Networks},
  no. 1288, 2013.

\bibitem{aly2008real}
M.~Aly, ``Real time detection of lane markers in urban streets,'' in
  \emph{Intelligent Vehicles Symposium (IV)}.\hskip 1em plus 0.5em minus
  0.4em\relax IEEE, 2008, pp. 7--12.

\bibitem{belongie2001shape}
S.~Belongie, J.~Malik, and J.~Puzicha, ``Shape context: A new descriptor for
  shape matching and object recognition,'' in \emph{Advances in neural
  information processing systems}, 2001, pp. 831--837.

\bibitem{chakravarty1982characteristic}
I.~Chakravarty and H.~Freeman, ``Characteristic views as a basis for
  three-dimensional object recognition,'' \emph{representations}, vol.~9,
  p.~10, 1982.

\bibitem{oshima1983object}
M.~Oshima and Y.~Shirai, ``Object recognition using three-dimensional
  information,'' \emph{PAMI}, no.~4, pp. 353--361, 1983.

\bibitem{besl1985three}
P.~J. Besl and R.~C. Jain, ``Three-dimensional object recognition,'' \emph{ACM
  Computing Surveys (CSUR)}, vol.~17, no.~1, pp. 75--145, 1985.

\bibitem{lowe1987three}
D.~G. Lowe, ``Three-dimensional object recognition from single two-dimensional
  images,'' \emph{Artificial intelligence}, vol.~31, no.~3, pp. 355--395, 1987.

\bibitem{lamdan1988object}
Y.~Lamdan, J.~T. Schwartz, and H.~J. Wolfson, ``Object recognition by affine
  invariant matching,'' in \emph{CVPR}.\hskip 1em plus 0.5em minus 0.4em\relax
  IEEE, 1988, pp. 335--344.

\bibitem{wilkes1992active}
D.~Wilkes and J.~K. Tsotsos, ``Active object recognition,'' in
  \emph{CVPR}.\hskip 1em plus 0.5em minus 0.4em\relax IEEE, 1992, pp. 136--141.

\bibitem{dickinson1994active}
S.~J. Dickinson, H.~I. Christensen, J.~Tsotsos, and G.~Olofsson, ``Active
  object recognition integrating attention and viewpoint control,'' in
  \emph{ECCV}.\hskip 1em plus 0.5em minus 0.4em\relax Springer, 1994, pp.
  2--14.

\bibitem{johnson1999using}
A.~E. Johnson and M.~Hebert, ``Using spin images for efficient object
  recognition in cluttered {3D} scenes,'' \emph{PAMI}, vol.~21, no.~5, pp.
  433--449, 1999.

\bibitem{swain1991color}
M.~J. Swain and D.~H. Ballard, ``Color indexing,'' \emph{IJCV}, vol.~7, no.~1,
  pp. 11--32, 1991.

\bibitem{gevers1999color}
T.~Gevers and A.~W. Smeulders, ``Color-based object recognition,''
  \emph{Pattern recognition}, vol.~32, no.~3, pp. 453--464, 1999.

\bibitem{pontil1998support}
M.~Pontil and A.~Verri, ``Support vector machines for {3D} object
  recognition,'' \emph{PAMI}, vol.~20, no.~6, pp. 637--646, 1998.

\bibitem{shotton2008semantic}
J.~Shotton, M.~Johnson, and R.~Cipolla, ``Semantic texton forests for image
  categorization and segmentation,'' in \emph{CVPR}.\hskip 1em plus 0.5em minus
  0.4em\relax IEEE, 2008, pp. 1--8.

\bibitem{torralba2004sharing}
A.~Torralba, K.~P. Murphy, and W.~T. Freeman, ``Sharing features: efficient
  boosting procedures for multiclass object detection,'' in \emph{CVPR},
  vol.~2.\hskip 1em plus 0.5em minus 0.4em\relax IEEE, 2004, pp. II--II.

\bibitem{lowe1999object}
D.~G. Lowe, ``Object recognition from local scale-invariant features,'' in
  \emph{ICCV}, vol.~2.\hskip 1em plus 0.5em minus 0.4em\relax Ieee, 1999, pp.
  1150--1157.

\bibitem{lowe2004distinctive}
D.~Lowe, ``Distinctive image features from scale-invariant keypoints,''
  \emph{IJCV}, vol.~60, no.~2, pp. 91--110, 2004.

\bibitem{bilen2014object}
H.~Bilen, M.~Pedersoli, V.~P. Namboodiri, T.~Tuytelaars, and L.~Van~Gool,
  ``Object classification with adaptable regions,'' in \emph{CVPR}, 2014, pp.
  3662--3669.

\bibitem{viola2001rapid}
P.~Viola and M.~Jones, ``Rapid object detection using a boosted cascade of
  simple features,'' in \emph{CVPR}, vol.~1.\hskip 1em plus 0.5em minus
  0.4em\relax IEEE, 2001, pp. I--I.

\bibitem{lienhart2002extended}
R.~Lienhart and J.~Maydt, ``An extended set of {H}aar-like features for rapid
  object detection,'' in \emph{International Conference on Image Processing},
  vol.~1.\hskip 1em plus 0.5em minus 0.4em\relax IEEE, 2002, pp. I--I.

\bibitem{torralba2003context}
A.~Torralba, K.~P. Murphy, W.~T. Freeman, and M.~A. Rubin, ``Context-based
  vision system for place and object recognition,'' in \emph{ICCV}.\hskip 1em
  plus 0.5em minus 0.4em\relax IEEE, 2003, p. 273.

\bibitem{felzenszwalb2010object}
P.~F. Felzenszwalb, R.~B. Girshick, D.~McAllester, and D.~Ramanan, ``Object
  detection with discriminatively trained part-based models,'' \emph{PAMI},
  vol.~32, no.~9, pp. 1627--1645, 2010.

\bibitem{yao2012describing}
J.~Yao, S.~Fidler, and R.~Urtasun, ``Describing the scene as a whole: Joint
  object detection, scene classification and semantic segmentation,'' in
  \emph{CVPR}.\hskip 1em plus 0.5em minus 0.4em\relax IEEE, 2012, pp. 702--709.

\bibitem{bo2014learning}
L.~Bo, X.~Ren, and D.~Fox, ``Learning hierarchical sparse features for
  {RGB-(D)} object recognition,'' \emph{The International Journal of Robotics
  Research}, vol.~33, no.~4, pp. 581--599, 2014.

\bibitem{gupta2014learning}
S.~Gupta, R.~Girshick, P.~Arbel{\'a}ez, and J.~Malik, ``Learning rich features
  from {RGB-D} images for object detection and segmentation,'' in
  \emph{ECCV}.\hskip 1em plus 0.5em minus 0.4em\relax Springer, 2014, pp.
  345--360.

\bibitem{uijlings2013selective}
J.~R. Uijlings, K.~E. Van De~Sande, T.~Gevers, and A.~W. Smeulders, ``Selective
  search for object recognition,'' \emph{IJCV}, vol. 104, no.~2, pp. 154--171,
  2013.

\bibitem{deselaers2012weakly}
T.~Deselaers, B.~Alexe, and V.~Ferrari, ``Weakly supervised localization and
  learning with generic knowledge,'' \emph{IJCV}, vol. 100, no.~3, pp.
  275--293, 2012.

\bibitem{chen20153d}
X.~Chen, K.~Kundu, Y.~Zhu, A.~G. Berneshawi, H.~Ma, S.~Fidler, and R.~Urtasun,
  ``{3D} object proposals for accurate object class detection,'' in
  \emph{Advances in Neural Information Processing Systems}, 2015, pp. 424--432.

\bibitem{he2015delving}
K.~He, X.~Zhang, S.~Ren, and J.~Sun, ``Delving deep into rectifiers: Surpassing
  human-level performance on {ImageNet} classification,'' in \emph{ICCV}, 2015,
  pp. 1026--1034.

\bibitem{krizhevsky2012imagenet}
A.~Krizhevsky, I.~Sutskever, and G.~E. Hinton, ``{ImageNet} classification with
  deep convolutional neural networks,'' in \emph{Advances in neural information
  processing systems}, 2012, pp. 1097--1105.

\bibitem{he2016deep}
K.~He, X.~Zhang, S.~Ren, and J.~Sun, ``Deep residual learning for image
  recognition,'' in \emph{CVPR}, 2016, pp. 770--778.

\bibitem{maturana2015voxnet}
D.~Maturana and S.~Scherer, ``Voxnet: A {3D} convolutional neural network for
  real-time object recognition,'' in \emph{IROS}.\hskip 1em plus 0.5em minus
  0.4em\relax IEEE, 2015, pp. 922--928.

\bibitem{wohlhart2015learning}
P.~Wohlhart and V.~Lepetit, ``Learning descriptors for object recognition and
  {3D} pose estimation,'' in \emph{CVPR}, 2015, pp. 3109--3118.

\bibitem{simonyan2014very}
K.~Simonyan and A.~Zisserman, ``Very deep convolutional networks for
  large-scale image recognition,'' \emph{arXiv preprint arXiv:1409.1556}, 2014.

\bibitem{szegedy2015going}
C.~Szegedy, W.~Liu, Y.~Jia, P.~Sermanet, S.~Reed, D.~Anguelov, D.~Erhan,
  V.~Vanhoucke, and A.~Rabinovich, ``Going deeper with convolutions,'' in
  \emph{CVPR}, 2015, pp. 1--9.

\bibitem{shankar2015deep}
S.~Shankar, V.~K. Garg, and R.~Cipolla, ``Deep-carving: Discovering visual
  attributes by carving deep neural nets,'' in \emph{CVPR}, 2015, pp.
  3403--3412.

\bibitem{oquab2014weakly}
M.~Oquab, L.~Bottou, I.~Laptev, J.~Sivic \emph{et~al.}, ``Weakly supervised
  object recognition with convolutional neural networks,'' in \emph{NIPS},
  2014.

\bibitem{erhan2014scalable}
D.~Erhan, C.~Szegedy, A.~Toshev, and D.~Anguelov, ``Scalable object detection
  using deep neural networks,'' in \emph{CVPR}, 2014, pp. 2147--2154.

\bibitem{girshick2014rich}
R.~Girshick, J.~Donahue, T.~Darrell, and J.~Malik, ``Rich feature hierarchies
  for accurate object detection and semantic segmentation,'' in \emph{CVPR},
  2014, pp. 580--587.

\bibitem{wei2014cnn}
Y.~Wei, W.~Xia, J.~Huang, B.~Ni, J.~Dong, Y.~Zhao, and S.~Yan, ``{CNN}:
  Single-label to multi-label,'' \emph{arXiv preprint arXiv:1406.5726}, 2014.

\bibitem{girshick2015fast}
R.~Girshick, ``Fast {R-CNN},'' in \emph{ICCV}, 2015, pp. 1440--1448.

\bibitem{caicedo2015active}
J.~C. Caicedo and S.~Lazebnik, ``Active object localization with deep
  reinforcement learning,'' in \emph{ICCV}, 2015, pp. 2488--2496.

\bibitem{redmon2016you}
J.~Redmon, S.~Divvala, R.~Girshick, and A.~Farhadi, ``You only look once:
  Unified, real-time object detection,'' in \emph{CVPR}, 2016, pp. 779--788.

\bibitem{ren2015faster}
S.~Ren, K.~He, R.~Girshick, and J.~Sun, ``Faster {R-CNN}: Towards real-time
  object detection with region proposal networks,'' in \emph{Advances in neural
  information processing systems}, 2015, pp. 91--99.

\bibitem{he2014spatial}
K.~He, X.~Zhang, S.~Ren, and J.~Sun, ``Spatial pyramid pooling in deep
  convolutional networks for visual recognition,'' in \emph{ECCV}.\hskip 1em
  plus 0.5em minus 0.4em\relax Springer, 2014, pp. 346--361.

\bibitem{kantorov2016contextlocnet}
V.~Kantorov, M.~Oquab, M.~Cho, and I.~Laptev, ``{ContextLocNet}: Context-aware
  deep network models for weakly supervised localization,'' in
  \emph{ECCV}.\hskip 1em plus 0.5em minus 0.4em\relax Springer, 2016, pp.
  350--365.

\bibitem{hu2016learning}
H.~Hu, G.-T. Zhou, Z.~Deng, Z.~Liao, and G.~Mori, ``Learning structured
  inference neural networks with label relations,'' in \emph{CVPR}, 2016, pp.
  2960--2968.

\bibitem{chen2015contextualizing}
Q.~Chen, Z.~Song, J.~Dong, Z.~Huang, Y.~Hua, and S.~Yan, ``Contextualizing
  object detection and classification,'' \emph{PAMI}, vol.~37, no.~1, pp.
  13--27, 2015.

\bibitem{liang2015recurrent}
M.~Liang and X.~Hu, ``Recurrent convolutional neural network for object
  recognition,'' in \emph{CVPR}, 2015, pp. 3367--3375.

\bibitem{gavrila1999real}
D.~M. Gavrila and V.~Philomin, ``Real-time object detection for" smart"
  vehicles,'' in \emph{ICCV}, vol.~1.\hskip 1em plus 0.5em minus 0.4em\relax
  IEEE, 1999, pp. 87--93.

\bibitem{papageorgiou2000trainable}
C.~Papageorgiou and T.~Poggio, ``A trainable system for object detection,''
  \emph{International Journal of Computer Vision}, vol.~38, no.~1, pp. 15--33,
  2000.

\bibitem{ess2009segmentation}
A.~Ess, T.~M{\"u}ller, H.~Grabner, and L.~J. Van~Gool, ``Segmentation-based
  urban traffic scene understanding.'' in \emph{BMVC}, vol.~1, 2009, p.~2.

\bibitem{wojek2013monocular}
C.~Wojek, S.~Walk, S.~Roth, K.~Schindler, and B.~Schiele, ``Monocular visual
  scene understanding: Understanding multi-object traffic scenes,''
  \emph{PAMI}, vol.~35, no.~4, pp. 882--897, 2013.

\bibitem{chen2016monocular}
X.~Chen, K.~Kundu, Z.~Zhang, H.~Ma, S.~Fidler, and R.~Urtasun, ``Monocular 3d
  object detection for autonomous driving,'' in \emph{CVPR}, 2016, pp.
  2147--2156.

\bibitem{fuerstenberg2002pedestrian}
K.~C. Fuerstenberg, K.~C. Dietmayer, and V.~Willhoeft, ``Pedestrian recognition
  in urban traffic using a vehicle based multilayer laserscanner,'' in
  \emph{Intelligent Vehicle Symposium (IV)}, vol.~1.\hskip 1em plus 0.5em minus
  0.4em\relax IEEE, 2002, pp. 31--35.

\bibitem{kidono2011pedestrian}
K.~Kidono, T.~Miyasaka, A.~Watanabe, T.~Naito, and J.~Miura, ``Pedestrian
  recognition using high-definition {LIDAR},'' in \emph{Intelligent Vehicles
  Symposium (IV)}.\hskip 1em plus 0.5em minus 0.4em\relax IEEE, 2011, pp.
  405--410.

\bibitem{broggi2009new}
A.~Broggi, P.~Cerri, S.~Ghidoni, P.~Grisleri, and H.~G. Jung, ``A new approach
  to urban pedestrian detection for automatic braking,'' \emph{IEEE
  Transactions on Intelligent Transportation Systems}, vol.~10, no.~4, pp.
  594--605, 2009.

\bibitem{beckwith1998passive}
D.~Beckwith and K.~Hunter-Zaworski, ``Passive pedestrian detection at
  unsignalized crossings,'' \emph{Transportation Research Record: Journal of
  the Transportation Research Board}, no. 1636, pp. 96--103, 1998.

\bibitem{nanda2002probabilistic}
H.~Nanda and L.~Davis, ``Probabilistic template based pedestrian detection in
  infrared videos,'' in \emph{Intelligent Vehicle Symposium (IV)},
  vol.~1.\hskip 1em plus 0.5em minus 0.4em\relax IEEE, 2002, pp. 15--20.

\bibitem{bertozzi2003shape}
M.~Bertozzi, A.~Broggi, R.~Chapuis, F.~Chausse, A.~Fascioli, and A.~Tibaldi,
  ``Shape-based pedestrian detection and localization,'' in \emph{Intelligent
  Transportation Systems (ITSC)}, vol.~1.\hskip 1em plus 0.5em minus
  0.4em\relax IEEE, 2003, pp. 328--333.

\bibitem{bertozzi2005stereo}
M.~Bertozzi, E.~Binelli, A.~Broggi, and M.~Rose, ``Stereo vision-based
  approaches for pedestrian detection,'' in \emph{CVPR}.\hskip 1em plus 0.5em
  minus 0.4em\relax IEEE, 2005, pp. 16--16.

\bibitem{richards1995vision}
C.~A. Richards, C.~E. Smith, and N.~P. Papanikolopoulos, ``Vision-based
  intelligent control of transportation systems,'' in \emph{the IEEE
  International Symposium on Intelligent Control}.\hskip 1em plus 0.5em minus
  0.4em\relax IEEE, 1995, pp. 519--524.

\bibitem{rohr1993incremental}
K.~Rohr, ``Incremental recognition of pedestrians from image sequences,'' in
  \emph{CVPR}.\hskip 1em plus 0.5em minus 0.4em\relax IEEE, 1993, pp. 8--13.

\bibitem{wohler1998time}
C.~W{\"o}hler, J.~K. Anlauf, T.~P{\"o}rtner, and U.~Franke, ``A time delay
  neural network algorithm for real-time pedestrian recognition,'' in
  \emph{International Conference on Intelligent Vehicle}.\hskip 1em plus 0.5em
  minus 0.4em\relax Citeseer, 1998.

\bibitem{broggi2000shape}
A.~Broggi, M.~Bertozzi, A.~Fascioli, and M.~Sechi, ``Shape-based pedestrian
  detection,'' in \emph{Intelligent Vehicles Symposium (IV)}.\hskip 1em plus
  0.5em minus 0.4em\relax IEEE, 2000, pp. 215--220.

\bibitem{oren1997trainable}
M.~Oren, C.~Papageorgiou, P.~Shinha, E.~Osuna, and T.~Poggio, ``A trainable
  system for people detection,'' in \emph{Proceedings of Image Understanding
  Workshop}, vol.~24, 1997.

\bibitem{dollar2009integral}
P.~Doll{\'a}r, Z.~Tu, P.~Perona, and S.~Belongie, ``Integral channel
  features,'' in \emph{BMVC}.\hskip 1em plus 0.5em minus 0.4em\relax British
  Machine Vision Association, 2009.

\bibitem{dalal2006human}
N.~Dalal, B.~Triggs, and C.~Schmid, ``Human detection using oriented histograms
  of flow and appearance,'' in \emph{ECCV}.\hskip 1em plus 0.5em minus
  0.4em\relax Springer, 2006, pp. 428--441.

\bibitem{yan2013robust}
J.~Yan, X.~Zhang, Z.~Lei, S.~Liao, and S.~Z. Li, ``Robust multi-resolution
  pedestrian detection in traffic scenes,'' in \emph{CVPR}, 2013, pp.
  3033--3040.

\bibitem{tian2015deep}
Y.~Tian, P.~Luo, X.~Wang, and X.~Tang, ``Deep learning strong parts for
  pedestrian detection,'' in \emph{ICCV}, 2015, pp. 1904--1912.

\bibitem{tian2015pedestrian}
Y.~Tian, P.~Luo, X.~g. Wang, and X.~Tang, ``Pedestrian detection aided by deep
  learning semantic tasks,'' in \emph{CVPR}, 2015, pp. 5079--5087.

\bibitem{schlosser2016fusing}
J.~Schlosser, C.~K. Chow, and Z.~Kira, ``Fusing {LIDAR} and images for
  pedestrian detection using convolutional neural networks,'' in
  \emph{ICRA}.\hskip 1em plus 0.5em minus 0.4em\relax IEEE, 2016, pp.
  2198--2205.

\bibitem{hu2017pushing}
Q.~Hu, P.~Wang, C.~Shen, A.~van~den Hengel, and F.~Porikli, ``Pushing the
  limits of deep cnns for pedestrian detection,'' \emph{IEEE Transactions on
  Circuits and Systems for Video Technology}, 2017.

\bibitem{brazil2017illuminating}
G.~Brazil, X.~Yin, and X.~Liu, ``Illuminating pedestrians via simultaneous
  detection \& segmentation,'' in \emph{ICCV}, 2017, pp. 4950--4959.

\bibitem{dollar2009pedestrian}
P.~Doll{\'a}r, C.~Wojek, B.~Schiele, and P.~Perona, ``Pedestrian detection: A
  benchmark,'' in \emph{CVPR}.\hskip 1em plus 0.5em minus 0.4em\relax IEEE,
  2009, pp. 304--311.

\bibitem{dollar2012pedestrian}
P.~Dollar, C.~Wojek, B.~Schiele, and P.~Perona, ``Pedestrian detection: An
  evaluation of the state of the art,'' \emph{PAMI}, vol.~34, no.~4, pp.
  743--761, 2012.

\bibitem{benenson2014ten}
R.~Benenson, M.~Omran, J.~Hosang, and B.~Schiele, ``Ten years of pedestrian
  detection, what have we learned?'' \emph{arXiv preprint arXiv:1411.4304},
  2014.

\bibitem{geronimo2010survey}
D.~Geronimo, A.~M. Lopez, A.~D. Sappa, and T.~Graf, ``Survey of pedestrian
  detection for advanced driver assistance systems,'' \emph{PAMI}, vol.~32,
  no.~7, pp. 1239--1258, 2010.

\bibitem{yang2015convolutional}
B.~Yang, J.~Yan, Z.~Lei, and S.~Z. Li, ``Convolutional channel features,'' in
  \emph{ICCV}, 2015, pp. 82--90.

\bibitem{zhang2016faster}
L.~Zhang, L.~Lin, X.~Liang, and K.~He, ``Is faster r-cnn doing well for
  pedestrian detection?'' in \emph{ECCV}.\hskip 1em plus 0.5em minus
  0.4em\relax Springer, 2016, pp. 443--457.

\bibitem{radford1989vehicle}
C.~Radford and D.~Houghton, ``Vehicle detection in open-world scenes using a
  {Hough} transform technique,'' in \emph{Third International Conference on
  Image Processing and its Applications}.\hskip 1em plus 0.5em minus
  0.4em\relax IET, 1989, pp. 78--82.

\bibitem{betke1996multiple}
M.~Betke, E.~Haritaoglu, and L.~S. Davis, ``Multiple vehicle detection and
  tracking in hard real-time,'' in \emph{Intelligent Vehicles Symposium
  (IV)}.\hskip 1em plus 0.5em minus 0.4em\relax IEEE, 1996, pp. 351--356.

\bibitem{bertozzi1997real}
M.~Bertozzi, A.~Broggi, and S.~Castelluccio, ``A real-time oriented system for
  vehicle detection,'' \emph{Journal of Systems Architecture}, vol.~43, no.
  1-5, pp. 317--325, 1997.

\bibitem{bertozzi2000stereo}
M.~Bertozzi, A.~Broggi, A.~Fascioli, and S.~Nichele, ``Stereo vision-based
  vehicle detection,'' in \emph{Intelligent Vehicles Symposium (IV)}.\hskip 1em
  plus 0.5em minus 0.4em\relax IEEE, 2000, pp. 39--44.

\bibitem{alessandretti2007vehicle}
G.~Alessandretti, A.~Broggi, and P.~Cerri, ``Vehicle and guard rail detection
  using {Radar} and vision data fusion,'' \emph{IEEE Transactions on
  Intelligent Transportation Systems}, vol.~8, no.~1, pp. 95--105, 2007.

\bibitem{bullock1993neural}
D.~Bullock, J.~Garrett, and C.~Hendrickson, ``A neural network for image-based
  vehicle detection,'' \emph{Transportation Research Part C: Emerging
  Technologies}, vol.~1, no.~3, pp. 235--247, 1993.

\bibitem{gupte2002detection}
S.~Gupte, O.~Masoud, R.~F. Martin, and N.~P. Papanikolopoulos, ``Detection and
  classification of vehicles,'' \emph{IEEE Transactions on intelligent
  transportation systems}, vol.~3, no.~1, pp. 37--47, 2002.

\bibitem{papageorgiou1999trainable}
C.~P. Papageorgiou and T.~Poggio, ``A trainable object detection system: Car
  detection in static images,'' 1999.

\bibitem{sun2005road}
Z.~Sun, G.~Bebis, and R.~Miller, ``On-road vehicle detection using evolutionary
  {Gabor} filter optimization,'' \emph{IEEE Transactions on Intelligent
  Transportation Systems}, vol.~6, no.~2, pp. 125--137, 2005.

\bibitem{niknejad2011vehicle}
H.~T. Niknejad, K.~Takahashi, S.~Mita, and D.~McAllester, ``Vehicle detection
  and tracking at nighttime for urban autonomous driving,'' in
  \emph{IROS}.\hskip 1em plus 0.5em minus 0.4em\relax IEEE, 2011, pp.
  4442--4447.

\bibitem{sivaraman2013looking}
S.~Sivaraman and M.~M. Trivedi, ``Looking at vehicles on the road: A survey of
  vision-based vehicle detection, tracking, and behavior analysis,'' \emph{IEEE
  Transactions on Intelligent Transportation Systems}, vol.~14, no.~4, pp.
  1773--1795, 2013.

\bibitem{lopez2008nighttime}
A.~L{\'o}pez, J.~Hilgenstock, A.~Busse, R.~Baldrich, F.~Lumbreras, and
  J.~Serrat, ``Nighttime vehicle detection for intelligent headlight control,''
  in \emph{Advanced Concepts for Intelligent Vision Systems}.\hskip 1em plus
  0.5em minus 0.4em\relax Springer, 2008, pp. 113--124.

\bibitem{cao2016robust}
L.~Cao, Q.~Jiang, M.~Cheng, and C.~Wang, ``Robust vehicle detection by
  combining deep features with exemplar classification,''
  \emph{Neurocomputing}, vol. 215, pp. 225--231, 2016.

\bibitem{chabot2017deep}
F.~Chabot, M.~Chaouch, J.~Rabarisoa, C.~Teuli{\`e}re, and T.~Chateau, ``Deep
  manta: A coarse-to-fine many-task network for joint {2D and 3D} vehicle
  analysis from monocular image,'' in \emph{CVPR}, 2017.

\bibitem{lange2016online}
S.~Lange, F.~Ulbrich, and D.~Goehring, ``Online vehicle detection using deep
  neural networks and lidar based preselected image patches,'' in
  \emph{Intelligent Vehicles Symposium (IV)}.\hskip 1em plus 0.5em minus
  0.4em\relax IEEE, 2016, pp. 954--959.

\bibitem{chen2017multi}
X.~Chen, H.~Ma, J.~Wan, B.~Li, and T.~Xia, ``Multi-view {3D} object detection
  network for autonomous driving,'' in \emph{CVPR}, 2017.

\bibitem{sun2006road}
Z.~Sun, G.~Bebis, and R.~Miller, ``On-road vehicle detection: A review,''
  \emph{IEEE transactions on pattern analysis and machine intelligence},
  vol.~28, no.~5, pp. 694--711, 2006.

\bibitem{sivaraman2013vehicle}
S.~Sivaraman and M.~M. Trivedi, ``Vehicle detection by independent parts for
  urban driver assistance,'' \emph{IEEE Transactions on Intelligent
  Transportation Systems}, vol.~14, no.~4, pp. 1597--1608, 2013.

\bibitem{priese1993traffic}
L.~Priese, V.~Rehrmann, R.~Schian, and R.~Lakmann, ``Traffic sign recognition
  based on color image evaluation,'' in \emph{Proceedings IEEE Intelligent
  Vehicles Symposium’93}.\hskip 1em plus 0.5em minus 0.4em\relax Citeseer,
  1993.

\bibitem{kehtarnavaz1995traffic}
N.~Kehtarnavaz and A.~Ahmad, ``Traffic sign recognition in noisy outdoor
  scenes,'' in \emph{Intelligent Vehicles Symposium (IV)}.\hskip 1em plus 0.5em
  minus 0.4em\relax IEEE, 1995, pp. 460--465.

\bibitem{de1997road}
A.~De~La~Escalera, L.~E. Moreno, M.~A. Salichs, and J.~M. Armingol, ``Road
  traffic sign detection and classification,'' \emph{IEEE transactions on
  industrial electronics}, vol.~44, no.~6, pp. 848--859, 1997.

\bibitem{arlicot2009circular}
A.~Arlicot, B.~Soheilian, and N.~Paparoditis, ``Circular road sign extraction
  from street level images using colour, shape and texture databases maps.'' in
  \emph{Workshop Laserscanning}, 2009, pp. 205--210.

\bibitem{seo2012recognizing}
Y.-W. Seo, D.~Wettergreen, and W.~Zhang, ``Recognizing temporary changes on
  highways for reliable autonomous driving,'' in \emph{International Conference
  on Systems, Man, and Cybernetics (SMC)}.\hskip 1em plus 0.5em minus
  0.4em\relax IEEE, 2012, pp. 3027--3032.

\bibitem{balali2015multi}
V.~Balali, E.~Depwe, and M.~Golparvar-Fard, ``Multi-class traffic sign
  detection and classification using google street view images,'' in
  \emph{Transportation Research Board 94th Annual Meeting}, 2015.

\bibitem{de2003traffic}
A.~De~la Escalera, J.~M. Armingol, and M.~Mata, ``Traffic sign recognition and
  analysis for intelligent vehicles,'' \emph{Image and vision computing},
  vol.~21, no.~3, pp. 247--258, 2003.

\bibitem{paclik2000road}
P.~Pacl{\i}k and J.~Novovicova, ``Road sign classification without color
  information,'' in \emph{the 6th Annual Conference of the Advanced School for
  Computing and Imaging}, 2000.

\bibitem{zhu2016traffic}
Y.~Zhu, C.~Zhang, D.~Zhou, X.~Wang, X.~Bai, and W.~Liu, ``Traffic sign
  detection and recognition using fully convolutional network guided
  proposals,'' \emph{Neurocomputing}, vol. 214, pp. 758--766, 2016.

\bibitem{crisman1991unscarf}
J.~D. Crisman and C.~E. Thorpe, ``{UNSCARF-A} color vision system for the
  detection of unstructured roads,'' in \emph{ICRA}.\hskip 1em plus 0.5em minus
  0.4em\relax IEEE, 1991, pp. 2496--2501.

\bibitem{broggi1995vision}
A.~Broggi and S.~Berte, ``Vision-based road detection in automotive systems: A
  real-time expectation-driven approach,'' \emph{Journal of Artificial
  Intelligence Research}, vol.~3, pp. 325--348, 1995.

\bibitem{he2004color}
Y.~He, H.~Wang, and B.~Zhang, ``Color-based road detection in urban traffic
  scenes,'' \emph{IEEE Transactions on Intelligent Transportation Systems},
  vol.~5, no.~4, pp. 309--318, 2004.

\bibitem{dahlkamp2006self}
H.~Dahlkamp, A.~Kaehler, D.~Stavens, S.~Thrun, and G.~R. Bradski,
  ``Self-supervised monocular road detection in desert terrain.'' in
  \emph{Robotics: science and systems}, vol.~38.\hskip 1em plus 0.5em minus
  0.4em\relax Philadelphia, 2006.

\bibitem{mohan2014deep}
R.~Mohan, ``Deep deconvolutional networks for scene parsing,'' \emph{arXiv
  preprint arXiv:1411.4101}, 2014.

\bibitem{laddha2016map}
A.~Laddha, M.~K. Kocamaz, L.~E. Navarro-Serment, and M.~Hebert,
  ``Map-supervised road detection,'' in \emph{Intelligent Vehicles Symposium
  (IV), 2016 IEEE}.\hskip 1em plus 0.5em minus 0.4em\relax IEEE, 2016, pp.
  118--123.

\bibitem{mendes2016exploiting}
C.~C.~T. Mendes, V.~Fr{\'e}mont, and D.~F. Wolf, ``Exploiting fully
  convolutional neural networks for fast road detection,'' in
  \emph{ICRA}.\hskip 1em plus 0.5em minus 0.4em\relax IEEE, 2016, pp.
  3174--3179.

\bibitem{oliveira2016efficient}
G.~L. Oliveira, W.~Burgard, and T.~Brox, ``Efficient deep methods for monocular
  road segmentation,'' in \emph{IROS}, 2016.

\bibitem{hillel2014recent}
A.~B. Hillel, R.~Lerner, D.~Levi, and G.~Raz, ``Recent progress in road and
  lane detection: a survey,'' \emph{Machine vision and applications}, vol.~25,
  no.~3, pp. 727--745, 2014.

\bibitem{Johnson10}
S.~Johnson and M.~Everingham, ``Clustered pose and nonlinear appearance models
  for human pose estimation,'' in \emph{BMVC}, 2010.

\bibitem{kazemi2013multi}
V.~Kazemi, M.~Burenius, H.~Azizpour, and J.~Sullivan, ``Multi-view body part
  recognition with random forests,'' in \emph{BMVC}.\hskip 1em plus 0.5em minus
  0.4em\relax British Machine Vision Association, 2013.

\bibitem{andriluka14cvpr}
M.~Andriluka, L.~Pishchulin, P.~Gehler, and B.~Schiele, ``2d human pose
  estimation: New benchmark and state of the art analysis,'' in \emph{CVPR},
  2014, pp. 3686--3693.

\bibitem{zhang2014facial}
Z.~Zhang, P.~Luo, C.~C. Loy, and X.~Tang, ``Facial landmark detection by deep
  multi-task learning,'' in \emph{ECCV}.\hskip 1em plus 0.5em minus 0.4em\relax
  Springer, 2014, pp. 94--108.

\bibitem{Ferrari08}
V.~Ferrari, M.~Marin-Jimenez, and A.~Zisserman, ``Progressive search space
  reduction for human pose estimation,'' in \emph{CVPR}, jun 2008.

\bibitem{Jammalamadaka12a}
N.~Jammalamadaka, A.~Zisserman, M.~Eichner, V.~Ferrari, and C.~V. Jawahar,
  ``Has my algorithm succeeded? an evaluator for human pose estimators,'' in
  \emph{ECCV}, 2012.

\bibitem{Charles13}
J.~Charles, T.~Pfister, D.~Magee, D.~Hogg, and A.~Zisserman, ``Domain
  adaptation for upper body pose tracking in signed {TV} broadcasts,'' in
  \emph{BMVC}, 2013.

\bibitem{gasparrini2014depth}
S.~Gasparrini, E.~Cippitelli, S.~Spinsante, and E.~Gambi, ``A depth-based fall
  detection system using a kinect sensor,'' \emph{Sensors}, vol.~14, no.~2, pp.
  2756--2775, 2014.

\bibitem{modec13}
B.~Sapp and B.~Taskar, ``Modec: Multimodal decomposable models for human pose
  estimation,'' in \emph{CVPR}, 2013.

\bibitem{Shafaei16}
A.~Shafaei and J.~J. Little, ``Real-time human motion capture with multiple
  depth cameras,'' in \emph{CRV}.\hskip 1em plus 0.5em minus 0.4em\relax
  Canadian Image Processing and Pattern Recognition Society (CIPPRS), 2016.

\bibitem{ferrari20092d}
V.~Ferrari, M.~Mar{\'\i}n-Jim{\'e}nez, and A.~Zisserman, ``{2D} human pose
  estimation in {TV} shows,'' \emph{Statistical and Geometrical Approaches to
  Visual Motion Analysis}, pp. 128--147, 2009.

\bibitem{sapp2011parsing}
B.~Sapp, D.~Weiss, and B.~Taskar, ``Parsing human motion with stretchable
  models,'' in \emph{CVPR}.\hskip 1em plus 0.5em minus 0.4em\relax IEEE, 2011,
  pp. 1281--1288.

\bibitem{sung2012unstructured}
J.~Sung, C.~Ponce, B.~Selman, and A.~Saxena, ``Unstructured human activity
  detection from {RGBD} images,'' in \emph{ICRA}.\hskip 1em plus 0.5em minus
  0.4em\relax IEEE, 2012, pp. 842--849.

\bibitem{kazemi2012using}
V.~Kazemi and J.~Sullivan, ``Using richer models for articulated pose
  estimation of footballers,'' in \emph{BMVC}, 2012.

\bibitem{koppula2013learning}
H.~S. Koppula, R.~Gupta, and A.~Saxena, ``Learning human activities and object
  affordances from {RGB-D} videos,'' \emph{The International Journal of
  Robotics Research}, vol.~32, no.~8, pp. 951--970, 2013.

\bibitem{dantone2013human}
M.~Dantone, J.~Gall, C.~Leistner, and L.~Van~Gool, ``Human pose estimation
  using body parts dependent joint regressors,'' in \emph{CVPR}, 2013, pp.
  3041--3048.

\bibitem{yu2013unconstrained}
T.-H. Yu, T.-K. Kim, and R.~Cipolla, ``Unconstrained monocular {3D} human pose
  estimation by action detection and cross-modality regression forest,'' in
  \emph{CVPR}, 2013, pp. 3642--3649.

\bibitem{escalera2013multi}
S.~Escalera, J.~Gonz{\`a}lez, X.~Bar{\'o}, M.~Reyes, O.~Lopes, I.~Guyon,
  V.~Athitsos, and H.~Escalante, ``Multi-modal {G}esture {R}ecognition
  {C}hallenge 2013: Dataset and results,'' in \emph{ICMI}.\hskip 1em plus 0.5em
  minus 0.4em\relax ACM, 2013, pp. 445--452.

\bibitem{zhang2013actemes}
W.~Zhang, M.~Zhu, and K.~G. Derpanis, ``From actemes to action: A
  strongly-supervised representation for detailed action understanding,'' in
  \emph{CVPR}, 2013, pp. 2248--2255.

\bibitem{h36m_pami}
C.~Ionescu, D.~Papava, V.~Olaru, and C.~Sminchisescu, ``Human3.6{M}: Large
  scale datasets and predictive methods for {3D} human sensing in natural
  environments,'' \emph{PAMI}, vol.~36, no.~7, pp. 1325--1339, jul 2014.

\bibitem{Antol2014}
S.~Antol, C.~L. Zitnick, and D.~Parikh, ``Zero-shot learning via visual
  abstraction,'' in \emph{ECCV}, 2014.

\bibitem{belagiannis20143d}
V.~Belagiannis, S.~Amin, M.~Andriluka, B.~Schiele, N.~Navab, and S.~Ilic,
  ``{3D} pictorial structures for multiple human pose estimation,'' in
  \emph{CVPR}, 2014, pp. 1669--1676.

\bibitem{Cherian14}
A.~Cherian, J.~Mairal, K.~Alahari, and C.~Schmid, ``Mixing body-part sequences
  for human pose estimation,'' in \emph{CVPR}, 2014.

\bibitem{cippitelli2015time}
E.~Cippitelli, S.~Gasparrini, E.~Gambi, S.~Spinsante, J.~W{\aa}hsl{\'e}ny,
  I.~Orhany, and T.~Lindhy, ``Time synchronization and data fusion for
  {RGB}-depth cameras and inertial sensors in {AAL} applications,'' in
  \emph{IEEE International Conference on Communication Workshop (ICCW)}.\hskip
  1em plus 0.5em minus 0.4em\relax IEEE, 2015, pp. 265--270.

\bibitem{WetzlerBMVC15}
A.~Wetzler, R.~Slossberg, and R.~Kimmel, ``Rule of thumb: Deep derotation for
  improved fingertip detection,'' in \emph{BMVC}.\hskip 1em plus 0.5em minus
  0.4em\relax BMVA Press, September 2015, pp. 33.1--33.12.

\bibitem{lopezquintero2015mvap}
M.~I. L{\'o}pez-Quintero, M.~J. Mar{\'i}n-Jim{\'e}nez, R.~Mu{\~{n}}oz-Salinas,
  F.~J. Madrid-Cuevas, and R.~Medina-Carnicer, ``{S}tereo {P}ictorial
  {S}tructure for {2D} articulated human pose estimation,'' \emph{Machine
  Vision and Applications}, vol.~27, no.~2, pp. 157--174, 2015.

\bibitem{haque2016viewpoint}
A.~Haque, B.~Peng, Z.~Luo, A.~Alahi, S.~Yeung, and L.~Fei-Fei, ``Towards
  viewpoint invariant 3d human pose estimation,'' in \emph{ECCV}, October 2016.

\bibitem{gasparrini2016proposal}
S.~Gasparrini, E.~Cippitelli, E.~Gambi, S.~Spinsante, J.~W{\aa}hsl{\'e}n,
  I.~Orhan, and T.~Lindh, ``Proposal and experimental evaluation of fall
  detection solution based on wearable and depth data fusion,'' in \emph{ICT
  innovations}.\hskip 1em plus 0.5em minus 0.4em\relax Springer, 2016, pp.
  99--108.

\bibitem{rodriguez2008action}
M.~D. Rodriguez, J.~Ahmed, and M.~Shah, ``Action match a spatio-temporal
  maximum average correlation height filter for action recognition,'' in
  \emph{CVPR}.\hskip 1em plus 0.5em minus 0.4em\relax IEEE, 2008, pp. 1--8.

\bibitem{marszalek09}
M.~Marsza{\l}ek, I.~Laptev, and C.~Schmid, ``Actions in context,'' in
  \emph{CVPR}, 2009.

\bibitem{schuldt2004recognizing}
C.~Schuldt, I.~Laptev, and B.~Caputo, ``Recognizing human actions: A local
  {SVM} approach,'' in \emph{ICPR}, vol.~3.\hskip 1em plus 0.5em minus
  0.4em\relax IEEE, 2004, pp. 32--36.

\bibitem{KarpathyCVPR14}
A.~Karpathy, G.~Toderici, S.~Shetty, T.~Leung, R.~Sukthankar, and L.~Fei-Fei,
  ``Large-scale video classification with convolutional neural networks,'' in
  \emph{CVPR}, 2014.

\bibitem{chaquet2013survey}
J.~M. Chaquet, E.~J. Carmona, and A.~Fern{\'a}ndez-Caballero, ``A survey of
  video datasets for human action and activity recognition,'' \emph{CVIU}, vol.
  117, no.~6, pp. 633--659, 2013.

\bibitem{wang2003silhouette}
L.~Wang, T.~Tan, H.~Ning, and W.~Hu, ``Silhouette analysis-based gait
  recognition for human identification,'' \emph{PAMI}, vol.~25, no.~12, pp.
  1505--1518, 2003.

\bibitem{ActionsAsSpaceTimeShapes_iccv05}
M.~Blank, L.~Gorelick, E.~Shechtman, M.~Irani, and R.~Basri, ``Actions as
  space-time shapes,'' in \emph{ICCV}, 2005, pp. 1395--1402.

\bibitem{weinland2006free}
D.~Weinland, R.~Ronfard, and E.~Boyer, ``Free viewpoint action recognition
  using motion history volumes,'' \emph{CVIU}, vol. 104, no.~2, pp. 249--257,
  2006.

\bibitem{kim2007tensor}
T.-K. Kim, S.-F. Wong, and R.~Cipolla, ``Tensor canonical correlation analysis
  for action classification,'' in \emph{CVPR}.\hskip 1em plus 0.5em minus
  0.4em\relax IEEE, 2007, pp. 1--8.

\bibitem{wang2007human}
Y.~Wang, K.~Huang, and T.~Tan, ``Human activity recognition based on r
  transform,'' in \emph{CVPR}.\hskip 1em plus 0.5em minus 0.4em\relax IEEE,
  2007, pp. 1--8.

\bibitem{ali2008floor}
S.~Ali and M.~Shah, ``Floor fields for tracking in high density crowd scenes,''
  in \emph{ECCV}.\hskip 1em plus 0.5em minus 0.4em\relax Springer, 2008, pp.
  1--14.

\bibitem{laptev:08}
I.~Laptev, M.~Marsza{\l}ek, C.~Schmid, and B.~Rozenfeld, ``Learning realistic
  human actions from movies,'' in \emph{CVPR}, 2008.

\bibitem{tran2008human}
D.~Tran and A.~Sorokin, ``Human activity recognition with metric learning,'' in
  \emph{ECCV}.\hskip 1em plus 0.5em minus 0.4em\relax Springer, 2008, pp.
  548--561.

\bibitem{UCF-Aerial}
\BIBentryALTinterwordspacing
UCF, ``{UCF} aerial action data set,'' Online, 2009. [Online]. Available:
  \url{http://crcv.ucf.edu/data/UCF\_Aerial\_Action.php}
\BIBentrySTDinterwordspacing

\bibitem{MSRData}
Microsoft, ``{MSR Action} dataset,''
  https://www.microsoft.com/en-us/download/details.aspx?id=52315, 2009.

\bibitem{choi2009they}
W.~Choi, K.~Shahid, and S.~Savarese, ``What are they doing?: Collective
  activity classification using spatio-temporal relationship among people,'' in
  \emph{ICCVW}.\hskip 1em plus 0.5em minus 0.4em\relax IEEE, 2009, pp.
  1282--1289.

\bibitem{PETS2009}
\BIBentryALTinterwordspacing
{RUCV Group}, Online, 2017-07-30. [Online]. Available:
  \url{http://www.cvg.reading.ac.uk/PETS2009/a.html}
\BIBentrySTDinterwordspacing

\bibitem{liu2009recognizing}
J.~Liu, J.~Luo, and M.~Shah, ``Recognizing realistic actions from videos “in
  the wild”,'' in \emph{CVPR}.\hskip 1em plus 0.5em minus 0.4em\relax IEEE,
  2009, pp. 1996--2003.

\bibitem{niebles2010modeling}
J.~C. Niebles, C.-W. Chen, and L.~Fei-Fei, ``Modeling temporal structure of
  decomposable motion segments for activity classification,'' in
  \emph{ECCV}.\hskip 1em plus 0.5em minus 0.4em\relax Springer, 2010, pp.
  392--405.

\bibitem{auvinet2010multiple}
E.~Auvinet, C.~Rougier, J.~Meunier, A.~St-Arnaud, and J.~Rousseau, ``Multiple
  cameras fall dataset,'' \emph{DIRO-Universit{\'e} de Montr{\'e}al, Tech.
  Rep}, vol. 1350, 2010.

\bibitem{blunsden2010behave}
S.~Blunsden and R.~Fisher, ``The {BEHAVE} video dataset: ground truthed video
  for multi-person behavior classification,'' \emph{Annals of the BMVA},
  vol.~4, no. 1-12, p.~4, 2010.

\bibitem{patron2010high}
A.~Patron-Perez, M.~Marszalek, A.~Zisserman, and I.~D. Reid, ``High five:
  Recognising human interactions in {TV} shows.'' in \emph{BMVC}, vol.~1.\hskip
  1em plus 0.5em minus 0.4em\relax Citeseer, 2010, p.~2.

\bibitem{UT-Interaction-Data}
\BIBentryALTinterwordspacing
M.~S. Ryoo and J.~K. Aggarwal, ``{UT}-{I}nteraction {D}ataset, {ICPR} contest
  on {S}emantic {D}escription of {H}uman {A}ctivities ({SDHA}),'' Online, 2010.
  [Online]. Available:
  \url{http://cvrc.ece.utexas.edu/SDHA2010/Human\_Interaction.html}
\BIBentrySTDinterwordspacing

\bibitem{UT-Tower-Data}
\BIBentryALTinterwordspacing
C.-C. Chen, M.~S. Ryoo, and J.~K. Aggarwal, ``{UT}-{T}ower {D}ataset: {A}erial
  {V}iew {A}ctivity {C}lassification {C}hallenge,'' Online, 2010. [Online].
  Available:
  \url{http://cvrc.ece.utexas.edu/SDHA2010/Aerial\_View\_Activity.html}
\BIBentrySTDinterwordspacing

\bibitem{denina2011videoweb}
G.~Denina, B.~Bhanu, H.~T. Nguyen, C.~Ding, A.~Kamal, C.~Ravishankar,
  A.~Roy-Chowdhury, A.~Ivers, and B.~Varda, ``Videoweb dataset for multi-camera
  activities and non-verbal communication,'' in \emph{Distributed Video Sensor
  Networks}.\hskip 1em plus 0.5em minus 0.4em\relax Springer, 2011, pp.
  335--347.

\bibitem{delaitre2010recognizing}
V.~Delaitre, I.~Laptev, and J.~Sivic, ``Recognizing human actions in still
  images: a study of bag-of-features and part-based representations,'' in
  \emph{BMVC}, 2010.

\bibitem{singh2010muhavi}
S.~Singh, S.~A. Velastin, and H.~Ragheb, ``Muhavi: A multicamera human action
  video dataset for the evaluation of action recognition methods,'' in
  \emph{Advanced Video and Signal Based Surveillance (AVSS)}.\hskip 1em plus
  0.5em minus 0.4em\relax IEEE, 2010, pp. 48--55.

\bibitem{oh2011large}
S.~Oh, A.~Hoogs, A.~Perera, N.~Cuntoor, C.-C. Chen, J.~T. Lee, S.~Mukherjee,
  J.~Aggarwal, H.~Lee, L.~Davis \emph{et~al.}, ``A large-scale benchmark
  dataset for event recognition in surveillance video,'' in \emph{CVPR}.\hskip
  1em plus 0.5em minus 0.4em\relax IEEE, 2011, pp. 3153--3160.

\bibitem{UCF-Arg}
\BIBentryALTinterwordspacing
UCF, ``{UCF-ARG} data set,'' Online, 2011. [Online]. Available:
  \url{http://crcv.ucf.edu/data/UCF-ARG.php}
\BIBentrySTDinterwordspacing

\bibitem{Kuehne11}
H.~Kuehne, H.~Jhuang, E.~Garrote, T.~Poggio, and T.~Serre, ``{HMDB}: A large
  video database for human motion recognition,'' in \emph{ICCV}, 2011.

\bibitem{reddy2013recognizing}
K.~K. Reddy and M.~Shah, ``Recognizing 50 human action categories of web
  videos,'' \emph{Machine Vision and Applications}, vol.~24, no.~5, pp.
  971--981, 2013.

\bibitem{ryoo2013first}
M.~S. Ryoo and L.~Matthies, ``First-person activity recognition: What are they
  doing to me?'' in \emph{CVPR}, June 2013.

\bibitem{soomro2012ucf101}
K.~Soomro, A.~R. Zamir, and M.~Shah, ``{UCF101}: A dataset of 101 human actions
  classes from videos in the wild,'' \emph{arXiv preprint arXiv:1212.0402},
  2012.

\bibitem{waltner14a}
G.~Waltner, T.~Mauthner, and H.~Bischof, ``{Improved Sport Activity Recognition
  using Spatio-temporal Context},'' in \emph{{Proc. DVS-Conference on Computer
  Science in Sport (DVS/GSSS)}}, 2014.

\bibitem{lee2015stare}
K.~Lee, D.~Ognibene, H.~J. Chang, T.-K. Kim, and Y.~Demiris, ``{STARE}:
  Spatio-temporal attention relocation for multiple structured activities
  detection,'' \emph{IEEE Transactions on Image Processing}, vol.~24, no.~12,
  pp. 5916--5927, 2015.

\bibitem{caba2015activitynet}
F.~Caba~Heilbron, V.~Escorcia, B.~Ghanem, and J.~Carlos~Niebles,
  ``{ActivityNet}: A large-scale video benchmark for human activity
  understanding,'' in \emph{CVPR}, 2015, pp. 961--970.

\bibitem{wang2011learning}
Y.~Wang, D.~Tran, and Z.~Liao, ``Learning hierarchical poselets for human
  parsing,'' in \emph{CVPR}.\hskip 1em plus 0.5em minus 0.4em\relax IEEE, 2011,
  pp. 1705--1712.

\bibitem{ohya1994human}
J.~Ohya and F.~Kishino, ``Human posture estimation from multiple images using
  genetic algorithm,'' in \emph{ICPR}, vol.~1.\hskip 1em plus 0.5em minus
  0.4em\relax IEEE, 1994, pp. 750--753.

\bibitem{mckenna1998real}
S.~J. McKenna and S.~Gong, ``Real-time face pose estimation,'' \emph{Real-Time
  Imaging}, vol.~4, no.~5, pp. 333--347, 1998.

\bibitem{ueda2003hand}
E.~Ueda, Y.~Matsumoto, M.~Imai, and T.~Ogasawara, ``A hand-pose estimation for
  vision-based human interfaces,'' \emph{IEEE Transactions on Industrial
  Electronics}, vol.~50, no.~4, pp. 676--684, 2003.

\bibitem{ng2002composite}
J.~Ng and S.~Gong, ``Composite support vector machines for detection of faces
  across views and pose estimation,'' \emph{Image and Vision Computing},
  vol.~20, no.~5, pp. 359--368, 2002.

\bibitem{agarwal20043d}
A.~Agarwal and B.~Triggs, ``{3D} human pose from silhouettes by relevance
  vector regression,'' in \emph{CVPR}, vol.~2.\hskip 1em plus 0.5em minus
  0.4em\relax IEEE, 2004, pp. II--II.

\bibitem{ramanan2007learning}
D.~Ramanan, ``Learning to parse images of articulated bodies,'' in
  \emph{Advances in neural information processing systems}, 2007, pp.
  1129--1136.

\bibitem{johnson2010pose}
S.~Johnson and M.~Everingham, ``Clustered pose and nonlinear appearance models
  for human pose estimation,'' in \emph{BMVC}, 2010, pp. 12.1--12.11.

\bibitem{ouyang2014multi}
W.~Ouyang, X.~Chu, and X.~Wang, ``Multi-source deep learning for human pose
  estimation,'' in \emph{CVPR}, 2014, pp. 2329--2336.

\bibitem{cherian2014mixing}
A.~Cherian, J.~Mairal, K.~Alahari, and C.~Schmid, ``Mixing body-part sequences
  for human pose estimation,'' in \emph{CVPR}, 2014, pp. 2353--2360.

\bibitem{pishchulin2013strong}
L.~Pishchulin, M.~Andriluka, P.~Gehler, and B.~Schiele, ``Strong appearance and
  expressive spatial models for human pose estimation,'' in \emph{ICCV}, 2013,
  pp. 3487--3494.

\bibitem{wei2016cpm}
S.-E. Wei, V.~Ramakrishna, T.~Kanade, and Y.~Sheikh, ``Convolutional pose
  machines,'' in \emph{CVPR}, 2016.

\bibitem{zhou2016sparseness}
X.~Zhou, M.~Zhu, S.~Leonardos, K.~G. Derpanis, and K.~Daniilidis, ``Sparseness
  meets deepness: {3D} human pose estimation from monocular video,'' in
  \emph{Proceedings of the IEEE Conference on Computer Vision and Pattern
  Recognition}, 2016, pp. 4966--4975.

\bibitem{toshev2014deeppose}
A.~Toshev and C.~Szegedy, ``{DeepPose}: Human pose estimation via deep neural
  networks,'' in \emph{CVPR}, 2014, pp. 1653--1660.

\bibitem{fan2015combining}
X.~Fan, K.~Zheng, Y.~Lin, and S.~Wang, ``Combining local appearance and
  holistic view: Dual-source deep neural networks for human pose estimation,''
  in \emph{CVPR}, 2015, pp. 1347--1355.

\bibitem{pfister2015flowing}
T.~Pfister, J.~Charles, and A.~Zisserman, ``Flowing {ConvNets} for human pose
  estimation in videos,'' in \emph{ICCV}, 2015, pp. 1913--1921.

\bibitem{cao2017realtime}
Z.~Cao, T.~Simon, S.-E. Wei, and Y.~Sheikh, ``Realtime multi-person {2D} pose
  estimation using part affinity fields,'' in \emph{CVPR}, 2017.

\bibitem{andriluka20142d}
M.~Andriluka, L.~Pishchulin, P.~Gehler, and B.~Schiele, ``{2D} human pose
  estimation: New benchmark and state of the art analysis,'' in \emph{CVPR},
  2014, pp. 3686--3693.

\bibitem{liu2015survey}
Z.~Liu, J.~Zhu, J.~Bu, and C.~Chen, ``A survey of human pose estimation: the
  body parts parsing based methods,'' \emph{Journal of Visual Communication and
  Image Representation}, vol.~32, pp. 10--19, 2015.

\bibitem{rautaray2015vision}
S.~S. Rautaray and A.~Agrawal, ``Vision based hand gesture recognition for
  human computer interaction: a survey,'' \emph{Artificial Intelligence
  Review}, vol.~43, no.~1, pp. 1--54, 2015.

\bibitem{poppe2010survey}
R.~Poppe, ``A survey on vision-based human action recognition,'' \emph{Image
  and vision computing}, vol.~28, no.~6, pp. 976--990, 2010.

\bibitem{weinland2011survey}
D.~Weinland, R.~Ronfard, and E.~Boyer, ``A survey of vision-based methods for
  action representation, segmentation and recognition,'' \emph{CVIU}, vol. 115,
  no.~2, pp. 224--241, 2011.

\bibitem{vishwakarma2013survey}
S.~Vishwakarma and A.~Agrawal, ``A survey on activity recognition and behavior
  understanding in video surveillance,'' \emph{The Visual Computer}, vol.~29,
  no.~10, pp. 983--1009, 2013.

\bibitem{aggarwal2014human}
J.~K. Aggarwal and L.~Xia, ``Human activity recognition from {3D} data: A
  review,'' \emph{Pattern Recognition Letters}, vol.~48, pp. 70--80, 2014.

\bibitem{guo2014survey}
G.~Guo and A.~Lai, ``A survey on still image based human action recognition,''
  \emph{Pattern Recognition}, vol.~47, no.~10, pp. 3343--3361, 2014.

\bibitem{ziaeefard2015semantic}
M.~Ziaeefard and R.~Bergevin, ``Semantic human activity recognition: a
  literature review,'' \emph{Pattern Recognition}, vol.~48, no.~8, pp.
  2329--2345, 2015.

\bibitem{herath2017going}
S.~Herath, M.~Harandi, and F.~Porikli, ``Going deeper into action recognition:
  A survey,'' \emph{Image and Vision Computing}, vol.~60, pp. 4--21, 2017.

\bibitem{sargano2017comprehensive}
A.~B. Sargano, P.~Angelov, and Z.~Habib, ``A comprehensive review on
  handcrafted and learning-based action representation approaches for human
  activity recognition,'' \emph{Applied Sciences}, vol.~7, no.~1, p. 110, 2017.

\bibitem{campbell1996invariant}
L.~W. Campbell, D.~A. Becker, A.~Azarbayejani, A.~F. Bobick, and A.~Pentland,
  ``Invariant features for 3-d gesture recognition,'' in \emph{Proceedings of
  the Second International Conference on Automatic Face and Gesture
  Recognition}.\hskip 1em plus 0.5em minus 0.4em\relax IEEE, 1996, pp.
  157--162.

\bibitem{curio2000walking}
C.~Curio, J.~Edelbrunner, T.~Kalinke, C.~Tzomakas, and W.~Von~Seelen, ``Walking
  pedestrian recognition,'' \emph{IEEE Transactions on intelligent
  transportation systems (ITS)}, vol.~1, no.~3, pp. 155--163, 2000.

\bibitem{niyogi1994analyzing}
S.~A. Niyogi, E.~H. Adelson \emph{et~al.}, ``Analyzing and recognizing walking
  figures in {XYT},'' in \emph{CVPR}, vol.~94, 1994, pp. 469--474.

\bibitem{bobick2001recognition}
A.~F. Bobick and J.~W. Davis, ``The recognition of human movement using
  temporal templates,'' \emph{PAMI}, vol.~23, no.~3, pp. 257--267, 2001.

\bibitem{ke2007event}
Y.~Ke, R.~Sukthankar, and M.~Hebert, ``Event detection in crowded videos,'' in
  \emph{ICCV}.\hskip 1em plus 0.5em minus 0.4em\relax IEEE, 2007, pp. 1--8.

\bibitem{ikizler2008recognizing}
N.~Ikizler, R.~G. Cinbis, S.~Pehlivan, and P.~Duygulu, ``Recognizing actions
  from still images,'' in \emph{ICPR}.\hskip 1em plus 0.5em minus 0.4em\relax
  IEEE, 2008, pp. 1--4.

\bibitem{yao2010modeling}
B.~Yao and L.~Fei-Fei, ``Modeling mutual context of object and human pose in
  human-object interaction activities,'' in \emph{CVPR}.\hskip 1em plus 0.5em
  minus 0.4em\relax IEEE, 2010, pp. 17--24.

\bibitem{feichtenhofer2015dynamically}
C.~Feichtenhofer, A.~Pinz, and R.~P. Wildes, ``Dynamically encoded actions
  based on spacetime saliency,'' in \emph{CVPR}, 2015, pp. 2755--2764.

\bibitem{yue2015beyond}
J.~Yue-Hei~Ng, M.~Hausknecht, S.~Vijayanarasimhan, O.~Vinyals, R.~Monga, and
  G.~Toderici, ``Beyond short snippets: Deep networks for video
  classification,'' in \emph{CVPR}, 2015, pp. 4694--4702.

\bibitem{yang2010recognizing}
W.~Yang, Y.~Wang, and G.~Mori, ``Recognizing human actions from still images
  with latent poses,'' in \emph{CVPR}.\hskip 1em plus 0.5em minus 0.4em\relax
  IEEE, 2010, pp. 2030--2037.

\bibitem{yamato1992recognizing}
J.~Yamato, J.~Ohya, and K.~Ishii, ``Recognizing human action in time-sequential
  images using hidden markov model,'' in \emph{CVPR}.\hskip 1em plus 0.5em
  minus 0.4em\relax IEEE, 1992, pp. 379--385.

\bibitem{efros2003recognizing}
A.~A. Efros, A.~C. Berg, G.~Mori, and J.~Malik, ``Recognizing action at a
  distance,'' in \emph{ICCV}.\hskip 1em plus 0.5em minus 0.4em\relax IEEE,
  2003, p. 726.

\bibitem{blank2005actions}
M.~Blank, L.~Gorelick, E.~Shechtman, M.~Irani, and R.~Basri, ``Actions as
  space-time shapes,'' in \emph{ICCV}, vol.~2.\hskip 1em plus 0.5em minus
  0.4em\relax IEEE, 2005, pp. 1395--1402.

\bibitem{ji20133d}
S.~Ji, W.~Xu, M.~Yang, and K.~Yu, ``{3D} convolutional neural networks for
  human action recognition,'' \emph{PAMI}, vol.~35, no.~1, pp. 221--231, 2013.

\bibitem{karpathy2014large}
A.~Karpathy, G.~Toderici, S.~Shetty, T.~Leung, R.~Sukthankar, and L.~Fei-Fei,
  ``Large-scale video classification with convolutional neural networks,'' in
  \emph{CVPR}, 2014, pp. 1725--1732.

\bibitem{simonyan2014two}
K.~Simonyan and A.~Zisserman, ``Two-stream convolutional networks for action
  recognition in videos,'' in \emph{Advances in neural information processing
  systems}, 2014, pp. 568--576.

\bibitem{feichtenhofer2016convolutional}
C.~Feichtenhofer, A.~Pinz, and A.~Zisserman, ``Convolutional two-stream network
  fusion for video action recognition,'' in \emph{CVPR}, 2016, pp. 1933--1941.

\bibitem{wang2016actions}
X.~Wang, A.~Farhadi, and A.~Gupta, ``Actions\~{}transformations,'' in
  \emph{CVPR}, 2016, pp. 2658--2667.

\bibitem{ma2016learning}
S.~Ma, L.~Sigal, and S.~Sclaroff, ``Learning activity progression in lstms for
  activity detection and early detection,'' in \emph{CVPR}, 2016, pp.
  1942--1950.

\bibitem{kar2016adascan}
A.~Kar, N.~Rai, K.~Sikka, and G.~Sharma, ``{AdaScan}: Adaptive scan pooling in
  deep convolutional neural networks for human action recognition in videos,''
  in \emph{CVPR}, 2017.

\bibitem{park2003recognition}
S.~Park and J.~Aggarwal, ``Recognition of two-person interactions using a
  hierarchical bayesian network,'' in \emph{First ACM SIGMM international
  workshop on Video surveillance}.\hskip 1em plus 0.5em minus 0.4em\relax ACM,
  2003, pp. 65--76.

\bibitem{ryoo2009spatio}
M.~S. Ryoo and J.~K. Aggarwal, ``Spatio-temporal relationship match: Video
  structure comparison for recognition of complex human activities,'' in
  \emph{ICCV}.\hskip 1em plus 0.5em minus 0.4em\relax IEEE, 2009, pp.
  1593--1600.

\bibitem{lan2012social}
T.~Lan, L.~Sigal, and G.~Mori, ``Social roles in hierarchical models for human
  activity recognition,'' in \emph{CVPR}.\hskip 1em plus 0.5em minus
  0.4em\relax IEEE, 2012, pp. 1354--1361.

\bibitem{deng2015deep}
Z.~Deng, M.~Zhai, L.~Chen, Y.~Liu, S.~Muralidharan, M.~J. Roshtkhari, and
  G.~Mori, ``Deep structured models for group activity recognition,'' in
  \emph{BMVC}, 2015.

\bibitem{deng2016structure}
Z.~Deng, A.~Vahdat, H.~Hu, and G.~Mori, ``Structure inference machines:
  Recurrent neural networks for analyzing relations in group activity
  recognition,'' in \emph{CVPR}, 2016, pp. 4772--4781.

\bibitem{richard2015shrp}
C.~M. Richard, J.~L. Brown, J.~L. Campbell, J.~S. Graving, J.~Milton, and
  I.~van Schalkwyk, ``{SHRP} 2 implementation assistance program concept to
  countermeasure-research to deployment using the {SHRP2} safety data {(NDS)}
  influence of roadway design features on episodic,'' 2015.

\bibitem{habibovic2013driver}
A.~Habibovic, E.~Tivesten, N.~Uchida, J.~B{\"a}rgman, and M.~L. Aust, ``Driver
  behavior in car-to-pedestrian incidents: An application of the driving
  reliability and error analysis method {(DREAM)},'' \emph{Accident Analysis \&
  Prevention}, vol.~50, pp. 554--565, 2013.

\bibitem{dozza2014introducing}
M.~Dozza and J.~Werneke, ``Introducing naturalistic cycling data: What factors
  influence bicyclists’ safety in the real world?'' \emph{Transportation
  research part F: traffic psychology and behaviour}, vol.~24, pp. 83--91,
  2014.

\bibitem{CAVIAR}
{EC Funded CAVIAR project}, ``Caviar: Context aware vision using image-based
  active recognition,'' http://homepages.inf.ed.ac.uk/rbf/CAVIARDATA1/, 2002.

\bibitem{kotseruba2016joint}
I.~Kotseruba, A.~Rasouli, and J.~K. Tsotsos, ``Joint attention in autonomous
  driving (jaad),'' \emph{arXiv preprint arXiv:1609.04741}, 2016.

\bibitem{schneider2013pedestrian}
N.~Schneider and D.~M. Gavrila, ``Pedestrian path prediction with recursive
  bayesian filters: A comparative study,'' in \emph{German Conference on
  Pattern Recognition}.\hskip 1em plus 0.5em minus 0.4em\relax Springer, 2013,
  pp. 174--183.

\bibitem{kooij2014context}
J.~F.~P. Kooij, N.~Schneider, F.~Flohr, and D.~M. Gavrila, ``Context-based
  pedestrian path prediction,'' in \emph{ECCV}.\hskip 1em plus 0.5em minus
  0.4em\relax Springer, 2014, pp. 618--633.

\bibitem{ohn2014head}
E.~Ohn-Bar, S.~Martin, A.~Tawari, and M.~M. Trivedi, ``Head, eye, and hand
  patterns for driver activity recognition,'' in \emph{ICPR}.\hskip 1em plus
  0.5em minus 0.4em\relax IEEE, 2014, pp. 660--665.

\bibitem{molchanov2015multi}
P.~Molchanov, S.~Gupta, K.~Kim, and K.~Pulli, ``Multi-sensor system for
  driver's hand-gesture recognition,'' in \emph{IEEE International Conference
  and Workshops on Automatic Face and Gesture Recognition (FG)}, vol.~1.\hskip
  1em plus 0.5em minus 0.4em\relax IEEE, 2015, pp. 1--8.

\bibitem{laugier2011probabilistic}
C.~Laugier, I.~E. Paromtchik, M.~Perrollaz, M.~Yong, J.-D. Yoder, C.~Tay,
  K.~Mekhnacha, and A.~N{\`e}gre, ``Probabilistic analysis of dynamic scenes
  and collision risks assessment to improve driving safety,'' \emph{IEEE
  Intelligent Transportation Systems Magazine}, vol.~3, no.~4, pp. 4--19, 2011.

\bibitem{li2016recognizing}
B.~Li, T.~Wu, C.~Xiong, and S.-C. Zhu, ``Recognizing car fluents from video,''
  in \emph{CVPR}, 2016, pp. 3803--3812.

\bibitem{kooij2014analysis}
J.~F. Kooij, N.~Schneider, and D.~M. Gavrila, ``Analysis of pedestrian dynamics
  from a vehicle perspective,'' in \emph{Intelligent Vehicles Symposium
  (IV)}.\hskip 1em plus 0.5em minus 0.4em\relax IEEE, 2014, pp. 1445--1450.

\bibitem{kohler2012early}
S.~K{\"o}hler, M.~Goldhammer, S.~Bauer, K.~Doll, U.~Brunsmann, and
  K.~Dietmayer, ``Early detection of the pedestrian's intention to cross the
  street,'' in \emph{Intelligent Transportation Systems (ITSC)}.\hskip 1em plus
  0.5em minus 0.4em\relax IEEE, 2012, pp. 1759--1764.

\bibitem{phan2013estimating}
M.~T. Phan, I.~Thouvenin, V.~Fremont, and V.~Cherfaoui, ``Estimating driver
  unawareness of pedestrian based on visual behaviors and driving behaviors,''
  2013.

\bibitem{bahram2016combined}
M.~Bahram, C.~Hubmann, A.~Lawitzky, M.~Aeberhard, and D.~Wollherr, ``A combined
  model-and learning-based framework for interaction-aware maneuver
  prediction,'' \emph{IEEE Transactions on Intelligent Transportation Systems},
  vol.~17, no.~6, pp. 1538--1550, 2016.

\bibitem{ohn2016looking}
E.~Ohn-Bar and M.~M. Trivedi, ``Looking at humans in the age of self-driving
  and highly automated vehicles,'' \emph{IEEE Transactions on Intelligent
  Vehicles}, vol.~1, no.~1, pp. 90--104, 2016.

\bibitem{schulz2015controlled}
A.~T. Schulz and R.~Stiefelhagen, ``A controlled interactive multiple model
  filter for combined pedestrian intention recognition and path prediction,''
  in \emph{Intelligent Transportation Systems (ITSC)}.\hskip 1em plus 0.5em
  minus 0.4em\relax IEEE, 2015, pp. 173--178.

\bibitem{hashimoto2015probability}
Y.~Hashimoto, Y.~Gu, L.-T. Hsu, and S.~Kamijo, ``Probability estimation for
  pedestrian crossing intention at signalized crosswalks,'' in
  \emph{International Conference on Vehicular Electronics and Safety
  (ICVES)}.\hskip 1em plus 0.5em minus 0.4em\relax IEEE, 2015, pp. 114--119.

\bibitem{brouwer2016comparison}
N.~Brouwer, H.~Kloeden, and C.~Stiller, ``Comparison and evaluation of
  pedestrian motion models for vehicle safety systems,'' in \emph{Intelligent
  Transportation Systems (ITSC)}.\hskip 1em plus 0.5em minus 0.4em\relax IEEE,
  2016, pp. 2207--2212.

\bibitem{trivedi2015trajectory}
M.~M. Trivedi, T.~B. Moeslund \emph{et~al.}, ``Trajectory analysis and
  prediction for improved pedestrian safety: Integrated framework and
  evaluations,'' in \emph{Intelligent Vehicles Symposium (IV)}.\hskip 1em plus
  0.5em minus 0.4em\relax IEEE, 2015, pp. 330--335.

\bibitem{quintero2014pedestrian}
R.~Quintero, I.~Parra, D.~F. Llorca, and M.~Sotelo, ``Pedestrian path
  prediction based on body language and action classification,'' in
  \emph{Intelligent Transportation Systems (ITSC)}.\hskip 1em plus 0.5em minus
  0.4em\relax IEEE, 2014, pp. 679--684.

\bibitem{volz2016data}
B.~V{\"o}lz, K.~Behrendt, H.~Mielenz, I.~Gilitschenski, R.~Siegwart, and
  J.~Nieto, ``A data-driven approach for pedestrian intention estimation,'' in
  \emph{Intelligent Transportation Systems (ITSC)}.\hskip 1em plus 0.5em minus
  0.4em\relax IEEE, 2016, pp. 2607--2612.

\bibitem{bandyopadhyay2013intention}
T.~Bandyopadhyay, C.~Z. Jie, D.~Hsu, M.~H. Ang~Jr, D.~Rus, and E.~Frazzoli,
  ``Intention-aware pedestrian avoidance,'' in \emph{Experimental
  Robotics}.\hskip 1em plus 0.5em minus 0.4em\relax Springer, 2013, pp.
  963--977.

\bibitem{bai2015intention}
H.~Bai, S.~Cai, N.~Ye, D.~Hsu, and W.~S. Lee, ``Intention-aware online pomdp
  planning for autonomous driving in a crowd,'' in \emph{ICRA}.\hskip 1em plus
  0.5em minus 0.4em\relax IEEE, 2015, pp. 454--460.

\bibitem{hariyono2016estimation}
J.~Hariyono, A.~Shahbaz, L.~Kurnianggoro, and K.-H. Jo, ``Estimation of
  collision risk for improving driver's safety,'' in \emph{Annual Conference of
  Industrial Electronics Society (IECON)}.\hskip 1em plus 0.5em minus
  0.4em\relax IEEE, 2016, pp. 901--906.

\bibitem{kwak2017pedestrian}
J.-Y. Kwak, B.~C. Ko, and J.-Y. Nam, ``Pedestrian intention prediction based on
  dynamic fuzzy automata for vehicle driving at nighttime,'' \emph{Infrared
  Physics \& Technology}, vol.~81, pp. 41--51, 2017.

\bibitem{pellegrini2009you}
S.~Pellegrini, A.~Ess, K.~Schindler, and L.~Van~Gool, ``You'll never walk
  alone: Modeling social behavior for multi-target tracking,'' in
  \emph{ICCV}.\hskip 1em plus 0.5em minus 0.4em\relax IEEE, 2009, pp. 261--268.

\bibitem{madrigal2014intention}
F.~Madrigal, J.-B. Hayet, and F.~Lerasle, ``Intention-aware multiple pedestrian
  tracking,'' in \emph{ICPR}.\hskip 1em plus 0.5em minus 0.4em\relax IEEE,
  2014, pp. 4122--4127.

\bibitem{kohler2013autonomous}
S.~K{\"o}hler, B.~Schreiner, S.~Ronalter, K.~Doll, U.~Brunsmann, and
  K.~Zindler, ``Autonomous evasive maneuvers triggered by infrastructure-based
  detection of pedestrian intentions,'' in \emph{Intelligent Vehicles Symposium
  (IV)}.\hskip 1em plus 0.5em minus 0.4em\relax IEEE, 2013, pp. 519--526.

\bibitem{kohler2015stereo}
S.~K{\"o}hler, M.~Goldhammer, K.~Zindler, K.~Doll, and K.~Dietmeyer,
  ``Stereo-vision-based pedestrian's intention detection in a moving vehicle,''
  in \emph{Intelligent Transportation Systems (ITSC)}.\hskip 1em plus 0.5em
  minus 0.4em\relax IEEE, 2015, pp. 2317--2322.

\bibitem{rangesh2018vehicles}
A.~Rangesh and M.~M. Trivedi, ``When vehicles see pedestrians with phones: A
  multi-cue framework for recognizing phone-based activities of pedestrians,''
  \emph{arXiv preprint arXiv:1801.08234}, 2018.

\bibitem{rasouli2017they}
A.~Rasouli, I.~Kotseruba, and J.~K. Tsotsos, ``Are they going to cross? a
  benchmark dataset and baseline for pedestrian crosswalk behavior,'' in
  \emph{ICCVW}, 2017, pp. 206--213.

\bibitem{volz2015feature}
B.~V{\"o}lz, H.~Mielenz, G.~Agamennoni, and R.~Siegwart, ``Feature relevance
  estimation for learning pedestrian behavior at crosswalks,'' in
  \emph{Intelligent Transportation Systems (ITSC)}.\hskip 1em plus 0.5em minus
  0.4em\relax IEEE, 2015, pp. 854--860.

\bibitem{schneemann2016context}
F.~Schneemann and P.~Heinemann, ``Context-based detection of pedestrian
  crossing intention for autonomous driving in urban environments,'' in
  \emph{IROS}.\hskip 1em plus 0.5em minus 0.4em\relax IEEE, 2016, pp.
  2243--2248.

\bibitem{park2016hi}
C.~Park, J.~Ond{\v{r}}ej, M.~Gilbert, K.~Freeman, and C.~O'Sullivan, ``{HI
  Robot}: Human intention-aware robot planning for safe and efficient
  navigation in crowds,'' in \emph{IROS}.\hskip 1em plus 0.5em minus
  0.4em\relax IEEE, 2016, pp. 3320--3326.

\bibitem{goldhammer2013early}
M.~Goldhammer, M.~Gerhard, S.~Zernetsch, K.~Doll, and U.~Brunsmann, ``Early
  prediction of a pedestrian's trajectory at intersections,'' in
  \emph{Intelligent Transportation Systems (ITSC)}.\hskip 1em plus 0.5em minus
  0.4em\relax IEEE, 2013, pp. 237--242.

\bibitem{gonzalez2016review}
D.~Gonz{\'a}lez, J.~P{\'e}rez, V.~Milan{\'e}s, and F.~Nashashibi, ``A review of
  motion planning techniques for automated vehicles,'' \emph{ITSC}, vol.~17,
  no.~4, pp. 1135--1145, 2016.

\bibitem{fraichard2001fuzzy}
T.~Fraichard and P.~Garnier, ``Fuzzy control to drive car-like vehicles,''
  \emph{Robotics and autonomous systems}, vol.~34, no.~1, pp. 1--22, 2001.

\bibitem{brooks1990elephants}
R.~A. Brooks, ``Elephants don't play chess,'' \emph{Robotics and autonomous
  systems}, vol.~6, no. 1-2, pp. 3--15, 1990.

\bibitem{mitchell1987planning}
J.~S. Mitchell, D.~W. Payton, and D.~M. Keirsey, ``Planning and reasoning for
  autonomous vehicle control,'' \emph{International Journal of Intelligent
  Systems}, vol.~2, no.~2, pp. 129--198, 1987.

\bibitem{ferguson2008reasoning}
D.~Ferguson, C.~Baker, M.~Likhachev, and J.~Dolan, ``A reasoning framework for
  autonomous urban driving,'' in \emph{Intelligent Vehicles Symposium, 2008
  IEEE}.\hskip 1em plus 0.5em minus 0.4em\relax IEEE, 2008, pp. 775--780.

\bibitem{davis1980meta}
R.~Davis, ``Meta-rules: Reasoning about control,'' \emph{Artificial
  intelligence}, vol.~15, no.~3, pp. 179--222, 1980.

\bibitem{cucchiara2000image}
R.~Cucchiara, M.~Piccardi, and P.~Mello, ``Image analysis and rule-based
  reasoning for a traffic monitoring system,'' \emph{ITSC}, vol.~1, no.~2, pp.
  119--130, 2000.

\bibitem{watson1994case}
I.~Watson and F.~Marir, ``Case-based reasoning: A review,'' \emph{The knowledge
  engineering review}, vol.~9, no.~4, pp. 327--354, 1994.

\bibitem{golding1991improving}
A.~R. Golding and P.~S. Rosenbloom, ``Improving rule-based systems through
  case-based reasoning.'' in \emph{AAAI}, vol.~1, 1991, pp. 22--27.

\bibitem{moorman1992case}
K.~Moorman and A.~Ram, ``A case-based approach to reactive control for
  autonomous robots,'' in \emph{Proceedings of the AAAI Fall symposium on AI
  for real-world autonomous mobile robots}, 1992, pp. 1--11.

\bibitem{urdiales2006purely}
C.~Urdiales, E.~J. Perez, J.~V{\'a}zquez-Salceda, M.~S{\`a}nchez-Marr{\`e}, and
  F.~Sandoval, ``A purely reactive navigation scheme for dynamic environments
  using case-based reasoning,'' \emph{Autonomous Robots}, vol.~21, no.~1, pp.
  65--78, 2006.

\bibitem{vacek2007using}
S.~Vacek, T.~Gindele, J.~M. Zollner, and R.~Dillmann, ``Using case-based
  reasoning for autonomous vehicle guidance,'' in \emph{IROS}.\hskip 1em plus
  0.5em minus 0.4em\relax IEEE, 2007, pp. 4271--4276.

\bibitem{rasouli2014visual}
A.~Rasouli and J.~K. Tsotsos, ``Visual saliency improves autonomous visual
  search,'' in \emph{Canadian Conference on Computer and Robot Vision
  (CRV)}.\hskip 1em plus 0.5em minus 0.4em\relax IEEE, 2014, pp. 111--118.

\bibitem{hertzberg2008ai}
J.~Hertzberg and R.~Chatila, ``{AI} reasoning methods for robotics,'' in
  \emph{Springer Handbook of Robotics}.\hskip 1em plus 0.5em minus 0.4em\relax
  Springer, 2008, pp. 207--223.

\bibitem{tsotsos1990analyzing}
J.~K. Tsotsos, ``Analyzing vision at the complexity level,'' \emph{Behavioral
  and brain sciences}, vol.~13, no.~3, pp. 423--445, 1990.

\bibitem{potapova2017survey}
E.~Potapova, M.~Zillich, and M.~Vincze, ``Survey of recent advances in 3d
  visual attention for robotics,'' \emph{The International Journal of Robotics
  Research}, vol.~36, no.~11, pp. 1159--1176, 2017.

\bibitem{ba2014multiple}
J.~Ba, V.~Mnih, and K.~Kavukcuoglu, ``Multiple object recognition with visual
  attention,'' \emph{arXiv preprint arXiv:1412.7755}, 2014.

\bibitem{xu2015show}
K.~Xu, J.~Ba, R.~Kiros, K.~Cho, A.~Courville, R.~Salakhudinov, R.~Zemel, and
  Y.~Bengio, ``Show, attend and tell: Neural image caption generation with
  visual attention,'' in \emph{International Conference on Machine Learning},
  2015, pp. 2048--2057.

\bibitem{rasouli2017integrating}
A.~Rasouli and J.~K. Tsotsos, ``Integrating three mechanisms of visual
  attention for active visual search,'' \emph{arXiv preprint arXiv:1702.04292},
  2017.

\bibitem{bylinskii2015towards}
Z.~Bylinskii, E.~DeGennaro, R.~Rajalingham, H.~Ruda, J.~Zhang, and J.~Tsotsos,
  ``Towards the quantitative evaluation of visual attention models,''
  \emph{Vision research}, vol. 116, pp. 258--268, 2015.

\bibitem{Chang2011fusing}
K.~Chang, T.~Liu, H.~Chen, and S.~Lai, ``Fusing generic objectness and visual
  saliency for salient object detection,'' in \emph{ICCV}, Barcelona, October
  2011.

\bibitem{Jiang2011salobj}
H.~Jiang, J.~Wang, Z.~Yuan, T.~Liu, N.~Zheng, and S.~Li, ``Automatic salient
  object segmentation based on context and shape prior,'' in \emph{The British
  Machine Vision Conference (BMVC)}, September 2011.

\bibitem{Itti1998rapidscene}
L.~Itti, C.~Koch, and E.~Niebur, ``A model of saliency-based visua attention
  for rapid scene analysis,'' \emph{Pattern Analysis and Machine Intelligence},
  vol.~20, no.~11, pp. 1254--1259, 1998.

\bibitem{Bruce2007aim}
N.~Bruce and J.~K. Tsotsos, ``Attention based on information maximization,''
  \emph{Journal of Vision}, vol.~7, no.~9, p. 950, 2007.

\bibitem{Li2015cnnsal}
G.~Li and Y.~Yu, ``Visual saliency based on multiscale deep features,'' in
  \emph{CVPR}, Boston, June 2015.

\bibitem{Zhao2015deepsal}
R.~Zhao, W.~Ouyang, H.~Li, and X.~Wang, ``Saliency detection by multi-context
  deep learning,'' in \emph{CVPR}, Boston, June 2015.

\bibitem{Cave1999top}
R.~K. Cave, ``The featuregate model of visual selection,'' \emph{Psychological
  research}, vol.~62, no. 2-3, pp. 182--194, 1999.

\bibitem{Zhu2014context}
J.~Zhu, Y.~Qiu, R.~Zhang, and J.~Huang, ``Top-down saliency detection via
  contextual pooling,'' \emph{Journal of Signal Processing Systems}, vol.~74,
  no.~1, pp. 33--46, 2014.

\bibitem{JYang2012tpdown}
J.~Yang and M.~Yangm, ``Top-down visual saliency via joint crf and dictionary
  learning,'' in \emph{CVPR}, June 2012.

\bibitem{He2016deeptop}
S.~He, R.~Lau, and Q.~Yang, ``Exemplar-driven top-down saliency detection via
  deep association,'' in \emph{CVPR}, Las Vegas, June 2016.

\bibitem{zhang2016top}
J.~Zhang, Z.~Lin, J.~Brandt, X.~Shen, and S.~Sclaroff, ``Top-down neural
  attention by excitation backprop,'' in \emph{European Conference on Computer
  Vision}.\hskip 1em plus 0.5em minus 0.4em\relax Springer, 2016, pp. 543--559.

\bibitem{wang2018video}
W.~Wang, J.~Shen, and L.~Shao, ``Video salient object detection via fully
  convolutional networks,'' \emph{{IEEE} Transactions on Image Processing},
  vol.~27, no.~1, pp. 38--49, 2018.

\bibitem{leifman2016learning}
G.~Leifman, D.~Rudoy, T.~Swedish, E.~Bayro-Corrochano, and R.~Raskar,
  ``Learning gaze transitions from depth to improve video saliency
  estimation,'' \emph{International Conference on Computer Vision (ICCV)},
  2017.

\bibitem{Borji2013quantanalysis}
A.~Borji, D.~N. Sihite, and L.~Itti, ``Quantitative analysis of humanmodel
  agreement in visual saliency modeling: A comparative study,'' \emph{IEEE
  Transactions on Image Processing}, vol.~22, no.~1, 2013.

\bibitem{filipe2013human}
S.~Filipe and L.~A. Alexandre, ``From the human visual system to the
  computational models of visual attention: a survey,'' \emph{Artificial
  Intelligence Review}, vol.~39, no.~1, pp. 1--47, 2013.

\bibitem{doshi2010attention}
A.~Doshi and M.~M. Trivedi, ``Attention estimation by simultaneous observation
  of viewer and view,'' in \emph{CVPRW}.\hskip 1em plus 0.5em minus 0.4em\relax
  IEEE, 2010, pp. 21--27.

\bibitem{tawari2014robust}
A.~Tawari and M.~M. Trivedi, ``Robust and continuous estimation of driver gaze
  zone by dynamic analysis of multiple face videos,'' in \emph{Intelligent
  Vehicles Symposium (IV)}.\hskip 1em plus 0.5em minus 0.4em\relax IEEE, 2014,
  pp. 344--349.

\bibitem{tawari2014looking}
A.~Tawari, S.~Sivaraman, M.~M. Trivedi, T.~Shannon, and M.~Tippelhofer,
  ``Looking-in and looking-out vision for urban intelligent assistance:
  Estimation of driver attentive state and dynamic surround for safe merging
  and braking,'' in \emph{Intelligent Vehicles Symposium (IV)}.\hskip 1em plus
  0.5em minus 0.4em\relax IEEE, 2014, pp. 115--120.

\bibitem{tanishige2014prediction}
R.~Tanishige, D.~Deguchi, K.~Doman, Y.~Mekada, I.~Ide, and H.~Murase,
  ``Prediction of driver's pedestrian detectability by image processing
  adaptive to visual fields of view,'' in \emph{Intelligent Transportation
  Systems (ITSC)}.\hskip 1em plus 0.5em minus 0.4em\relax IEEE, 2014, pp.
  1388--1393.

\bibitem{deng2016does}
T.~Deng, K.~Yang, Y.~Li, and H.~Yan, ``Where does the driver look?
  top-down-based saliency detection in a traffic driving environment,''
  \emph{{IEEE} Transactions on Intelligent Transportation Systems}, vol.~17,
  no.~7, pp. 2051--2062, 2016.

\bibitem{palazzi2017learning}
A.~Palazzi, F.~Solera, S.~Calderara, S.~Alletto, and R.~Cucchiara, ``Learning
  where to attend like a human driver,'' in \emph{Intelligent Vehicles
  Symposium (IV), 2017 IEEE}.\hskip 1em plus 0.5em minus 0.4em\relax IEEE,
  2017, pp. 920--925.

\bibitem{tawari2017computational}
A.~Tawari and B.~Kang, ``A computational framework for driver's visual
  attention using a fully convolutional architecture,'' in \emph{Intelligent
  Vehicles Symposium (IV), 2017 IEEE}.\hskip 1em plus 0.5em minus 0.4em\relax
  IEEE, 2017, pp. 887--894.

\bibitem{alletto2016dr}
S.~Alletto, A.~Palazzi, F.~Solera, S.~Calderara, and R.~Cucchiara, ``Dr (eye)
  ve: A dataset for attention-based tasks with applications to autonomous and
  assisted driving,'' in \emph{Conference on Computer Vision and Pattern
  Recognition Workshops (CVPRW)}, 2016, pp. 54--60.

\bibitem{amer2017cultural}
T.~Amer, K.~Ngo, and L.~Hasher, ``Cultural differences in visual attention:
  Implications for distraction processing,'' \emph{British Journal of
  Psychology}, vol. 108, no.~2, pp. 244--258, 2017.

\bibitem{bylinskii2016should}
Z.~Bylinskii, A.~Recasens, A.~Borji, A.~Oliva, A.~Torralba, and F.~Durand,
  ``Where should saliency models look next?'' in \emph{European Conference on
  Computer Vision}.\hskip 1em plus 0.5em minus 0.4em\relax Springer, 2016, pp.
  809--824.

\end{thebibliography}

\end{document}